\documentclass{article}

\usepackage{bbding}

\usepackage{hyperref}
\usepackage{url}

\usepackage{framed}
\usepackage{color}
\usepackage[utf8]{inputenc} 
\usepackage[T1]{fontenc}    
\usepackage{hyperref}       
\usepackage{url}            
\usepackage{booktabs}       
\usepackage{amsfonts}       
\usepackage{nicefrac}       
\usepackage{microtype}      
\usepackage{xcolor}         
\usepackage{color, colortbl}
\definecolor{greyC}{RGB}{180,180,180}
\definecolor{greyL}{RGB}{235,235,235}
\definecolor{citeColor}{RGB}{0,20,115}
\hypersetup{colorlinks,linkcolor={citeColor},citecolor={citeColor},urlcolor={citeColor}}
\usepackage{multicol}
\usepackage{multirow}
\usepackage{makecell}

\usepackage{amsmath}
\usepackage{amssymb}
\usepackage{mathtools}
\usepackage{amsthm}
\usepackage{algorithmic,algorithm}

\usepackage{soul}
\usepackage{microtype}
\usepackage{graphicx}
\usepackage{booktabs}
\usepackage{caption}

\usepackage{threeparttable}
\usepackage{subfigure}
\usepackage{dsfont}
\usepackage{enumerate}
\usepackage{amsmath,amsthm,amssymb}
\usepackage{multirow}

\usepackage{wrapfig}

\usepackage{framed}
\usepackage{color}
\definecolor{shadecolor}{rgb}{0.92,0.92,0.92}

\definecolor{mygray}{gray}{.9}
\newcommand{\gray}{\cellcolor{mygray}}

\usepackage[toc,page,header]{appendix}
\usepackage{minitoc}

\renewcommand{\algorithmicrequire}{\textbf{Input:}}
\renewcommand{\algorithmicensure}{\textbf{Output:}}

\newcommand{\cX}{\mathcal{X}}
\newcommand{\cY}{\mathcal{Y}}

\newcommand{\bx}{{x}}

\usepackage[preprint]{neurips_2024}

\theoremstyle{plain}
\newtheorem{theorem}{Theorem}[section]
\newtheorem{proposition}[theorem]{Proposition}

\theoremstyle{definition}
\newtheorem{definition}[theorem]{Definition}

\theoremstyle{remark}

\title{Decoupling the Class Label and the Target Concept \\ in Machine Unlearning}

\author{\\
\textbf{Jianing Zhu}$^{1,2} \thanks{Work done during an internship at RIKEN Center for Advanced Intelligence Project.}$ \quad
\textbf{Bo Han}$^{1}$ \quad
\textbf{Jiangchao Yao}$^{3,4}$ \\ \textbf{Jianliang Xu}$^{1}$ \quad
\textbf{Gang Niu}$^{2}$ \quad \textbf{Masashi Sugiyama}$^{2,5}$ 
\\
\\
$^{1}$Hong Kong Baptist University \quad
$^{2}$RIKEN Center for Advanced Intelligence Project \\
$^{3}$CMIC, Shanghai Jiao Tong University \quad
$^{4}$Shanghai AI Laboratory\\
$^{5}$The University of Tokyo \\
}

\begin{document}

\maketitle

\begin{abstract}
Machine unlearning as an emerging research topic for data regulations, aims to adjust a trained model to approximate a retrained one that excludes a portion of training data. Previous studies showed that \textit{class-wise} unlearning is successful in forgetting the knowledge of a target class, through gradient ascent on the forgetting data or fine-tuning with the remaining data. However, while these methods are useful, they are insufficient as the class label and the target concept are often considered to coincide. In this work, we expand the scope by considering the label domain mismatch and investigate three problems beyond the conventional \textit{all matched} forgetting, e.g., \textit{target mismatch}, \textit{model mismatch}, and \textit{data mismatch} forgetting. We systematically analyze the new challenges in restrictively forgetting the target concept and also reveal crucial forgetting dynamics in the representation level to realize these tasks. Based on that, we propose a general framework, namely, \textit{TARget-aware Forgetting} (TARF). It enables the additional tasks to actively forget the target concept while maintaining the rest part, by simultaneously conducting annealed gradient ascent on the forgetting data and selected gradient descent on the hard-to-affect remaining data. Empirically, various experiments under the newly introduced settings are conducted to demonstrate the effectiveness of our TARF.
\end{abstract}

\section{Introduction}

In response to data regulations~\cite{hoofnagle2019european}, e.g., ``the right to be forgotten''~\cite{rosen2011right}, 
machine unlearning~\cite{cao2015towards,shaik2023exploring,maini2024tofu} has emerged to eliminate the influence of training data from a trained model~\cite{fan2023salun,xu2023machine}. 
The intuitive goal is to forget the specific data as if the model had never used it during training~\cite{bourtoule2021machine}. To achieve that, a direct way~\cite{shaik2023exploring} is to retrain the model from scratch by excluding the data to be unlearned, which is called \textit{exact unlearning}. Considering its computational cost, prior works proposed \textit{approximate unlearning}~\cite{golatkar2020eternal,warnecke2021machine} to efficiently adjust the trained model for forgetting and approximate the behaviors of the retrained one. Recent studies~\cite{kurmanji2023towards,jia2023model,chen2023boundary,fan2023salun} showed that \textit{class-wise} unlearning is successful in forgetting the knowledge of a training class, through reverse optimization~\cite{thudi2022unrolling,izzo2021approximate,chen2023boundary} on the class data or fine-tuning on the remaining data~\cite{golatkar2020eternal,kurmanji2023towards} to realize catastrophic forgetting~\cite{french1999catastrophic,kirkpatrick2017overcoming}. 

\begin{figure*}[t!]
\begin{center}
\hspace{-0.17in}
\includegraphics[scale=0.12]{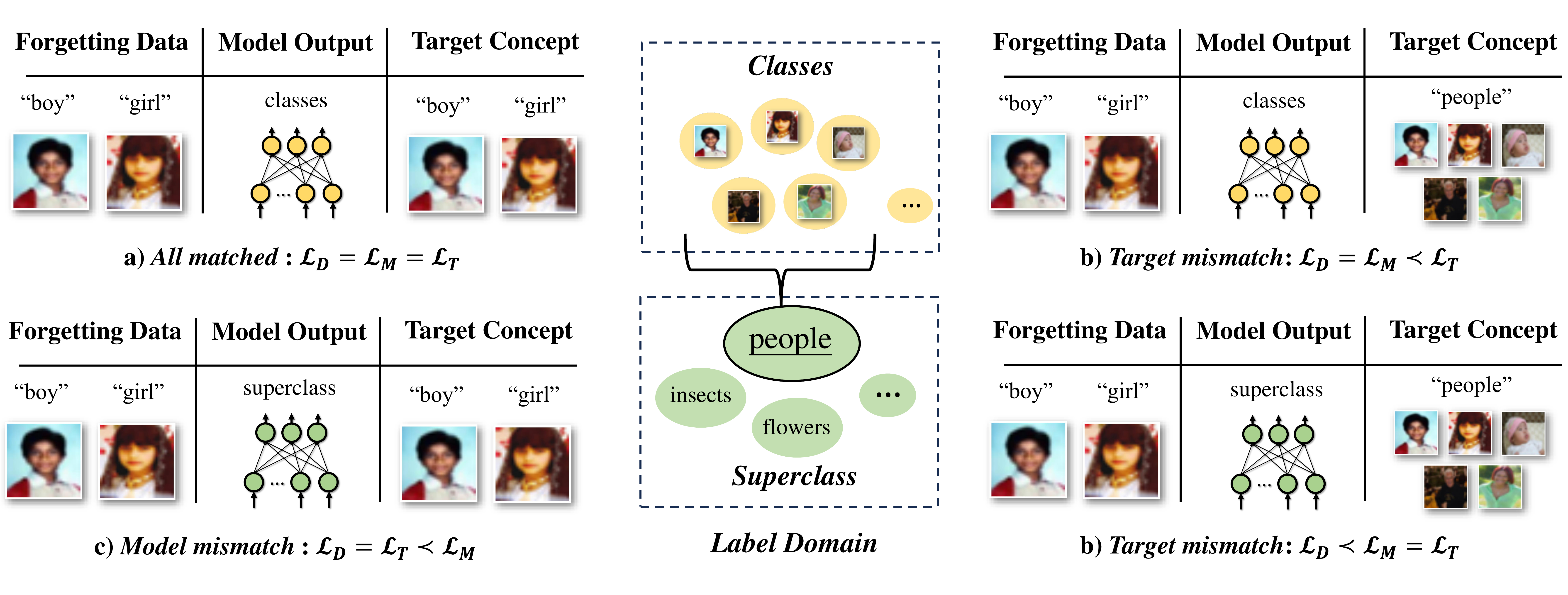}
\end{center}
\vspace{-4mm}
\scriptsize{Taking the \textit{CIFAR-100}~\cite{krizhevsky2009learning_cifar10} dataset, we instantiate four unlearning tasks given the same forgetting data with the class labels of ``boy'' and ``girl'': a) \textit{all matched forgetting} (conventional scenario): unlearn ``boy'' and ``girl'' with the model trained on the classes; b) \textit{target mismatch forgetting}: unlearn ``people'' with the model trained on the classes; c) \textit{model mismatch forgetting}: unlearn ``boy'' and ``girl'' with the model trained on the superclass; d) \textit{data mismatch forgetting}: unlearn ``people'' with the model trained on the superclass. More discussion is provided in Appendix~\ref{app:whole_mismatch}.}
\vspace{-1mm}
\caption{Illustrations of decoupling the class label and the target concept.}
\label{fig: setting_example}
\vspace{-6mm}
\end{figure*}

Despite the promising achievements, the previously studied scenario~\cite{ warnecke2021machine,golatkar2020eternal,chen2023boundary,jia2023model,fan2023salun} mainly assumed the target concept to coincide with the class label, overlooking that the practical unlearning request~\cite{cao2015towards,bommasani2021opportunities,hashimoto2018fairness, kovashka2016crowdsourcing} may violate the taxonomy of the pre-training tasks. In this work, we decouple them by considering the different label domains of the forgetting data $\mathcal{L}_D$, the model output $\mathcal{L}_M$, and the target concept $\mathcal{L}_T$ in unlearning. Given the relation defined for two label domains, e.g., being match $\mathcal{L}_1=\mathcal{L}_2$, superclass $\mathcal{L}_1\succ\mathcal{L}_2$, and subclass $\mathcal{L}_1\prec\mathcal{L}_2$, we introduce four practical scenarios. Assuming that the reported forgetting data should be included in the target concept, e.g., $\mathcal{L}_D\preceq\mathcal{L}_T$, we have \textit{all matched} $\mathcal{L}_D = \mathcal{L}_T = \mathcal{L}_M$; \textit{target mismatch} $\mathcal{L}_D = \mathcal{L}_M \prec \mathcal{L}_T$; \textit{model mismatch} $\mathcal{L}_D = \mathcal{L}_T \prec \mathcal{L}_M$; and \textit{data mismatch} $\mathcal{L}_D \prec \mathcal{L}_T = \mathcal{L}_M$. We further illustrated four instantiated tasks in Figure~\ref{fig: setting_example},  using the CIFAR-100~\cite{krizhevsky2009learning_cifar10} dataset with its classes and superclass.

\begin{figure*}[t!]
\begin{center}
\includegraphics[scale=0.09]{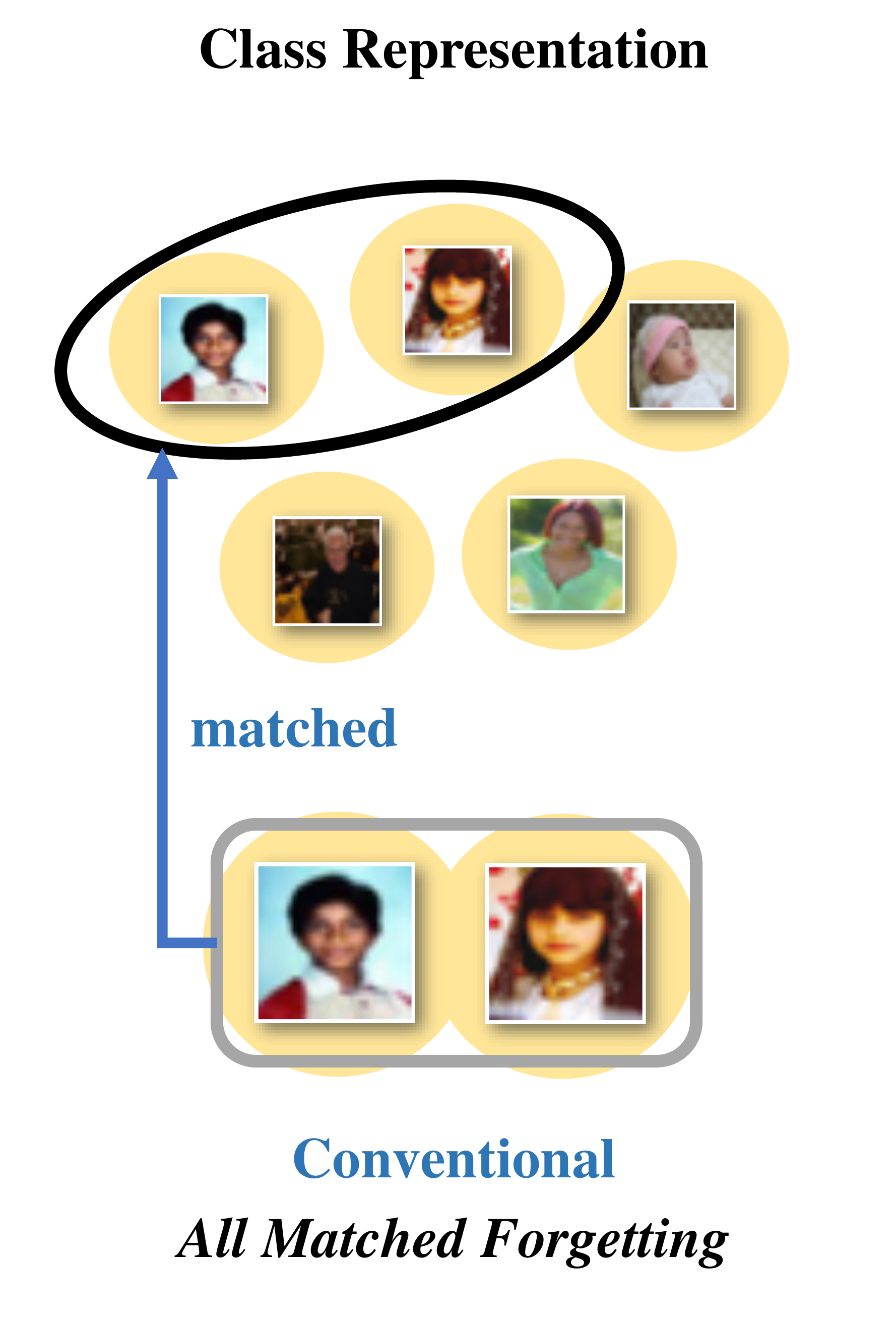}
\begin{minipage}[b]{0.29\linewidth}
\includegraphics[scale=0.16]{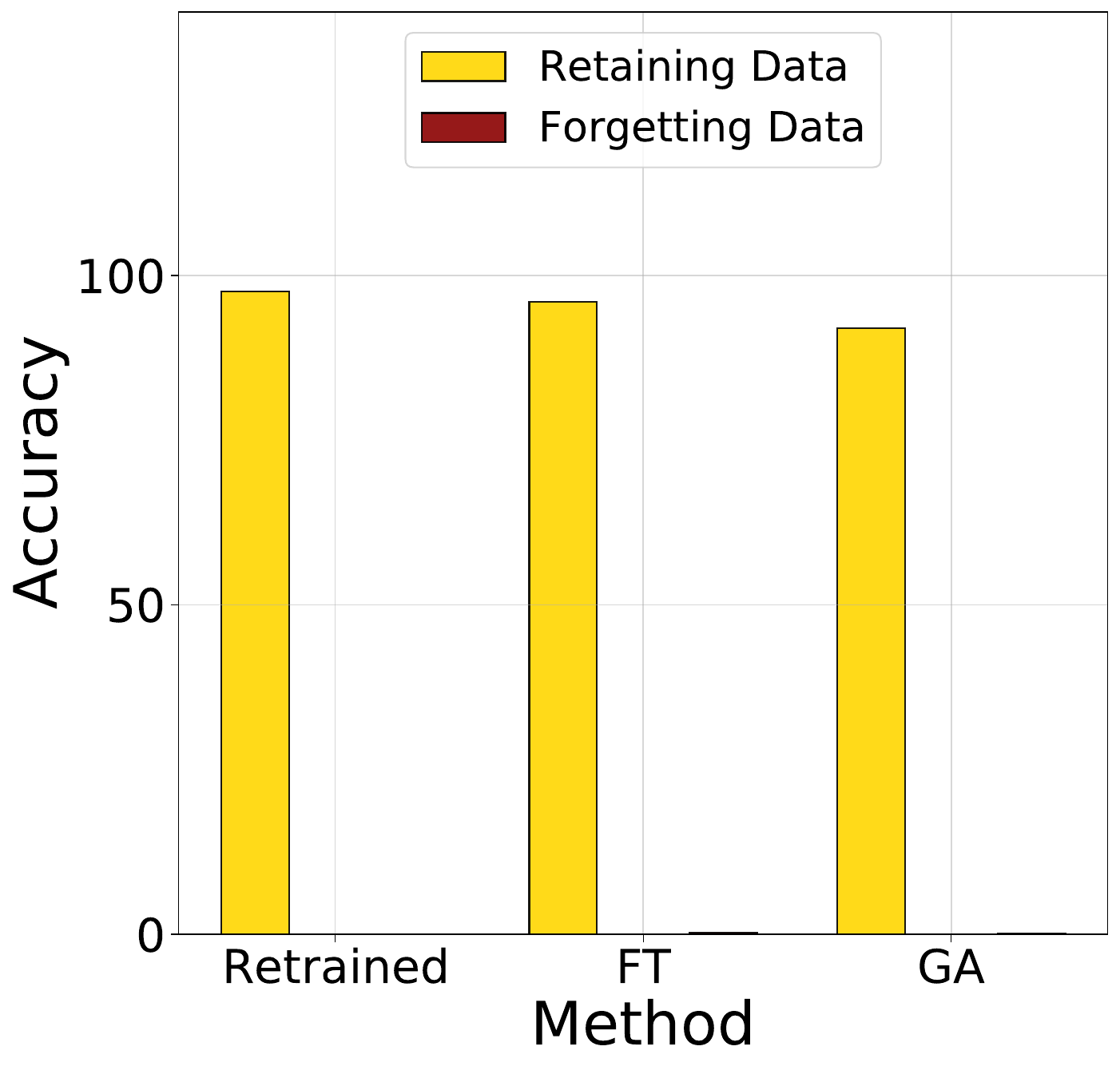}
\end{minipage}
\includegraphics[scale=0.09]{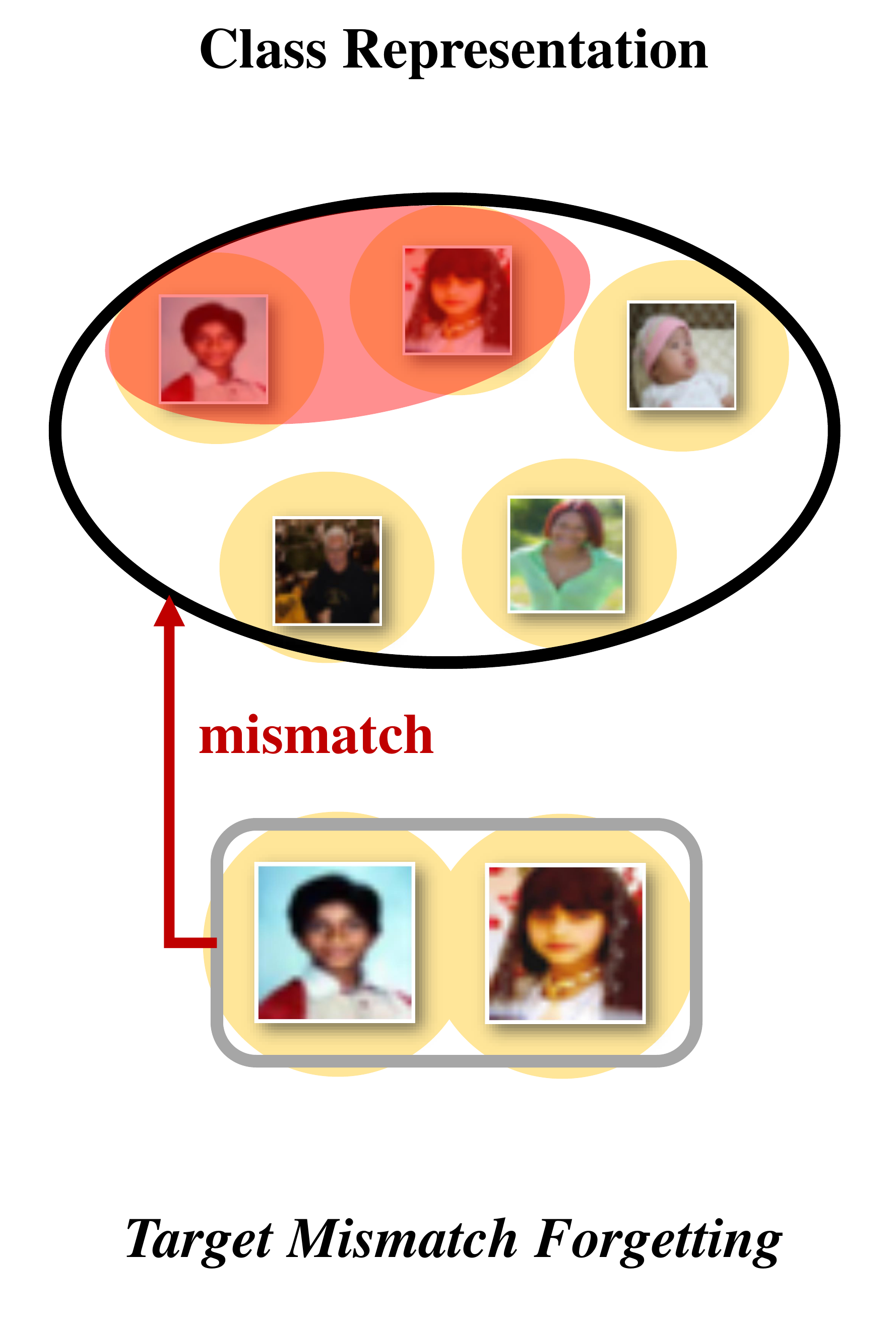}
\begin{minipage}[b]{0.29\linewidth}
\includegraphics[scale=0.16]{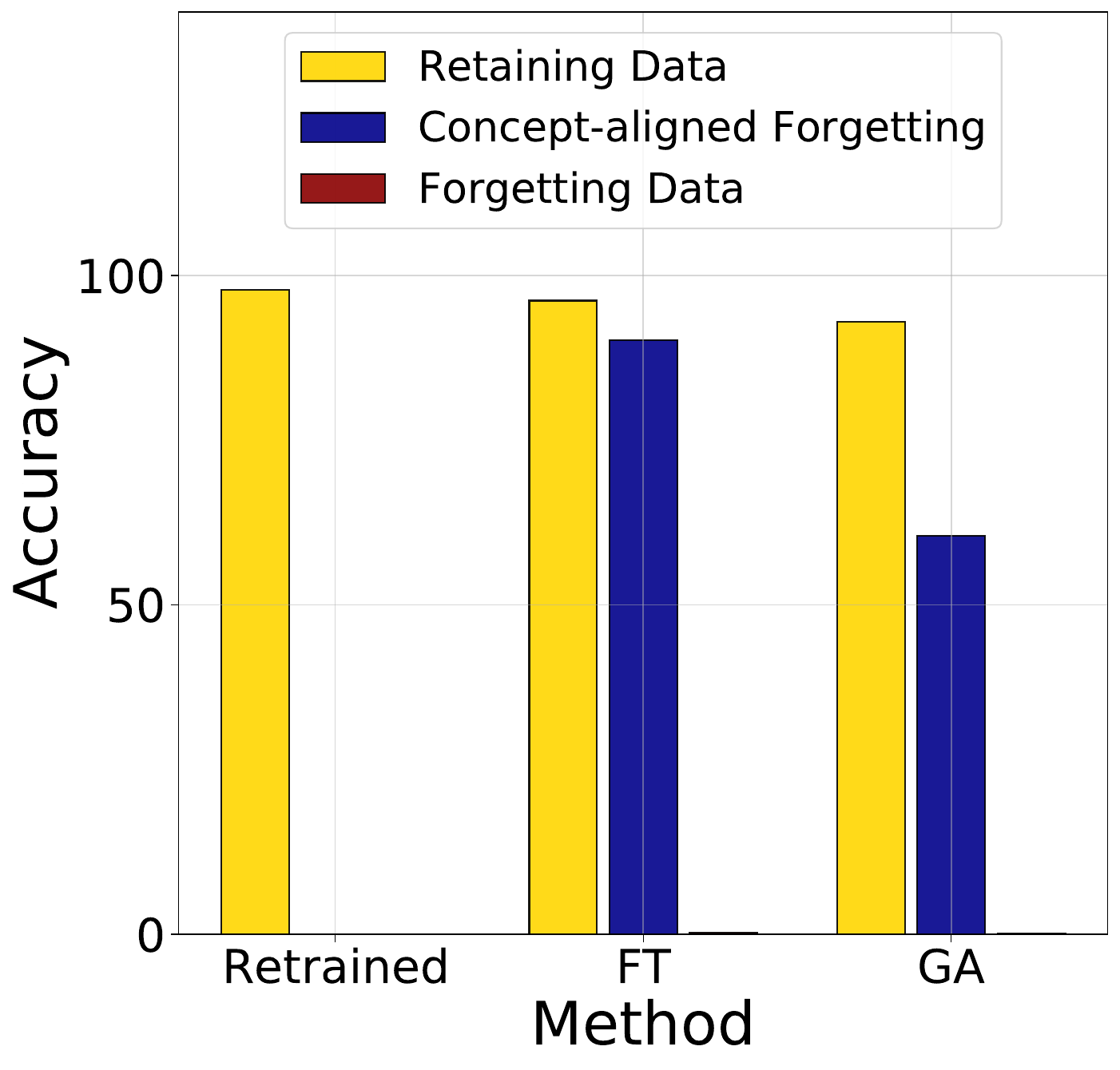}
\end{minipage}\\

\includegraphics[scale=0.09]{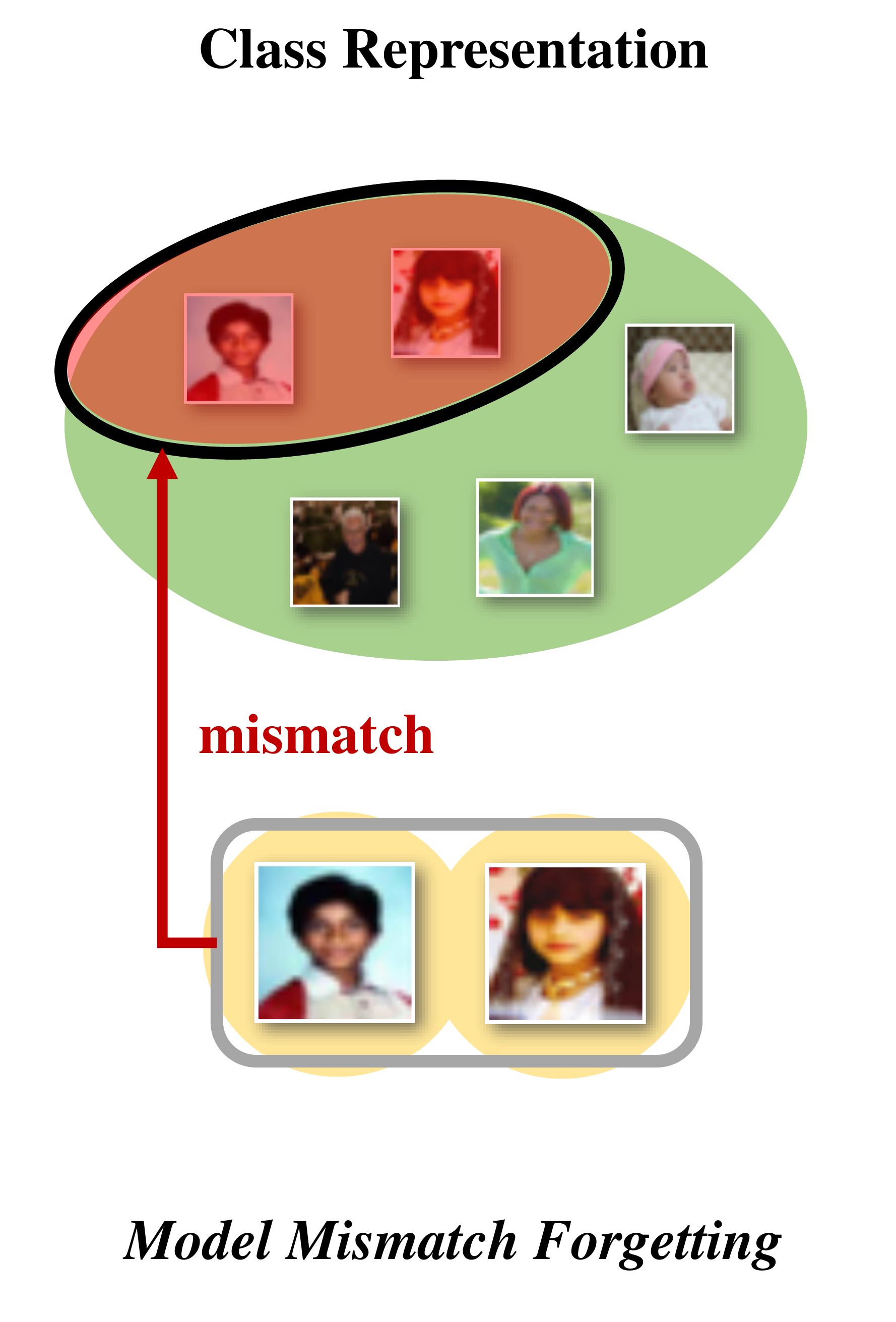}
\begin{minipage}[b]{0.29\linewidth}
\includegraphics[scale=0.16]{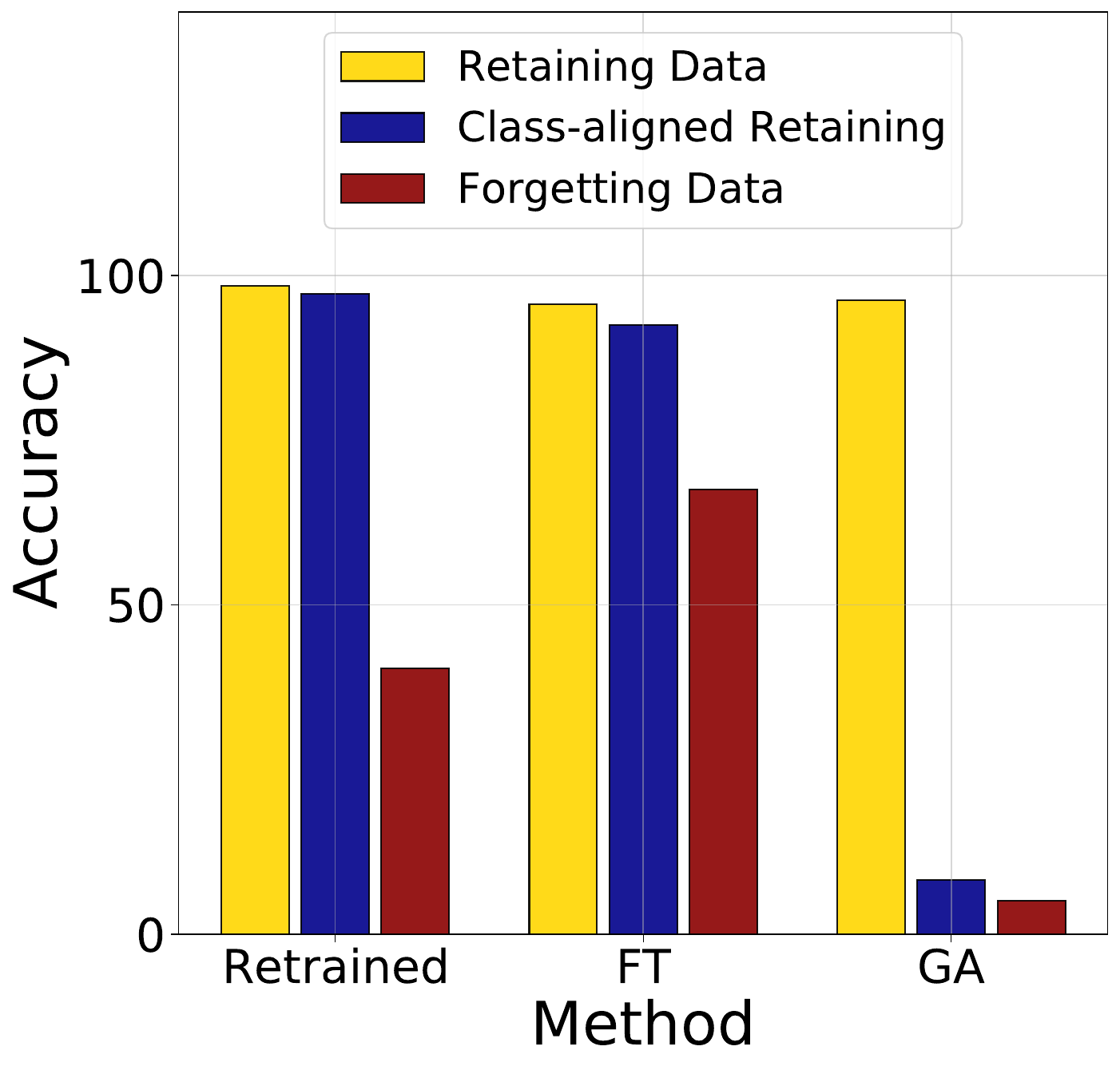}
\end{minipage}
\includegraphics[scale=0.09]{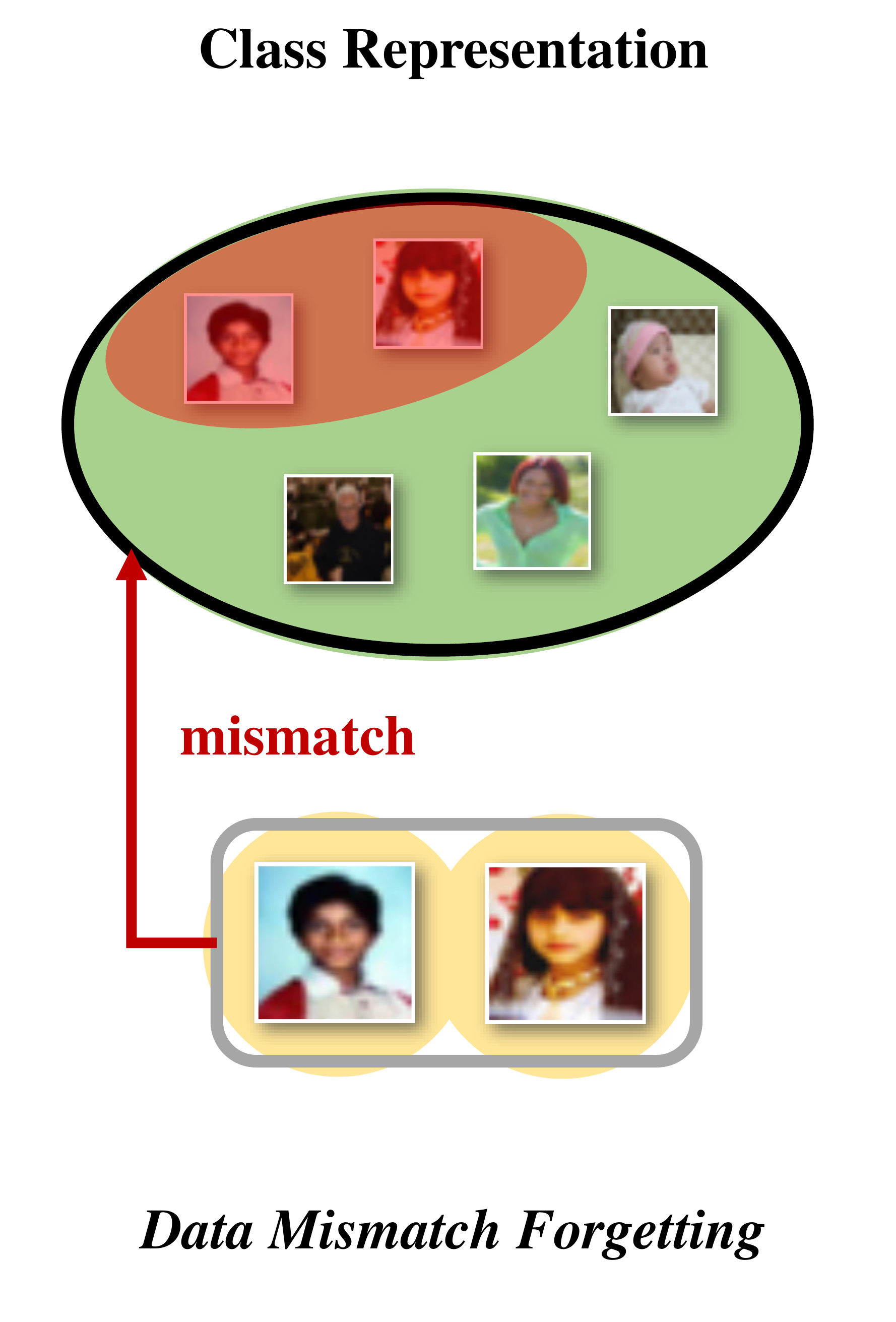}
\begin{minipage}[b]{0.29\linewidth}
\includegraphics[scale=0.16]{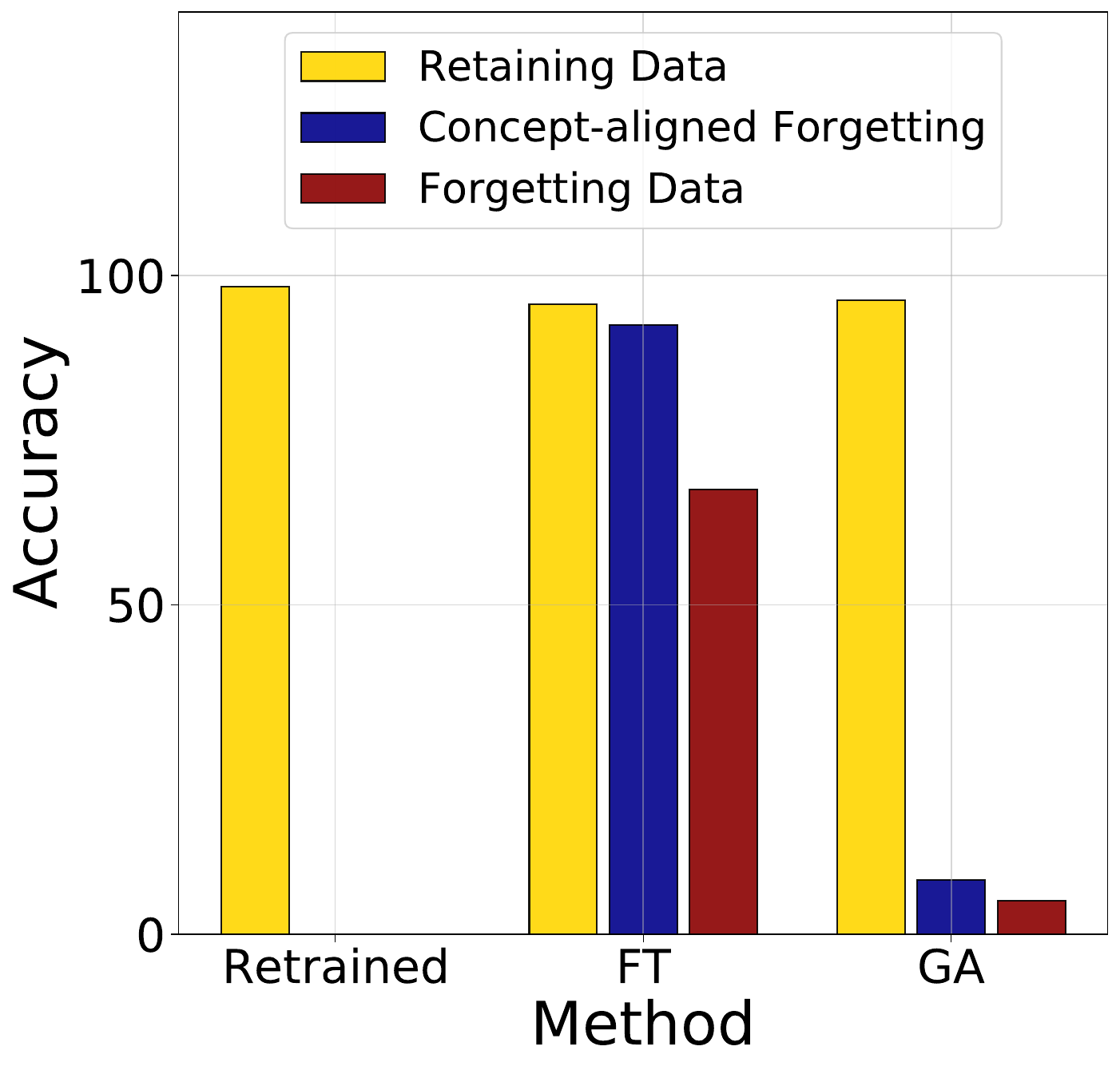}
\end{minipage}
\end{center}\vspace{-1mm}
\scriptsize{We present unlearning (e.g., Retrained, finetuning~\cite{golatkar2020eternal} (FT), and gradient ascent~\cite{thudi2022unrolling} (GA)) for the four tasks. In the conventional all matched forgetting, FT and GA can achieve similar performance on retaining and forgetting data like Retrained. In contrast, we can find that 
model mismatch forgetting can be affected by the trained model, coupling the behaviors on class-aligned retaining and forgetting data. In target or data mismatch forgetting, the class labels can not fully represent the target concepts, making the concept-aligned forgetting data hard to unlearn. In addition, we also provide the tSNE visualization of the unlearned results (e.g., Retrained, FT, and GA) in different tasks Appendix~\ref{app:exp_identification}.}
\caption{The challenges of restrictive unlearning with the mismatched label domains.
}
\vspace{-3mm}
\label{fig: challenge}
\end{figure*}

\begin{figure*}[t!]
    \begin{center}
    \subfigure[Distances in under-entangled representation]{
    \includegraphics[scale=0.14]{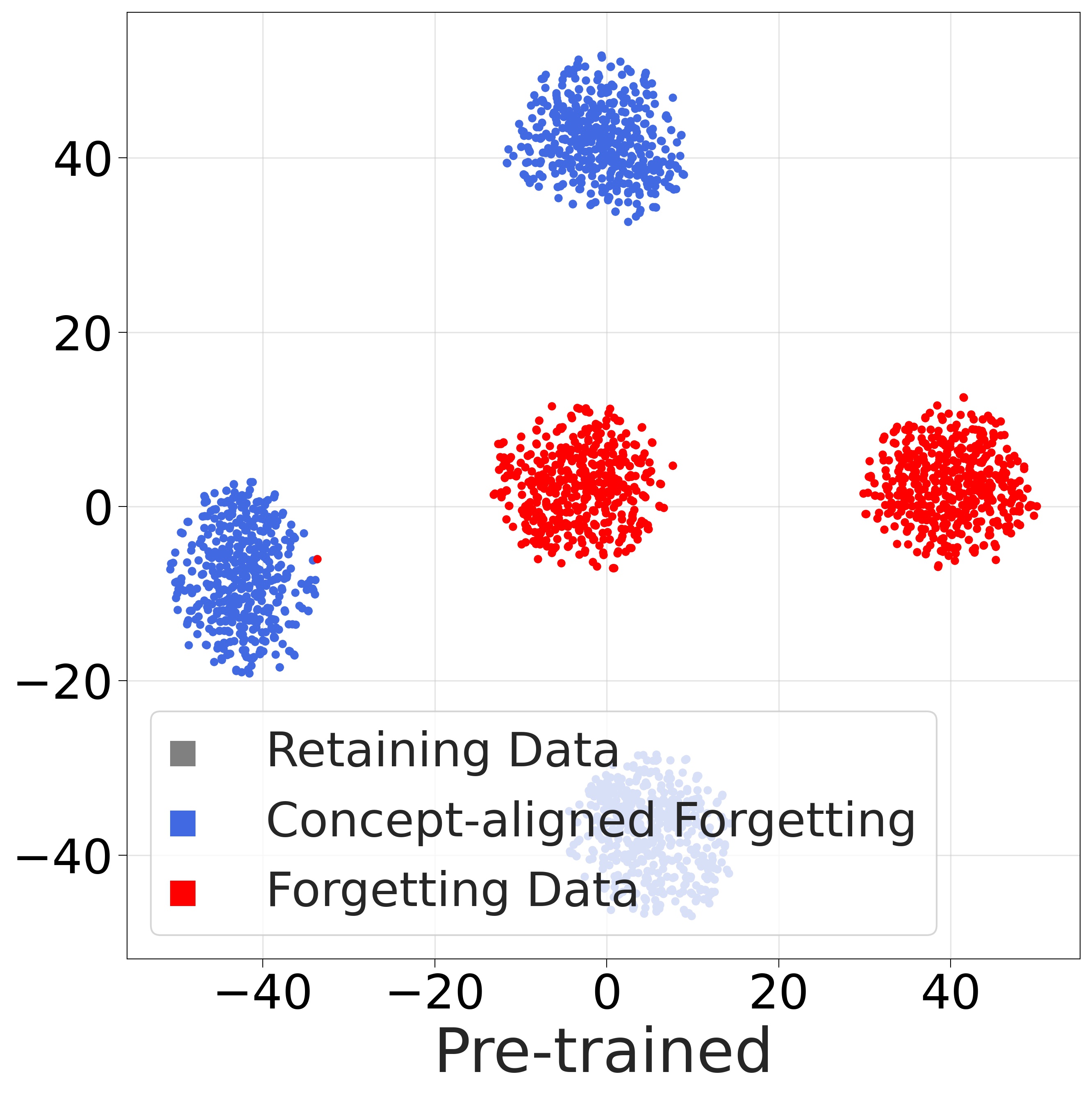}
    \includegraphics[scale=0.14]{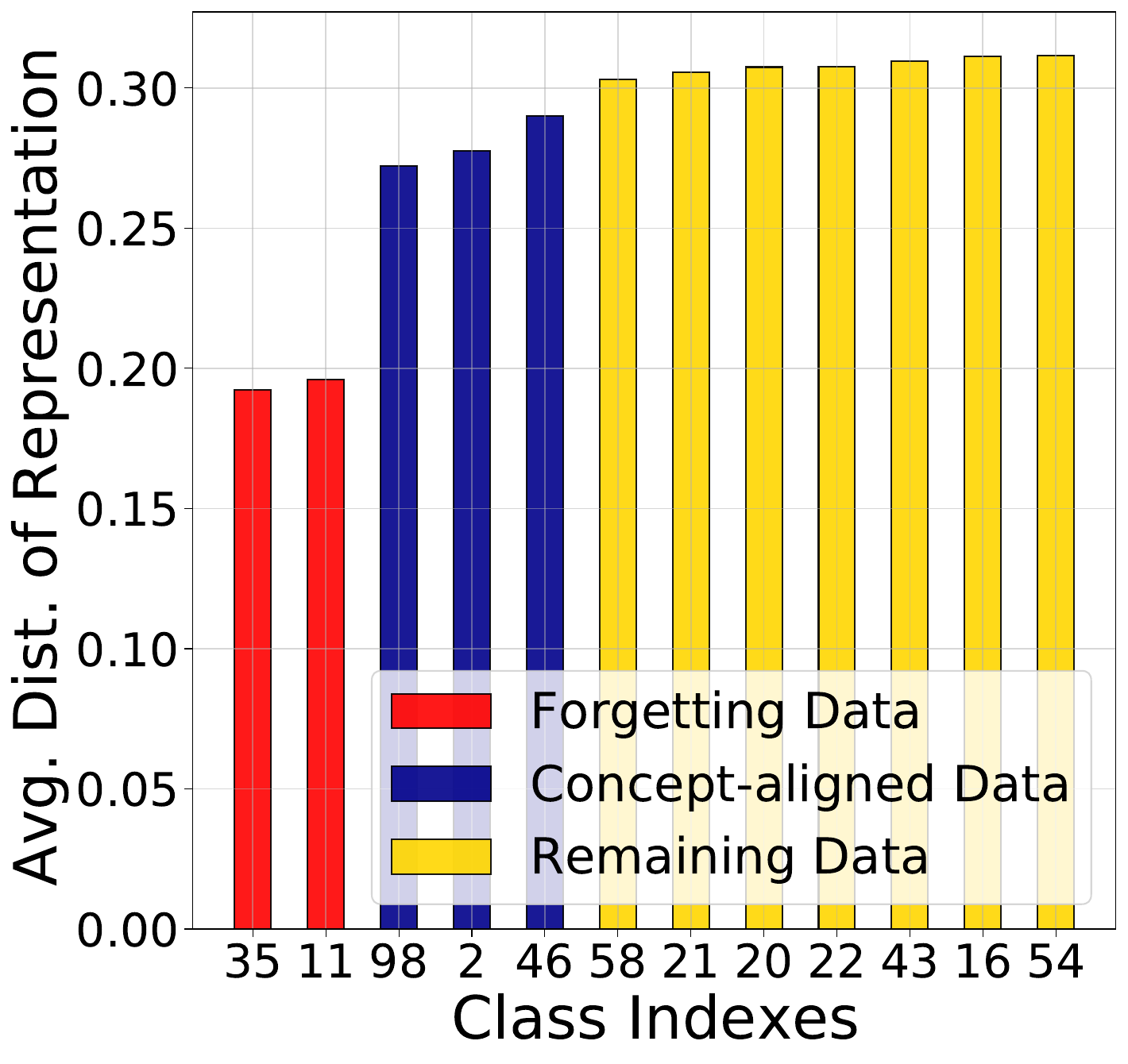}
    \label{fig:distance_a}
    }
    \subfigure[Distances in entangled representation]{
    \includegraphics[scale=0.14]{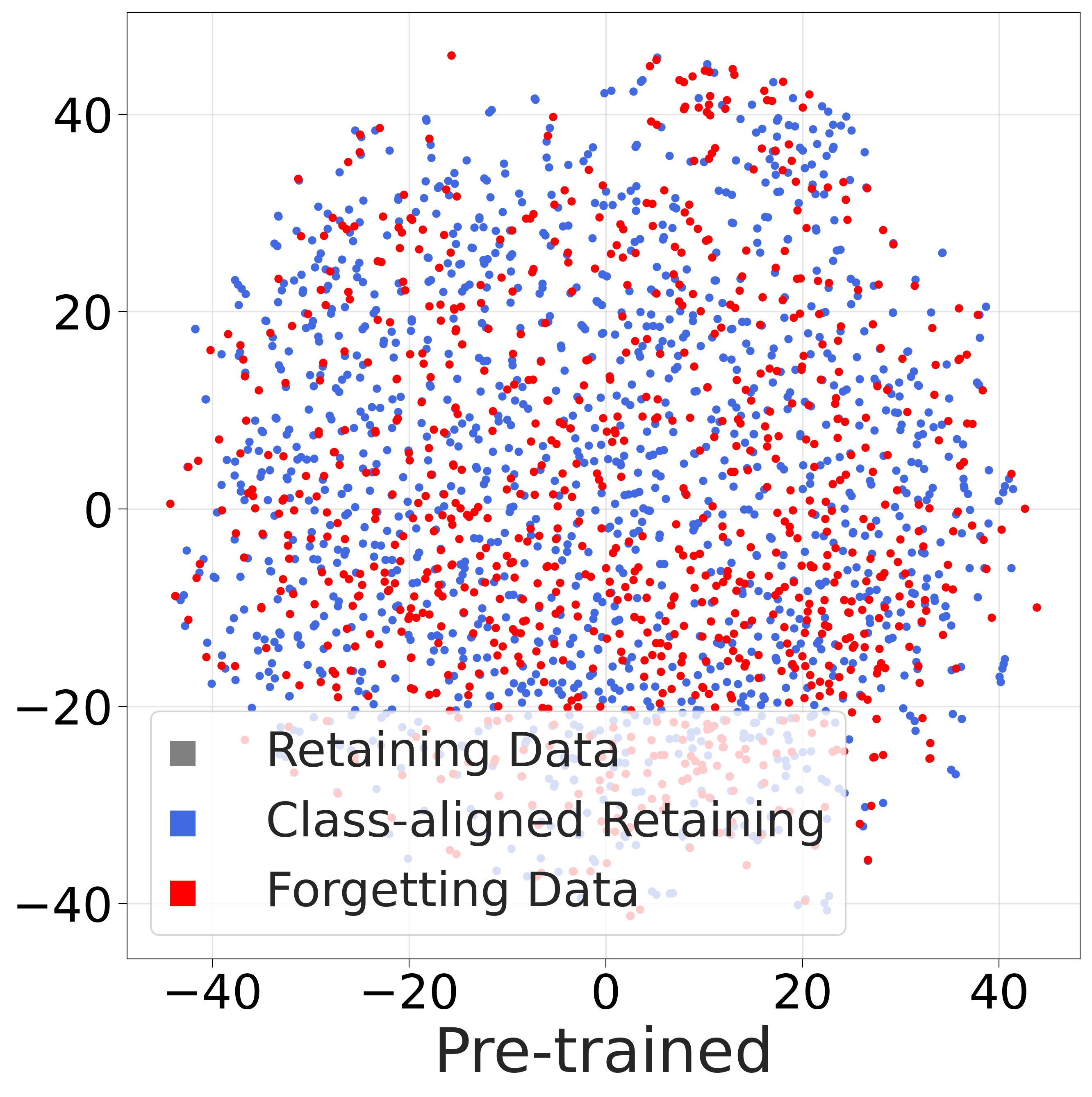}
    \includegraphics[scale=0.14]{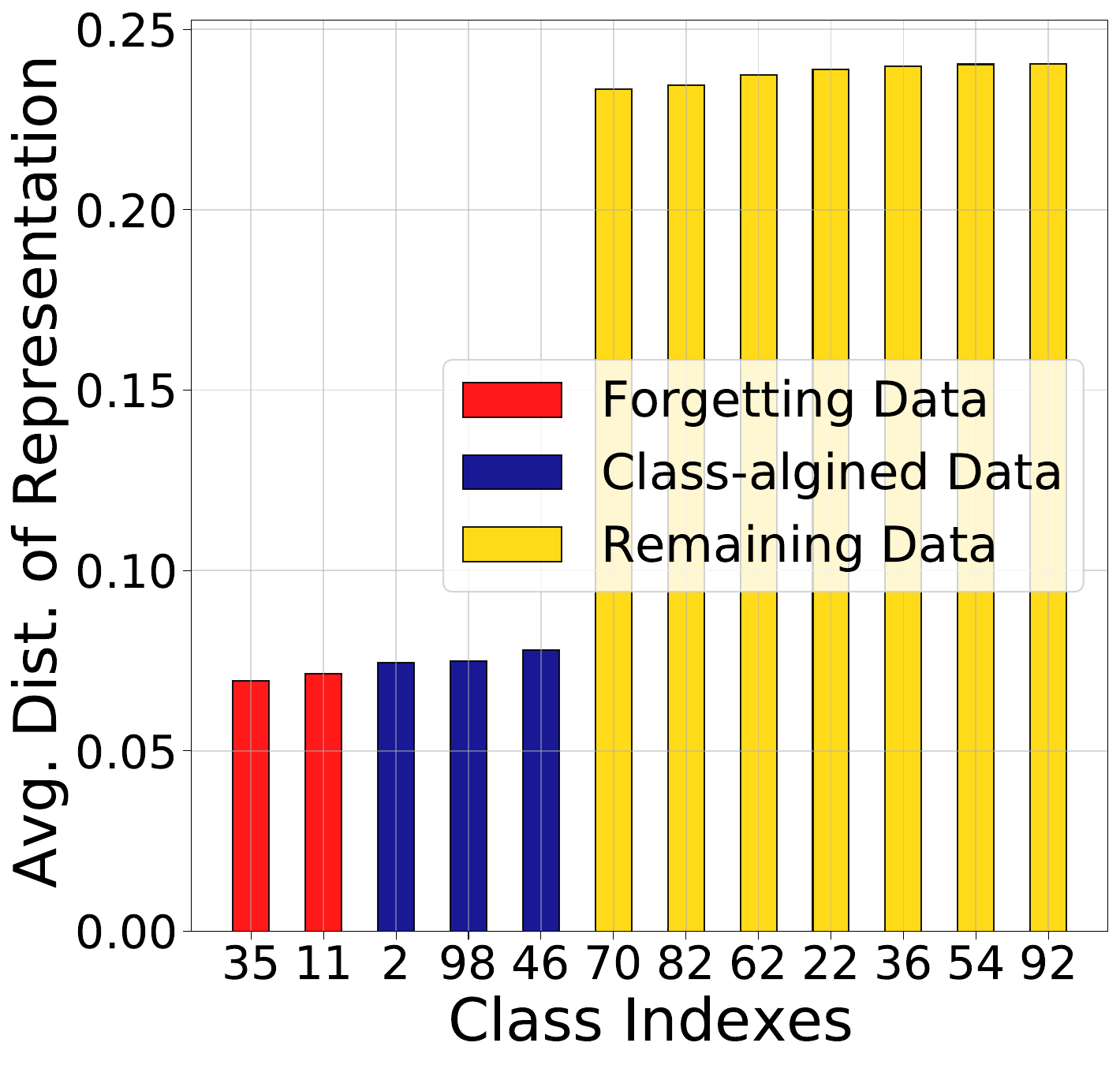}
    \label{fig:distance_b}
    }
    \\
    \subfigure[Forgetting in the model trained by classes]{
    \includegraphics[scale=0.14]{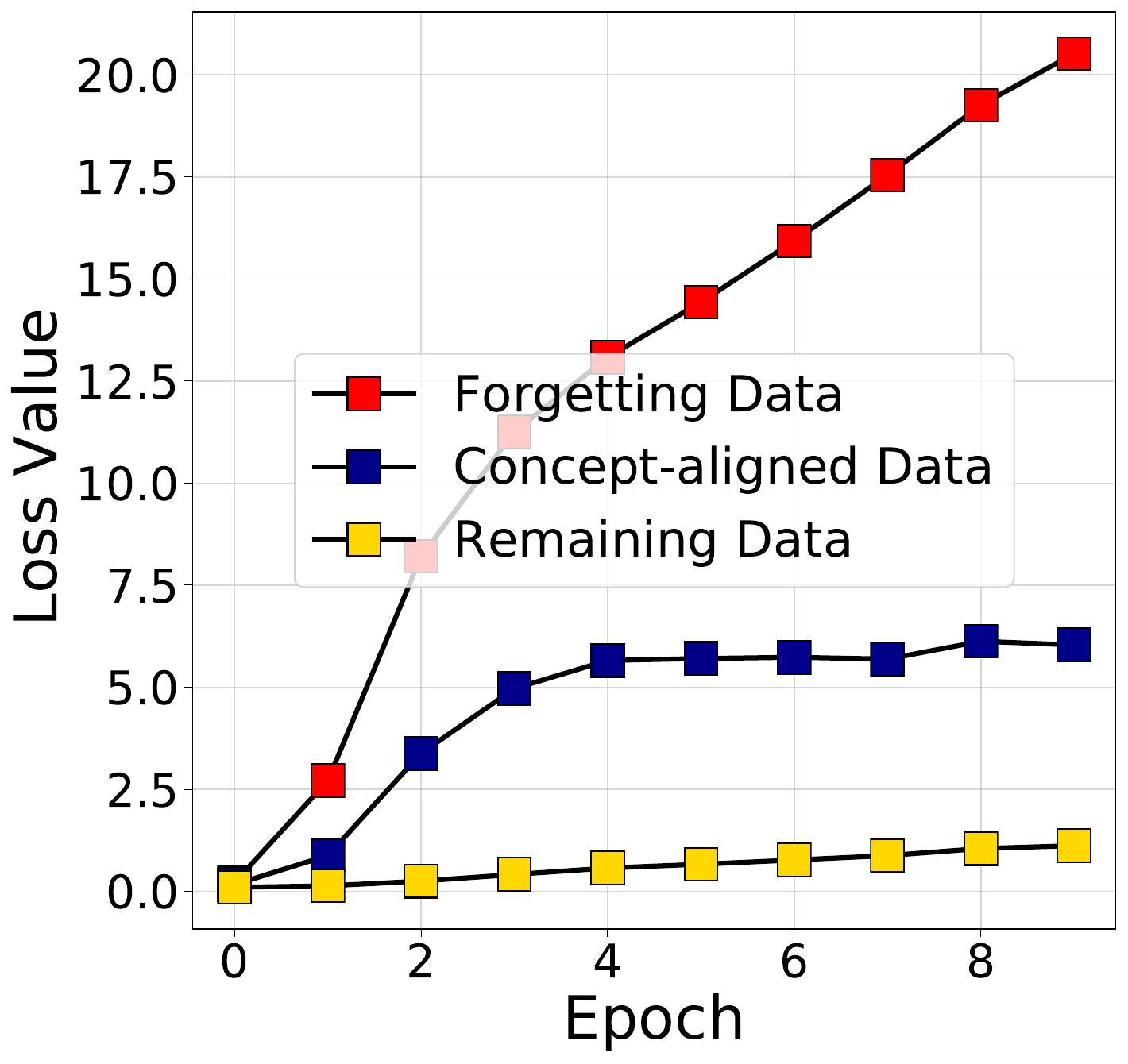}
    \includegraphics[scale=0.14]{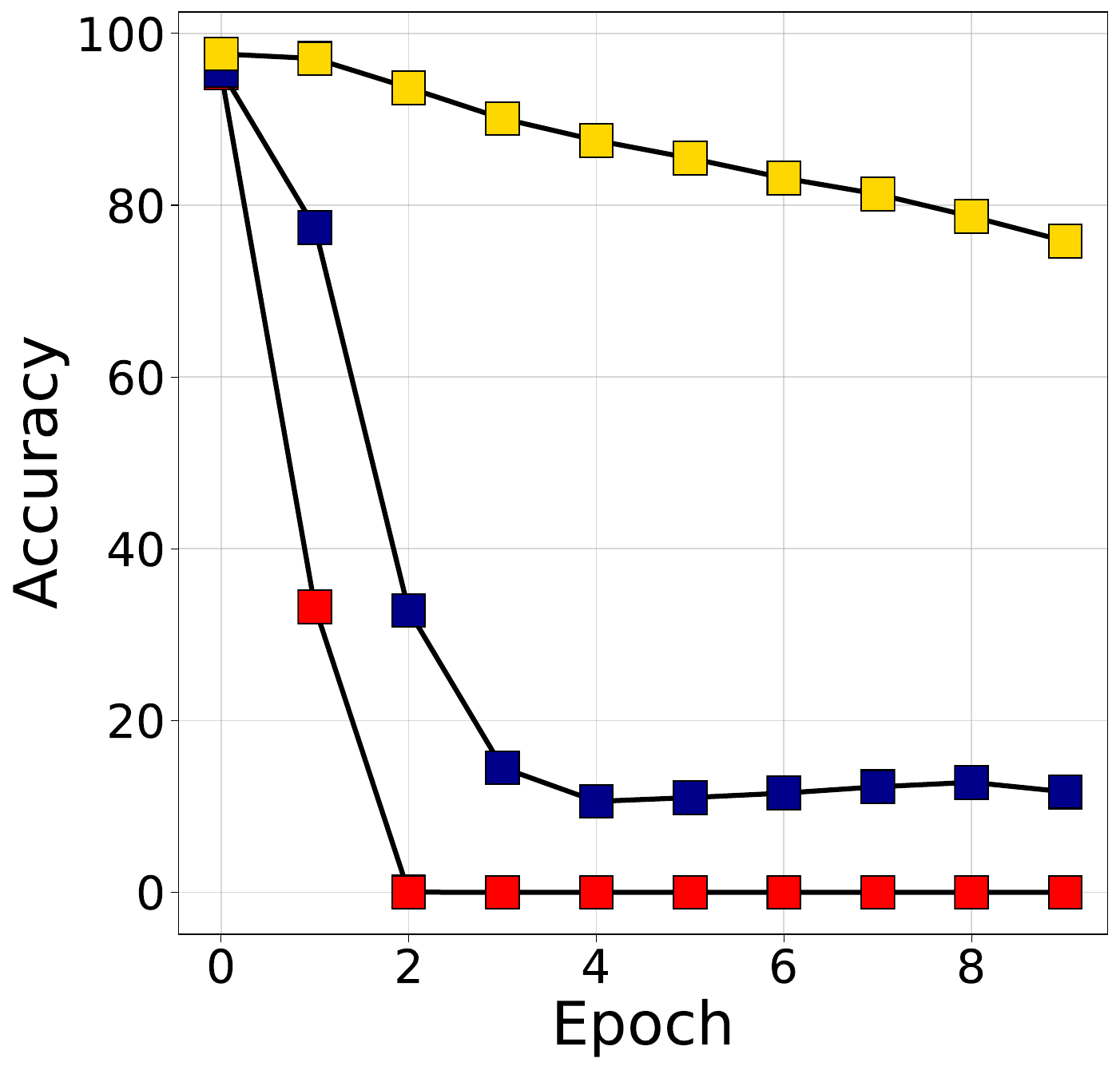}
    \label{fig:feature_a}
    }
    \subfigure[Forgetting in the model trained by superclass]{
    \includegraphics[scale=0.14]{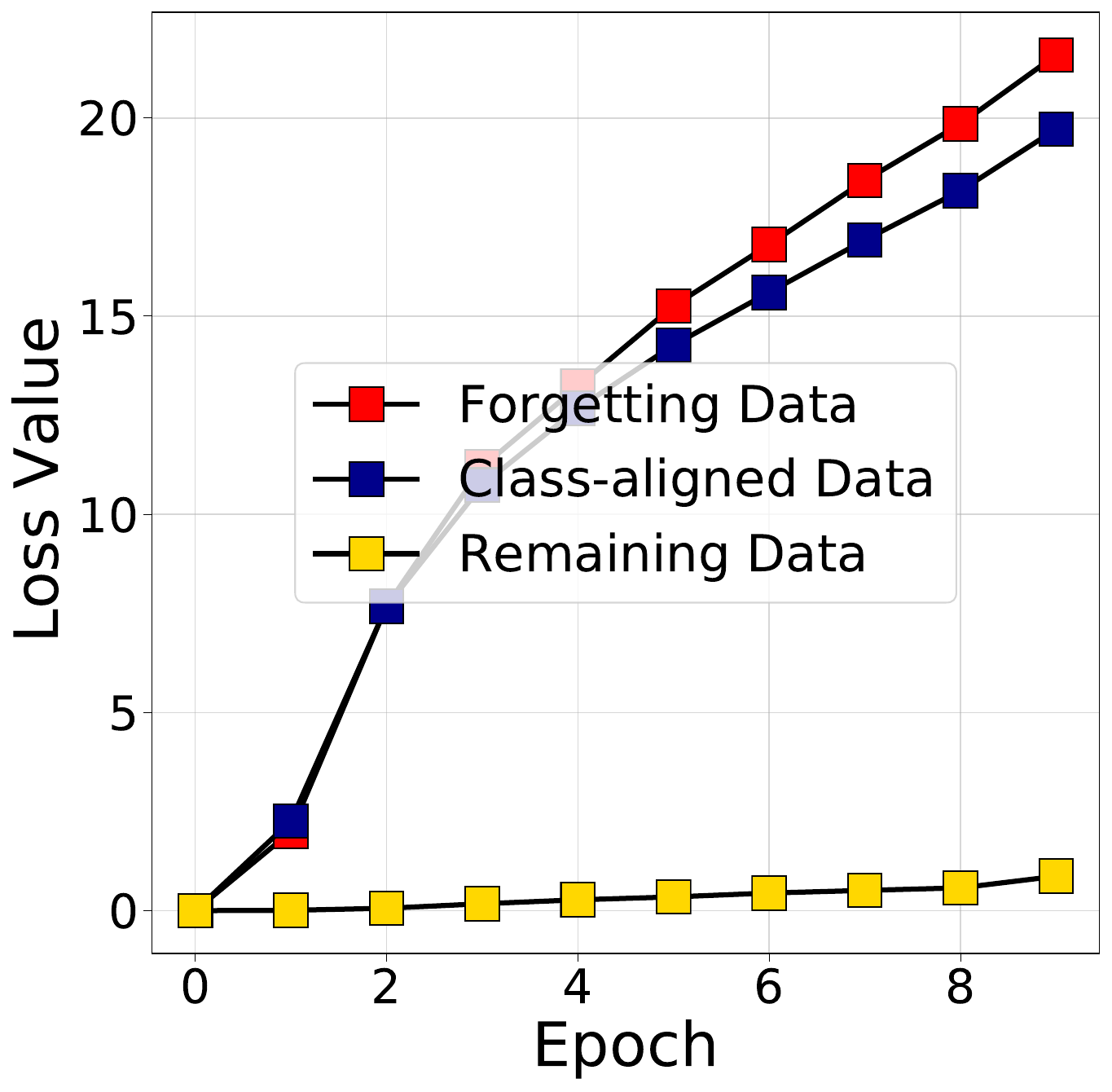}
    \includegraphics[scale=0.14]{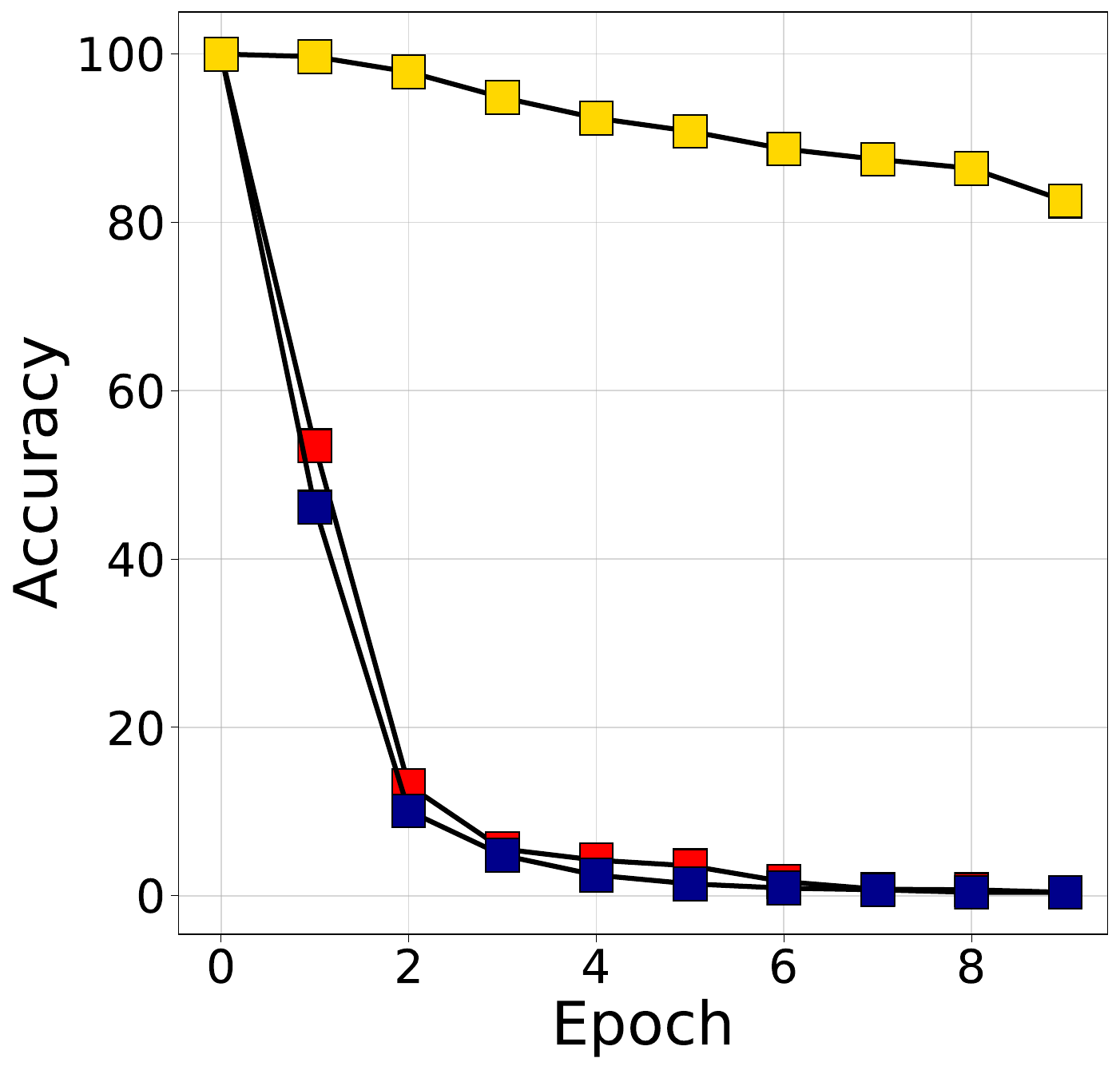}
    \label{fig:feature_b}
    }
    \\
    \end{center}
    \vspace{-2mm}
    \scriptsize{
    In the top row, we present the tSNE visualization~\cite{maaten2008visualizing_tsne} of the learned features from the pre-trained model trained by (a) class and (b) superclass, and their averaged Euclidean distance to cluster centers of the forgetting data. In the bottom row, we report the loss value and the accuracy of forgetting data, concept/class-aligned data, and the retaining data during GA in the model trained by (c) class and (d) superclass respectively.}
    \caption{Forgetting dynamics on entangled/under-entangled feature representations.
    }
    \label{fig:feature}
    \vspace{-2mm}
\end{figure*}

Investigating the aforementioned tasks, we identify new challenges in restrictive unlearning with the mismatched label domains (refer to Figure~\ref{fig: challenge}).
Unlike the accurate approximation in the conventional all matched task~\cite{warnecke2021machine,jia2023model,chen2023boundary}, the representative unlearning methods~\cite{warnecke2021machine,thudi2022unrolling} exhibit different performance gap with the retrained reference in the other tasks. Specifically, the under-entangled feature representation (when $\mathcal{L}_M\prec\mathcal{L}_T$) or the under-representative forgetting data (when $\mathcal{L}_D\prec\mathcal{L}_T$) results in insufficient forgetting, while the entangled feature representation (when $\mathcal{L}_T\preceq\mathcal{L}_M$) prevents the decomposition of target concept with the retaining part. The former requires target identification in the remaining training set, while the latter requires target separation over the entangled feature representation. Through exploration of forgetting dynamics (refer to Figure~\ref{fig:feature}), we demonstrate the feature distance reflected by representation gravity is a crucial factor for the feasibility of these tasks. 

Based on the above analysis, we propose a novel framework, namely, \textit{TARget-aware Forgetting} (TARF). In general, we consider two parts (refer to Eq.~\eqref{eq:4}), i.e., annealed forgetting and target-aware retaining, which collaboratively enable the target identification and separation (refer to Figure~\ref{fig:target_iden_separa}) for these forgetting tasks. 
Specifically, the algorithmic framework (refer to Figure~\ref{fig:framework}) incorporates an annealed gradient ascent and target-aware gradient descent in a dynamical manner, which can be viewed as three phases. The first actively unlearns the identified forgetting data, and constructs the contrast information to filter out the remaining data which is hard to be affected. Then, simultaneously learning the selected retaining data with gradient descent deconstructs the entangled feature representation. Ultimately, the learning objective can progressively approach standard retraining using the aligned retaining data. 
The main contributions of our work can be summarized as follows,
\begin{itemize}
    \item Conceptually, we introduce new settings that decouple the class label and the target concept, which investigate the label domain mismatch in class-wise unlearning (in Section~\ref{sec:method_motivation}).
    \item Empirically, we systematically reveal the challenges of restrictive unlearning with the mismatched label domains, and demonstrate that the representation gravity in forgetting dynamics is critical for achieving the forgetting target in the new tasks (in Section~\ref{sec:method_analyze}).
    \item Technically, we propose a general framework, namely, \textit{TARF}, to realize the target identification and separation in unlearning. It consists of annealed forgetting and target-aware retaining which collaboratively approximate retraining on the retaining data (in Section~\ref{sec:method_gradient}).
    \item Experimentally, we conduct extensive explorations to validate the effectiveness of our framework and perform various ablations to characterize algorithm properties (in Section~\ref{sec:exp}).
\end{itemize}

\section{Preliminaries}
\label{sec:pre}

In this section, we briefly introduce the problem settings of class-wise machine unlearning, and compare the differences between ours and the conventional setting considered in the previous research works. More details about the unlearning baselines considered in our work can refer to Appendix~\ref{app:baseline_info}.

\textbf{Unlearning basis and objective.} Following the literature~\cite{shaik2023exploring,xu2023machine}, we mainly consider the multi-class classification~\cite{goodfellow2016deep} as the original training task for class-wise unlearning. Let $\mathcal{X}\subset\mathbb{R}^d$ denote the input space and $\mathcal{Y}=\{1,\ldots, C\}$ denote the label space, where $C$ is the number of classes, the training dataset $\mathcal{D}=\{(x_i,y_i)\}^N_{i=1}$ consists of two subsets in machine unlearning, e.g., the forgetting dataset $\mathcal{D}_{\text{f}}$ and the retaining dataset $\mathcal{D}_{\text{r}} = \mathcal{D}\backslash\mathcal{D}_\text{f}$. Building upon the model $f_{\theta^*}: \mathcal{X}\rightarrow \mathcal{Y}$ trained on $\mathcal{D}$ with the loss function $\ell$, the general goal of this problem is to find an unlearned model $\theta^*_\text{un}$ which approximates the behaviors of the model $\theta^\mathrm{r}$ that retrained on $\mathcal{D}_\text{r}$,
\begin{align}
\begin{split}
\theta_\text{un}^* = \mathop{\arg\min}\limits_{\theta} \frac{1}{|\mathcal{D}|}\sum_{(x,y)\sim\mathcal{D}}\mathcal{R}(\theta,\theta^\mathrm{r},x,y) \quad \text{s.t.}~~\theta^\mathrm{r} = \mathop{\arg\min}\limits_{\theta}\underbrace{\frac{1}{|\mathcal{D}_\text{r}|}\sum_{(x,y)\sim\mathcal{D}_\text{r}}\ell(f_\theta(x),y)}_{L_\text{retrain}},
\end{split}
\end{align}
where $\mathcal{R}$ indicates a general risk measure for model behavior consistency~\cite{golatkar2020eternal,shaik2023exploring}, which can be instantiated by comprehensive evaluation metrics~\cite{jia2023model,fan2023salun} (e.g., unlearning accuracy (UA), retaining accuracy (RA), and others related to privacy) in experiments to pursue the unlearning efficacy and the model utility. The specific definition of evaluation metrics can be referred to in Section~\ref{sec:exp_setup}.

\textbf{Forgetting target.} In previous works~\cite{warnecke2021machine,chen2023boundary}, the target concept to be forgotten is mainly considered as all matched where $\mathcal{D}_\text{t}=\mathcal{D}\{y=y_\text{f}\}$ has the same label domains (exactly same labels) with the pre-training task and forgetting data $\mathcal{D}_\text{f}=\mathcal{D}\{y=y_\text{f}\}$. In contrast, we assume that the target concept can be decoupled from the class label in practical unlearning requests. As illustrated in Figure~\ref{fig: setting_example}, we further instantiate with three forgetting tasks given $\mathcal{D}_\text{f}=\mathcal{D}\{y=y_\text{f}\}$ with the superclass labels $\mathcal{Y}'$ of $\mathcal{Y}$ (classes): i) model mismatch forgetting, e.g., $\mathcal{D}_\text{t}=\mathcal{D}\{y=y_\text{t}\}$ and $y_\text{t} \subseteq y'_\text{f}$ where $y'_\text{f}\in\mathcal{Y}'$ given the model trained on $\mathcal{Y}'$; ii) target mismatch forgetting, e.g., $ \mathcal{D}_{\text{t}}=\mathcal{D}\{y=y'_f\}$ given the model trained on  $\mathcal{Y}$; iii) data mismatch forgetting, e.g., $\mathcal{D}_{\text{t}}=\mathcal{D}\{y=y'_\text{f}\}$ given the model trained on $\mathcal{Y}'$. 

\textbf{Dataset partition.} Given the decoupled target concept, we can find that the previous assumptions of $\mathcal{D}_\text{f} = \mathcal{D}_\text{t}$ and $\mathcal{D}_\text{r} = \mathcal{D}\backslash\mathcal{D}_\text{f}$ do not strictly hold. In model mismatch forgetting, the former is still held; in target mismatch forgetting and data mismatch forgetting, $\mathcal{D}_\text{f} \subseteq \mathcal{D}_\text{t}$ and the remaining data $\mathcal{D}_\text{un} = \mathcal{D}\backslash\mathcal{D}_\text{f}$ include both retaining dataset $\mathcal{D}_\text{r}\subseteq\mathcal{D}_\text{un}$ and the rest forgetting data $\mathcal{D}_\text{uf} = \mathcal{D}_\text{t}\backslash\mathcal{D}_\text{f}$. Specifically, we assume that the number of classes in $\mathcal{D}_\text{un}$ belonging to the target concept $\mathcal{D}_\text{t}$ is known in target mismatch forgetting, and the retrained model for every task is trained using $\mathcal{D}_\text{r} = \mathcal{D}\backslash\mathcal{D}_\text{t}$. More details about dataset partition and unlearning request construction are provided in Appendix~\ref{app:dataset_partition}.

\section{TARF: \textit{TARget-aware Forgetting}}
\label{sec:method}

In this section, we first introduce the motivation and reveal the challenges of restrictive unlearning when the class label and the target concept are decoupled (Section~\ref{sec:method_motivation}).
Second, we present systematic exploration through the perspectives of feature representation and forgetting dynamics (Section~\ref{sec:method_analyze}).
Lastly, we propose our novel framework, i.e., \textit{TARget-aware Forgetting} (TARF) (Section~\ref{sec:method_gradient}).

\subsection{Motivation: exploring mismatched taxonomy in unlearning}
\label{sec:method_motivation}

With the increasing concerns about trustworthiness, the unlearning requests emerged to erase the knowledge about an unwanted set (e.g., the biased~\cite{yao2023large} or inappropriate content~\cite{gandikota2023erasing}) from the trained model, especially in the era of foundation models~\cite{bommasani2021opportunities}. Recently, a series of studies~\cite{golatkar2020eternal,warnecke2021machine,jia2023model,fan2023salun,chen2023boundary} have several proposals on forgetting a training class of the models, and demonstrated it can be successfully achieved by partially scrubbing the class data or fine-tuning on the retaining data to realize catastrophic forgetting~\cite{french1999catastrophic,goodfellow2014explaining}, to achieve a state as if the model had never been trained with this class of data. However, a general scenario considered in previous works is that the target concept is aligned with the taxonomy of the pre-training tasks, which may not always hold in practical unlearning requests. This naturally motivates the following research question,
\begin{quote}
\centering
\textit{What if the class labels and target concept do not coincide in unlearning?}
\end{quote}
To investigate it, we consider different label domains of three critical aspects in class-wise unlearning, i.e., $\mathcal{L}_D$ of forgetting data, $\mathcal{L}_M$ of model outputs, and $\mathcal{L}_T$ of target concepts. Assuming that $\mathcal{L}_D\preceq\mathcal{L}_T$, we have either $\mathcal{L}_D=\mathcal{L}_T$ or $\mathcal{L}_D\prec\mathcal{L}_T$. When $\mathcal{L}_D=\mathcal{L}_T$, we have all matched if $\mathcal{L}_D=\mathcal{L}_M$ and model mismatch if $\mathcal{L}_T\prec\mathcal{L}_M$; When $\mathcal{L}_D\prec\mathcal{L}_T$, we have target mismatch if $\mathcal{L}_D=\mathcal{L}_M$ and data mismatch if $\mathcal{L}_T=\mathcal{L}_M$\footnote{Note that we also provide a detailed discussion in appendix~\ref{app:whole_mismatch} for all the potential cases, while some (e.g., $\mathcal{L}_T\prec\mathcal{L}_D$) are impractical and some (e.g., $\mathcal{L}_M\prec\mathcal{L}_T$) are similar to the major scenarios considered here.}. In Figure~\ref{fig: challenge}, we conduct the unlearning on the four forgetting tasks as instantiated in Figure~\ref{fig: setting_example}. As a result, the two representative unlearning methods, e.g., Gradient Ascent (GA)~\cite{thudi2022unrolling,thudi2022necessity} and Fine-tuning (FT) show different performance gaps compared with the retrained models except in the conventional setting. It can be found with illustrations that some concept-aligned forgetting data are under-represented by the forgetting data when $\mathcal{L}_D\prec\mathcal{L}_T$ and some class-aligned remaining data are entangled with the forgetting part when $\mathcal{L}_T\prec\mathcal{L}_M$, raising the new challenges.

\begin{figure*}[t!]
    \begin{center}
    \subfigure[\textit{Target identification}: accuracy changes, selection results, unlearning performance]{
    \includegraphics[scale=0.135]{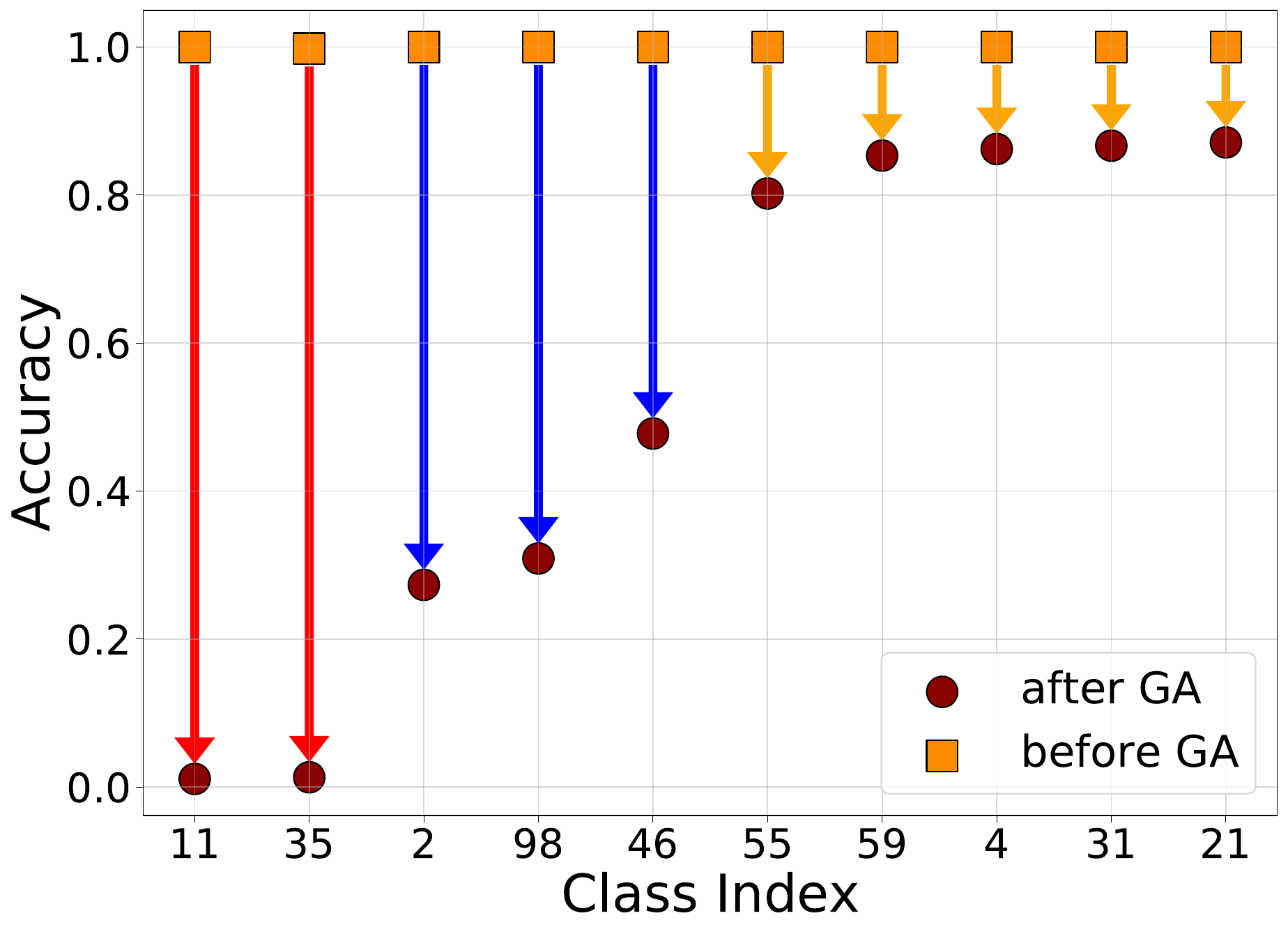}
    \includegraphics[scale=0.135]{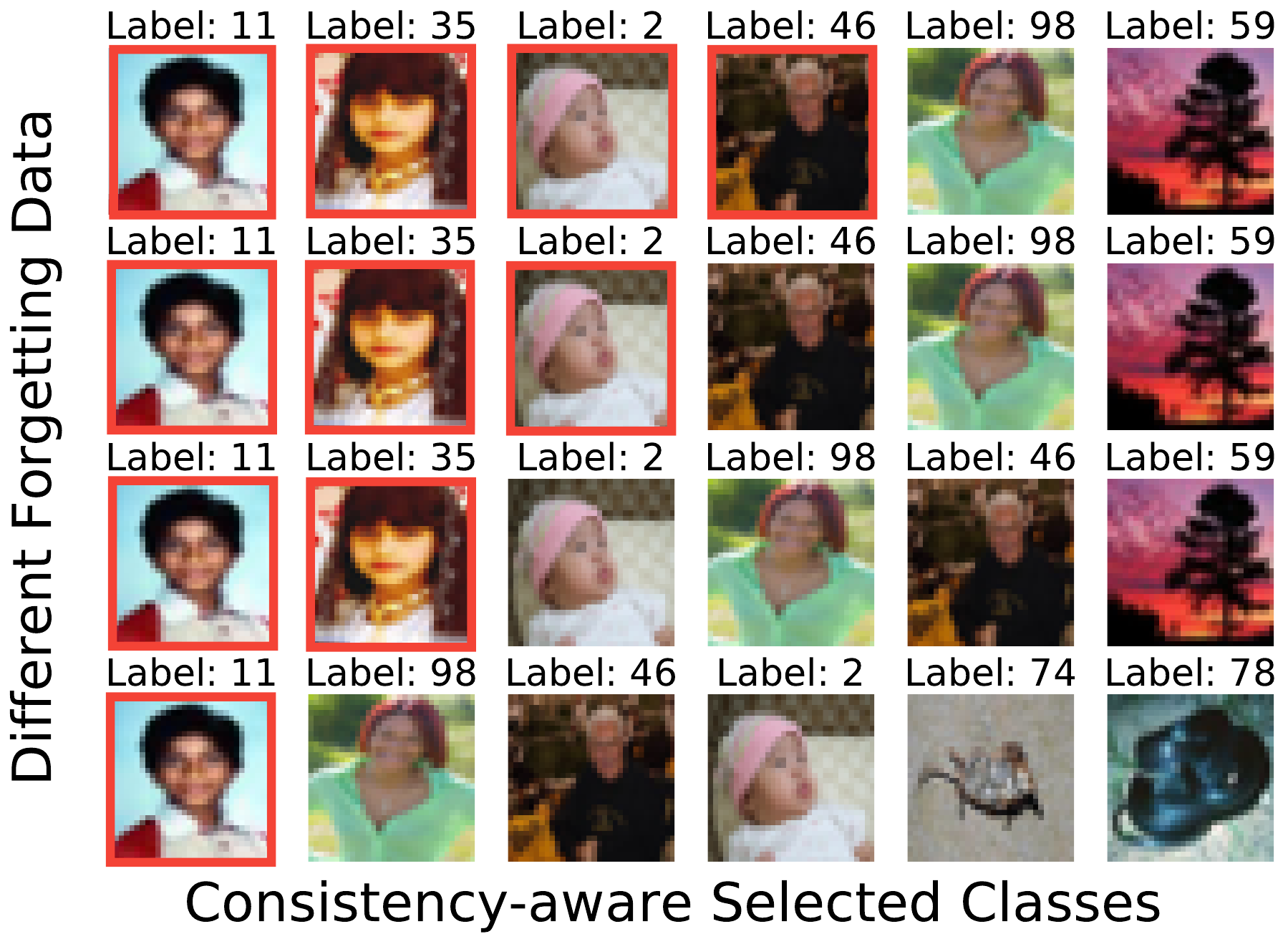}
    \includegraphics[scale=0.135]{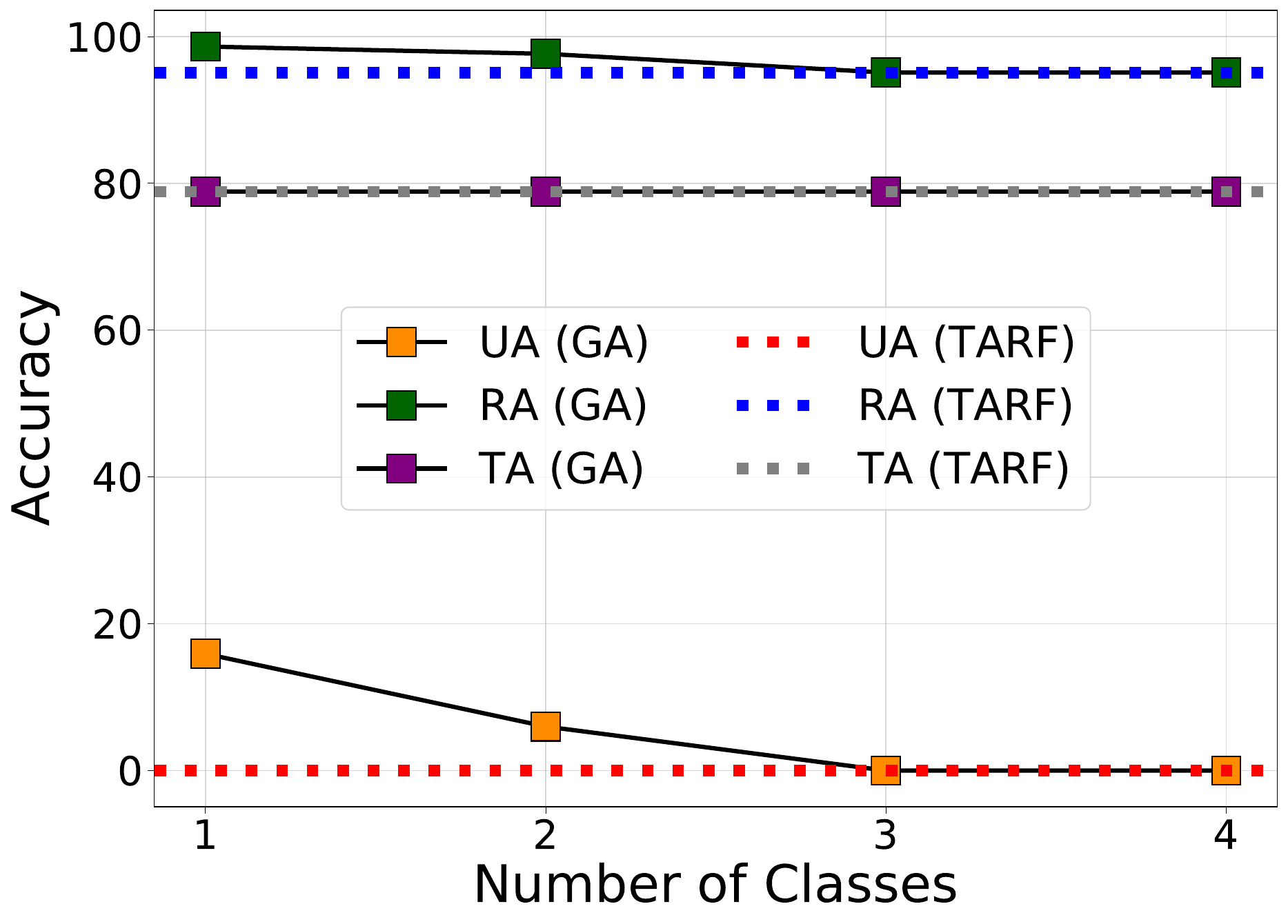}
    \label{fig:frameworka}
    }
    \\
    \vspace{-2mm}
    \subfigure[\textit{Target separation}: decomposed performance, decomposed position, reconstruction needs]{
    \includegraphics[scale=0.135]{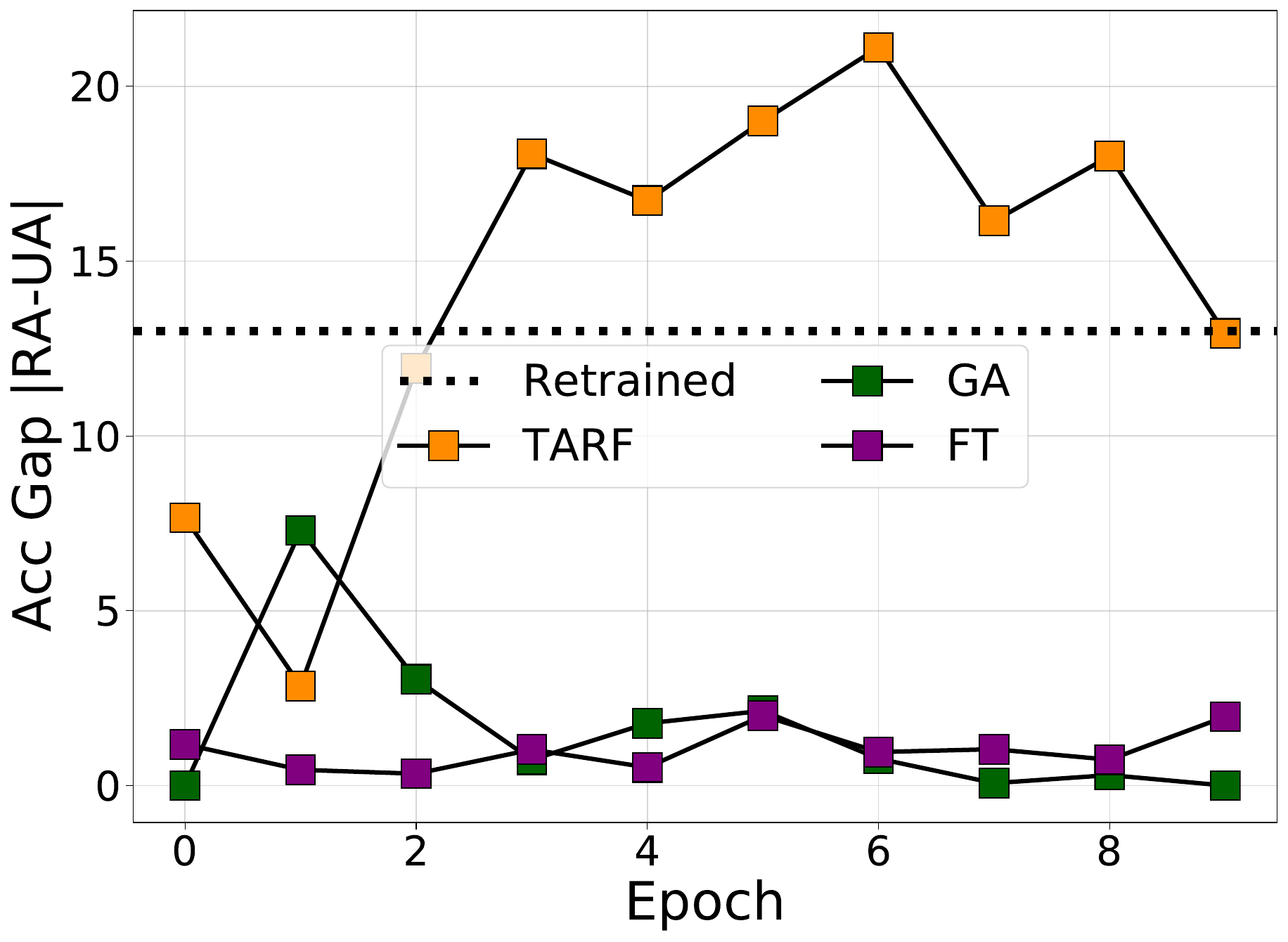}
    \includegraphics[scale=0.135]{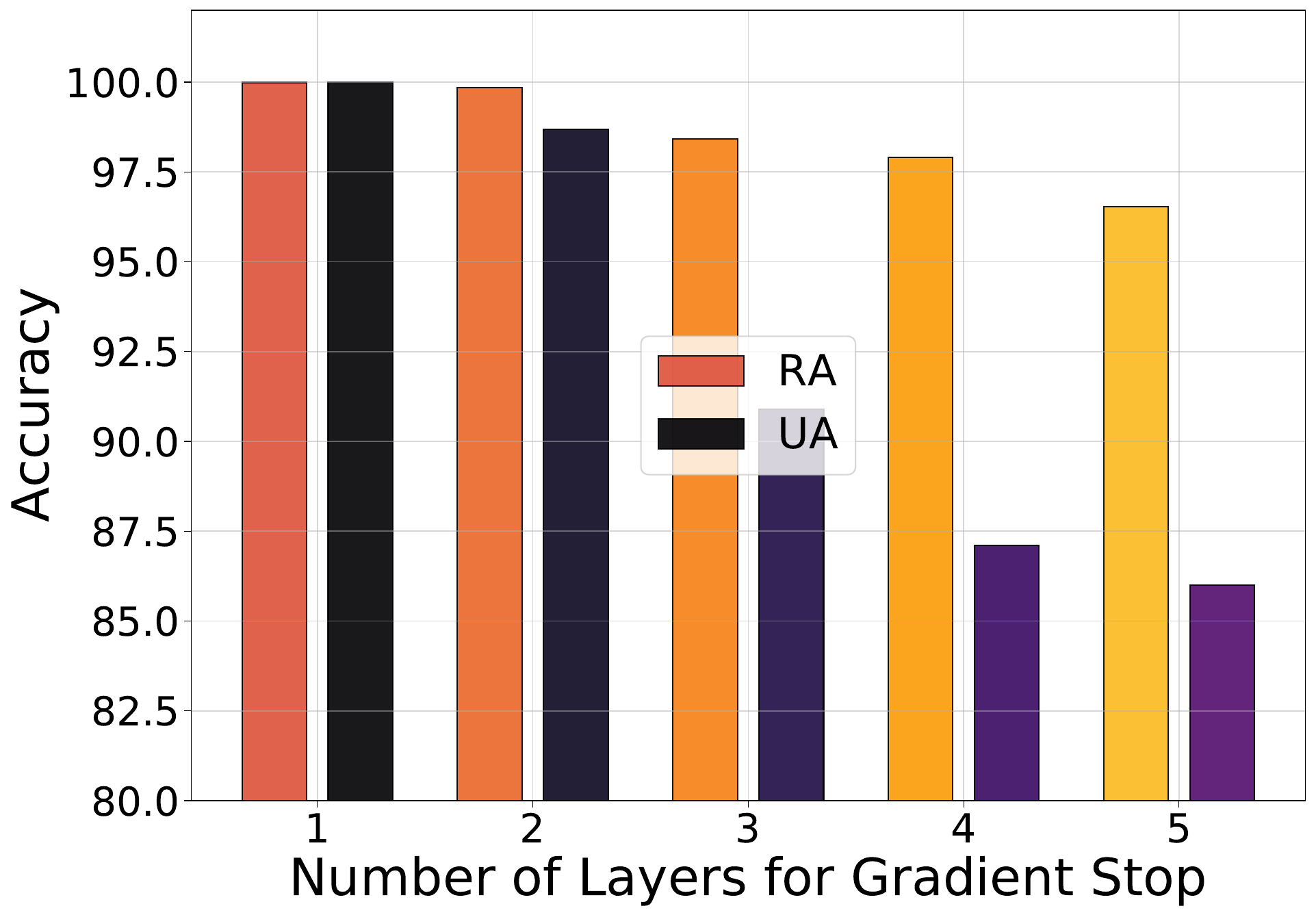}
    \includegraphics[scale=0.135]{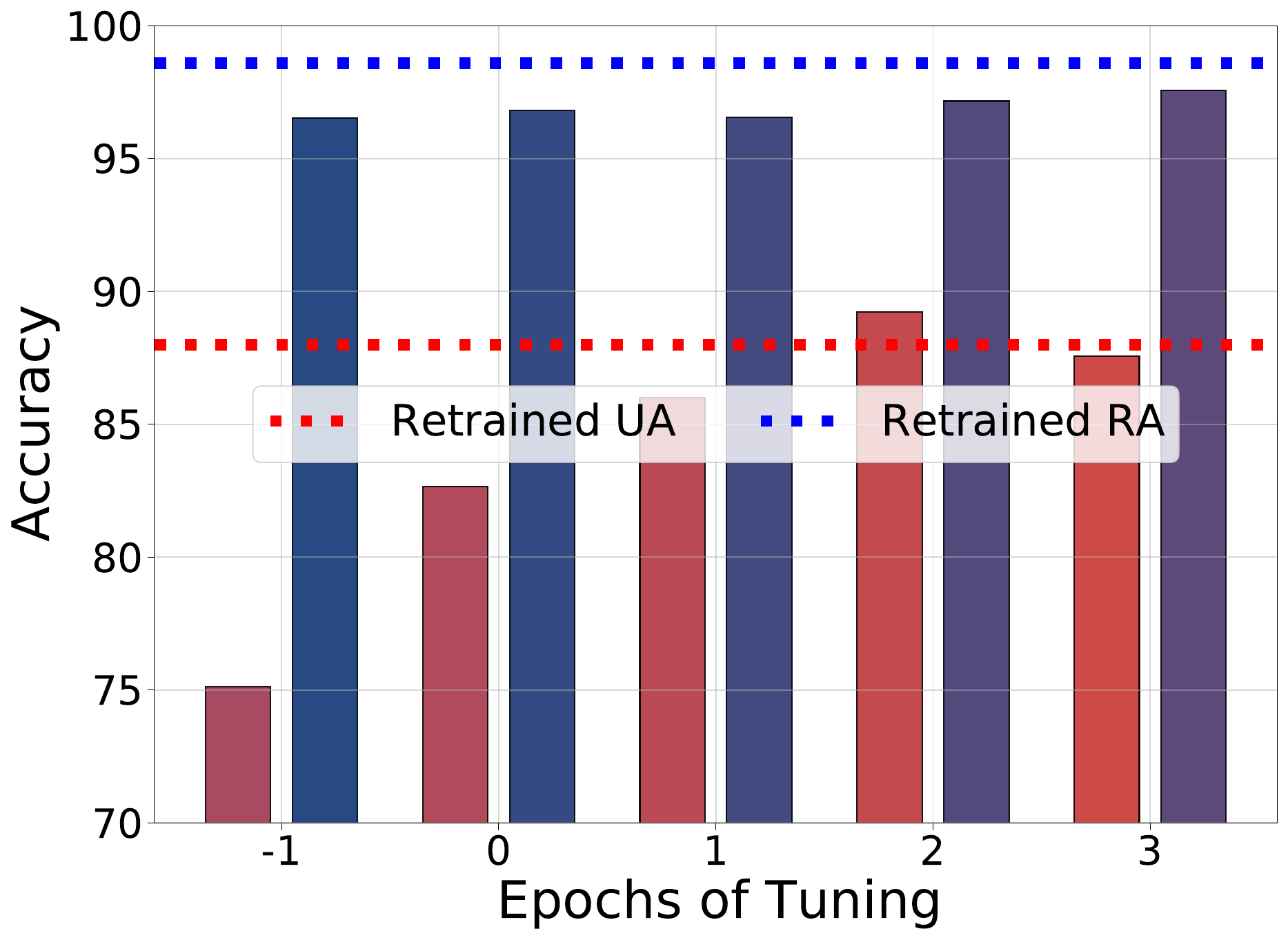}
    \label{fig:frameworkb}
    }
    \end{center}
    \vspace{-2mm}
    \scriptsize{From left to right (a): accuracy changes of forgetting data, concept-aligned data, and remaining data in target mismatch forgetting, the selection results using different forgetting data, unlearning performance using different forgetting data; (b): accuracy gap of remaining and forgetting part of the same class, unlearning with gradient stop at different layers to show the feature deconstruction, the need of reconstruction.}
    \caption{Target identification and target separation for unlearning under mismatch.
    }
    \vspace{-2mm}
    \label{fig:target_iden_separa}
\end{figure*}

\subsection{Systematic Exploration on Forgetting Dynamics}
\label{sec:method_analyze}

The mismatch of label domains affects the construction of model representation in unlearning, which requires us to explore it further to understand the performance gaps as well as the forgetting dynamics. 

\paragraph{Target or data mismatch.} In both target and data mismatch forgetting, we have $\mathcal{L}_D\prec\mathcal{L}_T$, which means the forgetting data can be a subset of the target concept, i.e., $\mathcal{D}_\text{f}\subset\mathcal{D}_\text{t}$. As indicated in Figure~\ref{fig: challenge}, partially relying on the forgetting or remaining data can not fully represent the target concept, and leaves concept-aligned data hard to unlearn. We can have the following proposition for $\mathcal{L}_D\prec\mathcal{L}_T$, 
\begin{proposition}[Insufficient representation] 
\label{proposition-1} Given $\mathcal{L}_D\prec\mathcal{L}_T$ that indicates $\mathcal{D}_\text{f}\subset\mathcal{D}_\text{t}$, and a cluster center $h^*$ of feature representations $h_{x\sim \mathcal{D}_\text{f}}(x)$ extracted at the pre-trained model $\theta$, as well 
as a distance measure $d(\cdot,\cdot)$, there exist a constant value $\zeta_1$ and sample $(x^u,y^u)\sim (\mathcal{D}_\text{t}\backslash\mathcal{D}_\text{f})$ that exhibits weak gravity following the sample  $(x,y)\sim\mathcal{D}_\text{f}$ on the forgetting dynamic $\epsilon=\mathbb{E}(\ell(f_\theta(x),y) - \ell(f_{\theta^t}(x),y))$, measured with the model $\theta^t$ unlearned from $\theta$ at a time interval $t$,
\begin{equation}
     (d(h(x^u), h^*)>\sup_{x\sim \mathcal{D}_\text{f}} d(h(x),h^*)) \;  \Rightarrow \; |(\ell(f_\theta(x^u),y^u) - \ell(f_{\theta^t}(x^u),y^u))-\epsilon| >\zeta_1.
\end{equation}
\end{proposition}
\paragraph{Model mismatch forgetting.} In this task, we have $\mathcal{L}_D=\mathcal{L}_T$ while $\mathcal{L}_T\prec\mathcal{L}_M$. 
Regarding the model trained by the superclass, it can be found in Figure~\ref{fig:distance_b} that the features of forgetting data and class-aligned retaining data are closely entangled, showing that the unlearning of the forgetting data can unavoidably affect the representation of the other part. In contrast, it is also notable in the left-bottom of Figure~\ref{fig: challenge} that the accuracy gap between forgetting data and class-aligned retaining data is expected to be large in the retrained reference. In formal, we summarize the observation as follows,
\begin{proposition}[Decomposition lacking] 
\label{proposition-2}
Given $\mathcal{L}_T\prec\mathcal{L}_M$ that indicates the broader representation region for $\mathcal{D}_{\rm z}:=\mathcal{D}_{\rm r,z}\cup\mathcal{D}_{\rm t}$ within the same class $z$, where the sample $x^m\sim (\mathcal{D}_{\rm 
 z}\backslash\mathcal{D}_{\rm f})$ exhibits strong gravity following the forgetting dynamic $\epsilon=\mathbb{E}(\ell(f_\theta(x),y) - \ell(f_{\theta^t}(x),y))$ with a constant value $\zeta_2$,
\begin{equation}
    (d(h(x^m), h^*)\leq\sup_{x\sim \mathcal{D}_{\rm z}} d(h(x),h^*))\;
    \Rightarrow\; |(\ell(f_\theta(x^m),y^m) - \ell(f_{\theta^t}(x^m),y^m))-\epsilon| < \zeta_2.
\end{equation}
\end{proposition}

\paragraph{Forgetting dynamics with representation distance.} Despite the issues revealed by previous propositions under label domain mismatch, the forgetting performance varied obviously on different representations. We can find that GA achieves better forgetting efficacy on the data mismatch forgetting as the feature representation of the forgetting data and concept-aligned data is entangled. Through a closer check on the geometry distance of representations and the model behaviors during forgetting in Figure~\ref{fig:feature}, we have the following that reveals a crucial factor for achieving these tasks,

\begin{definition}[Representation gravity] 
\label{proposition-3}
Given the empirically demonstrated Proposition~\ref{proposition-1} and Proposition~\ref{proposition-2}, we can have the formal indicator $I_\text{con}(x,y,\theta)$ to reflect the representation similarity $d(h(x),h^*)$ in the model $\theta^t$, which is a crucial factor for the feasibility of unlearning under mismatch,
\begin{equation}\label{eq:consistency}
    I_\text{con}(x,y,\theta) = |\ell(f_\theta(x),y) - \ell(f_{\theta^t}(x),y)|.
\end{equation}
\end{definition}
It is empirically demonstrated in Figure~\ref{fig:feature} where we present the average representation distance to the cluster center of forgetting data, the corresponding changes in accuracy and loss values show that the smaller the distance in representation level, the similar forgetting dynamics the model would have on prediction. As the Propositions~\ref{proposition-1} and~\ref{proposition-2} reveal the issues of insufficient representation and decomposition missing, we can utilize the representation gravity to identify the other unidentified forgetting data in the remaining set, and reveal the needs of deconstructing the pre-entangled representation by simultaneously considering the forgetting and retaining objectives. On the left panels of Figures~\ref{fig:frameworka} and~\ref{fig:frameworkb}, we show the empirical feasibility of the above two goals.

\begin{figure*}[t!]
    \begin{center}
    \includegraphics[scale=0.23]{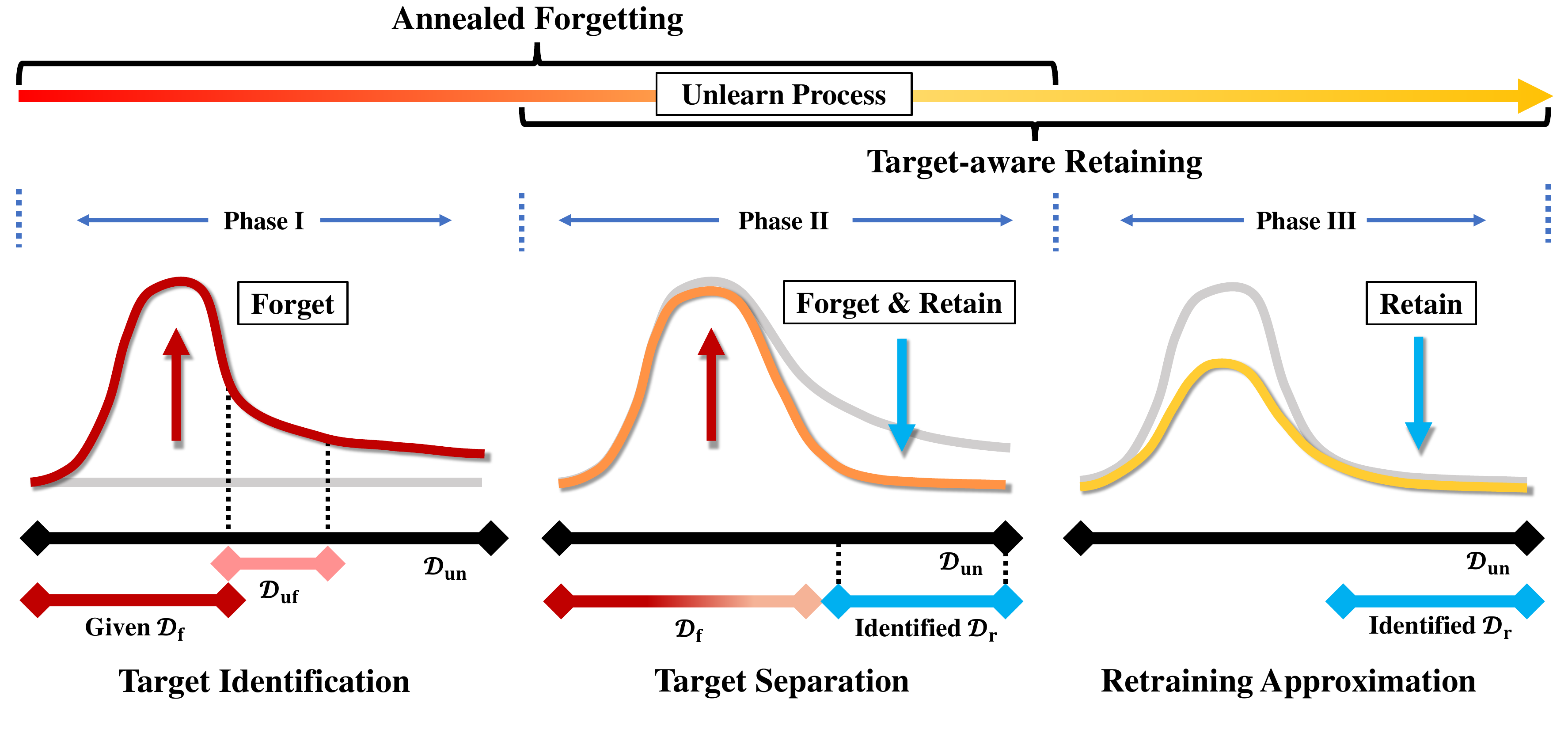}
    \end{center}
    \vspace{-4mm}
    \scriptsize{The overall framework consists of two objective parts, e.g., annealed forgetting and target-aware retaining, which can be regarded as three phases to enable all the class-wise unlearning tasks through the view of the unlearning process.  (a) Phase I utilizes the gradient ascent to construct dynamic information for all class data; (b) Phase II simultaneously considers gradient ascent on forgetting data and gradient decent on remaining data that is hard to affect to separate target concept; (c) Phase III conducts gradient descent on the selected data to approximate the retraining.}
    \caption{Overview of the proposed framework TARF.
    }
    \vspace{-2mm}
    \label{fig:framework}
\end{figure*}

\subsection{Algorithm Framework of TARF}
\label{sec:method_gradient}

Based on the previous analysis, we introduce the whole framework of \textit{TARget-aware Forgetting} (TARF), to enable the four class-wise unlearning tasks. Given the identified forgetting data, we illustrate the overall process in Figure~\ref{fig:framework}, and introduce its dynamic learning objective as follows:
\begin{align}\label{eq:4}
\begin{split}
    L_\text{TARF} = \underbrace{k(t)\cdot\bigg (-\frac{1}{|\mathcal{D}_{\text{f}}|}\sum_{(x,y)\sim \mathcal{D}_{\text{f}}}\ell(f(x),y)\bigg )}_{\text{\textbf{Annealed Forgetting}}~ L_{\text{f}}(k)}+ \underbrace{\frac{1}{|\mathcal{D}_\text{un}|}\sum_{(x,y)\sim \mathcal{D}_\text{un}}\ell(f(x),y)\cdot\tau(x,y,t)}_{\text{\textbf{Target-aware Retaining}}~ L_\text{u}(\tau)},
\end{split}
\end{align}
where $k(t)$ serves as an annealing strategy to control the strength of the forgetting part. Along with training, we expect the overall objective to approximate the retraining ones $L_\text{TARF}\rightarrow L_\text{retrain}$ through,
\begin{align}
    \begin{split}
    L_{\text{f}}(k)\rightarrow 0,\quad
    L_\text{u}(\tau)\rightarrow L_\text{retrain},
    \end{split}
\end{align}
given the initially provided forgetting data $\mathcal{D}_\text{f}$ and the remaining set $\mathcal{D}_\text{un}$. Specifically, we design the two dynamic hyperparameters $k(t)$ and $\tau(x,y,t)$ as follows to achieve that, 
\begin{equation}\label{eq:k_t}
    k(t) = \frac{k\cdot(T-t-t_0)}{T},\quad t\in[0,T];\quad \tau(x, y, t)=\left \{
    \begin{aligned}
    0& && I_\text{con}(x,y,\theta_{t_1})>\beta ~\text{or}~ t<t_1,\\
    1& && I_\text{con}(x,y,\theta_{t_1})<\beta ~\text{and}~ t\geq t_1,
    \end{aligned}
    \right.
\end{equation}
where $T$ indicates the total training time (e.g., epochs), and the value of $k(t)$ decreases with the training process, $\beta$ can be estimated by the information of forgetting dynamics about the specific unlearning request and the rank of loss values at $t_1$, $t_0$ and $t_1$ respectively control the end time of active forgetting and the begin time of retaining part. The overall process can be regarded as, 

\paragraph{Phase I: Target Identification.} Before $t_1$, since $\tau(x,y,t)=0$, Eq.~\eqref{eq:4} can be formalized as,
$L_\text{TARF-Phase-I} = k(t)\cdot (-\frac{1}{|\mathcal{D}_{\text{f}}|}\sum_{(x,y)\sim \mathcal{D}_{\text{f}}}\ell(f(x),y) ),$
in which the retaining part is waiting for the dynamic information revealed by this phase. As shown in Figures~\ref{fig:feature}, the concept-aligned forgetting data in $\mathcal{D}_\text{uf}$ can be identified due to the similar semantic features with the forgetting data. For the specific selection, we utilize the class label information in our main tasks. In the middle and right of Figure~\ref{fig:frameworka}, we show the selected classes and validate unlearning efficacy with the identification.

\paragraph{Phase II: Target Separation.} After phase I, the retaining part is engaged with the forgetting part with the identified data $\mathcal{D}_\text{uf}$ and the remaining retaining data $\mathcal{D}_\text{r}$. 
By simultaneously considering the forgetting and retaining part as Eq.~\eqref{eq:4}, $L_\text{TARF-Phase-II}$ encourages the model to deconstruct the target concept and reconstruct the feature representation of the retaining part, which can effectively decouple the pre-entangled feature in the model mismatch forgetting. In the middle of Figure~\ref{fig:frameworkb}, we adopt a gradient stop at each layer of target separation to check the position of feature decomposition, showing that unlearning distorts the underlying representation at the deeper layers of the model.

\paragraph{Phase III: Retraining Approximation.} After $t_0$, we focus on retaining the current model, which approximates the retraining objective as follows,
$L_\text{TARF-Phase-III} = \frac{1}{|\mathcal{D}_\text{un}|}\sum_{(x,y)\sim \mathcal{D}_\text{un}}\ell(f(x),y)\cdot\tau(x,y,t),$
where we use $\tau$ at $t_1$ to indicate the identified hard-to-effect retaining data, and continually reconstruct the representations. Since the general goal of unlearning considered in our work is similar to retraining, this phase can prevent excessive forgetting. In the right of Figure~\ref{fig:frameworkb}, we compare the performance using different lengths of this phases with the retrained reference,

\section{Experiments}
\label{sec:exp}

In this section, we present the comprehensive experimental results. To begin with, we introduce the basic setups for the unlearning and the evaluation (Section~\ref{sec:exp_setup}). Then, we validate the effectiveness of our method in four unlearning scenarios with the decoupled class label and the target concept (Section~\ref{sec:exp_main}). To better understand its properties, we conduct various experiments on the ablation study and provide further discussions (Section~\ref{sec:exp_ablation}). More details and results are provided in Appendix~\ref{app:exp}.

\subsection{Experimental Setup}
\label{sec:exp_setup}

\paragraph{Datasets and models.} In our experiments, we explore machine unlearning for the conventional image classification tasks~\cite{goodfellow2016deep}. Since the introduced unlearning settings need a coarse-to-fine label structure, we adopt the benchmarked dataset, e.g., \textit{CIFAR-10/CIFAR-100}~\cite{krizhevsky2009learning_cifar10} with their superclass information in the major experiments. 
Specifically, we train two models based on the original classes and its superclass respectively, and instantiate four tasks (as illustrated in Figure~\ref{fig: setting_example}) of unlearning with the decoupled class label and the target concept. The detailed information is summarized in Table~\ref{tab:exp_set}. Following previous works~\cite{warnecke2021machine,jia2023model,fan2023salun}, we use ResNet-18~\cite{he2016deep} as the major architecture to obtain the original trained models with standard learning~\cite{goodfellow2016deep}, and then set it to be the basis for unlearning.

\begin{table*}[t!]
    \caption{Main Results ($\%$). Comparison with the unlearning baselines. All methods are trained on the same backbone, i.e., the basis of unlearning initialization is the same (except for retraining from scratch). Bold numbers are superior results. $\downarrow$ indicates smaller values are better. (The complete results under multiple runs are summarized with mean and std values in Appendix~\ref{app:exp})}
    \centering
    \footnotesize
    \renewcommand\arraystretch{0.92}
    \resizebox{\textwidth}{!}{
    \begin{tabular}{c|l|cccccc|cccccc}
        \toprule[1.5pt]
        Type / $\mathcal{D}$ & Dataset & \multicolumn{6}{c|}{\textbf{CIFAR-10}} & \multicolumn{6}{c}{\textbf{CIFAR-100}} \\
        \cmidrule{2-14}
        &  Method / Metrics & UA & RA & TA  & MIA & Gap$\downarrow$ & RTE$\downarrow$ & UA & RA & TA  & MIA & Gap$\downarrow$ & RTE$\downarrow$ \\
        \midrule[0.6pt]
        \multirow{10}*{\shortstack{\textbf{All matched}}}
        &\gray Retrained &\gray0.00 & \gray99.51 & \gray94.69 & \gray100.00 & \gray- & \gray43.3 & \gray0.00 & \gray97.85 & \gray76.03 & \gray100.00 & \gray- & \gray43.2  \\
        & FT~\cite{warnecke2021machine} & 1.07 & 98.62 & 92.36 & 100.00 & 1.07 & 4.43 & 0.67 & 96.32 & 72.34 & 100.00 & 1.47 & 5.02  \\
        & RL~\cite{toneva2018empirical} & 4.13 & 97.65 & 91.23 & 100.00 & 2.36 & 4.88 & 1.00 & 96.09 & 72.00 & 100.00 & 1.70 & 4.96 \\
        & GA~\cite{ishida2020we} & 0.49 & 95.24 & 88.17 & 99.78 & 2.88 & \textbf{0.25} & 1.33 & 94.74 & 68.56 & 99.89 & 3.01 & \textbf{0.06}  \\
        & IU~\cite{izzo2021approximate} &0.22 & 88.15 & 82.38 & 99.96 & 5.99 & 0.45 & 0.00 & 37.61 & 29.58 & 100.00 & 26.67 & 0.51  \\
        & BS~\cite{chen2023boundary} &25.04 & 87.94 & 80.90 & 88.67 & 15.43 & 0.82 & 12.88 & 97.11 & 68.66 & 99.33 & 5.42 & 0.78  \\
        & $L_{1}$-sparse~\cite{jia2023model} &0.00 & 94.20 & 89.77 & 100.00 & 2.56 & 4.39 & 0.00 & 82.00 & 65.08 & 100.00 & 6.70 & 4.39  \\
        & SalUn~\cite{fan2023salun} & 0.00 & 91.32 & 86.87 & 100.00 & 4.00 & 5.65 & 0.0 & 75.34 & 62.14 & 100.00 & 9.10 & 5.75  \\
        & SCRUB~\cite{kurmanji2023towards} & 0.00 & 12.92 & 12.92 & 0.00 & 67.09 & 2.88 & 0.00 & 99.98 & 76.75 & 100.00 & \textbf{0.71} & 3.23 \\
        \cmidrule{2-14}
        & \textbf{TARF} (ours) & 0.00 & 98.23 & 91.95 & 100.00 & \textbf{1.01} & 4.21 & 0.00 & 96.90 & 72.53 & 100.00 & 1.11 & 4.68  \\
        \midrule[0.6pt]
        \multirow{10}*{\shortstack{\textbf{Model}\\\textbf{mismatch}}}
         &\gray Retrained & \gray87.76 & \gray99.58 & \gray95.91 & \gray20.57 & \gray- & \gray43.8 & \gray88.22 & \gray98.58 & \gray78.50 & \gray25.78 & \gray- & \gray43.8 \\
        & FT~\cite{warnecke2021machine} & 94.67 & 98.53 & 93.56 & 9.56 & 5.33 & 4.29 & 92.67 & 95.02 & 79.34 & 16.33 & 4.58 & 4.86  \\
        & RL~\cite{toneva2018empirical} & 53.69 & 97.85 & 92.39 & 96.60 & 28.84 & 4.82 & 80.11 & 95.83 & 79.83 & 99.00 & 21.35 & 4.93 \\
        & GA~\cite{ishida2020we} & 5.76 & 86.99 & 82.20 & 94.98 & 45.68 & \textbf{0.25} & 6.78 & 94.83 & 76.96 & 97.78 & 39.68 & \textbf{0.06}  \\
        & IU~\cite{izzo2021approximate} & 23.69 & 87.34 & 82.57 & 89.87 & 39.74 & 0.44 & 34.67 & 96.83 & 79.08 & 86.44 & 29.14 & 0.49  \\
        & BS~\cite{chen2023boundary} & 10.29 & 50.77 & 49.39 & 95.96 & 62.05 & 0.79 & 18.11 & 95.90 & 72.28 & 95.22 & 37.14 & 0.89  \\
        & $L_{1}$-sparse~\cite{jia2023model}& 93.11 & 94.76 & 91.63 & 14.44 & 5.15 & 4.24 & 82.11 & 85.17 & 75.22 & 20.00 & 7.15 & 5.00  \\
        & SalUn~\cite{fan2023salun} & 8.91 & 93.95 & 84.38 & 99.32 & 43.69 & 6.04 & 66.33 & 78.83 & 70.78 & 77.00 &25.15 & 5.97  \\
        & SCRUB~\cite{kurmanji2023towards} & 48.62 & 27.86 & 28.29 & 48.62 &  51.63 & 3.06 & 0.00 & 11.25 & 10.00 & 98.89 & 79.29 & 4.12 \\
        \cmidrule{2-14}
        & \textbf{TARF} (ours) & 91.11 & 97.49 & 92.49 & 17.82 & \textbf{2.90} & 4.31 & 86.67 & 97.05 & 80.07 & 26.00 & \textbf{1.21} & 4.81 \\
        \midrule[0.6pt]
        \multirow{10}*{\shortstack{\textbf{Target}\\\textbf{mismatch}}}
        &\gray Retrained &\gray 0.00 &\gray 99.38 & \gray93.85 & \gray100.00 & \gray- & \gray52.1 & \gray0.00 & \gray97.85 & \gray73.72 & \gray100.00 & \gray- & \gray53.2 \\
        & FT~\cite{warnecke2021machine} & 50.43 & 98.47 & 91.65 & 50.44 & 25.78 & 4.38 & 58.18 & 96.32 & 72.53 & 46.76 & 28.54 & 5.00  \\
        & RL~\cite{toneva2018empirical} &51.25 & 97.56 & 90.90 & 56.23 & 24.95 & 4.79 & 58.89 & 96.05 & 72.20 & 46.98 & 28.81 & 4.93 \\
        & GA~\cite{ishida2020we} & 40.82 & 97.01 & 89.51 & 64.32 & 20.80 & \textbf{0.26} & 21.38 & 96.64 & 70.22 & 90.67 & 8.86 & \textbf{0.05}  \\
        & IU~\cite{izzo2021approximate} & 44.51 & 88.07 & 81.80 & 58.73 & 27.29 & 0.44 & 30.62 & 37.19 & 29.58 & 63.69 & 42.93& 0.50  \\
        & BS~\cite{chen2023boundary} & 53.62 & 88.65 & 75.39 & 76.33 & 26.62 & 0.82 & 40.44 & 98.32 & 68.66 & 85.16 & 15.20 & 0.97 \\
        & $L_{1}$-sparse~\cite{jia2023model} & 49.47 & 93.61 & 88.83 & 51.24 & 27.26 & 4.38 & 46.97 & 82.11 &65.08 & 54.00 & 29.33& 4.78  \\
        & SalUn~\cite{fan2023salun} & 46.63 & 91.08 & 86.31 & 60.94 & 25.38 & 5.90 & 59.64 & 75.52 & 62.37 & 65.96 & 27.35 & 5.81  \\
        & SCRUB~\cite{kurmanji2023towards}& 43.57 & 3.64 & 3.66 & 56.26 & 62.62 & 2.89 & 59.64 & 99.99 & 75.32 & 44.89 & 29.90 & 3.52 \\
        \cmidrule{2-14}
        & \textbf{TARF} (ours) & 0.06 & 97.57 & 90.81 & 100.00 & \textbf{1.23} & 4.23 &  0.31 & 97.35 & 73.68 & 100.00 & \textbf{0.21} & 4.85 \\
        \midrule[0.6pt]
        \multirow{10}*{\shortstack{\textbf{Data}\\\textbf{mismatch}}}
        & \gray Retrained & \gray0.00 & \gray99.54 & \gray95.56 & \gray100.00 & \gray- & \gray52.1 & \gray0.00 & \gray98.50 & \gray80.15 & \gray100.00 & \gray- & \gray53.2 \\
        & FT~\cite{warnecke2021machine} & 96.79 & 98.49 & 93.26 & 6.48 & 48.41 & 4.32 & 82.62 & 95.66 & 79.77 & 37.24 & 37.15 & 4.93  \\
        & RL~\cite{toneva2018empirical} & 76.47 & 97.68 & 91.93 & 49.81 & 33.04 & 4.76 & 89.78 & 96.82 & 79.90 & 70.76 & 30.49 & 4.97 \\
        & GA~\cite{ishida2020we} & 8.69 & 96.41 & 90.78 & 93.03 & 5.89 & \textbf{0.25} & 6.00 & 97.65 & 79.23 & 98.04 & 2.43 & \textbf{0.05}  \\
        & IU~\cite{izzo2021approximate} & 22.84 & 95.50 & 89.54 & 88.57 & 11.08 & 0.44 & 31.51 & 98.96 & 78.20 & 88.09 & 11.46 & 0.48  \\
        & BS~\cite{chen2023boundary} & 16.70 & 61.21 & 49.76 & 92.24 & 22.37 & 0.82 & 15.38 & 98.50 & 72.28 & 96.22 & 6.76 & 0.96  \\
        & $L_{1}$-sparse~\cite{jia2023model} &95.76 & 94.31 & 91.08 & 9.52 & 48.99 & 4.78 & 84.53 & 85.13 & 75.22 & 17.02 & 46.45 & 5.03 \\
        & SalUn~\cite{fan2023salun} & 51.77 & 93.87 & 90.46 & 63.52 & 24.75 & 5.72 & 72.93 & 78.87 & 71.04 & 54.13 & 36.89 & 5.72 \\
        & SCRUB~\cite{kurmanji2023towards} & 59.46 & 22.55 & 23.13 & 59.46 & 62.35 & 2.94 & 0.00 & 11.61 & 11.00 & 98.90 & 39.29 & 3.68 \\
        \cmidrule{2-14}
        & \textbf{TARF} (ours) & 0.00 & 98.17 & 93.09 & 100.00& \textbf{0.96} & 4.22 & 0.00 & 95.01 & 78.98 & 100.00 & \textbf{1.17} & 4.78 \\
        \bottomrule[1.5pt]
    \end{tabular}
    }
    \label{tab:mu_main}
    \vspace{-6mm}
\end{table*}

\paragraph{Evaluation metrics.} The general target of class-wise unlearning considered in this work is to approximate the Retrained model~\cite{warnecke2021machine}. To give a comprehensive evaluation, we adopt 5 specific evaluation metrics in classification tasks following previous works~\cite{jia2023model,fan2023salun}. 
\begin{wrapfigure}{r}{0.4\textwidth}
    \begin{center}
    \vspace{-3mm}
    \subfigure{
    \hspace{0.05in}
    \rotatebox{90}
    {\scriptsize{SD}}
    \hspace{0.1in}
    \begin{minipage}{\linewidth}
        \includegraphics[scale=0.056]{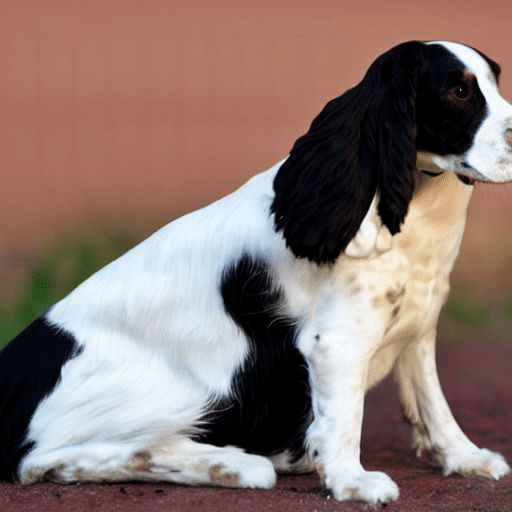}
    \includegraphics[scale=0.056]{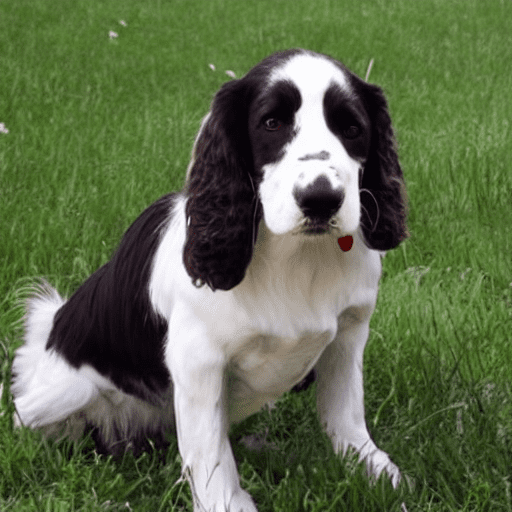}
    \includegraphics[scale=0.056]{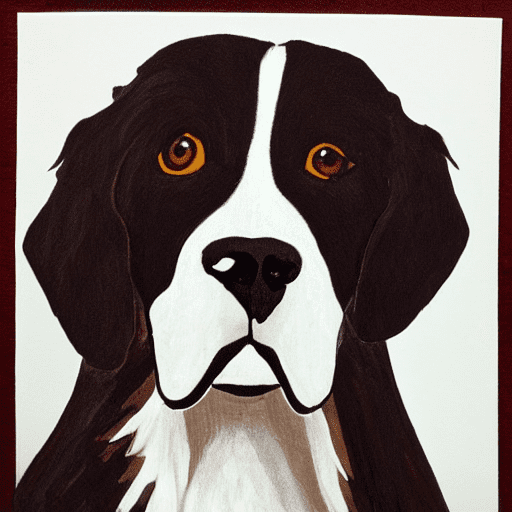}
    \includegraphics[scale=0.056]{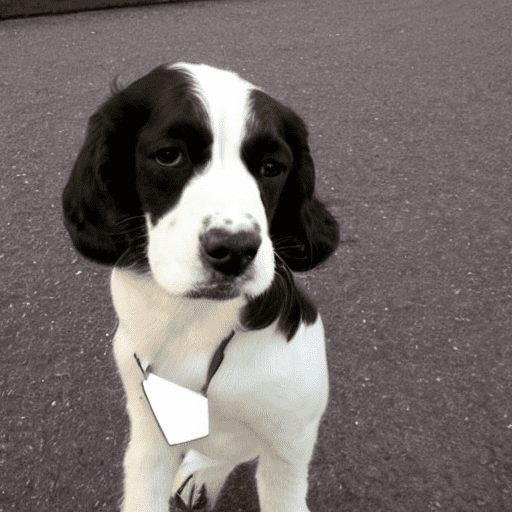}
    \\
    \includegraphics[scale=0.056]{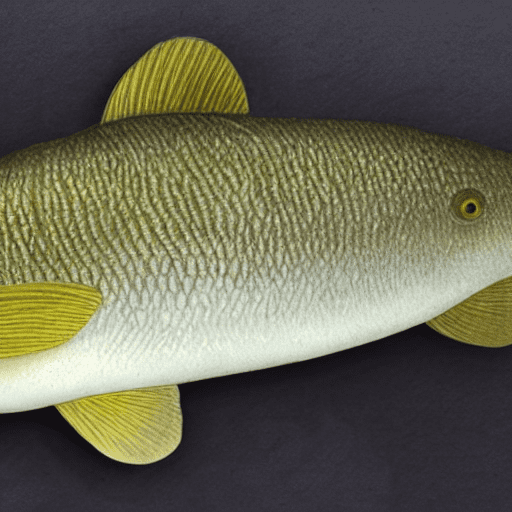}
    \includegraphics[scale=0.056]{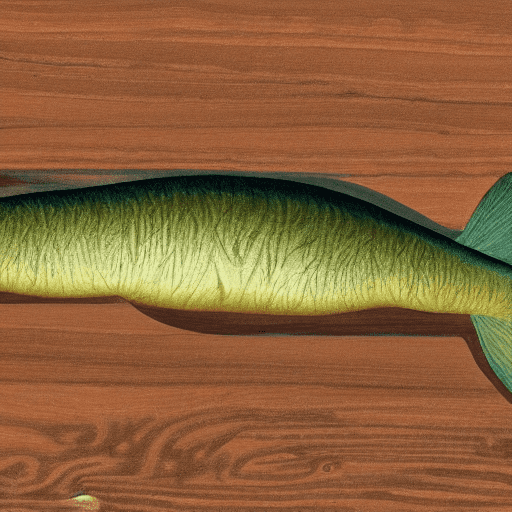}
    \includegraphics[scale=0.056]{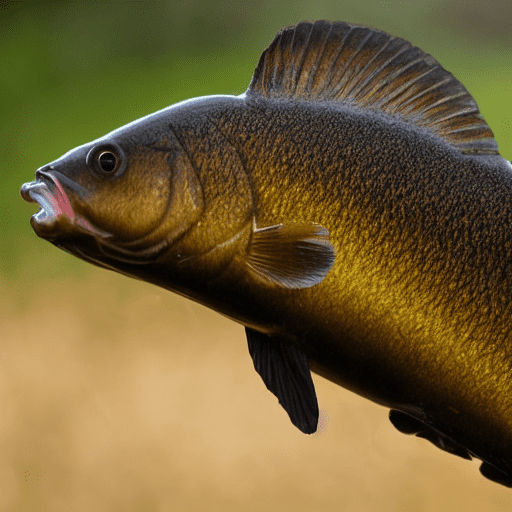}
    \includegraphics[scale=0.056]{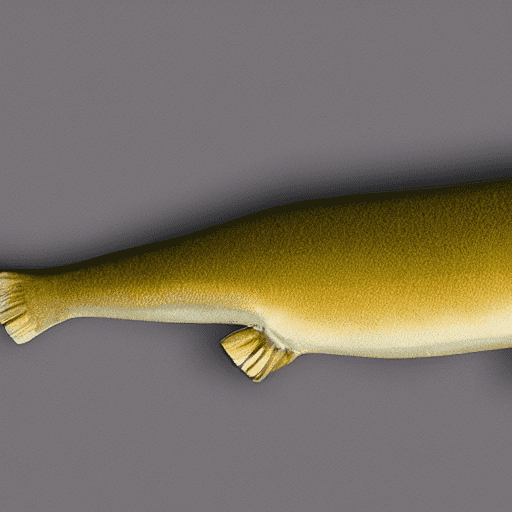}
    \end{minipage}
    }
    \\
    \vspace{-3mm}
    \subfigure{
    \hspace{0.05in}
    \rotatebox{90}
    {\scriptsize{CL}}
    \hspace{0.1in}
    \begin{minipage}{\linewidth}
        \includegraphics[scale=0.056]{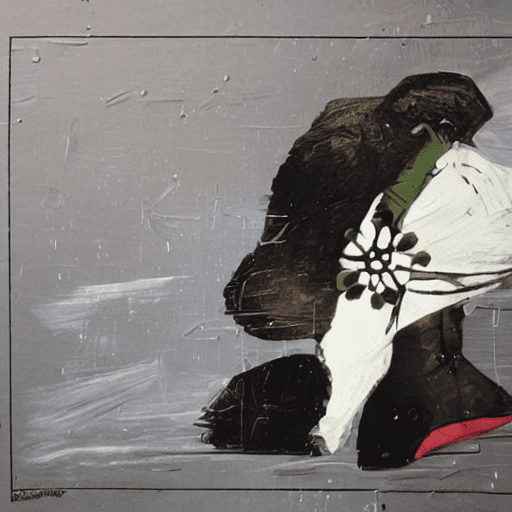}
    \includegraphics[scale=0.056]{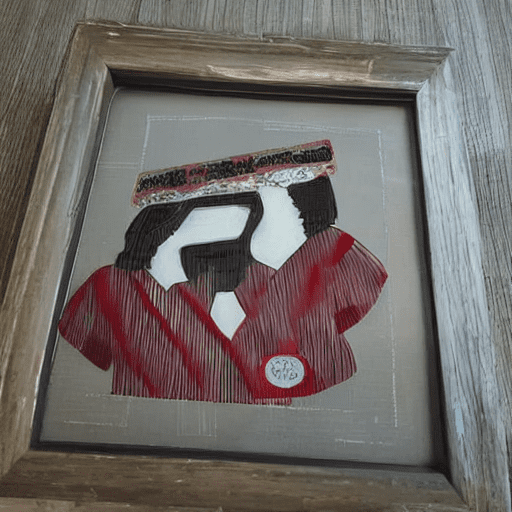}
    \includegraphics[scale=0.056]{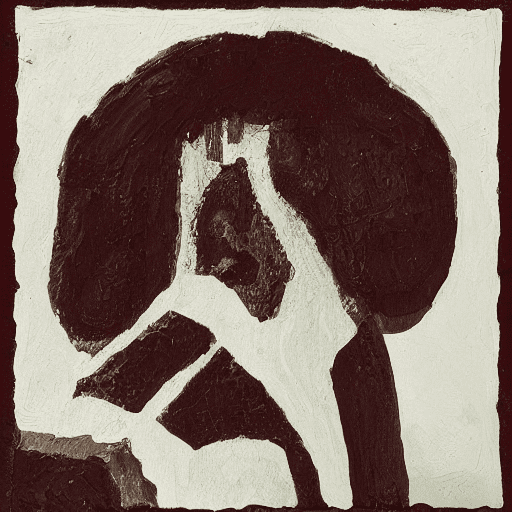}
    \includegraphics[scale=0.056]{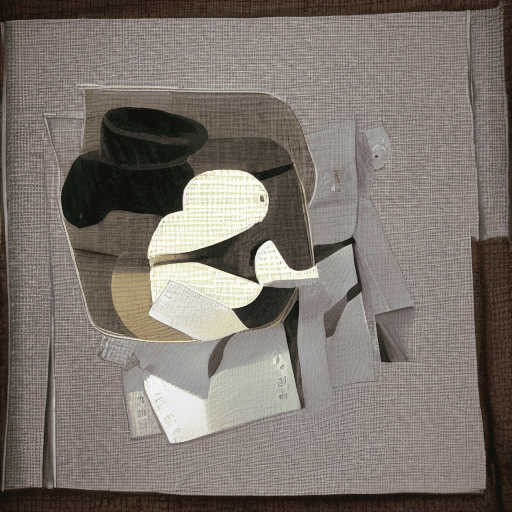}\\
    \includegraphics[scale=0.056]{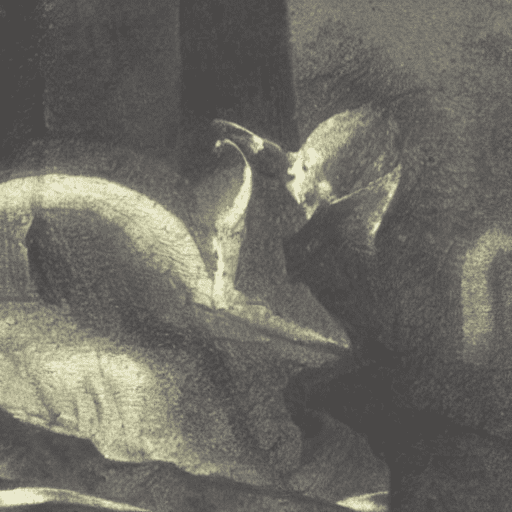}
    \includegraphics[scale=0.056]{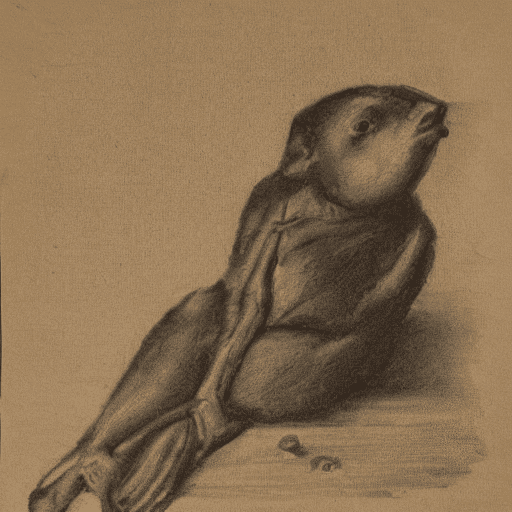}
    \includegraphics[scale=0.056]{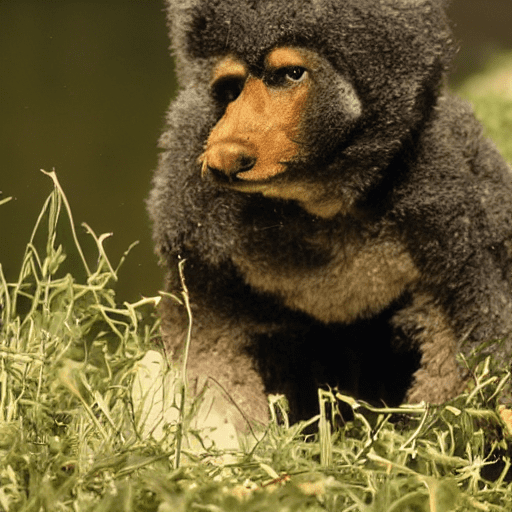}
    \includegraphics[scale=0.056]{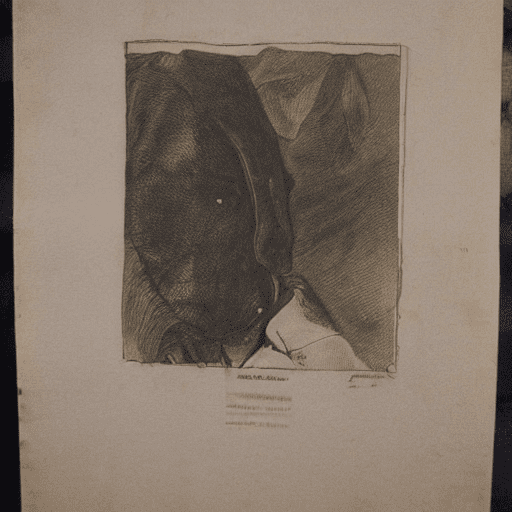}
    \end{minipage}
    }
    \\
    \vspace{-3mm}
    \subfigure{
    \hspace{0.05in}
    \rotatebox{90}
    {\scriptsize{TARF}}
    \hspace{0.1in}
    \begin{minipage}{\linewidth}
    \includegraphics[scale=0.056]{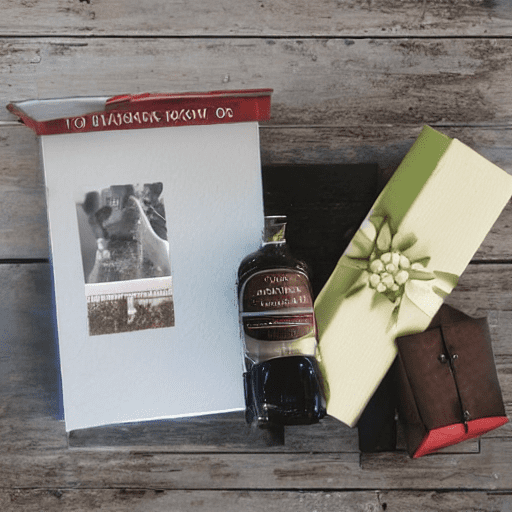}
    \includegraphics[scale=0.056]{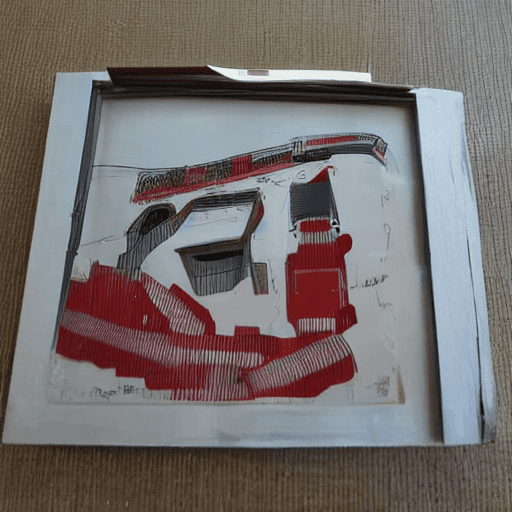}
    \includegraphics[scale=0.056]{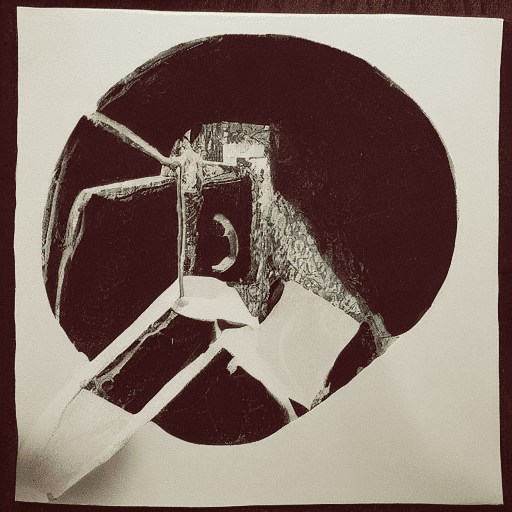}
    \includegraphics[scale=0.056]{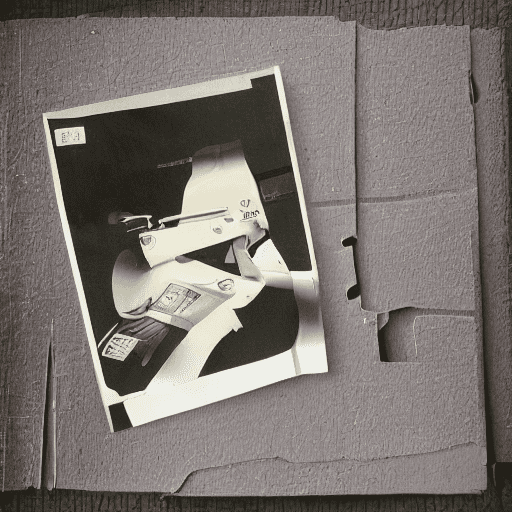}
    \\
    \includegraphics[scale=0.056]{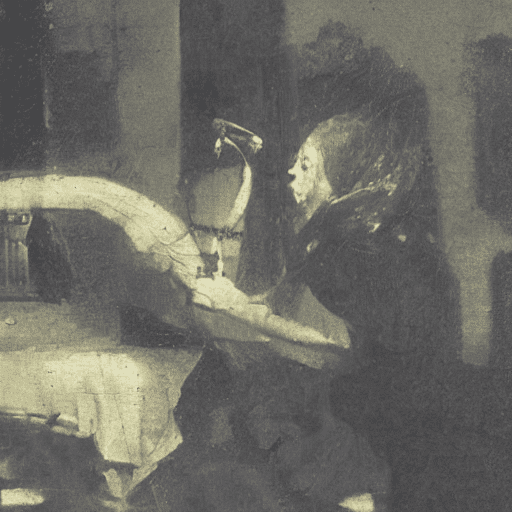}
    \includegraphics[scale=0.056]{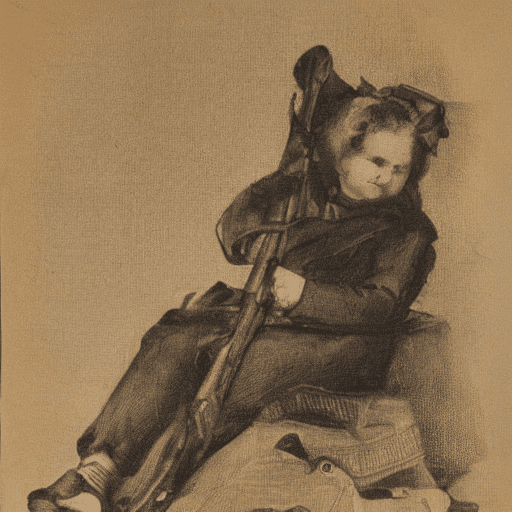}
    \includegraphics[scale=0.056]{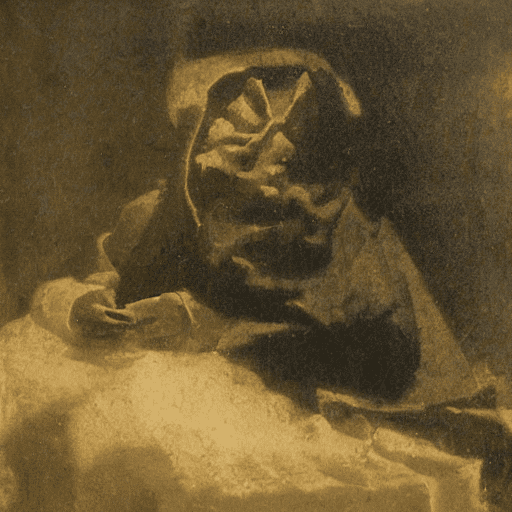}
    \includegraphics[scale=0.056]{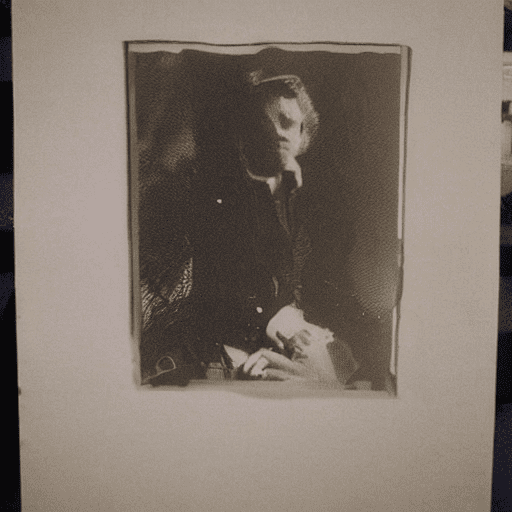}
    \end{minipage}
    }
    \vspace{-3mm}
    \end{center}
    \caption{Image generation results of original and unlearned stable diffusion. More results are in Tables~\ref{tab:img_tench} and~\ref{tab:img_english_springer}.
    }
    \vspace{-8mm}
    \label{fig:ablation_diffusion}
\end{wrapfigure}
We utilize Unlearning Accuracy (UA) to evaluate the accuracy of the unlearning targeted subset; Retaining Accuracy (RA) to evaluate the accuracy of the retaining subset; Testing Accuracy (TA) to evaluate the generalization ability of the model; Membership Inference Attack (MIA) to evaluate the efficacy of unlearning by the confidence-based predictor. All the above will be compared with that of the Retrained model and summarized in a "Gap" value (averaged gap with Retrained) to indicate the overall performance (the lower the better), and we also adopt Run-time Efficiency (RTE) to show the computational time cost.

\paragraph{Unlearning baselines.} For demonstrating the effectiveness of our methods in the four unlearning scenarios, we consider four representative baselines for comparison with the retrained model (Retrained), e.g., Finetuning (FT)~\cite{warnecke2021machine,goodfellow2016deep}, Random Labels (RL)~\cite{toneva2018empirical}, Gradient Ascent (GA)~\cite{ishida2020we}, Influence Unlearning (IU)~\cite{izzo2021approximate}, and also consider three recent advanced methods, e..g, Boundary Shrink (BS)~\cite{chen2023boundary}, $L_{1}$-sparse unlearning ($L_{1}$-sparse)~\cite{jia2023model}, Saliency Unlearning (SalUn)~\cite{fan2023salun}, and  SCalable Remembering and Unlearning unBound (SCRUB)~\cite{kurmanji2023towards}. All the methods are compared with the same trained models and the training data. We leave more details of baselines in Appendix~\ref{app:baseline_info}.

\subsection{Main Results}
\label{sec:exp_main}

In this part, we present the main comparison results In Table~\ref{tab:mu_main} with those considered baselines in the four unlearning tasks, evaluated with the four detailed metrics and the overall performance gap with retrained references. 

As the performance reference, all the retrained models (termed Retrained) are trained with the fully aligned retaining data. In general, we can find the previous unlearning methods achieved satisfactory performance on the conventional all matched forgetting, but did not perform well on the other three newly considered tasks with the label domain mismatch. Specifically, since the previous methods partially rely on forgetting data or remaining data, it results in ineffective or excessive forgetting due to the insufficient representation or decomposition missing. For example, FT can retain a similar RA with the Retrained but be less effective in forgetting, while GA reaches the lowest UA across different tasks but sacrifices too much model performance on the retaining dataset. In contrast, our TARF can consistently perform better (or comparable with the best method) through simultaneous gradient ascent and target-aware gradient descent to restrict the forgetting regions on the four unlearning tasks. In addition, we also extend our framework to unlearn the stable diffusion~\cite{ho2020denoising} in data mismatch forgetting, showing the efficacy of TARF, and full results are provided in Appendix~\ref{app:algo}.

\subsection{Ablations and Further Exploration}
\label{sec:exp_ablation}

In this part, we provide further exploration of the three class-wise unlearning tasks and conduct various ablation studies to characterize TARF. More results and discussion are provided in Appendix~\ref{app:exp}.

\begin{figure*}[t!]
    \begin{center}
    \includegraphics[scale=0.145]{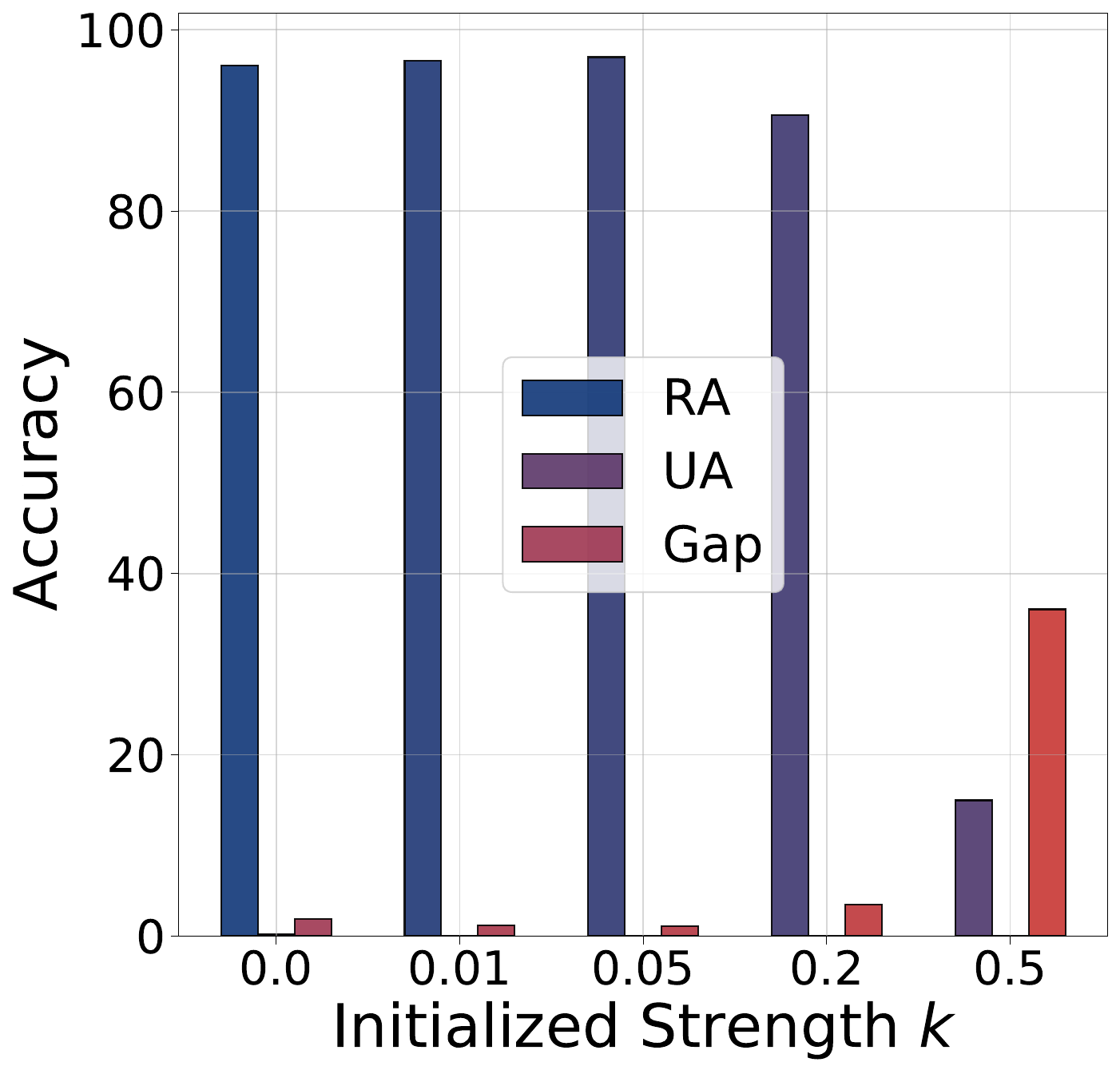}
    \includegraphics[scale=0.145]{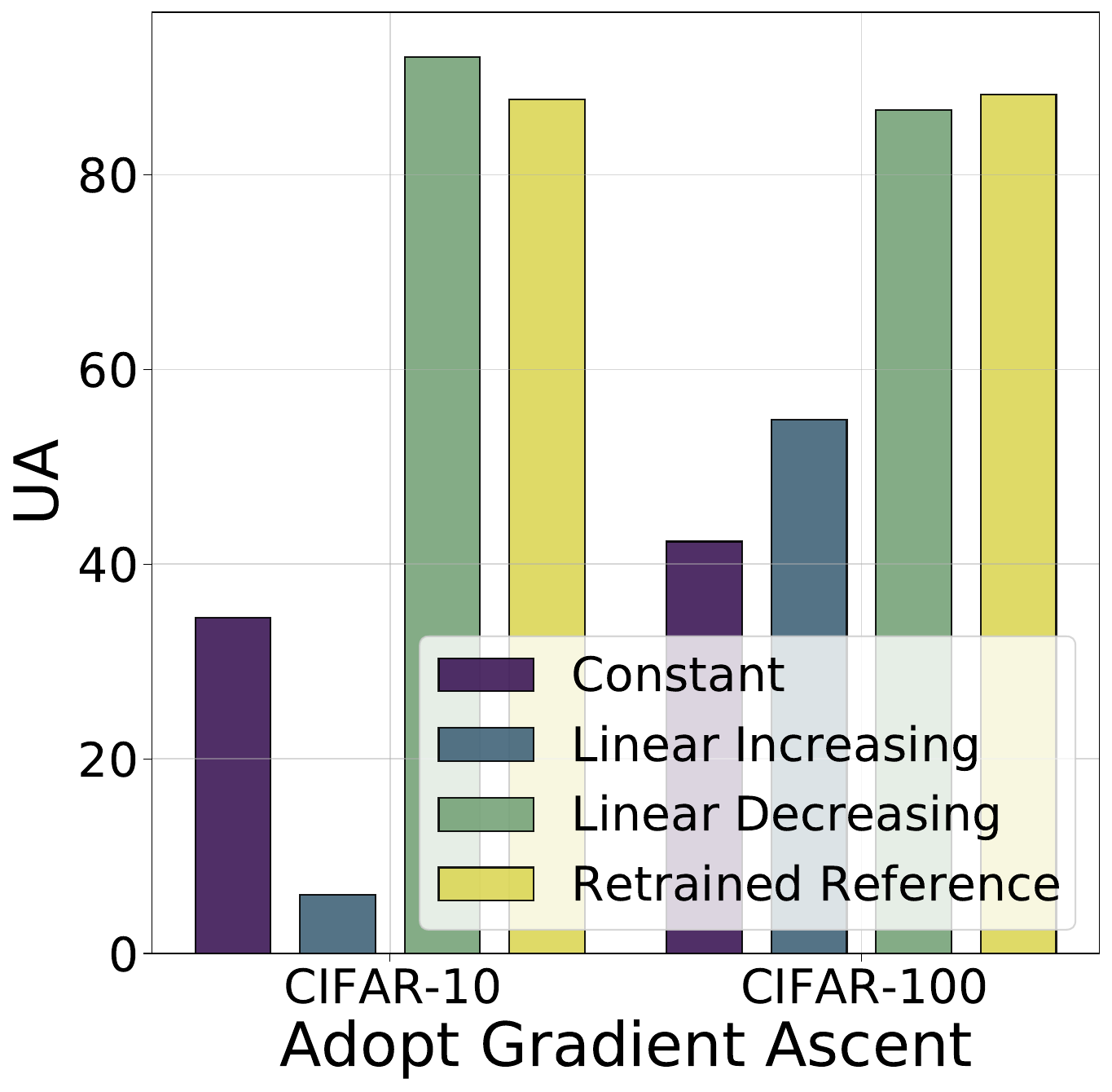}
    \includegraphics[scale=0.145]{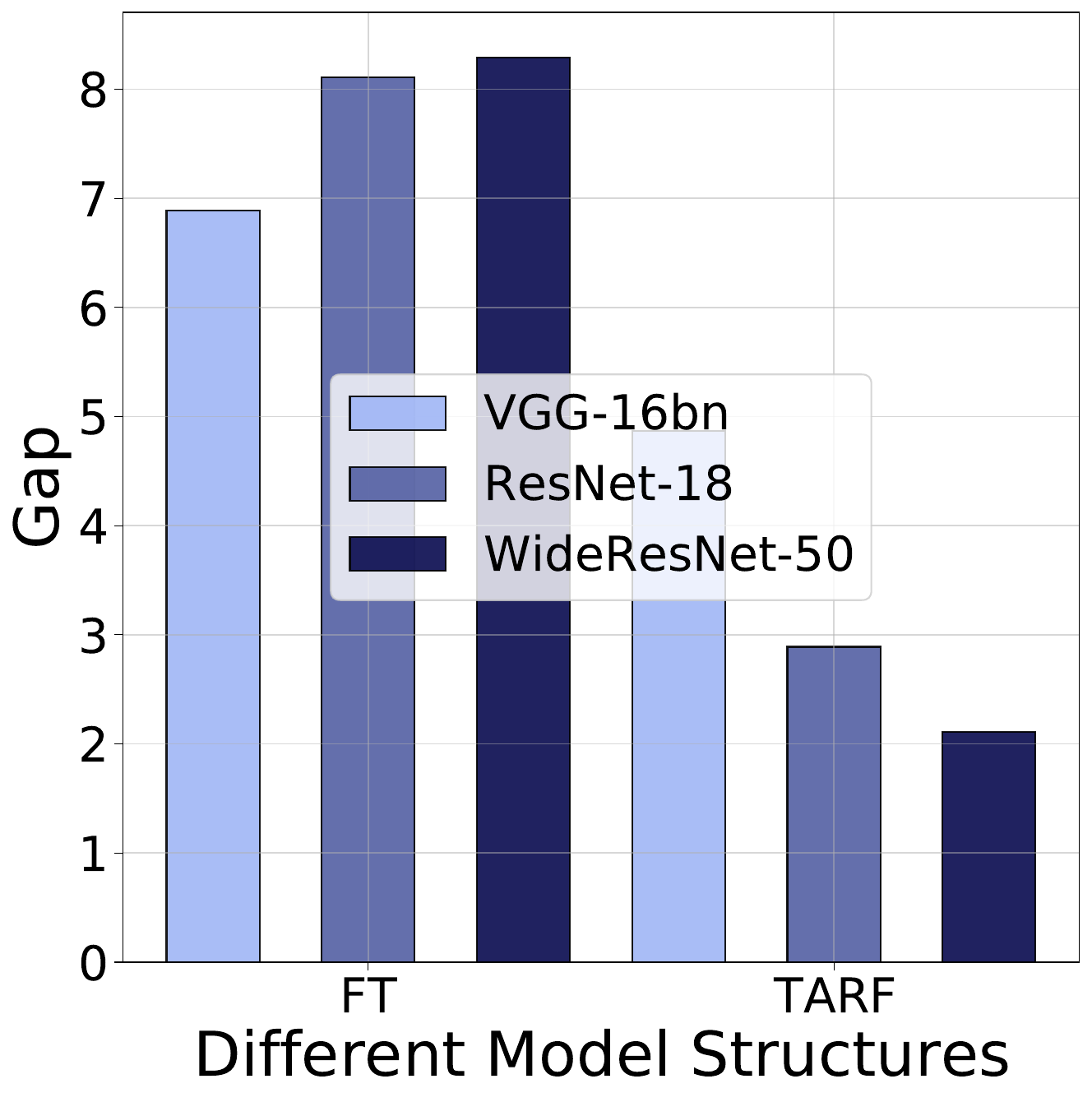}
    \includegraphics[scale=0.145]{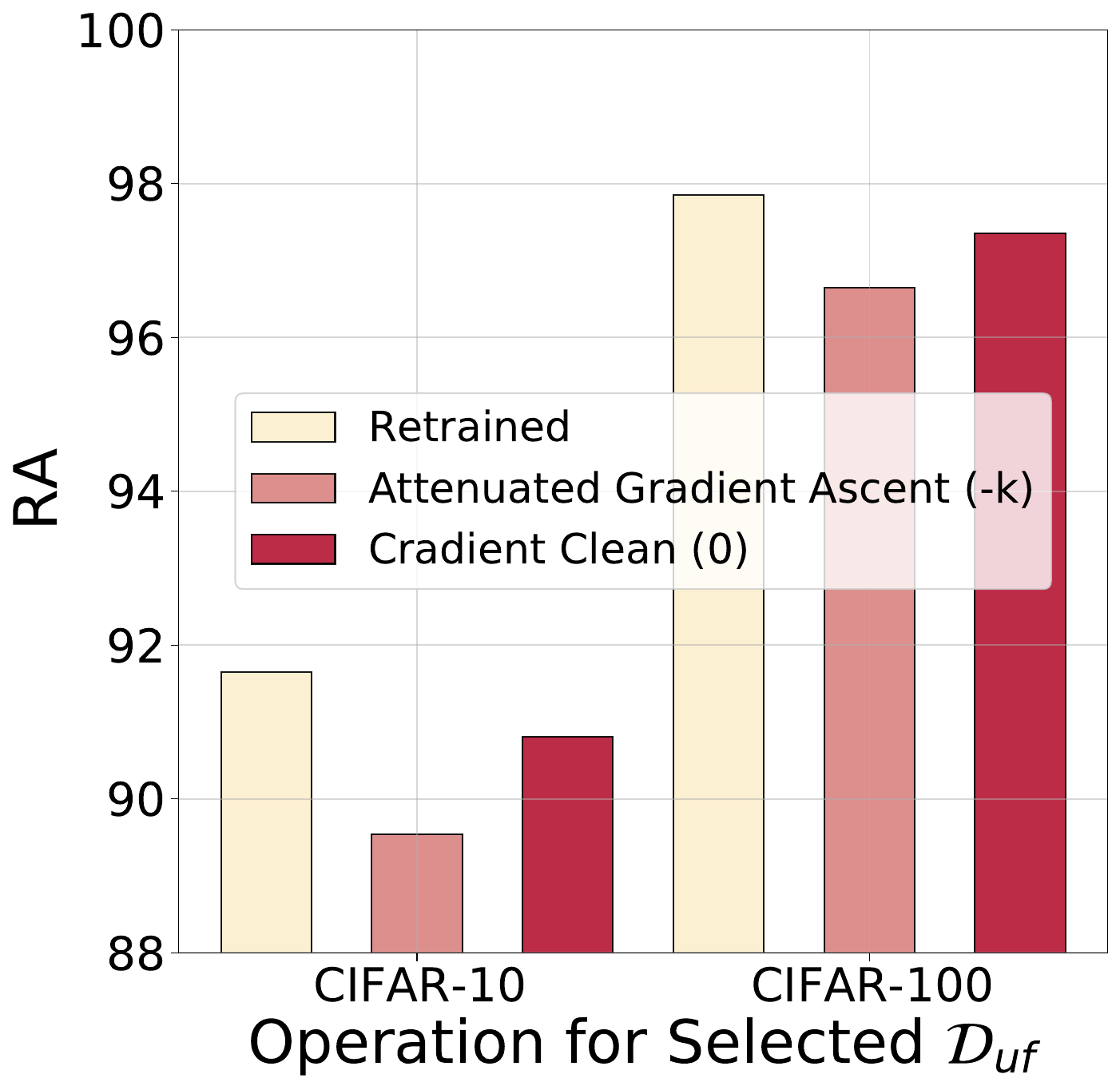}
    \end{center}
    \vspace{-3mm}
    \caption{Ablation studies: \textit{Left:} performance using different initialized $k$ on all matched forgetting; \textit{middle-left:} effects of constant or different dynamic gradient ascent controlled by $k(t)$; \textit{middle-right:} comparison of forgetting with different model structures; \textit{right:} comparison of using different operations on the selected forgetting data. More experimental details can refer to Appendix~\ref{app:exp}.
    }
    \vspace{-4mm}
    \label{fig:ablation}
\end{figure*}

\paragraph{Weighted control on annealed gradient ascent.} 
To analyze the annealed gradient ascent, we present the results on the left of Figure~\ref{fig:ablation} to show the effects of initialized strength $k$ on the all matched forgetting task using the CIFAR-100 dataset. The results show that an appropriate $k$ (e.g., about $0.05$) can help the model to achieve a satisfactory performance. However, the larger $k$ results in lower retaining performance and higher Gap value as the strength increases on the feature deconstruction.

\paragraph{Constant or dynamic gradient ascent for forgetting.} In the middle-left of Figure~\ref{fig:ablation}, we study whether we need the learning-rate-reduced $k$ for the forgetting part. Specifically, we compare it with using constant $k$ and learning-rate-increased $k$ on two model mismatch forgetting tasks. The results demonstrate that annealed gradient ascent can achieve more similar performance with the Retrained on forgetting data. The gradient ascent is considered simultaneously with gradient descent for restricting the forgetting region, while we adopt the annealed one since the unlearning target is to approximate the retrained model instead of continually maximizing the loss of forgetting data.

\paragraph{Unlearning on models trained by different structures.} 
In the middle-right of Figure~\ref{fig:ablation}, we investigate forgetting on the models pre-trained using different structures, e.g., ResNet-18~\cite{he2016deep}, VGG-16bn~\cite{simonyan2014very}, and WideResNet-50~\cite{zagoruyko2016wide}. The results of TARF on the model mismatch forgetting demonstrate that our TARF can achieve the lower performance gap than FT, evaluated with the retrained reference. With the increasing model capacity on the original training tasks, we can also find the model with a smaller capacity makes it harder to decompose the entangled feature representation.

\paragraph{Differernt operations on the selected forgetting data.} 
In the right of Figure~\ref{fig:ablation}, we present the ablation on the specific gradient operation on the selected forgetting data $\mathcal{D}_\text{uf}$. We compare using the gradient ascent ($-k(t)$) and cleaning ($0$) with the Retrained reference in target mismatch forgetting. Except for the similar forgetting efficacy achieved by the three trials, major differences exist in the performance evaluated by RA. The results show operation of gradient cleaning may be a better choice for $\mathcal{D}_\text{uf}$ to not deconstruct the feature representation too much and affect the retaining accuracy.

\section{Conclusion}
\label{sec:con}

In this work, we decouple the class label and target concept in class-wise unlearning. By introducing the label domain mismatch among forgetting data, model output, and target concept, we uncover three additional tasks beyond the conventional all matched forgetting, e.g., target mismatch, model mismatch, and data mismatch forgetting. We identify the insufficient representation and decomposition lacking of restrictively forgetting the target concept, and reveal the crucial forgetting dynamics in the representation level for the feasibility of these unlearning requests. Based on that, we propose the TARF that assigns an annealed gradient ascent on the identified forgetting data and the normal gradient descent on the selected retaining data. By collaboratively considering the forgetting/retaining target, TARF is more accurate in unlearning while maintaining the rest. We hope our work can provide new insights and draw more attention toward the practical scenarios of machine unlearning.


\bibliographystyle{plain}

\clearpage

\appendix

\section*{Appendixes}

The whole Appendix is structured in the following manner. In Appendix~\ref{app:repro_state}, we introduce the critical aspects of reproducibility. In Appendix~\ref{app:relate_work}, we provide a detailed discussion with related works of machine unlearning and other aspects. In Appendix~\ref{app:baseline_info}, we review the representative baseline methods in machine unlearning, which are considered in our experimental comparisons. In Appendix~\ref{app:whole_mismatch}, we introduce the complete scenarios considering the mismatch issues in machine unlearning, going beyond the four basic scenarios presented in the main text. 
In Appendix~\ref{app:algo}, we formally present the algorithm implementation of our proposed TARF with its variant, and further explanation of the rationality of TARF in unlearning. In Appendix~\ref{app:exp}, we provide additional experimental results to characterize forgetting dynamics and the properties of TARF. In Appendix~\ref{app:broader_limitation}, we discuss the potential broader impact and limitations of our work.

\section{Reproducibility Statement}
\label{app:repro_state}

Below we summarize  some critical aspects to facilitate reproducible results:

\begin{itemize}
    \item \textbf{Datasets.}  The datasets we used are all publicly accessible~\cite{krizhevsky2009learning_cifar10}, which is introduced in Section~\ref{sec:exp_setup}. For our newly introduced unlearning scenarios, the specific dataset construction is implemented as described in Section~\ref{sec:exp_setup} and Appendix~\ref{app:dataset_partition}.
    \item \textbf{Assumption.} Following the previous work~\cite{warnecke2021machine,jia2023model,fan2023salun}, We set our experiments to a tuning scenario where a well-trained model is available, and all the training samples are available but limited samples are labeled as "to be unlearned".
    \item \textbf{Baselines.} We provide the details about a series of unlearning methods considered in our work and also introduce the specific way of obtaining the pre-trained model for unlearning. 
    \item \textbf{Environment.} All experiments are conducted with multiple runs on NVIDIA Tesla V100-SXM2-32GB GPUs with Python 3.8 and PyTorch 1.8. 
\end{itemize}

\section{Discussion about Related Work}
\label{app:relate_work}

In this section, we discuss the related literature on machine unlearning, and provide more detailed comparisons of some work with their approaches and motivations. 

\subsection{Machine Unlearning}

Machine unlearning targets to adjust a trained model to scrub the data influence~\cite{koh2017understanding,shaik2023exploring,xu2023machine}. It is initially proposed to protect data privacy~\cite{cao2015towards,bourtoule2021machine,ginart2019making}, and a series of studies explore probabilistic methods through the differential privacy~\cite{ginart2019making,guo2018curriculumnet,neel2021descent,ullah2021machine,sekhari2021remember}. Except for the traditional classification model, recently some pioneering works also investigate unlearning for those generative models like diffusion model~\cite{gandikota2023erasing,zhang2023forgetmenot} or large language model~\cite{yao2023large, wang2024unlearning}. Although having achieved some promising progress, the strong algorithmic assumptions hinder the practical effectiveness~\cite{jia2023model}. Current research~\cite{golatkar2020eternal,thudi2022necessity,thudi2022unrolling,fan2023salun,chen2023boundary} focus more on developing more effective and efficient unlearning methods to approximate the Retrained model, with the given trained model. As for the assumption on data generation, prior works~\cite{golatkar2020eternal,warnecke2021machine,jia2023model,chen2023boundary} mainly consider all matched forgetting targets, with similar features on the original training tasks. As for the assumption on label generation, most prior works~\cite{bourtoule2021machine,graves2021amnesiac,thudi2022unrolling,jia2023model,fan2023salun,fan2024challenging} assume the accessibility on the fully identified forgetting dataset, and the complementary is the remaining dataset. One recent work~\cite{yoon2022few} considers unlearning with only a few forgetting samples but requires another generative model to generate approximated data. Our work considers a more practical scenario in which we can conduct mismatched forgetting and use limited identified forgetting data with the unlabeled remaining set. More details and the intuition of related baseline methods are introduced in Appendix~\ref{app:baseline_info}.

\subsection{Positive-unlabeled learning}

Positive-unlabeled learning~\cite{du2014analysis,menon2015learning} tries to learn a binary classifier from a few labeled positive samples with the rest unlabeled ones. A series of PU algorithms~\cite{liu2002partially,du2015modelling,du2015convex} are developed to train an accurate binary classifier, and can be roughly divided into two categories~\cite{bekker2020learning}. The first branch is cost-sensitive learning, which is related to importance weighting~\cite{liu2015classification}. Given the estimated class prior, these methods~\cite{du2015convex,kiryo2017positive,chen2020self} can develop an unbiased or consistent risk estimator for PU learning. Another branch of PU learning adopts two heuristic steps to perform binary classification. Such methods~\cite{liu2002partially,yu2004pebl} first identify reliable negative and positive examples from the unlabeled data, and then conduct semi-supervised learning. The model trained using cost-sensitive learning can also be a recognizer for positive or negative samples~\cite{hsieh2019classification}. Different from PU learning focusing on binary classification tasks, our work tries to enable more practical scenarios in class-wise unlearning~\cite{shaik2023exploring} where the class labels and target concepts are decoupled and we consider the issues of label domain mismatch.

\section{Details about Considered Baselines and Metrics}
\label{app:baseline_info}

In this section, we provide details about the considered representative baselines for machine unlearning methods, as well as their general intuitions with specific objectives. For the specific hyperparameters adopted in different methods,  we keep the same setting with previous related works~\cite{jia2023model,fan2023salun}, and the specific values will be listed in detail in our source codes.

\paragraph{Finetune (FT).} 
Utilizing the catastrophic forgetting~\cite{kirkpatrick2017overcoming} in the model (e.g., existed in the continual learning), FT~\cite{warnecke2021machine} fine-tunes the given trained model partially on $\mathcal{D}_\text{r}$ with few training epochs to obtain the $\theta_\text{un}^*$ with the following objective function,
\begin{equation}
    L_\text{FT} = \frac{1}{|\mathcal{D}_\text{r}|}\sum_{(x,y)\sim \mathcal{D}_\text{r}}\ell(f(x),y).
\end{equation} 

\paragraph{Gradient Ascent (GA).} Different from the normal gradient descent, GA reverses the gradient signal on $\mathcal{D}_\text{f}$ to conduct maximization with ascended gradients, resulting in the increasing loss of the forgetting data to obtain the $\theta_\text{un}^*$. The objective is given as follows,
\begin{equation}
    L_\text{GA} = -\frac{1}{|\mathcal{D}_\text{f}|}\sum_{(x,y)\sim \mathcal{D}_\text{f}}\ell(f(x),y).
\end{equation}
With reverse optimization to maximize the loss on the specific data, the model can approximate $\theta^*$ by directly forgetting the learned knowledge represented by the forgetting data.

\paragraph{Random Label (RL).} Similar to GA, RL~\cite{golatkar2020eternal} assign the random labels $Y^*$ on the forgetting data in $\mathcal{D}_\text{f}$ and fine-tune the given model with it to obtain the unlearned model $\theta_\text{un}^*$,
\begin{equation}
    L_\text{RL} = \frac{1}{|\mathcal{D}_\text{f}|}\sum_{(x,y)\sim \mathcal{D}_\text{f}}\ell(f(x),y^*).
\end{equation}
Instead of using the original training label on the forgetting data in $\mathcal{D}_\text{f}$, RL can destroy the learned feature by using the random label $y^*$ on $\mathcal{D}_\text{f}$, which violate the minimized loss value.

\paragraph{Influence Unlearning (IU).} IU adopts the influence function~\cite{koh2017understanding} to estimate the change if the training point is removed from the training loss. It is designed for random data unlearning~\cite{shaik2023exploring} with the provable guarantee on the unlearning effects. In general, IU estimates the change in model parameters of $\theta_\text{un}^*-\theta$ and adds the weight perturbation to the given model to obtain the unlearned one. However, it usually requires additional model information and training assumptions for the theoretical guarantee and may suffer hyperparameter tuning with inaccurate hessian estimation~\cite{jia2023model,fan2023salun}. 

\paragraph{Boundary Shrink (BS).} BS~\cite{chen2023boundary} is recently proposed for class-wise unlearning, especially on the all matched forgetting. It focuses on the decision spaces~\cite{goodfellow2016deep} of the given trained model. The critical idea is to shift the original decision boundary to imitate the decision behavior of the model retrained from scratch. Motivated by adversarial attacks~\cite{madry2017towards}, it proposes a neighbor searching method to identify the nearest but incorrect class labels $y_\text{near}$ for $\mathcal{D}_\text{f}$ to guide the model to unlearn the existing class and shift the decision boundary. Using the adversarial attack to find the nearest incorrect label, the objective of BS can be formulated as follows,
\begin{equation}
    L_\text{BS} = \frac{1}{|\mathcal{D}_\text{f}|}\sum_{(x,y)\sim \mathcal{D}_\text{f}}\ell(f(x),y_\text{near}),
\end{equation}
where $y_\text{near}$ is obtained by first perturbing the forgetting data and getting the newly predicted result as,
\begin{align}
    \begin{split}
        x' = x &+ \epsilon\cdot\text{sign}(\nabla\ell(f(x),y))\\
        y_\text{near} &\leftarrow \text{softmax}(f(x'))
    \end{split}
\end{align}

\paragraph{$L_1$-sparse.} 
Developed based on the conventional FT, $L_1$-sparse~\cite{jia2023model} investigate the model sparsity on machine unlearning. It figures out that model sparsification can benefit the unlearning performance on different perspectives via first pruning and then conducting unlearning. By carrying out pruning and unlearning simultaneously, $L_1$-sparse proposes the sparsity-aware unlearning utilizing the $L_1$ norm-based penalty. The objective is as follows with a hyperparameter $\gamma$,
\begin{equation}
    L_\text{$L_{1}$-sparse}=\frac{1}{|\mathcal{D}_\text{r}|}\sum_{(x,y)\sim \mathcal{D}_\text{r}}\ell(f(x),y) + \gamma||\theta^*||,
\end{equation}
and the general sparsity-aware penalty can also be added to different unlearning methods. In this work, we mainly compare the $L_1$-sparse FT as the previous work~\cite{jia2023model,fan2023salun} considered.

\paragraph{SalUn.} 
With the concern on unlearning stability and cross-domain applicability, SalUn~\cite{fan2023salun} introduces the concept of weight saliency in machine unlearning. This innovation directs the attention of unlearning into specific model weights for specific data that need to be unlearned. In general, it first generates the gradient-based weight saliency map inspired by model sparsification~\cite{jia2023model} with gradient-value thresholding, where the specific generation method is defined as,
\begin{equation}
    {\rm m}_s = \mathbf{1}(|\nabla_\theta\ell(\theta;\mathcal{D}_f)|_\theta=\theta_o|\geq\gamma),\quad \theta_u={\rm m}_s\odot(\delta\theta+\theta_o)+(1-{\rm m})\odot\theta_o,
\end{equation}
in which $\mathbf{1}(g \geq \gamma)$ is an element-wise indicator function that yields a value of $\mathbf{1}$ for the i-th element if  and 0 otherwise, $|\cdot|$ is an element-wise absolute value operation, and $ \gamma> 0$ is a hard threshold.
and then conducts saliency-based unlearning using the generated saliency map. Specifically, SalUn adopts RL~\cite{golatkar2020eternal} to fine-tune the forgetting data in $\mathcal{D}_\text{f}$ on the salience map, and the extended objective is given as follows, 
\begin{equation}
    L_\text{SalUn} = \frac{1}{|\mathcal{D}_\text{f}|}\sum_{(x,y)\sim \mathcal{D}_\text{f}}\ell_{\theta_u}(f(x),y^*)+\alpha\frac{1}{|\mathcal{D}_\text{r}|}\sum_{(x,y)\sim\mathcal{D}_\text{r}}\ell(f(x),y),
\end{equation}
More detailed operations can refer to~\cite{fan2023salun}, and we keep the same hyperparameter used in~\cite{fan2023salun} to conduct the class-wise unlearning tasks.

\paragraph{SCRUB.} SCRUB is a newly proposed unlearning algorithm based on a novel casting of the problem into a teacher-student framework~\cite{kurmanji2023towards}. It is designed to meet the desiderata of unlearning: efficiently forgetting without hurting the model utility. As the general target of SCRUB in forgetting is application-dependent, it is proposed with a recipe that works across applications: SCRUB is first to strive for maximal forget error, which is desirable in some scenarios like removing bias or restricted contents but not in others like user privacy protection. To address the latter case, SCRUB is integrated with a rewinding procedure that can reduce the forget set error appropriately when required. 

Given the original model $\theta^o$ as the teacher model, the goal of SCRUB is formatting as training a student model $\theta^u$ that selectively obeys the teacher. The overall objective can be divided into two folds, the first is to remember $\mathcal{D}_\text{r}$ under the teacher model's guide while the second is to forget $\mathcal{D}_\text{f}$ by disobeying the teacher model's guide. To measure the degree to which the student model obeys the teacher model, SCRUB utilizes the following distance measure,
\begin{equation}
    d(x;\theta^u)=D_{\rm KL}(p(f(x;\theta^o))||p(f(x;\theta^u))),
\end{equation}
where $D_{\rm KL}$ is the KL-divergence and the overall measures of the distance between the student model's and teacher model's prediction distribution. With the aforementioned distance, the objective of SCRUB is as follows,
\begin{equation}
    L_\text{SCRUB} =\min_{\theta^u}\frac{\alpha}{N_r}\sum_{x_r\in\mathcal{D}_r} d(x_r;\theta^u) + \frac{\gamma}{N_r}\sum_{(x_r,y_r)\in\mathcal{D}_r} \ell(f(x_r;\theta^u),y_r) - \frac{1}{N_f}\sum_{x_f\in\mathcal{D}_f}d(x_f;\theta^u),
\end{equation}
where the first two parts can be regarded as a variant of distillation from a teacher model on $\mathcal{D}_\text{r}$ and the third part is encouraging the student model to disobey the teacher model to forget the target data.



\section{Full Discussion about Label Domain Mismatch}
\label{app:whole_mismatch}

In this section, we discuss the full scenarios of label domain mismatch in class-wise unlearning~\cite{shaik2023exploring, warnecke2021machine,golatkar2020eternal,chen2023boundary,jia2023model,fan2023salun}. Specifically, we will start by why focusing on class-wise unlearning, and then discuss the motivation for investigating its label domain mismatch, with the newly introduced setting being friendly for empirical analysis and further research. Finally, we provide detailed information on our instantiated four tasks using the benchmarked CIFAR-10 and CIFAR-100 datasets~\cite{krizhevsky2009learning_cifar10}.

\begin{figure*}[t!]
\begin{center}
\hspace{-0.18in}
\includegraphics[scale=0.122]{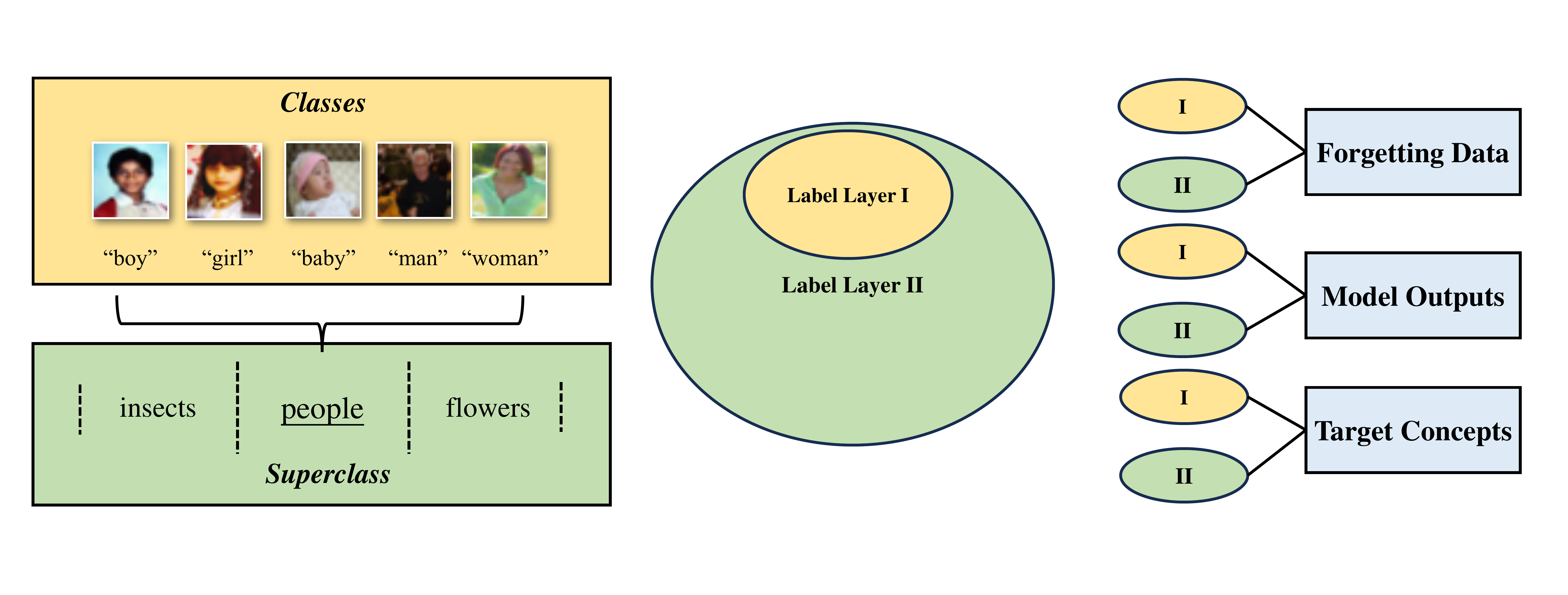}
\end{center}
\vspace{-6mm}
\footnotesize{The left panel shows an example of two-layer label domains; The middle panel is the Venn diagram to show the hierarchical relation; The right panel illustrates the potentials of three critical class-wise unlearning aspects.}
\caption{\textbf{Label domain mismatch with the two-layer illustration.}}
\label{fig: app_two_layer}
\end{figure*}

\begin{table}[t!]
    \centering
    \caption{Mismatching in the label domain of three critical aspects with a two-layer label structure.}
    \vspace{2mm}
  \begin{tabular}{c|c|c|c|c}
    \toprule[1.5pt]
    No. & Forgetting data & Model output & Target concept & Comment \\
    \midrule[0.6pt]
    1 & Class label & Class label & Class label & \textbf{All matched}\\
    2 & Class label & Class label & Superclass & \textbf{Target mismatch}\\
    3 & Class label & Superclass & Class label & \textbf{Model mismatch} \\
    4 & Class label & Superclass & Superclass & \textbf{Data mismatch} \\
    \midrule[0.6pt]
    5 & Superclass & Class label & Class label & Impractical since $\mathcal{L}_D\succ\mathcal{L}_T$ \\
    6 & Superclass & Class label & Superclass & Similar to all matched \\
    7 & Superclass & Superclass & Class label & Impractical since $\mathcal{L}_D\succ\mathcal{L}_T$\\
    8 & Superclass & Superclass & Superclass & \textbf{All matched}\\
    \bottomrule[1.5pt]
  \end{tabular}
    \label{tab:app_exp_set_two_layer}
\end{table}

To begin with, machine unlearning~\cite{cao2015towards,thudi2022necessity,xu2023machine,shaik2023exploring} is originally proposed in response to "the right to be forgotten" to protect the data privacy, and recently deep machine unlearning is a timely research topic associated with foundation models which use massive of data to train~\cite{kurmanji2023towards,bommasani2021opportunities}. The ensuing data regulation concerns have also expanded the original privacy-protecting goal to more general needs and scenarios~\cite{yao2023large,maini2024tofu,gandikota2023erasing}. As stated in~\cite{shah2023unlearning,kurmanji2023towards,jia2023model}, unlearning a subset of the training set has received increasing attention (like removing sensitive information, and inappropriate content). However, the previous scenarios mainly consider the coinciding class labels with the target concept to be unlearned. Although achieving promising results in forgetting, it is still not enough in practice.

\begin{wraptable}{r}{0.48\textwidth}
  \centering
  \vspace{-4mm}
  \resizebox{\linewidth}{!}{
  \begin{tabular}{c|c}
    \toprule[1.5pt]
    \textit{Label Domain $\mathcal{L}$} & Relation of Data $\mathcal{L}_D$, Model $\mathcal{L}_M$, and Target $\mathcal{L}_T$ \\
    \midrule[0.6pt]
    All matched & $\mathcal{L}_D \quad = \quad \mathcal{L}_T \quad = \quad \mathcal{L}_M$ \\
    Target mismatch & $\mathcal{L}_M \quad = \quad \mathcal{L}_D \quad \prec\quad \mathcal{L}_T$ \\
    Model mismatch & $\mathcal{L}_D \quad= \quad\mathcal{L}_T \quad\prec\quad \mathcal{L}_M$ \\
    Data mismatch & $\mathcal{L}_D \quad\prec\quad \mathcal{L}_T \quad=\quad \mathcal{L}_M$ \\
    \bottomrule[1.5pt]
  \end{tabular}
  }
  \caption{considering \textbf{label domain} relations of three critical aspects in class-wise unlearning.}
  \vspace{-4mm}
  \label{tab:mismatch}
\end{wraptable}
Considering the problem setups of unlearning, we have three critical aspects, e.g., the well-trained machine learning model $\theta$, and the reported data $\mathcal{D}_\text{f}$ to be unlearned, as well as the target concept. In previous studies, the three aspects are mainly considered to be under the same label taxonomy. In other words, the unlearning tasks are aligned with the pre-training task, where the latter trains a multi-class classification model, and the former aims to unlearn a training class. However, in practice, the unlearning request may violate the taxonomy of the pre-training tasks, while the specific target concepts always exhibit a unified property for specific forgetting data. It naturally motivates us to consider different label domains of the three aspects of unlearning. As listed in Table~\ref{tab:mismatch}, the label domain of data $\mathcal{L}_D$, the label domain of model output $\mathcal{L}_M$, and the label domain of target concept $\mathcal{L}_T$. To begin with, we have a practical assumption on the relation between label domains of forgetting data and target concept, i.e., $\mathcal{L}_D\preceq\mathcal{L}_T$, indicating that the reported forgetting data should be included in the target concept (as intuitively illustrated in the middle panel of Figure~\ref{fig: app_two_layer}). Considering $\mathcal{L}_D=\mathcal{L}_T$, we can have two possibilities on $\mathcal{L}_M$, e.g., $\mathcal{L}_M=\mathcal{L}_T$ and $\mathcal{L}_M\neq\mathcal{L}_T$, where the former is regarded as all matched when $\mathcal{L}_D=\mathcal{L}_M=\mathcal{L}_T$ and the latter is the model mismatch. To be more specific, we consider model mismatch forgetting as $\mathcal{L}_D=\mathcal{L}_T\prec\mathcal{L}_M$, since $\mathcal{L}_M\prec\mathcal{L}_T$ will have no additional effects on the unlearning when $\mathcal{L}_D=\mathcal{L}_T$ and we can regard it as similar to the all matched case. Considering $\mathcal{L}_D\prec\mathcal{L}_T$, we can have the target mismatch forgetting when $\mathcal{L}_D=\mathcal{L}_M$ and data mismatch forgetting when $\mathcal{L}_M=\mathcal{L}_T$.

We summarize the mainly considered mismatch cases in Table~\ref{tab:mismatch}, which can serve as a general reference for further research on constructing the unlearning tasks. In the following, we further explain the procedure of task instantiating and discuss the other potential scenarios with the typical two-layer label structure considered in the main text and an additional three-layer label structure.

\subsection{A two-layer label structure of mismatch}

In Figure~\ref{fig: app_two_layer}, we first show the illustration of a two-layer label structure and the three aspects of unlearning, i.e., forgetting data, model outputs, and target concept. Without losing generality, we utilize the class labels and superclass information (refer to the official information in CIFAR-100~\cite{krizhevsky2009learning_cifar10}) for consideration. Then we have a two-layer label structure representing different knowledge regions.

Given two potential label domains in each aspect, we can totally get the 8 scenarios list in Table~\ref{tab:app_exp_set_two_layer}. The first 4 scenarios are mainly considered and detailedly introduced in the main text. For the rest 4 scenarios (i.e., No. 5-8), we consider some (i.e., No. 5 and No. 7) to be impractical as the label domain of forgetting data is larger than the target concept, which means that the unlearning requests identify more forgetting data than the true target concept. It should be more reasonable that only limited forgetting data are identified by server users or internal examiner~\cite{kovashka2016crowdsourcing} in real-world applications. Therefore, we mainly consider the forgetting data $\mathcal{D}_\text{f}$ belongs a part of or equals to the overall data $\mathcal{D}_\text{t}$ of the target concept. As for No. 6 and No. 8 cases, the former is similar to the conventional all matched forgetting since the forgetting data has the same label domains with the target concept while the model output has a fine-grained label domain (e.g., class label) that will not affect the unlearning, and the latter is exactly same as the all matched forgetting considered in No.1.

\subsection{A three-layer label structure of mismatch}

\begin{figure*}[t!]
\begin{center}
\hspace{-0.18in}
\includegraphics[scale=0.122]{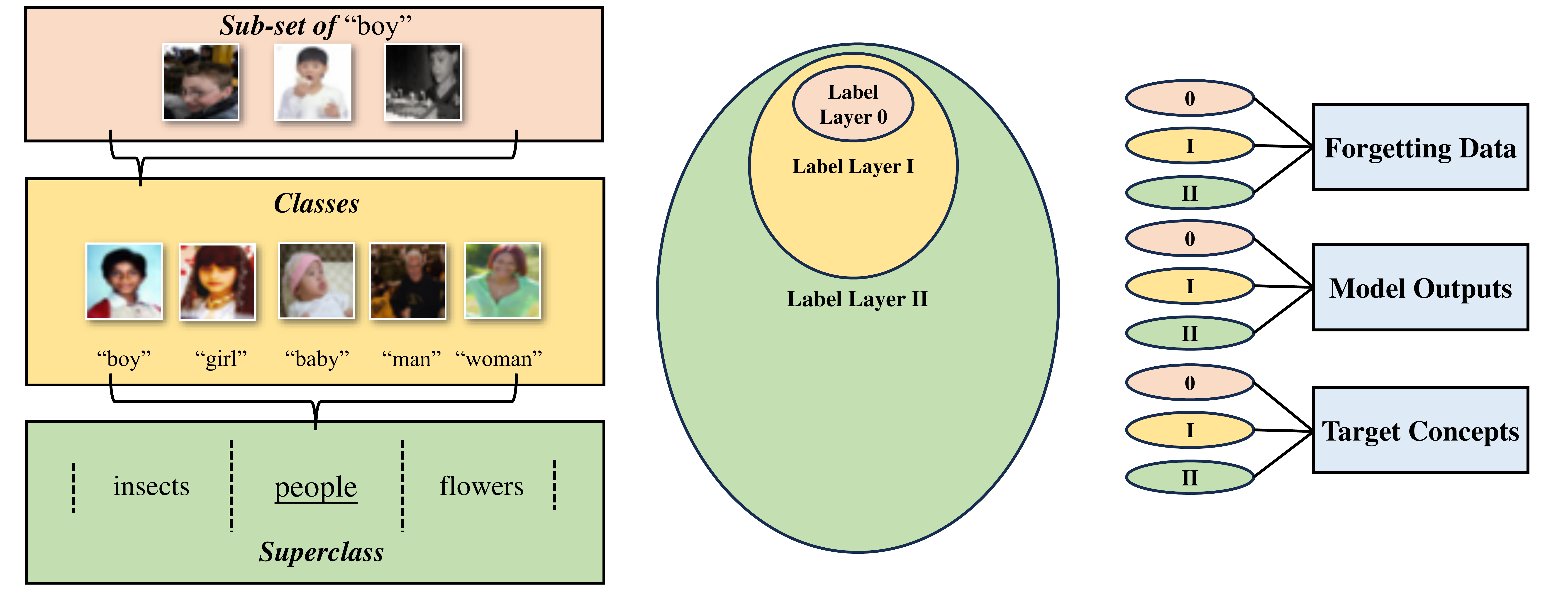}
\end{center}
\footnotesize{The left panel shows an example of three-layer label domains extended from the ordinary setting considered in our main text, where the sub-set is sampled from the "boy" class; The middle panel is the Venn diagram to show the hierarchical relation; The right panel illustrates the potentials of three critical class-wise unlearning aspects.}
\caption{\textbf{Label domain mismatch with the three-layer illustration.}}
\label{fig: app_three_layer}
\vspace{-4mm}
\end{figure*}

Since in more extreme cases, some unlearning requests would exhibit only several instances of forgetting an abstract concept not aligned with the pre-training tasks. We then consider an extra label layer (e.g., the sub-set level inside a class) to construct a three-layer structure beyond the previous one. In Figure~\ref{fig: app_three_layer}, we illustrate it with some samples and a Venn diagram.

Considering each aspect can have three potential label domains, we can totally get 27 scenarios in Table~\ref{tab:app_exp_set_three_layer}. In general, we have three rough categories for analysis. First, due to the aforementioned constraint that the target concept should include the forgetting data, we consider several cases (e.g., No. 14, 17, 18, 20, 22-24, and 26-27) to be impractical. Second, the three-layer structure also includes a group of scenarios that also existed in the two-layer structure, so No. 1-4 and 19 are the same as the four scenarios (i.e., all matched, target mismatch, model mismatch, and data mismatch). Third, for the rest scenarios, we regarded them to be novel cases than those considered in the main text.

To be more specific, there are two groups of cases in the third part. For No. 5, 6, and 7, since they also can be represented using a two-layer structure, the forgetting dynamics are similar to that in target, model, and data mismatch forgetting. By contrast, in No. 8, 9, and 16, all three label domains exist in the three aspects of class-wise unlearning, which is worthy of further discussion.

\begin{table}[t!]
    \centering
    \caption{Mismatching in the label domain of three critical aspects with a three-layer label structure.}
    \vspace{2mm}
  \resizebox{0.8\linewidth}{!}{
  \begin{tabular}{c|c|c|c|c}
    \toprule[1.5pt]
    No. & Forgetting data & Model output & Target concept & Comment \\
    \midrule[0.6pt]
    1 & Sub-set & Sub-set & Sub-set & \textbf{All matched}\\
    2 & Sub-set & Sub-set & Class label & \textbf{Target mismatch}\\
    3 & Sub-set & Class label & Sub-set & \textbf{Model mismatch} \\
    4 & Sub-set & Class label & Class label & \textbf{Data mismatch} \\
    5 & Sub-set & Sub-set & Superclass & \gray Different \\
    6 & Sub-set & Superclass & Sub-set & \gray Different \\
    7 & Sub-set & Superclass & Superclass & \gray Different\\
    8 & Sub-set & Class label & Superclass & \gray Different\\
    9 & Sub-set & Superclass & Class label & \gray Different\\
    \midrule[0.6pt]
    10 & Class label & Class label & Class label & \textbf{All matched}\\
    11 & Class label & Class label & Superclass & \textbf{Target mismatch}\\
    12 & Class label & Superclass & Class label & \textbf{Model mismatch} \\
    13 & Class label & Superclass & Superclass & \textbf{Data mismatch} \\
    14 & Class label & Sub-set & Sub-set & Impractical since $\mathcal{L}_D\succ\mathcal{L}_T$ \\
    15 & Class label & Sub-set & Class label & Similar to all matched \\
    16 & Class label & Sub-set & Superclass & \gray Different \\
    17 & Class label & Class label & Sub-set & Impractical since $\mathcal{L}_D\succ\mathcal{L}_T$ \\
    18 & Class label & Superclass & Sub-set & Impractical since $\mathcal{L}_D\succ\mathcal{L}_T$ \\
    \midrule[0.6pt]
    19 & Superclass & Superclass & Superclass & \textbf{All matched} \\
    20 & Superclass & Class label & Class label & Impractical since $\mathcal{L}_D\succ\mathcal{L}_T$\\
    21 & Superclass & Class label & Superclass & Similar to all matched \\
    22 & Superclass & Superclass & Class label & Impractical since $\mathcal{L}_D\succ\mathcal{L}_T$ \\
    23 & Superclass & Sub-set & Sub-set & Impractical since $\mathcal{L}_D\succ\mathcal{L}_T$ \\
    24 & Superclass & Sub-set & Class label & Impractical since $\mathcal{L}_D\succ\mathcal{L}_T$ \\
    25 & Superclass & Sub-set & Superclass & Similar to all matched \\
    26 & Superclass & Class label & Sub-set & Impractical since $\mathcal{L}_D\succ\mathcal{L}_T$ \\
    27 & Superclass & Superclass & Sub-set & Impractical since $\mathcal{L}_D\succ\mathcal{L}_T$ \\
    \bottomrule[1.5pt]
  \end{tabular}
  }
    \label{tab:app_exp_set_three_layer}
\end{table}

\begin{figure*}[t!]
\begin{center}
\includegraphics[scale=0.12]{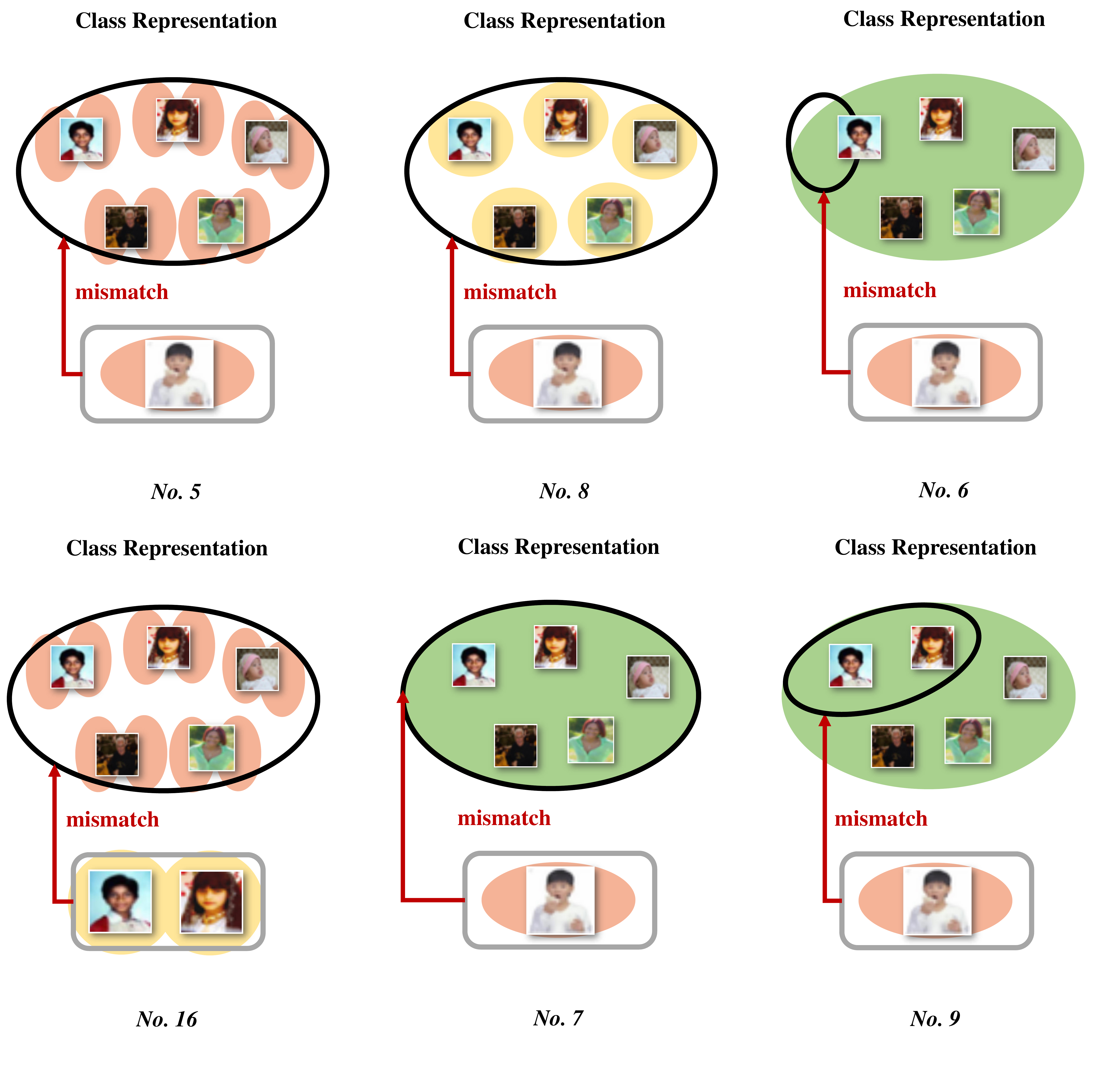}
\end{center}
\vspace{-4mm}
\caption{Illustration of 6 scenarios different from the four major tasks according to Table~\ref{tab:app_exp_set_three_layer}.}
\label{fig: app_additional_six}
\end{figure*}

\subsection{Further exploration on the other 6 different scenarios}
\label{app:other_scenarios}

In this part, we further discuss the 6 different scenarios discovered by constructing the three-layer label structure. We illustrated these forgetting tasks in Figure~\ref{fig: app_additional_six} and discuss them as follows,

- \textbf{No. 5\&16} In the two scenarios, the model output has the most fine-grained label domain (e.g., sub-set as illustrated in Figure~\ref{fig: app_three_layer}) for representation. At the same time, the target concept is broader than both model output and identified forgetting data. Different from the aforementioned target mismatch, the mismatch degree of this task is larger (e.g., superclass level) than the previous one (e.g., class level). In other words, the model output further loses the entanglement of feature representation of the samples belonging to target concept (compared with the original setups of target mismatch).
To simulate the case, we employ the same model pre-trained by class in target mismatch, but enlarge the target concept (consists of 7 classes with similar semantic features, instead of the original 5) and change the forgetting data (2 class as the given forgetting data in No.5 and 3 classes in No.16).

- \textbf{No. 8\&7.} Similar to the previous No. 5, the target concept in these tasks is also broader than the label domains of the identified forgetting data. However, in these two scenarios, the model output is varied which controls the entanglement of target samples. To construct these two forgetting tasks, we respectively adopt the models pre-trained by class labels and superclass, and use the same forgetting data (with 1 class) to investigate the performance change using our TARF and other baselines.

- \textbf{No. 6\&9.} In the last two scenarios, the forgetting tasks are more similar to the previous model mismatch forgetting. However, the distinguishable difference is that the label domain of model outputs can be much different from the identified forgetting data. In the No. 6 task, 
the target concept is aligned with the identified forgetting data, while since the remaining data is more than the original model mismatch forgetting, the task separation could be harder than the previous. In the No. 9 task, we can find that it is a complex scenario where the target concept is broader than the forgetting data but included in the same superclass. In both tasks, we use 1 class data as the forgetting data.

To further understand the properties of unlearning in these tasks, we conducted additional experiments and summarized the results in Table~\ref{tab:mu_main_app_other_scenarios}. We can find the empirical results well demonstrate the conceptual conjectures in the previous discussion, and the representative baselines exhibit varied performance gap with the Retrained reference. Among them, our TARF can consistently achieve the better performance regarding to the Gap.

\begin{table*}[t]
    \caption{Results (\%) of unlearning with different model structures. All methods are trained on the same backbone, i.e., the basis of unlearning initialization is the same (except for retraining from scratch). Values are percentages. Bold numbers are superior results. $\downarrow$ indicates smaller are better. }
    \vspace{2mm}
    \centering
    \footnotesize
    \renewcommand\arraystretch{1.0}
    \resizebox{\textwidth}{!}{
    \begin{tabular}{l|c|cccc|c|c|cccc|c}
        \toprule[1.5pt]
         \textbf{CIFAR-100} &Metric & UA & RA & TA  & MIA & Gap$\downarrow$ & Metric &  UA & RA & TA  & MIA & Gap$\downarrow$ \\
        \midrule[0.6pt]
        
        \gray Retrained
        & \multirow{5}*{\textbf{No. 5}} & \gray0.00 & \gray97.85 & \gray73.72 & \gray100.00 & \gray- & \multirow{5}*{\shortstack{\textbf{No. 16}}} & \gray0.00 & \gray97.85 & \gray73.72 & \gray100.00 & \gray- \\
        FT~\cite{warnecke2021machine}& &67.52& 96.43& 72.96& 41.14&32.72 & & 53.11&94.64 &71.23 &52.70 &26.53  \\
        RL~\cite{toneva2018empirical}& & 68.57& 96.12 & 72.58& 41.17& 33.15& &53.90 & 96.94& 73.07& 53.56& 25.48 \\
        GA~\cite{ishida2020we}&& 38.03 &97.00 &70.98 &76.92 &16.75  & &32.24 &95.73 &69.99&77.62 &15.12  \\
        \textbf{TARF} (ours) & &0.00 &96.58 & 72.03 &100.00 & \textbf{0.74}& & 0.00 &96.98 &72.87 &100.00 &\textbf{0.43} \\
        \midrule[0.6pt]
        \gray Retrained
        & \multirow{5}*{\textbf{No. 8}} & \gray0.00 & \gray97.85 & \gray73.72 & \gray100.00 & \gray- & \multirow{5}*{\shortstack{\textbf{No. 7}}} & \gray0.00 & \gray98.50 & \gray80.15 & \gray100.00 & \gray- \\
        FT~\cite{warnecke2021machine} & & 74.09& 97.19& 74.01&36.71& 34.58&&95.16 &94.98 &78.68 &13.06 &46.77  \\
        RL~\cite{toneva2018empirical}& &76.04 &96.76 &72.88 &36.00 & 35.49& & 91.51& 96.98&80.11 & 47.24&36.46  \\
        GA~\cite{ishida2020we}& & 49.47 &98.92 &72.94 & 77.96&  18.34& & 15.91 & 98.64 & 80.27 & 93.82 & 5.59 \\
        \textbf{TARF} (ours) & &  0.00 &96.22 &72.43 & 100.00 & \textbf{0.73} & &0.00 &96.54 &79.23 &100.00 &\textbf{0.65} \\
        \midrule[0.6pt]
        \gray Retrained
        & \multirow{5}*{\textbf{No. 6}} & \gray88.22 & \gray98.52 & \gray84.42 & \gray22.22 & \gray- & \multirow{5}*{\shortstack{\textbf{No. 9}}} & \gray88.22 & \gray98.58 & \gray78.50 & \gray25.78 & \gray- \\
        FT~\cite{warnecke2021machine} & & 94.33 &95.00 &78.77 &13.67 &5.96& & 91.78&95.02 &78.90 &18.44 &3.72  \\
        RL~\cite{toneva2018empirical}& &84.22 &96.96 &80.18 &65.77 &13.34& & 96.97& 70.22&80.24 & 94.67&26.94  \\
        GA~\cite{ishida2020we}& &18.44 &96.06 & 78.20&92.67 &37.23 &  &19.11 &95.27 &77.56&91.56 &34.79  \\
        \textbf{TARF} (ours) & &92.21 & 98.43 & 82.32 & 19.17 & \textbf{2.31} & &89.12 &  97.23 & 79.21 &24.32 &\textbf{1.11}  \\
        \bottomrule[1.5pt]
    \end{tabular}
    }
    \label{tab:mu_main_app_other_scenarios}
\end{table*}

\subsection{Specific Information of the instantiated Tasks}
\label{app:dataset_partition}

For the four major scenarios (i.e., conventional all matched forgetting, target mismatch forgetting, model mismatch forgetting, and data mismatch forgetting) considered in our work, we provide the dataset construction and partition details in this section. Note that we focus on class-wise unlearning in this work, which is different from random data forgetting that uniformly samples the forgetting target of all classes in the training dataset.

To ease the research investigation and empirical verification, we adopt the commonly used~\cite{kurmanji2023towards,jia2023model,fan2023salun,fan2024challenging} benchmark CIFAR-10 and CIFAR-100 for constructing the pre-training task for unlearning. Specifically, the official class labels are kept as classes for ordinary setup, and we provide the superclass information referring to the pre-defined lists~\cite{krizhevsky2009learning_cifar10} of CIFAR-100. Since there is no official superclass information for CIFAR-10 dataset, we manually grouped the classes of CIFAR-10 according to their semantic feature similarity and finalized 5 superclass clusters consisting of 2 classes in each. The full structured label layers information is summarized in Tables~\ref{tab:app_exp_set_cifar100} and~\ref{tab:app_exp_set_cifar10}. For all the unlearning scenarios where the label domain of model output is the superclass, we will first use the superclass information to train the 20-class and 5-class classification models respectively. For the specific data partition in unlearning requests, we randomly sampled two classes in CIFAR-100 and one class in CIFAR-10 as forgetting data and kept the setup across the four forgetting tasks as well as other experiments. For other additional experimental setups, we will state them at the near positions.

\begin{table}[ht]
    \centering
    \caption{Basic setup about unlearning scenarios. More illustrations can be found in Appendix~\ref{app:dataset_partition}.}
    \vspace{2mm}
    \renewcommand\arraystretch{0.9}
    \resizebox{\textwidth}{!}{
    \begin{tabular}{c|c|c|c|c|c|c}
    \toprule[1.5pt]
    Dataset & Forgetting Data & Setup & All matched & Model mismatch & Target mismatch & Data mismatch \\
    \midrule[0.6pt]
        \multirow{2}*{CIFAR-10} & \multirow{2}*{``automobile''} & Training Class & 10 & 5 & 10 & 5 \\
     \cmidrule{3-7}
     ~ & ~ & Target Concept  & ``automobile'' & ``automobile'' & ``vehicle'' & ``vehicle'' \\
    \midrule[0.6pt]
     \multirow{2}*{CIFAR-100} & \multirow{2}*{``boy'', ``girl''} & Training Class & 100 & 20 & 100 & 20 \\
     \cmidrule{3-7}
     ~& ~& Target Concept & ``boy'', ``girl'' & ``boy'', ``girl'' & ``people'' & ``people'' \\
    \bottomrule[1.5pt]
    \end{tabular}
    }
    \vspace{-5mm}
    \label{tab:exp_set}
\end{table}

\begin{table}[t!]
    \centering
    \caption{Full list of the 20-class classification on CIFAR-100 with its official superclass labels~\cite{krizhevsky2009learning_cifar10}.}
    \vspace{2mm}
    \begin{tabular}{c|c}
    \toprule[1.5pt]
    Superclass (20) & Classes (5 for each superclass)\\
    \midrule[0.6pt]
        aquatic mammals &	beaver, dolphin, otter, seal, whale\\
        fish &	aquarium fish, flatfish, ray, shark, trout\\
        flowers &	orchids, poppies, roses, sunflowers, tulips\\
        food containers &	bottles, bowls, cans, cups, plates \\
        fruit and vegetables &	apples, mushrooms, oranges, pears, sweet peppers\\
household electrical devices &	clock, computer keyboard, lamp, telephone, television \\
household furniture &	bed, chair, couch, table, wardrobe \\
insects &	bee, beetle, butterfly, caterpillar, cockroach \\
large carnivores &	bear, leopard, lion, tiger, wolf \\
large man-made outdoor things &	bridge, castle, house, road, skyscraper \\
large natural outdoor scenes &	cloud, forest, mountain, plain, sea \\
large omnivores and herbivores &	camel, cattle, chimpanzee, elephant, kangaroo \\
medium-sized mammals &	fox, porcupine, possum, raccoon, skunk \\
non-insect invertebrates &	crab, lobster, snail, spider, worm\\
people &	baby, boy, girl, man, woman \\
reptiles &	crocodile, dinosaur, lizard, snake, turtle\\
small mammals &	hamster, mouse, rabbit, shrew, squirrel \\
trees&	maple, oak, palm, pine, willow \\
vehicles 1 &	bicycle, bus, motorcycle, pickup truck, train \\
vehicles 2 &	lawn-mower, rocket, streetcar, tank, tractor \\
    \bottomrule[1.5pt]
    \end{tabular}
    \label{tab:app_exp_set_cifar100}
\end{table}

\begin{table}[t!]
\centering
\caption{Full list of the 5-class classification on CIFAR-10 with its manually set superclass~\cite{krizhevsky2009learning_cifar10}.}
    \label{tab:app_exp_set_cifar10}
\vspace{2mm}
    \begin{tabular}{c|c}
    \toprule[1.5pt]
    Superclass (5) & Classes (2 for each superclass)\\
    \midrule[0.6pt]
        1 & airplane, bird\\
        2 &	automobile, truck\\
        3 &	cat, dog\\
        4 &	deer, frog \\
        5 &	horse, ship\\
    \bottomrule[1.5pt]
    \end{tabular}
\end{table}

\begin{table}[t!]
\centering
\captionof{table}{Specific training set data partition corresponding to four major forgetting tasks.}
\vspace{2mm}
    \begin{tabular}{c|c|c}
    \toprule[1.5pt]
    Forgetting Tasks & \multicolumn{2}{c}{Data Partition}\\
    \midrule[0.6pt]
    & Identified & Unidentified\\
    \midrule[0.6pt]
        All matched & $\mathcal{D}_\text{f}=\mathcal{D}_\text{t}$ & $\mathcal{D}_\text{un}=\mathcal{D}_\text{r}$\\
        Target mismatch & $\mathcal{D}_\text{f}\subset\mathcal{D}_\text{t}$&$\mathcal{D}_\text{un}=\mathcal{D}_\text{uf}\cup\mathcal{D}_\text{r}$\\
        Model mismatch &$\mathcal{D}_\text{f}=\mathcal{D}_\text{t}$ & $\mathcal{D}_\text{un}=\mathcal{D}_\text{r}$\\
        Data mismatch &	$\mathcal{D}_\text{f}\subset\mathcal{D}_\text{t}$&$\mathcal{D}_\text{un}=\mathcal{D}_\text{uf}\cup\mathcal{D}_\text{r}$ \\
    \bottomrule[1.5pt]
    \end{tabular}
    \label{tab:app_exp_set_partition}
\end{table}

\clearpage
\section{Algorithm Implementation and Explanation}
\label{app:algo}

In this section, we present the pseudo-code of our proposed TARF and its variant, as well as additional discussions to enhance the understanding of our methods. Here we summarize the detailed procedure of algorithm implementation in Algorithm~\ref{alg:TARF} and Algorithm~\ref{alg:TARF_instance}. In detail, Algorithm~\ref{alg:TARF} identifies the potential target using the class labels, while Algorithm~\ref{alg:TARF_instance} can use the instance level information. 

As introduced in Section~\ref{sec:method_gradient}, the objective of our TARF is defined as follows,
\begin{align}\label{eq:5_app}
\begin{split}
    L_\text{TARF} = \underbrace{k(t)\cdot\bigg (-\frac{1}{|\mathcal{D}_{\text{f}}|}\sum_{(x,y)\sim \mathcal{D}_{\text{f}}}\ell(f(x),y)\bigg )}_{\text{Annealed Forgetting}~ L_{\text{f}}(k)}+ \underbrace{\frac{1}{|\mathcal{D}_\text{un}|}\sum_{(x,y)\sim \mathcal{D}_\text{un}}\ell(f(x),y)\cdot\tau(x,y,t)}_{\text{Target-aware Retaining}~ L_\text{u}(\tau)},
\end{split}
\end{align}
where $k(t)$ and $\tau(x,y,t)$ are two training-time-related hyperparameters to deal with the mismatch issues raised in our new settings. Specifically, we set a learning-rate-reduced $k(t)$ as,
\begin{equation}
    k(t) = k\cdot(T-t-t_0)/T,\quad t\in[0,T],
\end{equation}
where $T$ indicates the total training time (e.g., epochs), and the value of $k(t)$ decreases with the training process. On the other hand, we have the following indicator to measure the model prediction consistency with the training data$I_\text{con}(x,y,\theta) = |\ell_{f_\theta}(x,y) - \ell_{f_{\theta^*}}(x,y)|$,
with which we set $\tau(x,y,t)$ as follows,
\begin{equation}\label{eq:eq8_app}
    \tau(x, y, t)=\left \{
    \begin{aligned}
    0& && I_\text{con}(x,y,\theta_{t_1})>\beta ~\text{or}~ t<t_1&~~^*\text{Unconf. Retain,}\\
    1& && I_\text{con}(x,y,\theta_{t_1})<\beta ~\text{and}~ t\geq t_1 &~~^*\text{Conf. Retain,}
    \end{aligned}
    \right.
\end{equation}
where $t_1$ is a time stamp to control the start of pursuing the retaining part. The overall two dynamic hyperparameters can divide the whole unlearning process into three phases as illustrated in Figure~\ref{fig:framework}.

\paragraph{Annealed Forgetting.} For the forgetting target, we adopt the gradient ascent on the given forgetting data to unlearn it. However, to approximate the retrained model, the intuition is not to pursue the maximization of the risk on this part of the data but to destroy the learned feature on the given model. So we introduce a learning-rate-reduced $k(t)$ to realize the annealed gradient ascent where $t_0=1$ is adopted for target or data mismatch forgetting, and the value of $k(t)$ decreases with the training process. Resulting in destroyed features, gradient ascent on this part of data also constructs the dynamic information for differentiating the data of different consistency on its loss values, making the risks of the concept-aligned data higher than the rest, and helping to filter retaining data.

\paragraph{Target-aware Retaining.} For the retaining part, we need to selectively learn the data from the remaining set, since the complementary dataset may be biased with unidentified forgetting data. Compared with other remaining data, the concept-aligned data is easy to be affected by similar feature representation as indicated in Figure~\ref{fig:feature_a}. Thus, we can have $\tau(x,y,t)$ 
where we can divide the remaining set into unconfident/confident parts to note the estimated retaining data like Figure~\ref{fig:frameworka}.  $t_1=2$ is adopted at target and data mismatch tasks, and $\beta$ can be estimated by the prior information about the specific unlearning request and the rank of loss values. By simultaneously conducting gradient ascent on forgetting data and selective gradient descent on confident retaining data, we can better restrict the forgetting region and deconstruct the entangled feature representation (refer to the middle of Figure~\ref{fig:frameworkb} where we reveal the feature decomposition in deeper layers of model structure using ResNet). Finally, with the partial objective of retaining, it can approximate the retrained reference (refer to the right of Figure~\ref{fig:frameworkb}).

\begin{algorithm}[t!]
   \caption{TARF}
   \label{alg:TARF}
   {\bf Input:} 
   Training dataset $D = \{(\bx_i, y_i, s_i) \}^{n}_{i=1}$, where $s_i=1$ indicates the identified forgetting dat, otherwise the data is recognized to be unlabeled for unlearning,
   learning rate $\eta$, number of epochs $T$, batch size $m$, number of batches $M$, data ${x}\in \cX$, label $y \in \cY$, original trained model $\theta$, loss function $\ell$, initialized indicator value $\tau$ with the threshold $\beta$. \\
   {\bf Output:} model $\theta^{T}$;
\begin{algorithmic}[1]
    \FOR {mini-batch $=1$, $\dots$, $M$ }
    \STATE Sample a mini-batch $\{(\bx_i, y_i) \}^{m}_{i=1}$ from $D$
    \STATE $\{\ell(x_i,y_i)\}_{i=1}^m\leftarrow \theta.\text{forward}(f_\theta,\{(x_i, y_i) \}^{m}_{i=1})$,
    \STATE Collect the initial training accuracy in each class based on $\{\ell(x_i,y_i)\}_{i=1}^m$,
    \ENDFOR
    \FOR{epoch $= 1$, $\dots$, $T$}
    \STATE Update $k(t)$ according to Eq.~\eqref{eq:k_t}, 
    \IF{epoch < $t_0$} \STATE $\tau\leftarrow 0$
    \ELSE \STATE compute $\beta$ in Eq.~\eqref{eq:k_t} according to the rank of class accuracy difference, and update $\tau$.
    \ENDIF
    \FOR {mini-batch $=1$, $\dots$, $M$ }
    \STATE Sample a mini-batch $\{(\bx_i, y_i,s_i) \}^{m}_{i=1}$ from $D$
        \STATE Assign different weights for identified target samples and the rest retaining data,
        \STATE $L_\text{TARF} = k(t)\cdot\bigg (-\frac{1}{|\mathcal{D}_{\text{f}}|}\sum_{(x,y)\sim \mathcal{D}_{\text{f}}}\ell(f(x),y)\bigg )+ \frac{1}{|\mathcal{D}_\text{un}|}\sum_{(x,y)\sim \mathcal{D}_\text{un}}\ell(f(x),y)\cdot\tau(x,y,t),$
         
         \STATE $\mathbf{\theta} \gets \mathbf{\theta} - \eta \nabla_{\mathbf{\theta}} L_\text{TARF}(D, D_\text{f},f,\tau)$
        \ENDFOR
    \ENDFOR
\end{algorithmic}
\end{algorithm}

\paragraph{Case study for Unlearning Generation with Stable Diffusion.} To demonstrate the compatibility, we also extend the core idea of this work and investigate the performance of TARF on the specific text-to-image generation with stable diffusion~\cite{gandikota2023erasing,fan2023salun}, and presented in Tables~\ref{tab:img_tench} and~\ref{tab:img_english_springer}.

In detail, we aim to unlearn the image generation of a class with its specific prompt like ``a photo of a tench''. To simulate the practical unlearning request, we construct the given dataset consisting of limited forgetting data and the unidentified remaining data to conduct unlearning, which corresponds to the data mismatch forgetting task that similar to the right-bottom illustration of Figure~\ref{fig: challenge}. Then we compare the image generation on the original stable diffusion, the unlearned model with certain label (CL) mismatching~\cite{fan2023salun}, and that with our TARF. For this exploration, we adopt the instance-wise identification during the forgetting process as described in Algorithm~\ref{alg:TARF_instance}, to unlearn the target concept with the given limited forgetting data and pursue retaining the selected remaining data with lower loss values. The results in Tables~\ref{tab:img_tench} and~\ref{tab:img_english_springer} show the potential of our framework in the expansion on image generation without the perfect information in unlearning.

\begin{algorithm}[t!]
   \caption{TARF-I: generalized version on instance-wise identification}
   \label{alg:TARF_instance}
   {\bf Input:} 
    Training dataset $D = \{(\bx_i, y_i, s_i) \}^{n}_{i=1}$, where $s_i=1$ indicates the identified forgetting dat, otherwise the data is recognized to be unlabeled for unlearning,
   learning rate $\eta$, number of epochs $T$, batch size $m$, number of batches $M$, data ${x}\in \cX$, label $y \in \cY$, original trained model $\theta$, loss function $\ell$, initialized indicator value $\tau$ with the threshold $\beta$. \\
   {\bf Output:} model $\theta^{T}$;
\begin{algorithmic}[1]
    \FOR {mini-batch $=1$, $\dots$, $M$ }
    \STATE Sample a mini-batch $\{(\bx_i, y_i) \}^{m}_{i=1}$ from $D$
    \STATE $\{\ell(x_i,y_i)\}_{i=1}^m\leftarrow \theta.\text{forward}(f_\theta,\{(x_i, y_i) \}^{m}_{i=1})$,
    \STATE Collect the initial loss values in each training samples based on $\{\ell(x_i,y_i)\}_{i=1}^m$,
    \ENDFOR
    \FOR{epoch $= 1$, $\dots$, $T$}
    \STATE Update $k(t)$ according to Eq.~\eqref{eq:k_t}, 
    \IF{epoch < $t_0$} \STATE $\tau\leftarrow 0$
    \ELSE \STATE compute $\beta$ in Eq.~\eqref{eq:k_t} according to the rank of difference in instance loss values, and update $\tau$.
    \ENDIF
    \FOR {mini-batch $=1$, $\dots$, $M$ }
    \STATE Sample a mini-batch $\{(\bx_i, y_i,s_i) \}^{m}_{i=1}$ from $D$
        \STATE Assign different weights for identified target samples and the rest retaining data,
        \STATE $L_\text{TARF} = k(t)\cdot\bigg (-\frac{1}{|\mathcal{D}_{\text{f}}|}\sum_{(x,y)\sim \mathcal{D}_{\text{f}}}\ell(f(x),y)\bigg )+ \frac{1}{|\mathcal{D}_\text{un}|}\sum_{(x,y)\sim \mathcal{D}_\text{un}}\ell(f(x),y)\cdot\tau(x,y,t),$
         
         \STATE $\mathbf{\theta} \gets \mathbf{\theta} - \eta \nabla_{\mathbf{\theta}} L_\text{TARF}(D, D_\text{f},f,\tau)$
        \ENDFOR
    \ENDFOR
\end{algorithmic}
\end{algorithm}

\begin{table}[t!]
  \centering
  \caption{Image generation results of unlearned Stable Diffusion in the \textbf{Data mismatch forgetting}, compared with the original stable diffusion, certain label (CL) unlearning~\cite{fan2023salun}, and our \textbf{TARF}. The specific prompt used in the image generation is "a photo of tench".}
  \vspace{2mm}
  \footnotesize
  \resizebox{\textwidth}{!}{
  \begin{tabular}{ c | c | c }
    \toprule[1.5pt]
     {\shortstack{\textbf{Original}\\ Stable Diffusion}} &{\shortstack{\textbf{Unlearned}\\by CL~\cite{fan2023salun}}} & {\shortstack{\textbf{Unlearned}\\ by \textbf{TARF}}} \\
\midrule[0.6pt]
    \begin{minipage}[b]{0.32\columnwidth}
		\centering
		\raisebox{-.5\height}{\includegraphics[width=0.48\linewidth]{0_0_sd.png}}
  \raisebox{-.5\height}{\includegraphics[width=0.48\linewidth]{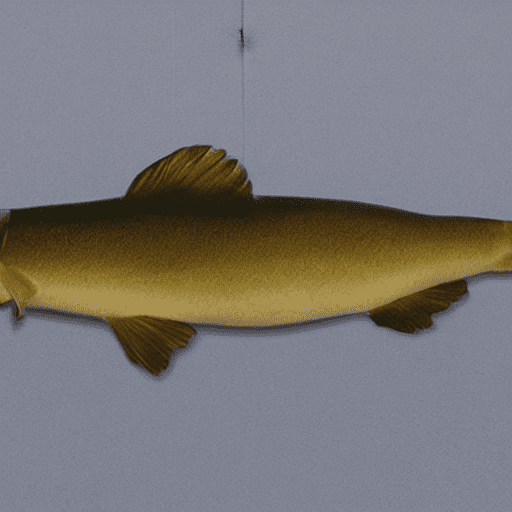}}\\
  \raisebox{-.5\height}{\includegraphics[width=0.48\linewidth]{0_2_sd.png}}
  \raisebox{-.5\height}{\includegraphics[width=0.48\linewidth]{0_3_sd.png}}\\
  \raisebox{-.5\height}{\includegraphics[width=0.48\linewidth]{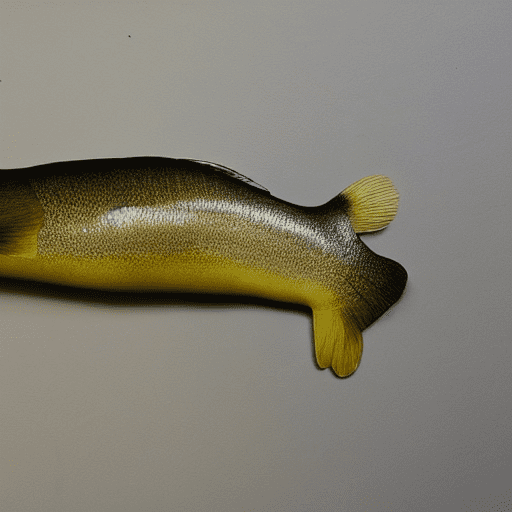}}
  \raisebox{-.5\height}{\includegraphics[width=0.48\linewidth]{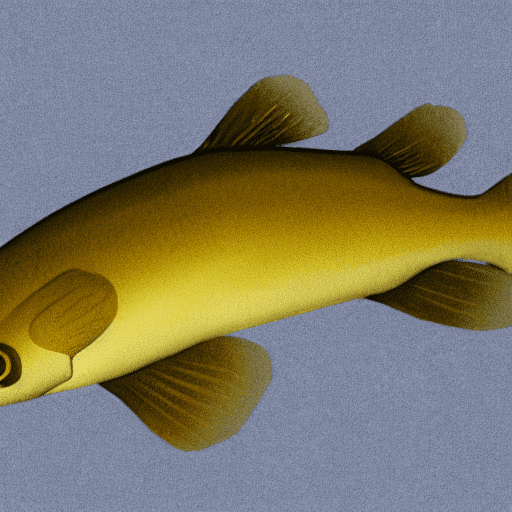}}\\
  \raisebox{-.5\height}{\includegraphics[width=0.48\linewidth]{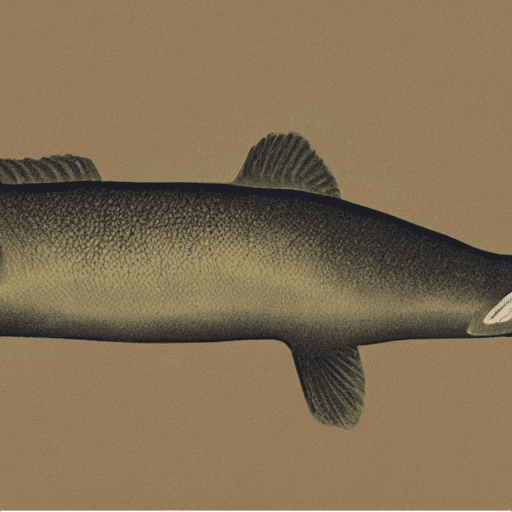}}
  \raisebox{-.5\height}{\includegraphics[width=0.48\linewidth]{0_7_sd.png}}\\
  \raisebox{-.5\height}{\includegraphics[width=0.48\linewidth]{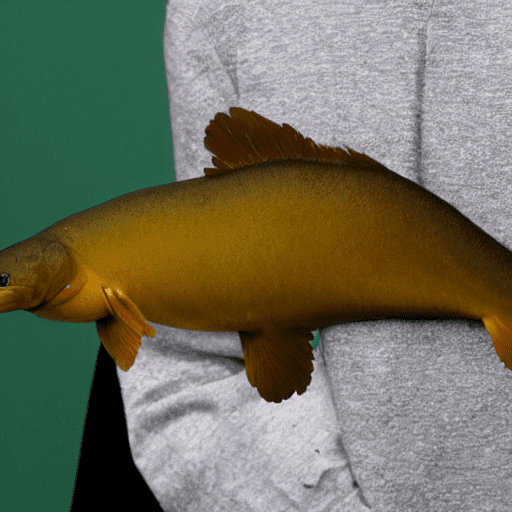}}
  \raisebox{-.5\height}{\includegraphics[width=0.48\linewidth]{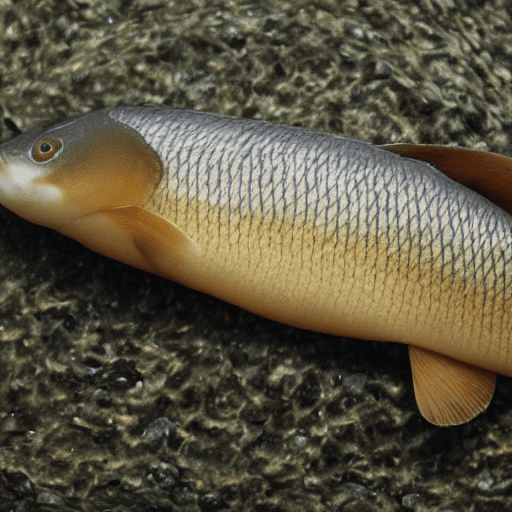}}\\
  \raisebox{-.5\height}{\includegraphics[width=0.48\linewidth]{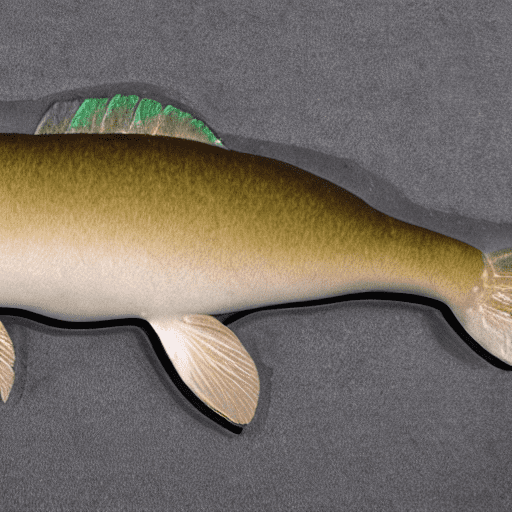}}
  \raisebox{-.5\height}{\includegraphics[width=0.48\linewidth]{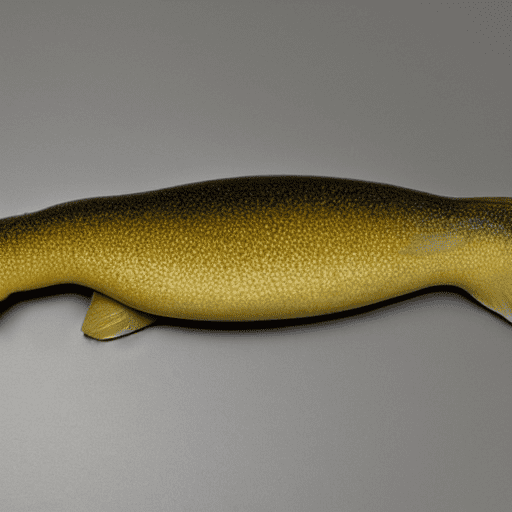}}\\
  \raisebox{-.5\height}{\includegraphics[width=0.48\linewidth]{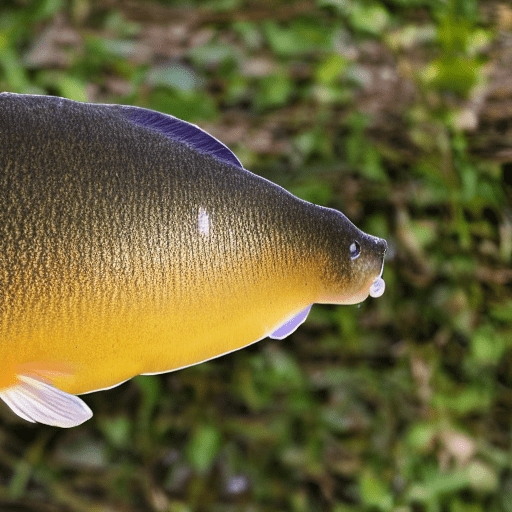}}
  \raisebox{-.5\height}{\includegraphics[width=0.48\linewidth]{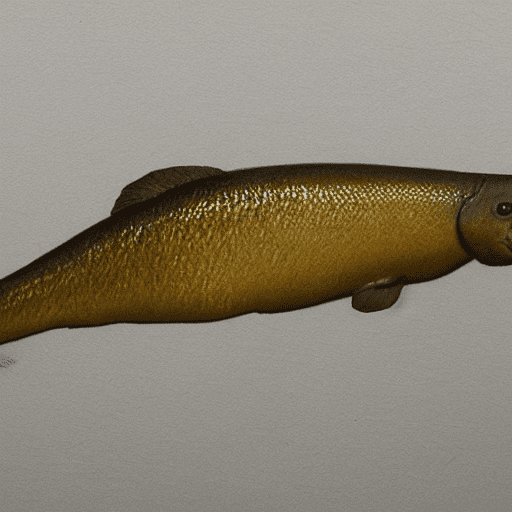}}
  \\
  \raisebox{-.5\height}{\includegraphics[width=0.48\linewidth]{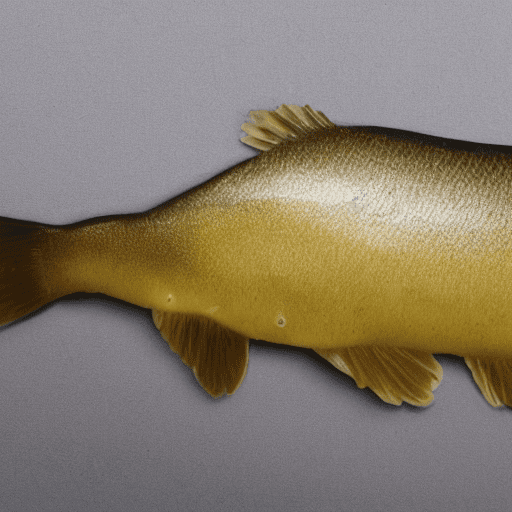}}
  \raisebox{-.5\height}{\includegraphics[width=0.48\linewidth]{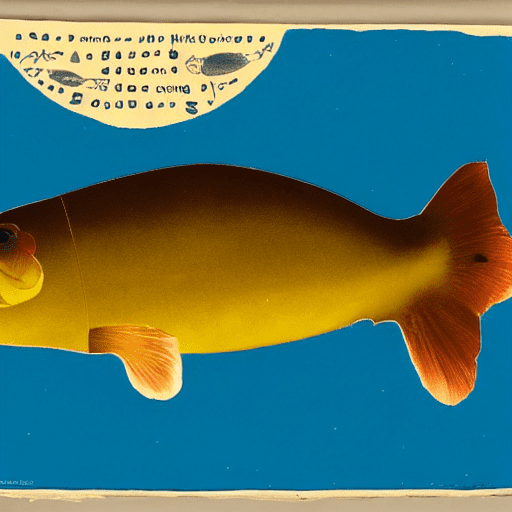}}
  \\
  \raisebox{-.5\height}{\includegraphics[width=0.48\linewidth]{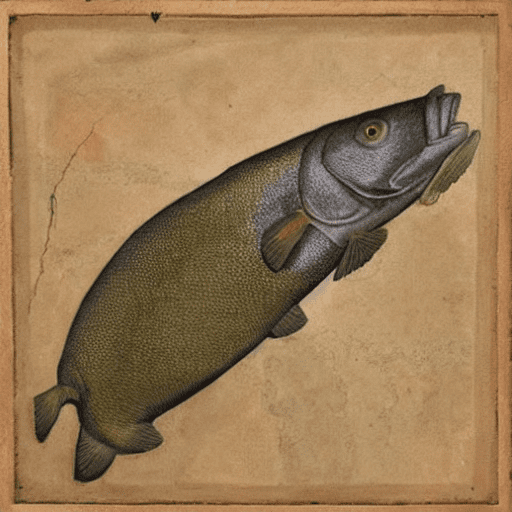}}
  \raisebox{-.5\height}{\includegraphics[width=0.48\linewidth]{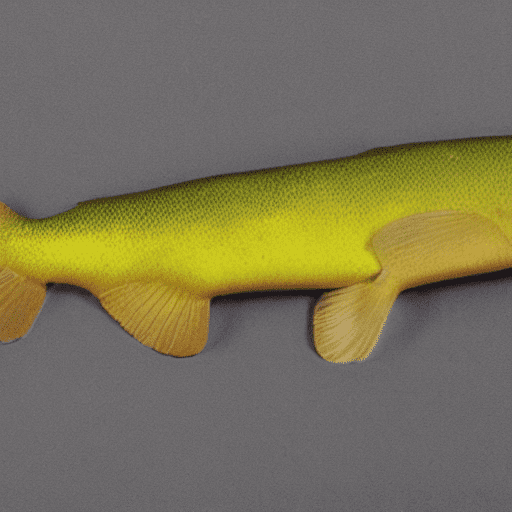}}
	\end{minipage}
    & 
    \begin{minipage}[b]{0.32\columnwidth}
		\centering
		\raisebox{-.5\height}{\includegraphics[width=0.48\linewidth]{0_0_cl.png}}
  \raisebox{-.5\height}{\includegraphics[width=0.48\linewidth]{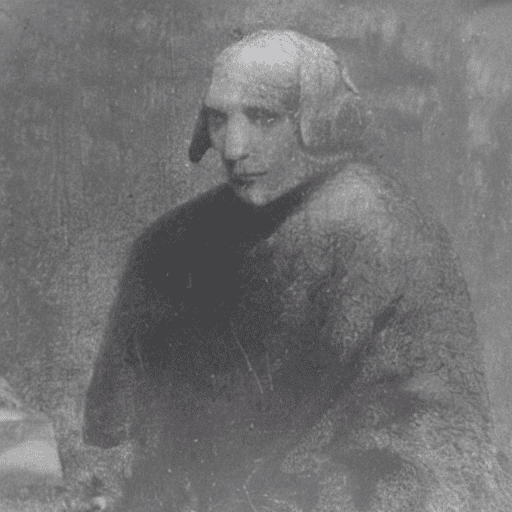}}\\
  \raisebox{-.5\height}{\includegraphics[width=0.48\linewidth]{0_2_cl.png}}
  \raisebox{-.5\height}{\includegraphics[width=0.48\linewidth]{0_3_cl.png}}\\
  \raisebox{-.5\height}{\includegraphics[width=0.48\linewidth]{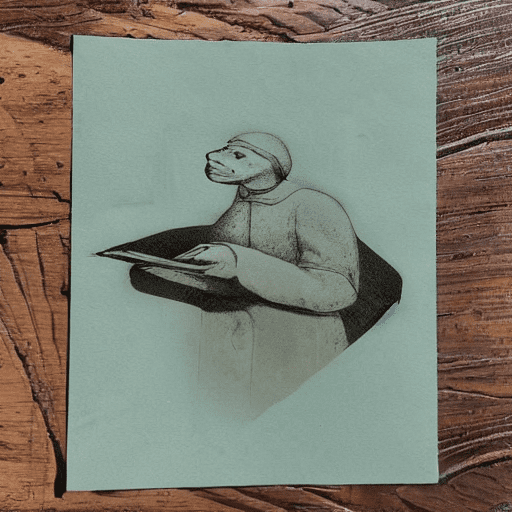}}
  \raisebox{-.5\height}{\includegraphics[width=0.48\linewidth]{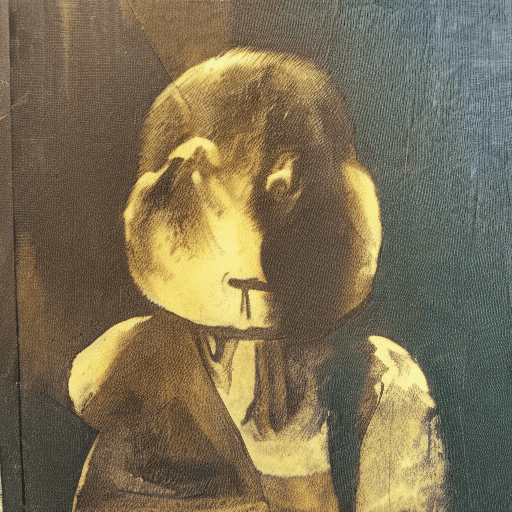}}\\
  \raisebox{-.5\height}{\includegraphics[width=0.48\linewidth]{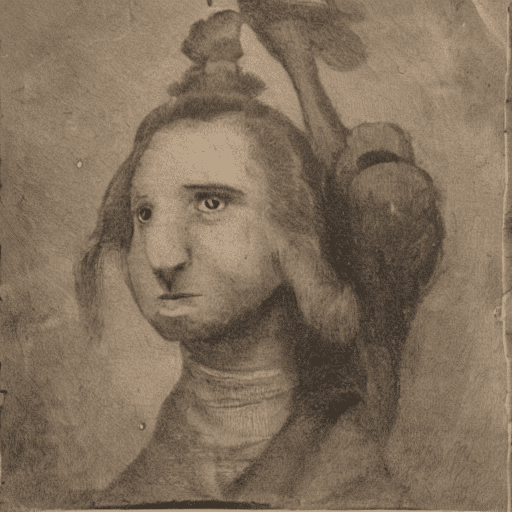}}
  \raisebox{-.5\height}{\includegraphics[width=0.48\linewidth]{0_7_cl.png}}\\
  \raisebox{-.5\height}{\includegraphics[width=0.48\linewidth]{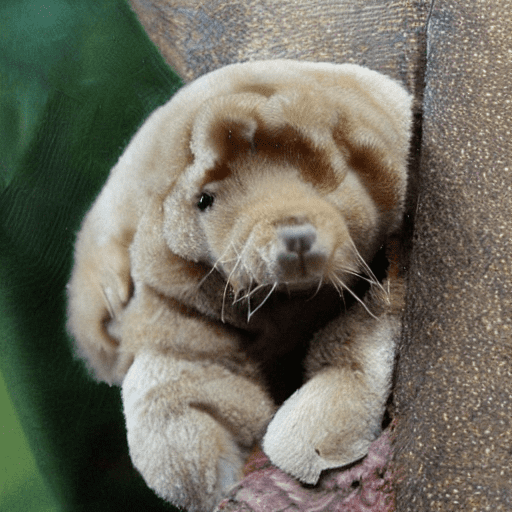}}
  \raisebox{-.5\height}{\includegraphics[width=0.48\linewidth]{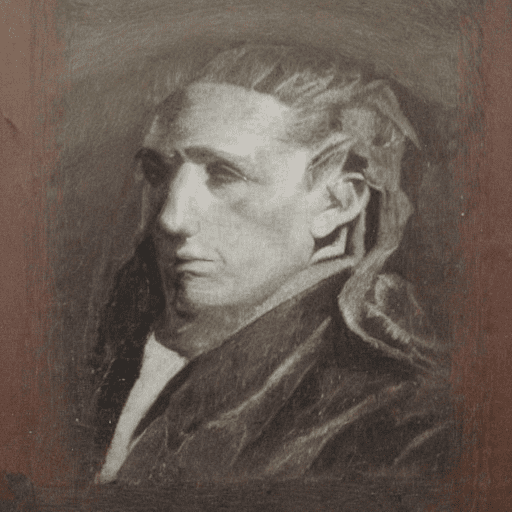}}\\
  \raisebox{-.5\height}{\includegraphics[width=0.48\linewidth]{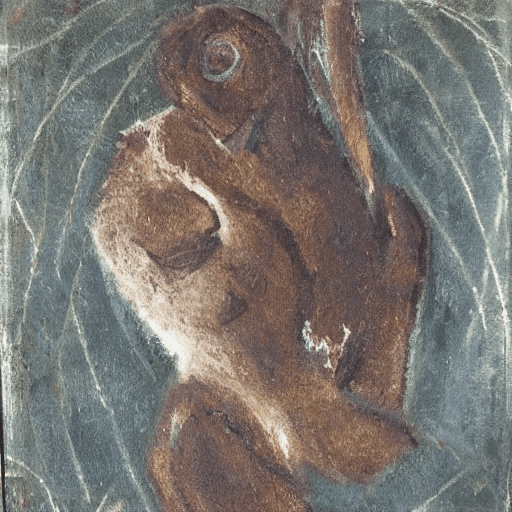}}
  \raisebox{-.5\height}{\includegraphics[width=0.48\linewidth]{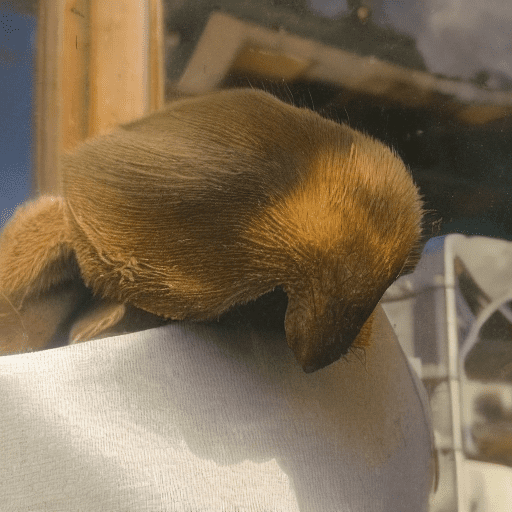}}\\
  \raisebox{-.5\height}{\includegraphics[width=0.48\linewidth]{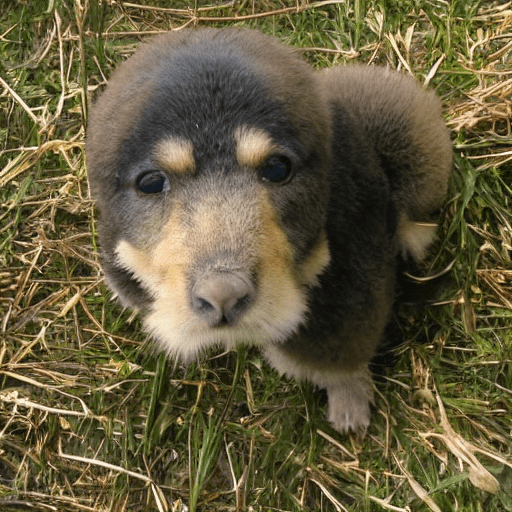}}
  \raisebox{-.5\height}{\includegraphics[width=0.48\linewidth]{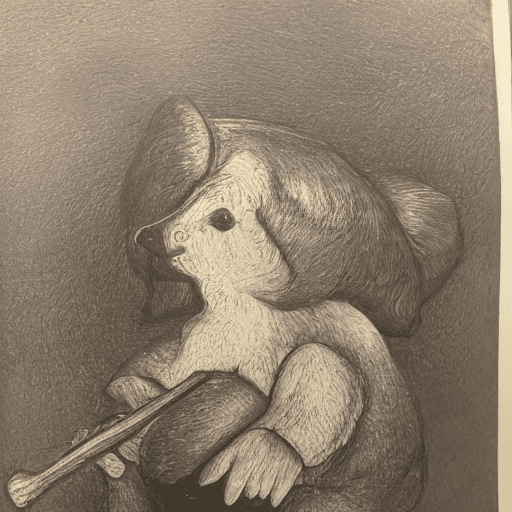}}
  \\
  \raisebox{-.5\height}{\includegraphics[width=0.48\linewidth]{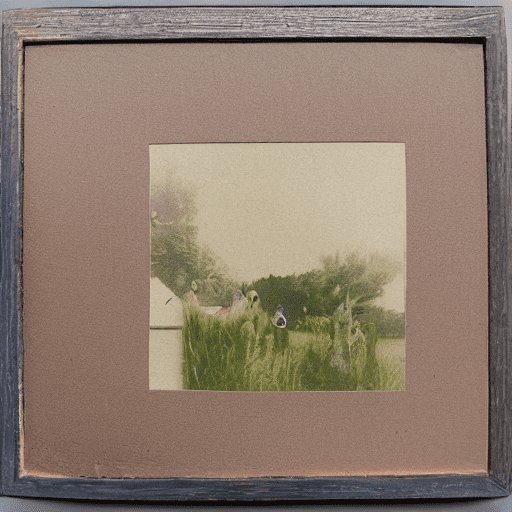}}
  \raisebox{-.5\height}{\includegraphics[width=0.48\linewidth]{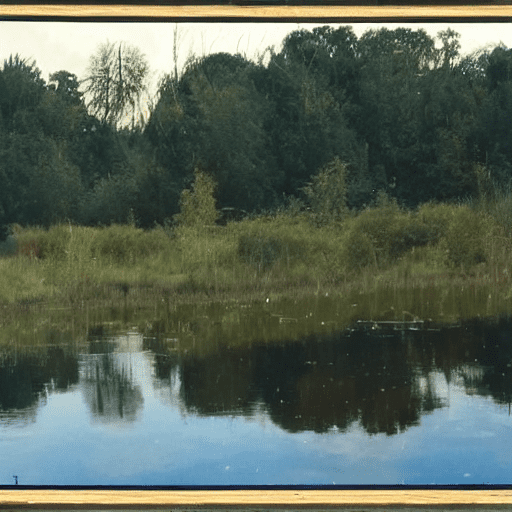}}
  \\
  \raisebox{-.5\height}{\includegraphics[width=0.48\linewidth]{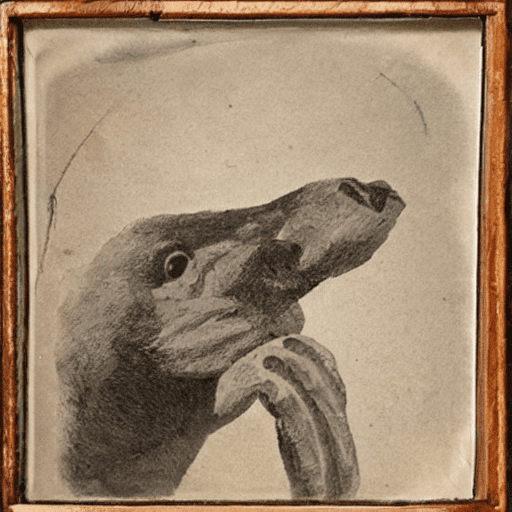}}
  \raisebox{-.5\height}{\includegraphics[width=0.48\linewidth]{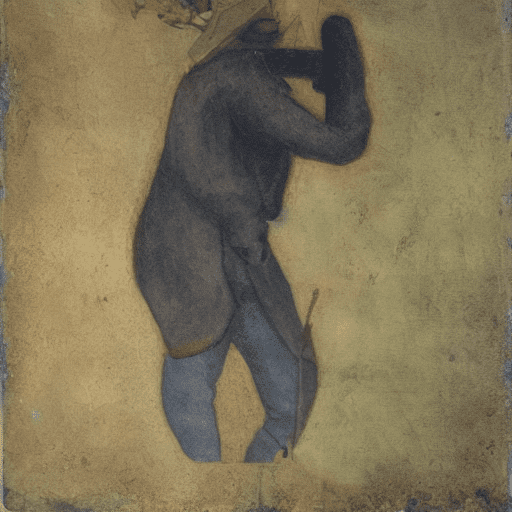}}
	\end{minipage}
    & 
    \begin{minipage}[b]{0.32\columnwidth}
		\centering
		\raisebox{-.5\height}{\includegraphics[width=0.48\linewidth]{0_0.png}}
  \raisebox{-.5\height}{\includegraphics[width=0.48\linewidth]{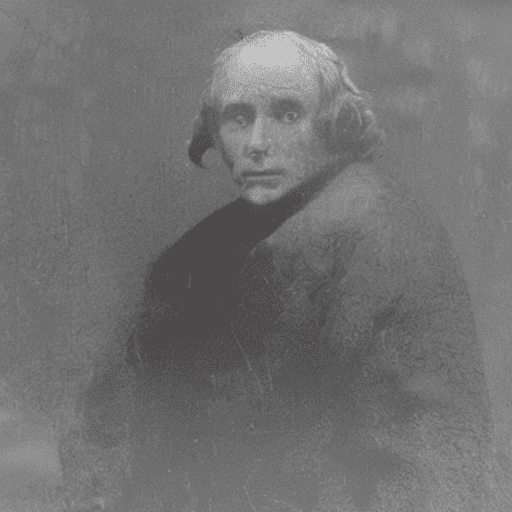}}\\
  \raisebox{-.5\height}{\includegraphics[width=0.48\linewidth]{0_2.png}}
  \raisebox{-.5\height}{\includegraphics[width=0.48\linewidth]{0_3.png}}\\
  \raisebox{-.5\height}{\includegraphics[width=0.48\linewidth]{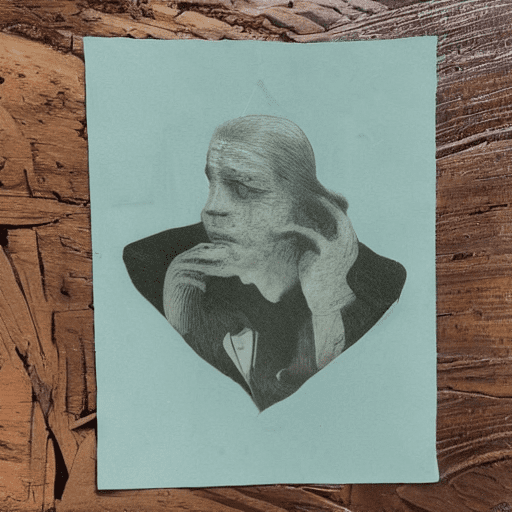}}
  \raisebox{-.5\height}{\includegraphics[width=0.48\linewidth]{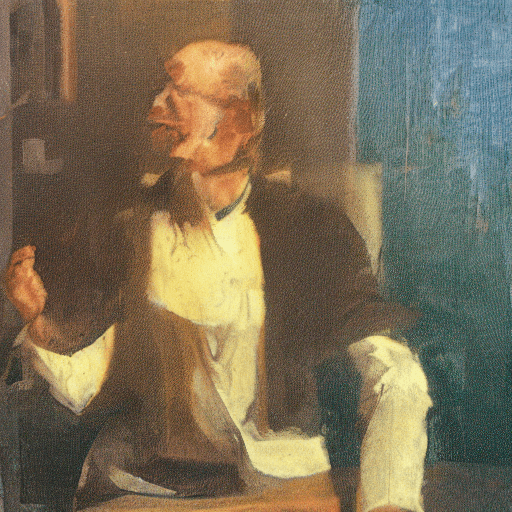}}\\
  \raisebox{-.5\height}{\includegraphics[width=0.48\linewidth]{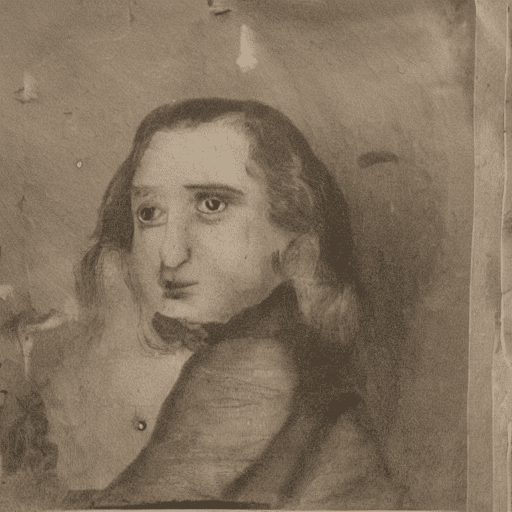}}
  \raisebox{-.5\height}{\includegraphics[width=0.48\linewidth]{0_7.png}}\\
  \raisebox{-.5\height}{\includegraphics[width=0.48\linewidth]{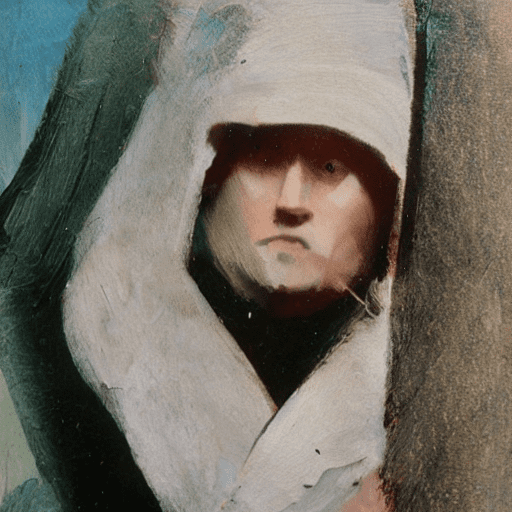}}
  \raisebox{-.5\height}{\includegraphics[width=0.48\linewidth]{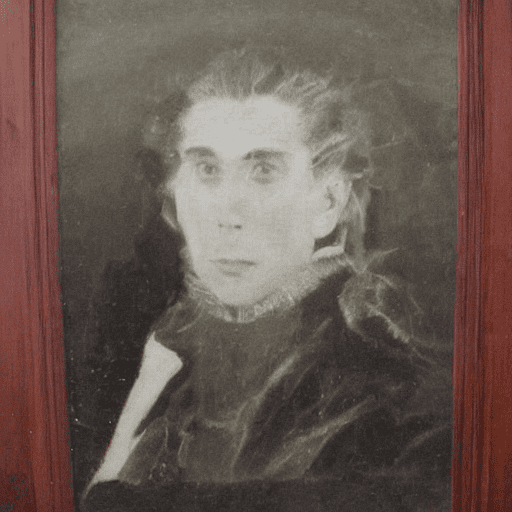}}\\
  \raisebox{-.5\height}{\includegraphics[width=0.48\linewidth]{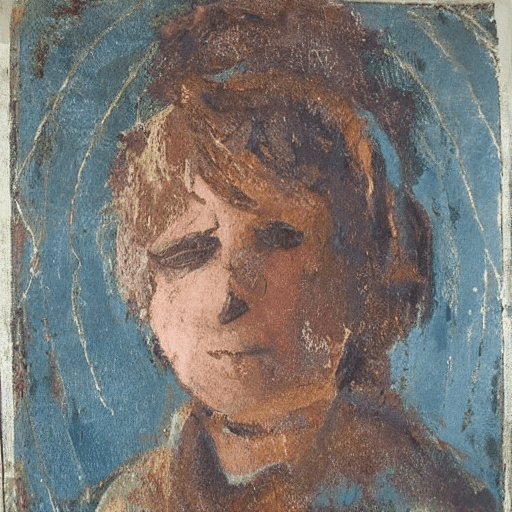}}
  \raisebox{-.5\height}{\includegraphics[width=0.48\linewidth]{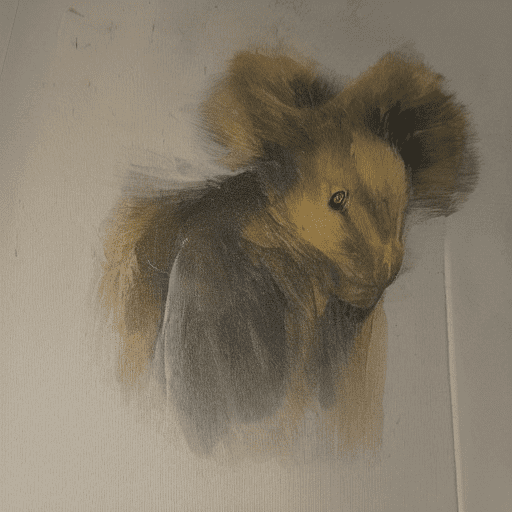}}\\
  \raisebox{-.5\height}{\includegraphics[width=0.48\linewidth]{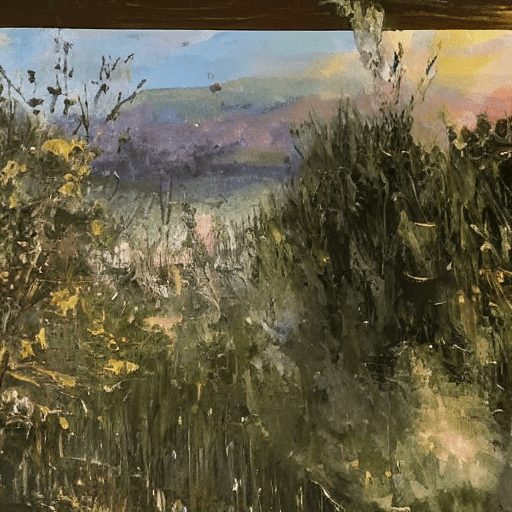}}
  \raisebox{-.5\height}{\includegraphics[width=0.48\linewidth]{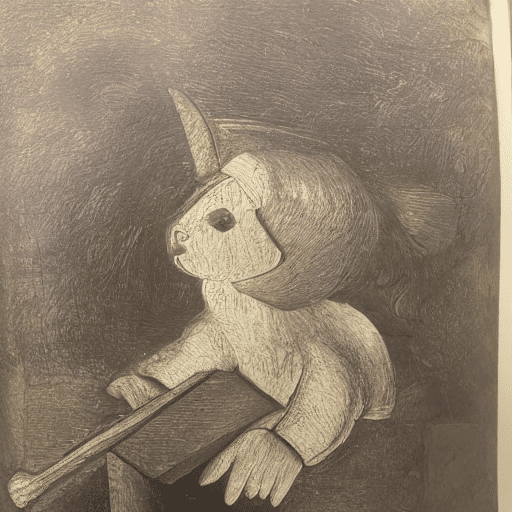}}
  \\
  \raisebox{-.5\height}{\includegraphics[width=0.48\linewidth]{0_14_cl.png}}
  \raisebox{-.5\height}{\includegraphics[width=0.48\linewidth]{0_15_cl.png}}
  \\
  \raisebox{-.5\height}{\includegraphics[width=0.48\linewidth]{0_16_cl.png}}
  \raisebox{-.5\height}{\includegraphics[width=0.48\linewidth]{0_17_cl.png}}
	\end{minipage}
    \\ 
    \bottomrule[1.5pt]
  \end{tabular}}
  \label{tab:img_tench}
\end{table}

\begin{table}[t!]
  \centering
  \caption{Image generation results of unlearned Stable Diffusion in the \textbf{Data mismatch forgetting}, compared with the original stable diffusion, certain label (CL) unlearning~\cite{fan2023salun}, and our \textbf{TARF}. The specific prompt used in the image generation is "a photo of English springer".}
  \vspace{2mm}
  \footnotesize
  \resizebox{\textwidth}{!}{
  \begin{tabular}{ c | c | c }
    \toprule[1.5pt]
     {\shortstack{\textbf{Original}\\ Stable Diffusion}} &{\shortstack{\textbf{Unlearned}\\by CL~\cite{fan2023salun}}} & {\shortstack{\textbf{Unlearned}\\ by \textbf{TARF}}} \\
\midrule[0.6pt]
    \begin{minipage}[b]{0.32\columnwidth}
		\centering
		\raisebox{-.5\height}{\includegraphics[width=0.48\linewidth]{1_0_sd.png}}
  \raisebox{-.5\height}{\includegraphics[width=0.48\linewidth]{1_1_sd.png}}\\
  \raisebox{-.5\height}{\includegraphics[width=0.48\linewidth]{1_2_sd.png}}
  \raisebox{-.5\height}{\includegraphics[width=0.48\linewidth]{1_3_sd.png}}\\
  \raisebox{-.5\height}{\includegraphics[width=0.48\linewidth]{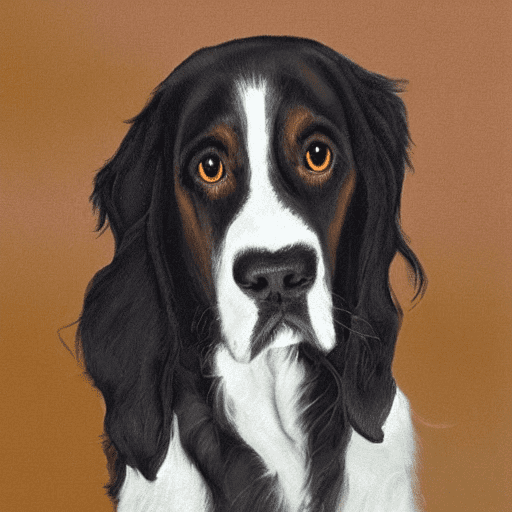}}
  \raisebox{-.5\height}{\includegraphics[width=0.48\linewidth]{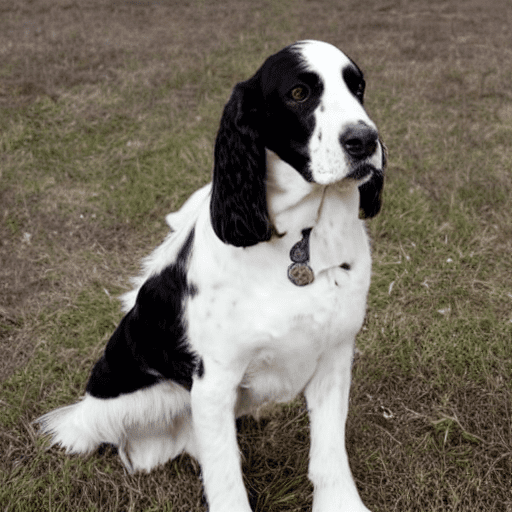}}\\
  \raisebox{-.5\height}{\includegraphics[width=0.48\linewidth]{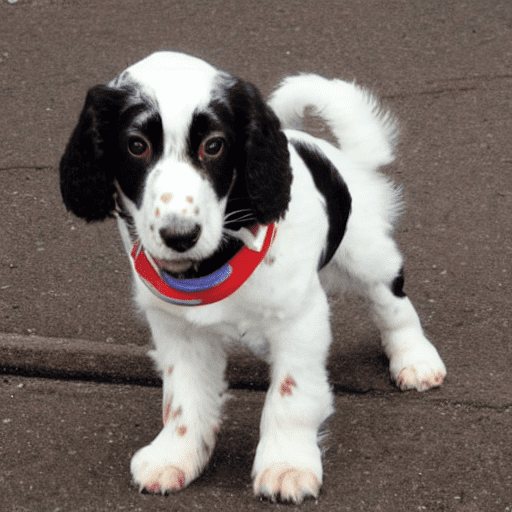}}
  \raisebox{-.5\height}{\includegraphics[width=0.48\linewidth]{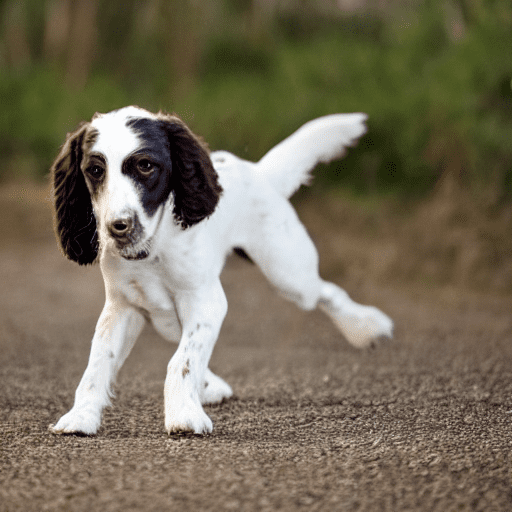}}\\
  \raisebox{-.5\height}{\includegraphics[width=0.48\linewidth]{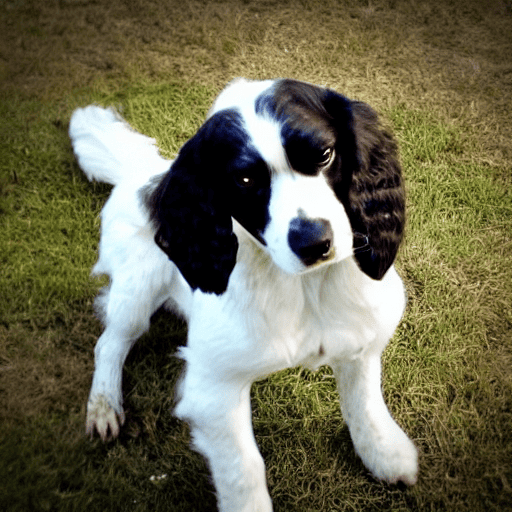}}
  \raisebox{-.5\height}{\includegraphics[width=0.48\linewidth]{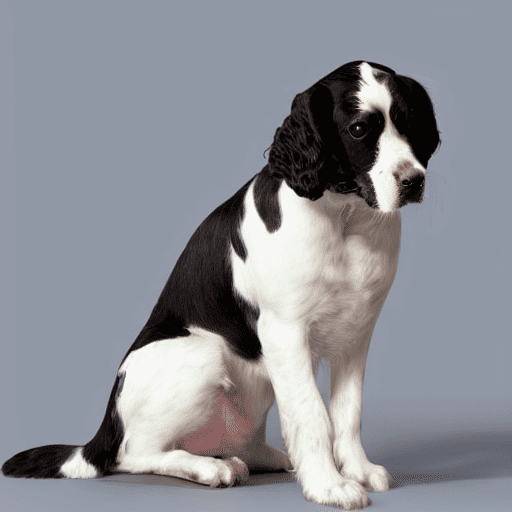}}\\
  \raisebox{-.5\height}{\includegraphics[width=0.48\linewidth]{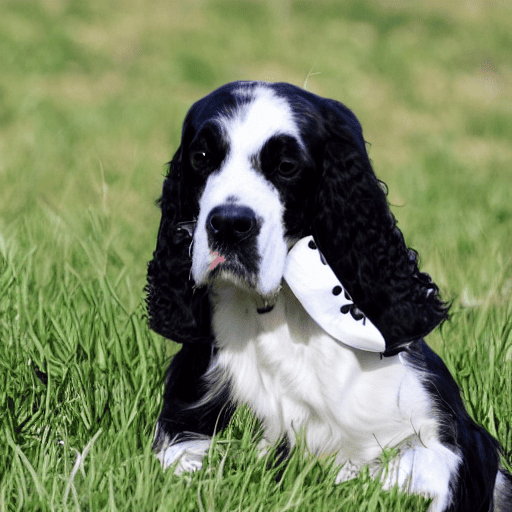}}
  \raisebox{-.5\height}{\includegraphics[width=0.48\linewidth]{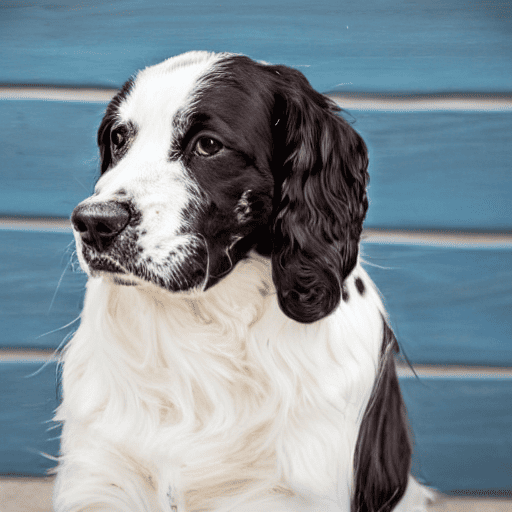}}\\
  \raisebox{-.5\height}{\includegraphics[width=0.48\linewidth]{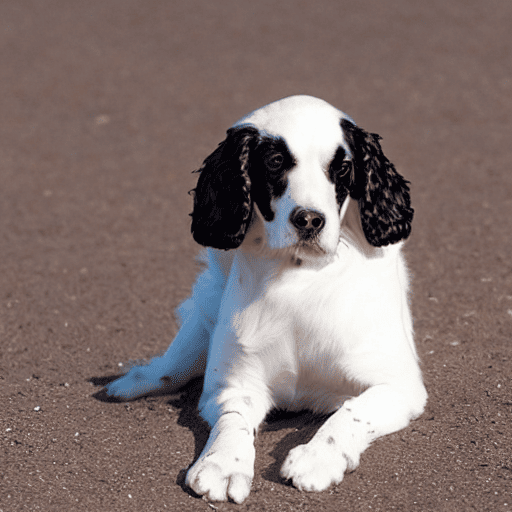}}
  \raisebox{-.5\height}{\includegraphics[width=0.48\linewidth]{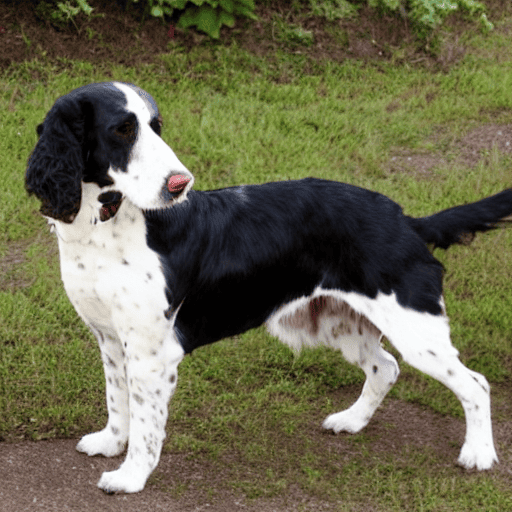}}
  \\
  \raisebox{-.5\height}{\includegraphics[width=0.48\linewidth]{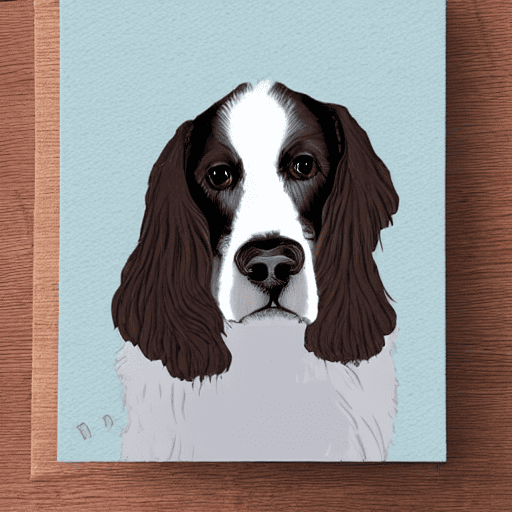}}
  \raisebox{-.5\height}{\includegraphics[width=0.48\linewidth]{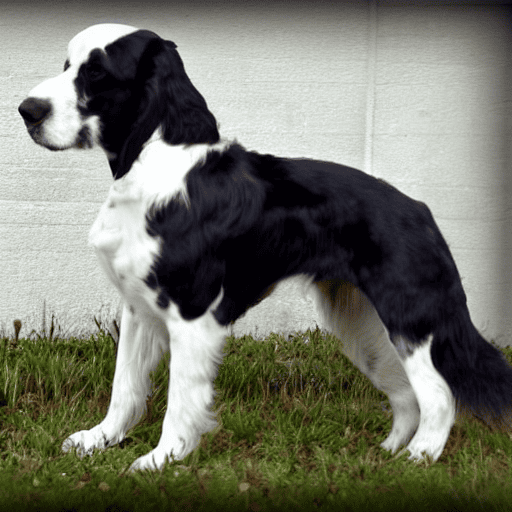}}
  \\
  \raisebox{-.5\height}{\includegraphics[width=0.48\linewidth]{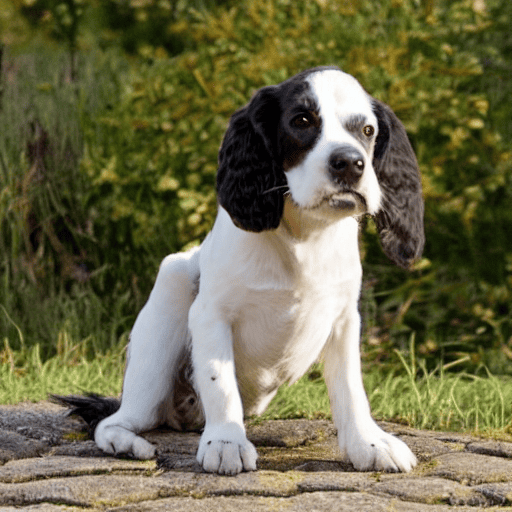}}
  \raisebox{-.5\height}{\includegraphics[width=0.48\linewidth]{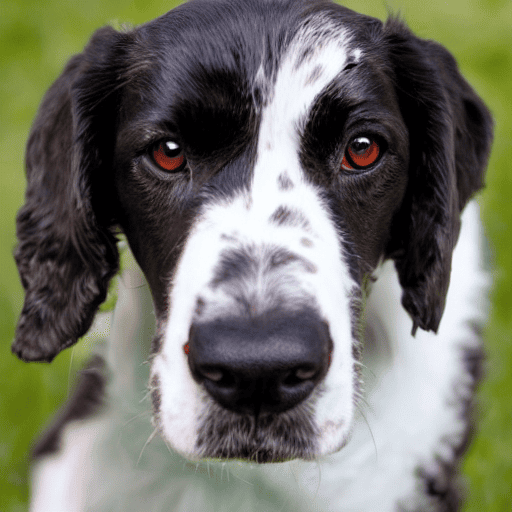}}
	\end{minipage}
    & 
    \begin{minipage}[b]{0.32\columnwidth}
		\centering
		\raisebox{-.5\height}{\includegraphics[width=0.48\linewidth]{0_0_cl.png}}
  \raisebox{-.5\height}{\includegraphics[width=0.48\linewidth]{1_1_cl.png}}\\
  \raisebox{-.5\height}{\includegraphics[width=0.48\linewidth]{1_2_cl.png}}
  \raisebox{-.5\height}{\includegraphics[width=0.48\linewidth]{1_3_cl.png}}\\
  \raisebox{-.5\height}{\includegraphics[width=0.48\linewidth]{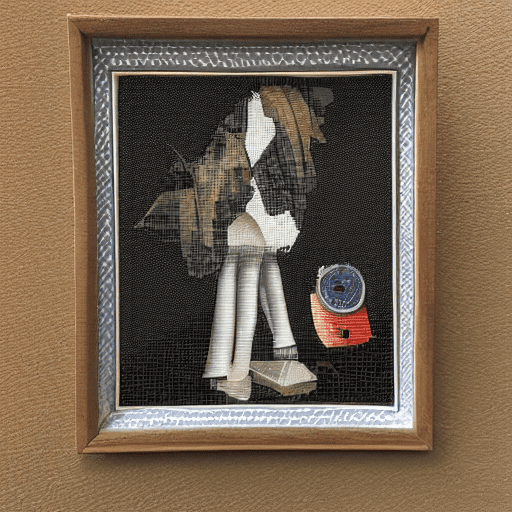}}
  \raisebox{-.5\height}{\includegraphics[width=0.48\linewidth]{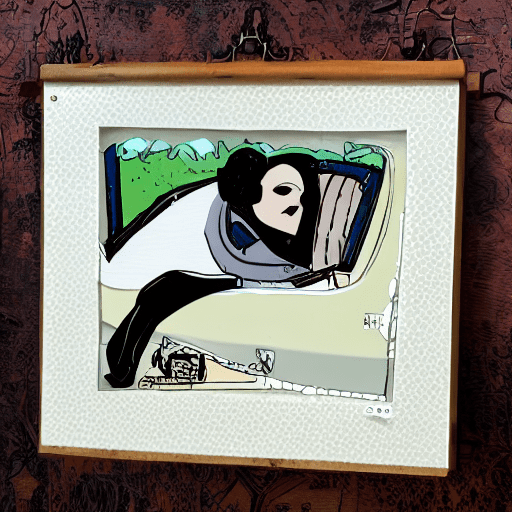}}\\
  \raisebox{-.5\height}{\includegraphics[width=0.48\linewidth]{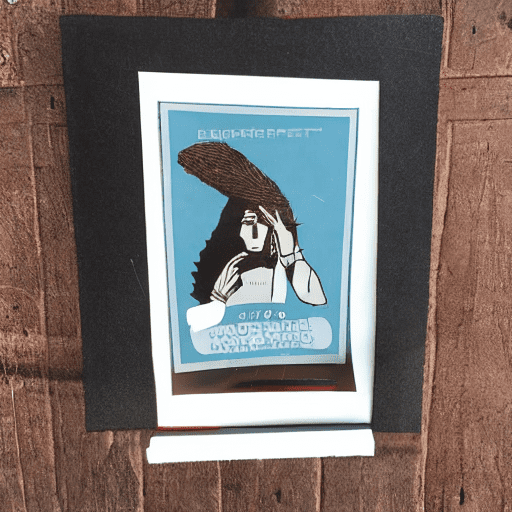}}
  \raisebox{-.5\height}{\includegraphics[width=0.48\linewidth]{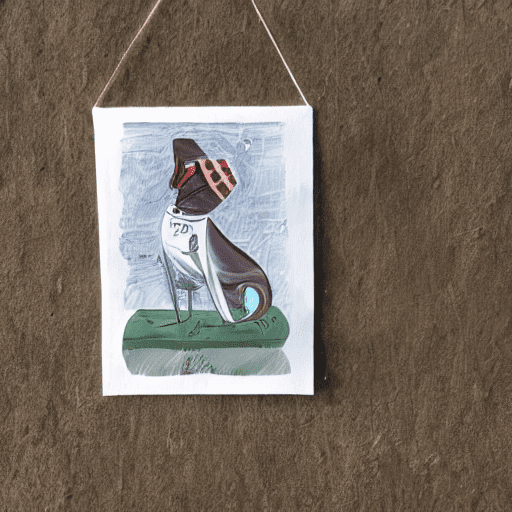}}\\
  \raisebox{-.5\height}{\includegraphics[width=0.48\linewidth]{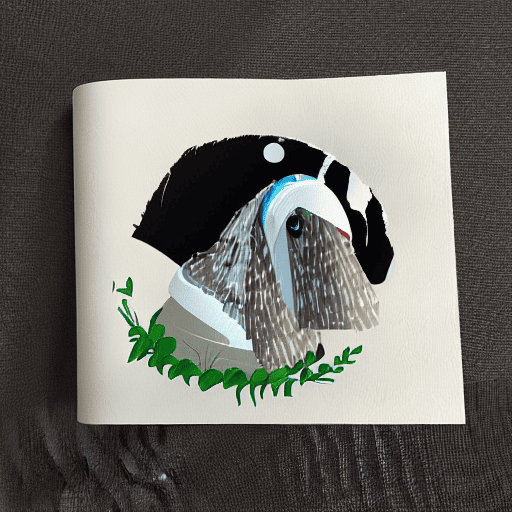}}
  \raisebox{-.5\height}{\includegraphics[width=0.48\linewidth]{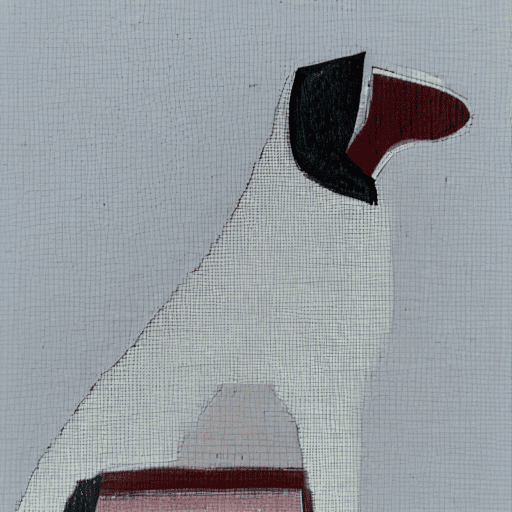}}\\
  \raisebox{-.5\height}{\includegraphics[width=0.48\linewidth]{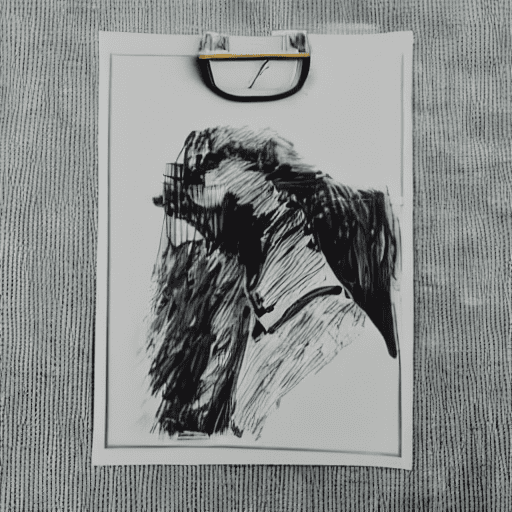}}
  \raisebox{-.5\height}{\includegraphics[width=0.48\linewidth]{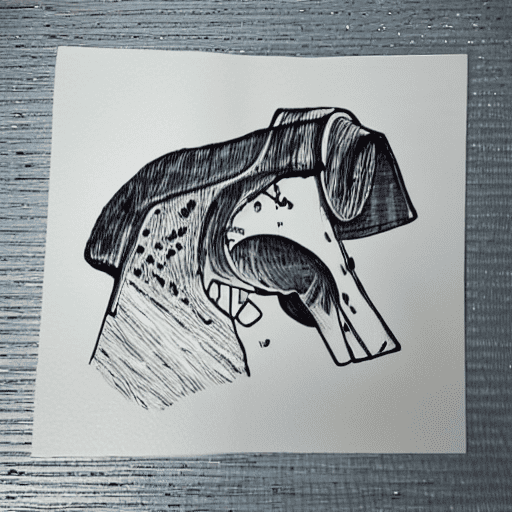}}\\
  \raisebox{-.5\height}{\includegraphics[width=0.48\linewidth]{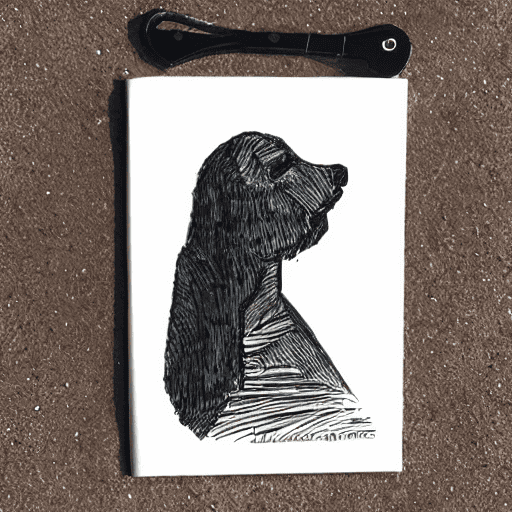}}
  \raisebox{-.5\height}{\includegraphics[width=0.48\linewidth]{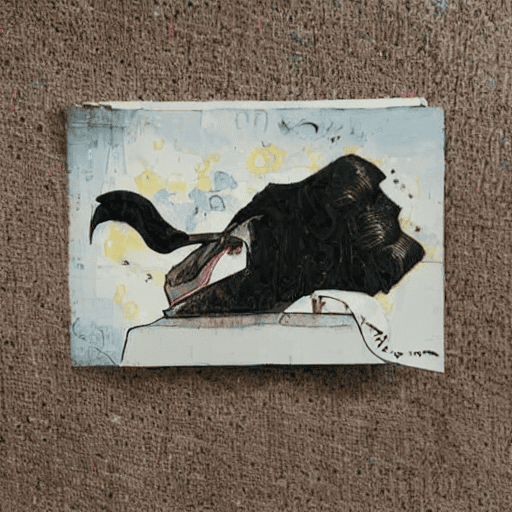}}
  \\
  \raisebox{-.5\height}{\includegraphics[width=0.48\linewidth]{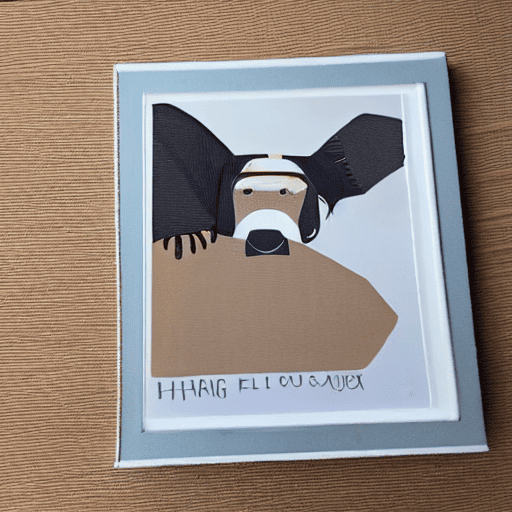}}
  \raisebox{-.5\height}{\includegraphics[width=0.48\linewidth]{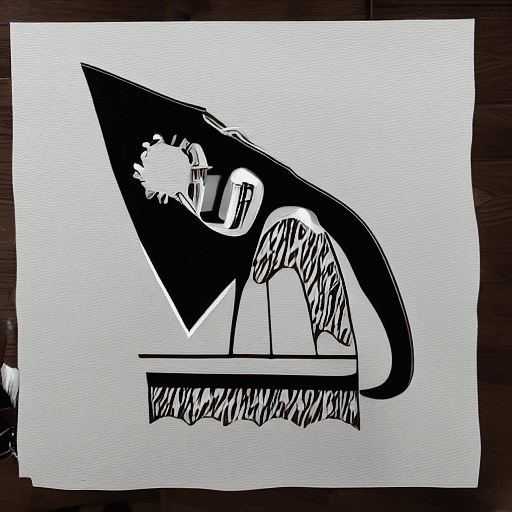}}
  \\
  \raisebox{-.5\height}{\includegraphics[width=0.48\linewidth]{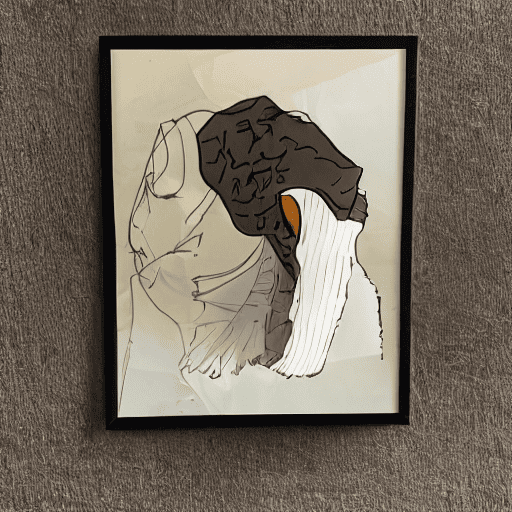}}
  \raisebox{-.5\height}{\includegraphics[width=0.48\linewidth]{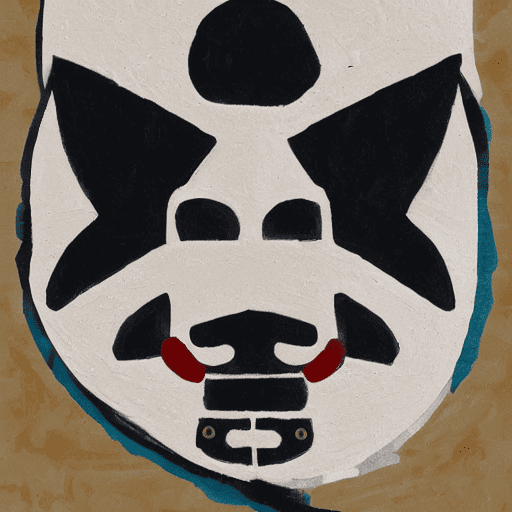}}
	\end{minipage}
    & 
    \begin{minipage}[b]{0.32\columnwidth}
		\centering
		\raisebox{-.5\height}{\includegraphics[width=0.48\linewidth]{0_0.png}}
  \raisebox{-.5\height}{\includegraphics[width=0.48\linewidth]{1_1.png}}\\
  \raisebox{-.5\height}{\includegraphics[width=0.48\linewidth]{1_2.png}}
  \raisebox{-.5\height}{\includegraphics[width=0.48\linewidth]{1_3.png}}\\
  \raisebox{-.5\height}{\includegraphics[width=0.48\linewidth]{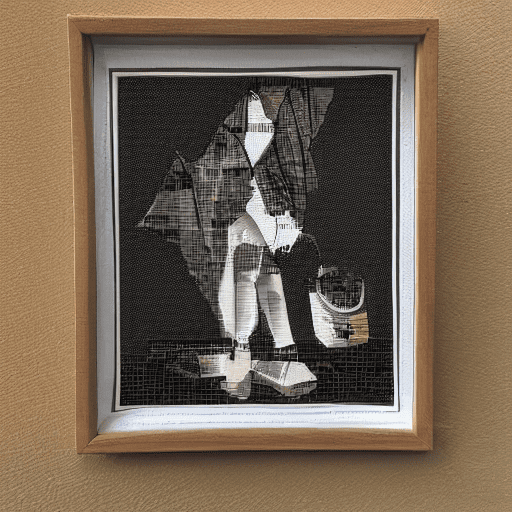}}
  \raisebox{-.5\height}{\includegraphics[width=0.48\linewidth]{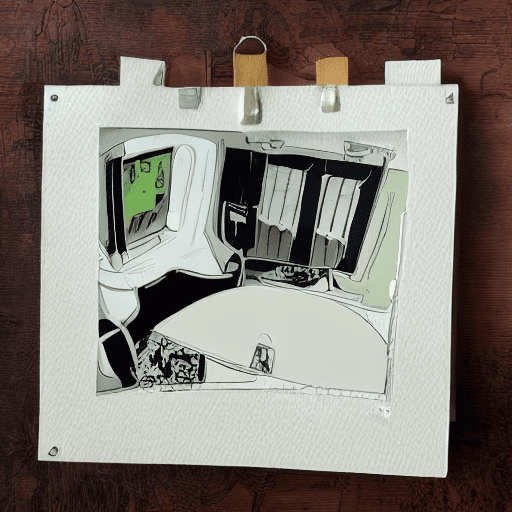}}\\
  \raisebox{-.5\height}{\includegraphics[width=0.48\linewidth]{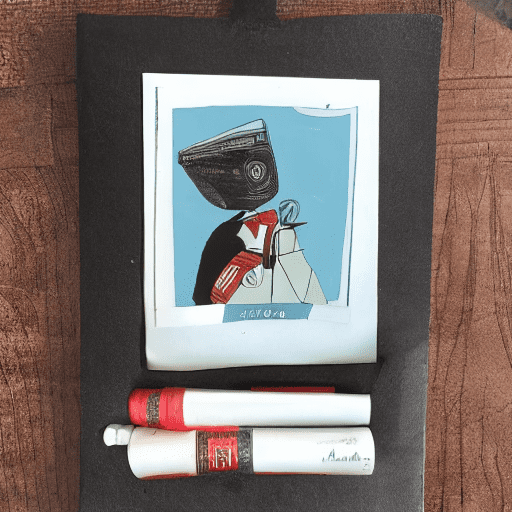}}
  \raisebox{-.5\height}{\includegraphics[width=0.48\linewidth]{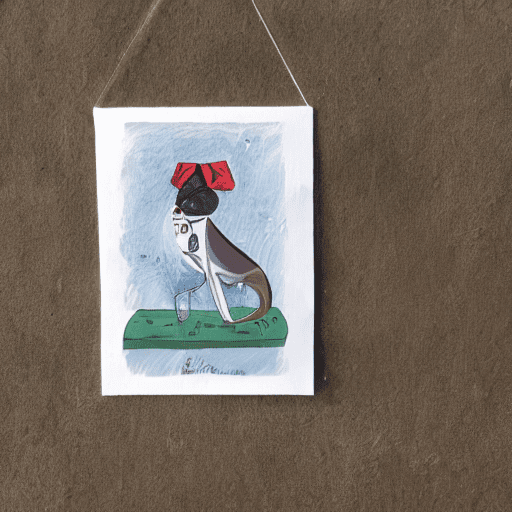}}\\
  \raisebox{-.5\height}{\includegraphics[width=0.48\linewidth]{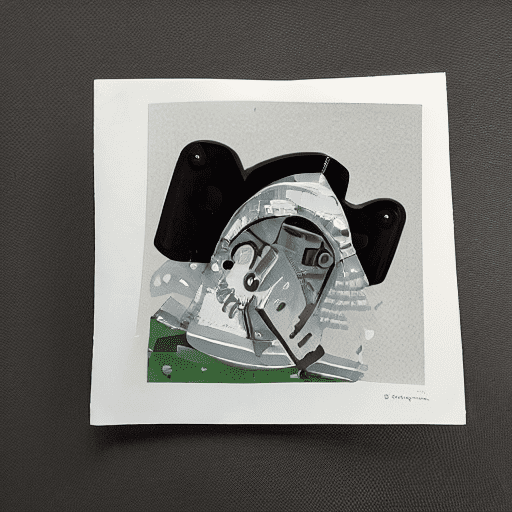}}
  \raisebox{-.5\height}{\includegraphics[width=0.48\linewidth]{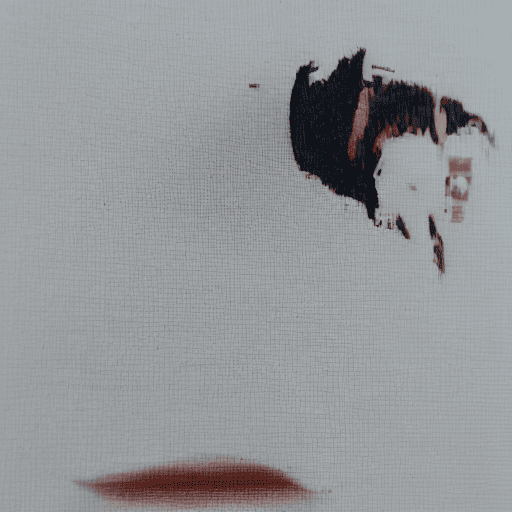}}\\
  \raisebox{-.5\height}{\includegraphics[width=0.48\linewidth]{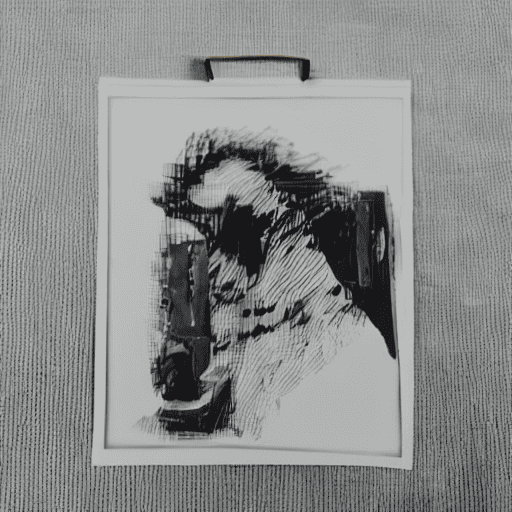}}
  \raisebox{-.5\height}{\includegraphics[width=0.48\linewidth]{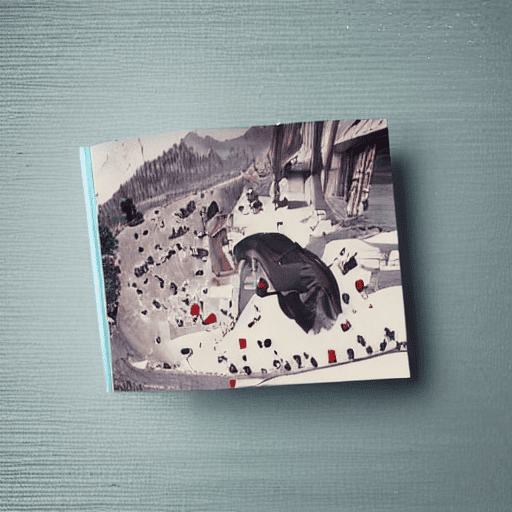}}\\
  \raisebox{-.5\height}{\includegraphics[width=0.48\linewidth]{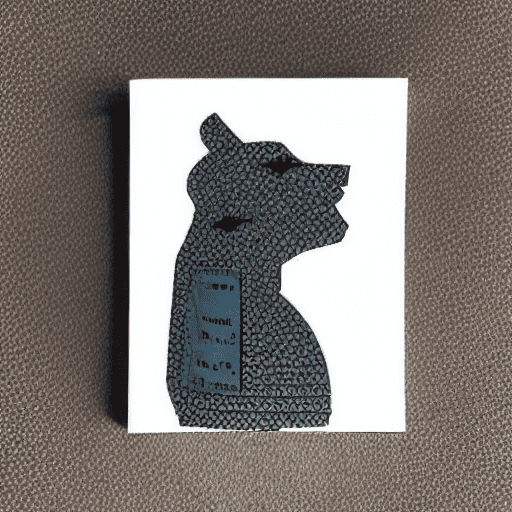}}
  \raisebox{-.5\height}{\includegraphics[width=0.48\linewidth]{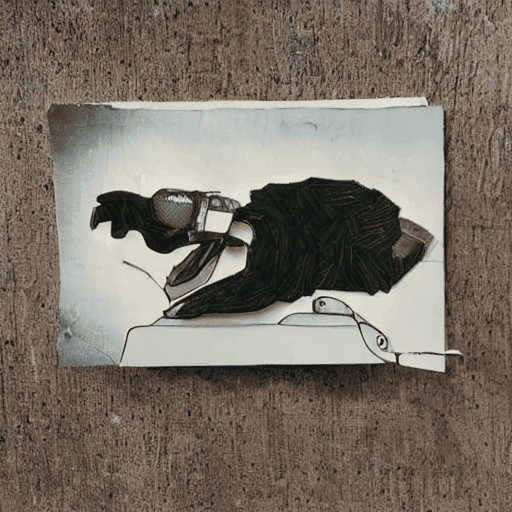}}
  \\
  \raisebox{-.5\height}{\includegraphics[width=0.48\linewidth]{1_14_cl.png}}
  \raisebox{-.5\height}{\includegraphics[width=0.48\linewidth]{1_15_cl.png}}
  \\
  \raisebox{-.5\height}{\includegraphics[width=0.48\linewidth]{1_16_cl.png}}
  \raisebox{-.5\height}{\includegraphics[width=0.48\linewidth]{1_17_cl.png}}
	\end{minipage}
    \\ 
    \bottomrule[1.5pt]
  \end{tabular}}
  \label{tab:img_english_springer}
\end{table}

\clearpage
\section{Additional Experimental Results}
\label{app:exp}

In this section, we provide additional experimental results of our work. 

- In Appendix~\ref{app:add_exp_setup}, we summarize the additional experimental setups. 

- In Appendix~\ref{app:exp_identification}, we discuss the crucial target identification in unlearning.

- In Appendix~\ref{app:exp_model}, we present unlearning with different model structures.

- In Appendix~\ref{app:full_exp_re}, we present the full results under multiple runs with the four forgetting tasks.

\subsection{Extra Experimental Setups}
\label{app:add_exp_setup}

We introduce additional experimental details in the specific unlearning tasks. In our TARF, In general, we set $t_1=1$ for all the target identification parts, and we adopt $k=0.04$, $t_0=2$ in model mismatch forgetting, and  $k=0.02$, $t_0=2$ for all matched, target mismatch and data mismatch forgetting in the unlearning request on CIFAR-10 classification task; for the CIFAR-100 classification task, we adopt $k=0.5$, $t_0=2$ in model mismatch forgetting, and $k=0.05$, $t_0=2$ for all matched, target mismatch and data mismatch forgetting. For the other hyperparameters, we follow the previous works~\cite{jia2023model,kurmanji2023towards,fan2023salun} to set the specific values. All the forgetting trails use 10 epochs for the total unlearning process except for GA (use 5 epochs) and IU (use the specific fixed step for optimization). The specific parameters and the pre-trained models (unlearn base) will be provided in our codes.

\subsection{Discussion about Target Identification in unlearning}
\label{app:exp_identification}

\begin{figure*}[t!]
    \begin{center}
    \subfigure[Inter-classes distance in the model trained by classes]{
    \includegraphics[scale=0.14]{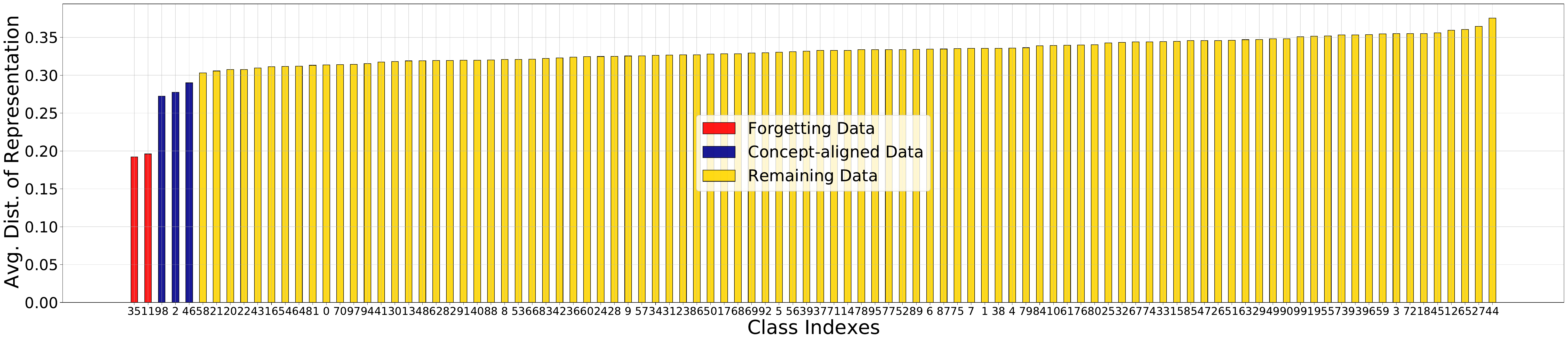}
    }\\
    \subfigure[Inter-superclass distance in the model trained by superclass]{
    \includegraphics[scale=0.14]{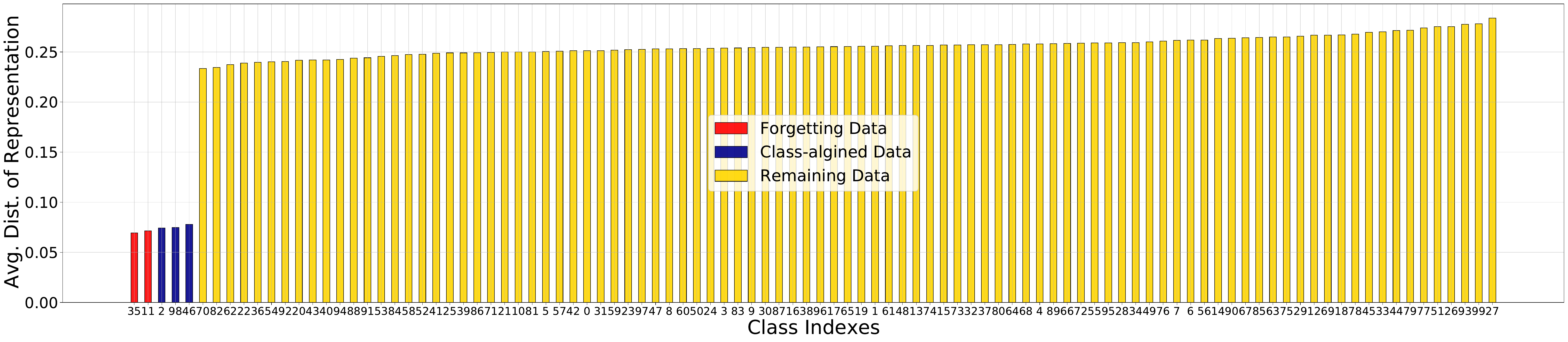}
    }
    \end{center}
    \footnotesize{The distance is calculated at the feature representation extracted from the penultimate layer of the model for each class, which measures the averaged Euclidean distance to the cluster center (averaged by the forgetting data).}
    \caption{\textbf{Inter-class distance and Inter-superclass distance for the unlearning assumption.}
    }
    \label{fig:app_task_assumption}
\end{figure*}

\begin{figure*}[t!]
    \begin{center}
    \subfigure[aquatic mammals]{
    \includegraphics[scale=0.1]{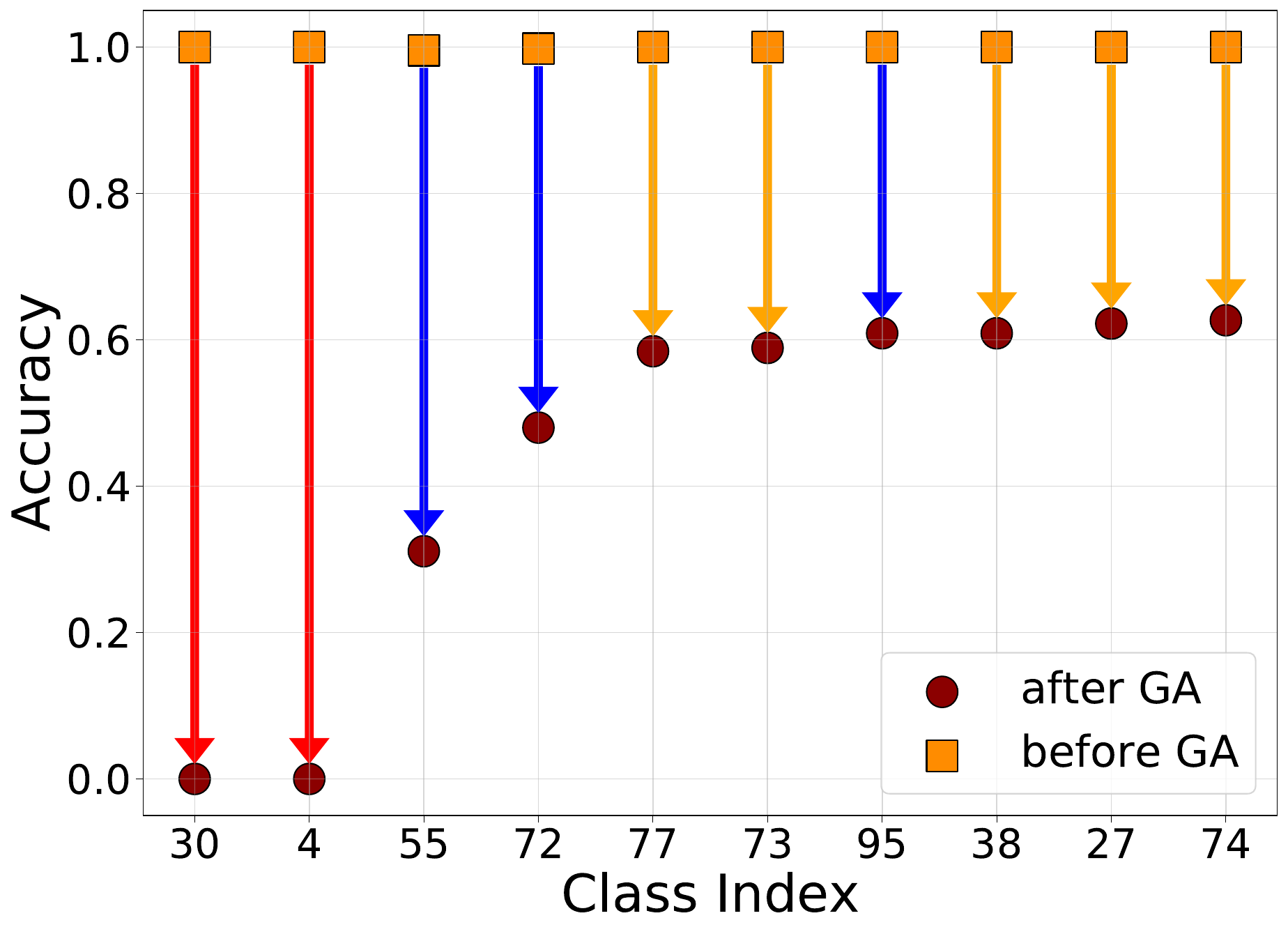}
    }
    \subfigure[fish]{
    \includegraphics[scale=0.1]{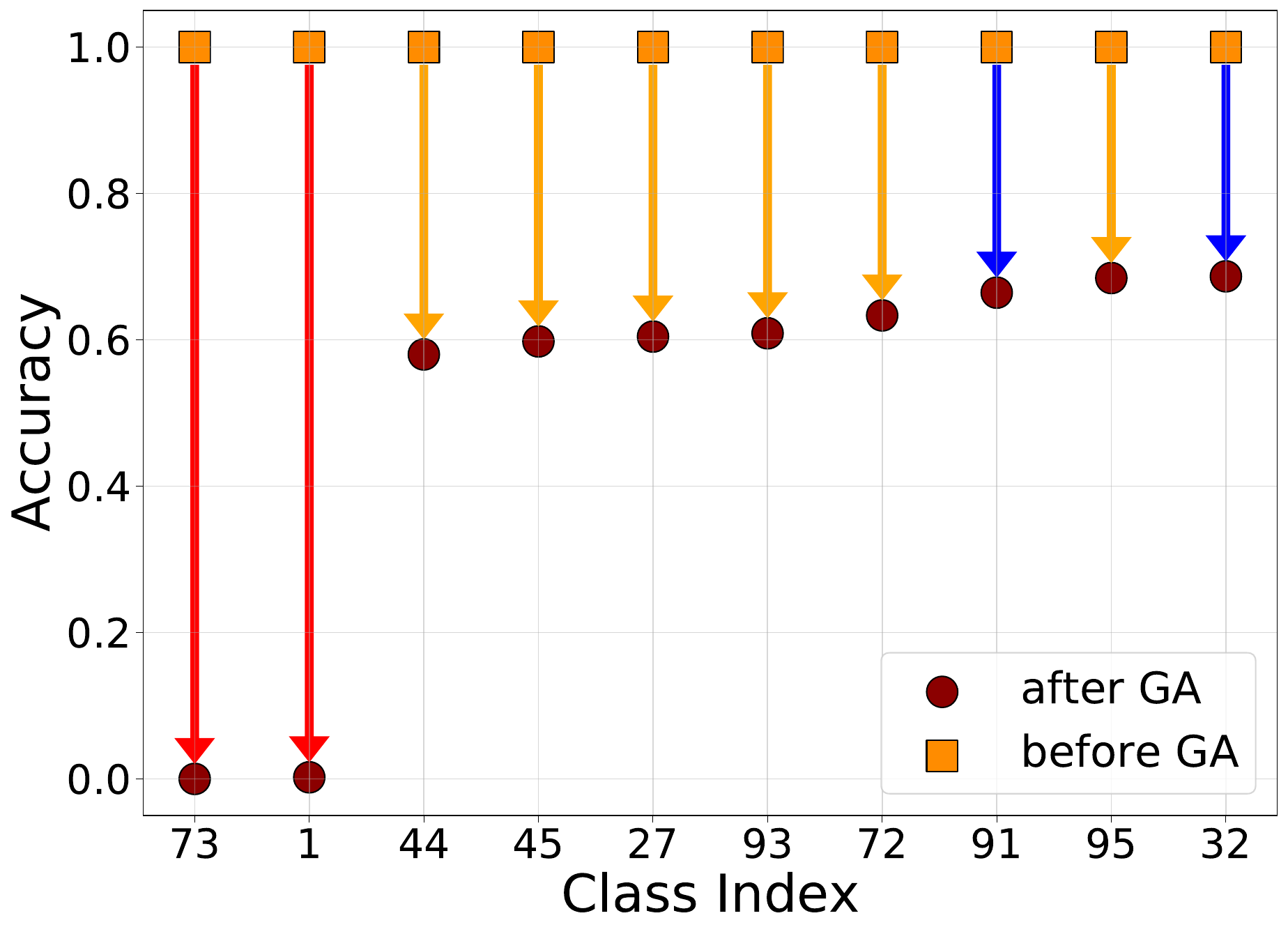}
    }
    \subfigure[flowers]{
    \includegraphics[scale=0.1]{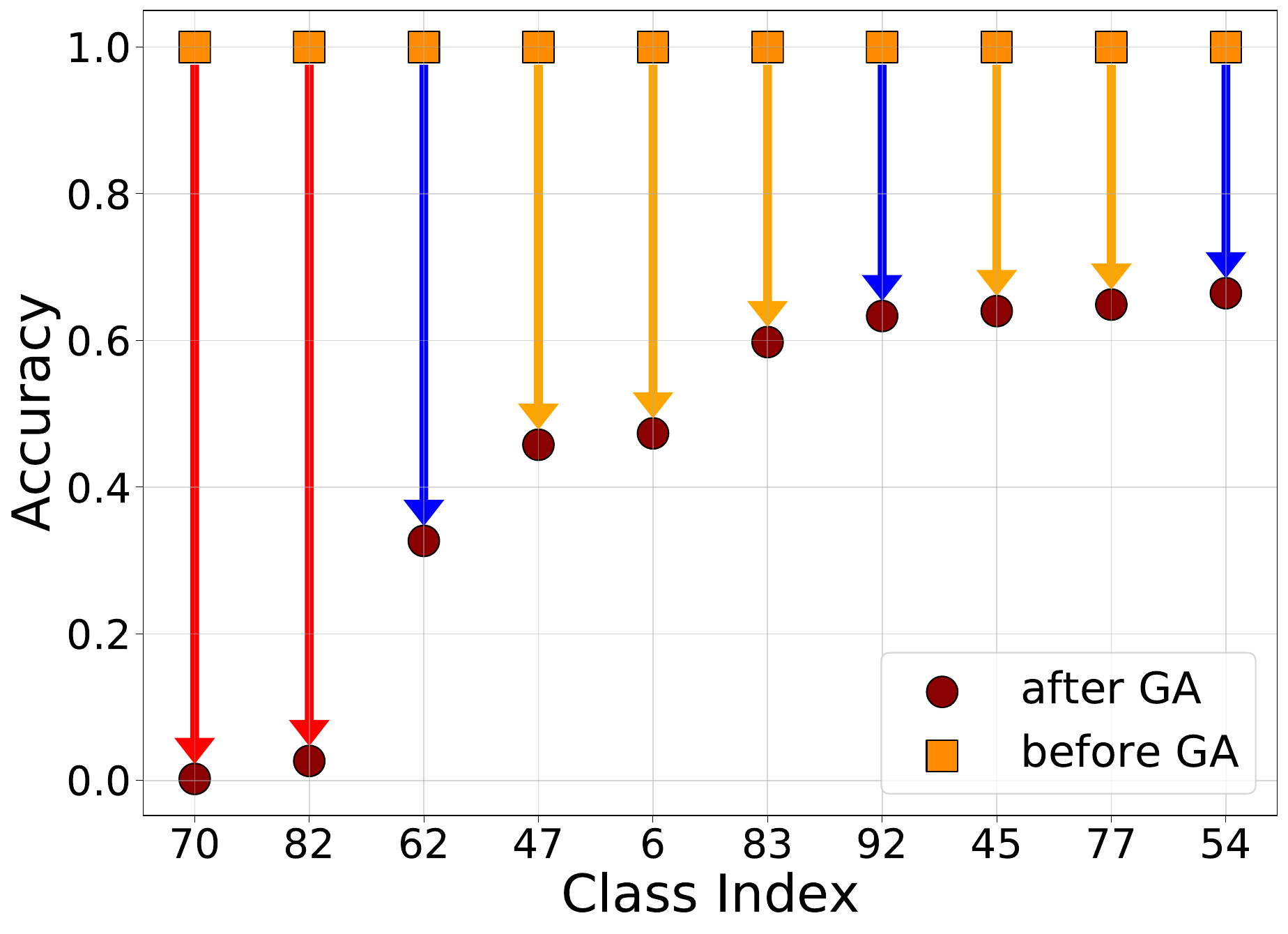}
    }
    \subfigure[food containers]{
    \includegraphics[scale=0.1]{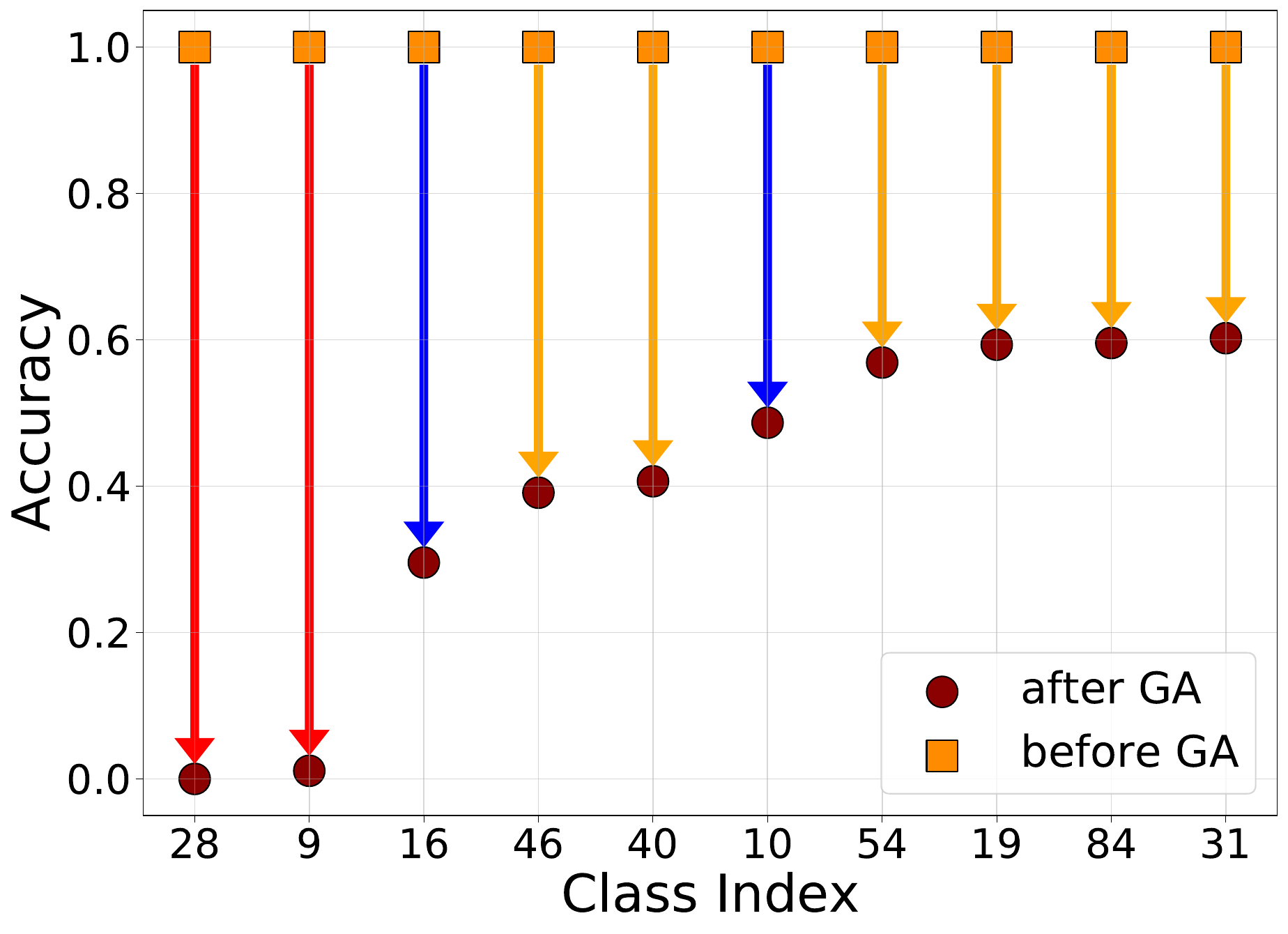}
    }\\
    \subfigure[fruit and vegetables]{
    \includegraphics[scale=0.1]{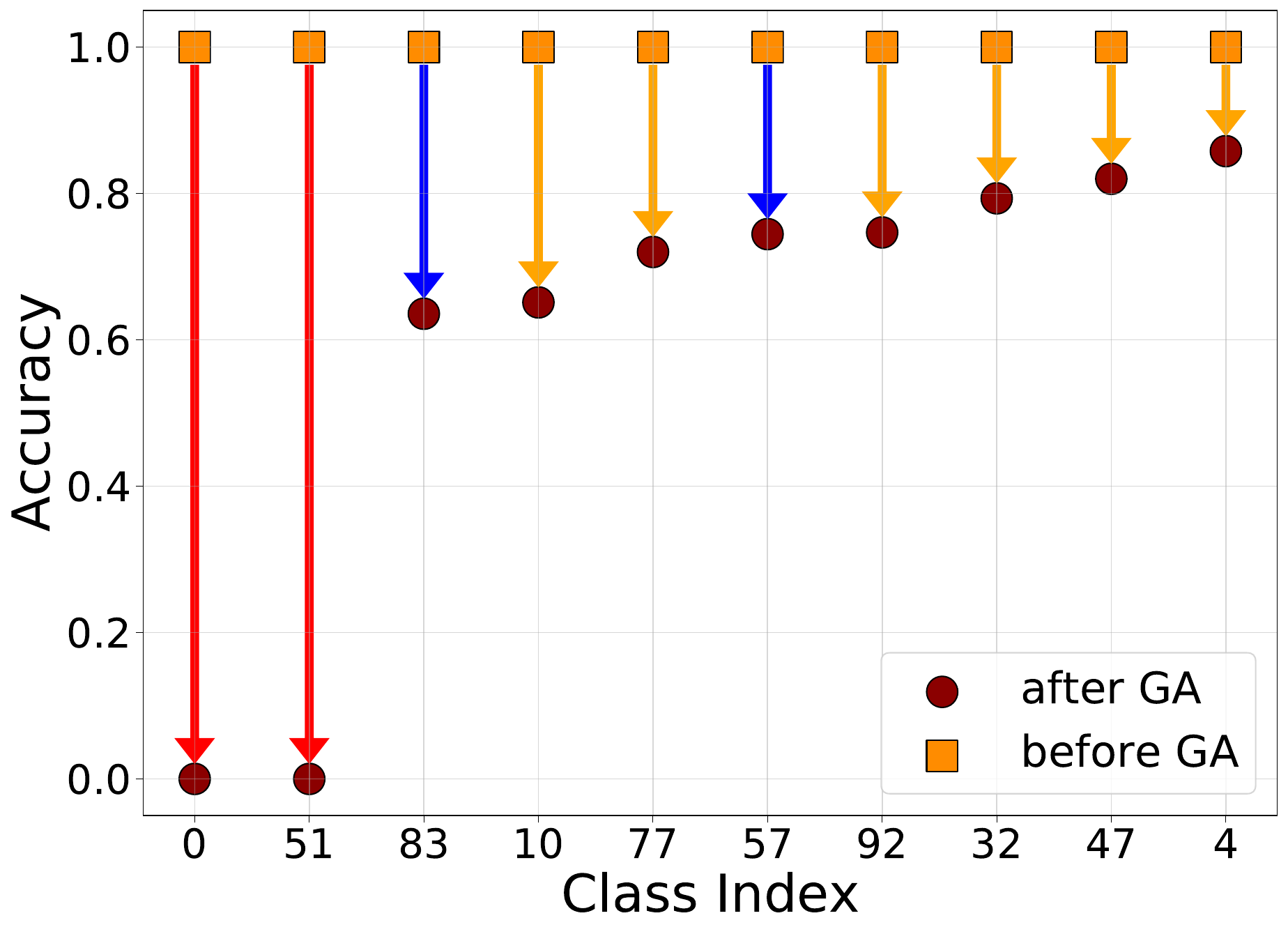}
    }
    \subfigure[electrical devices]{
    \includegraphics[scale=0.1]{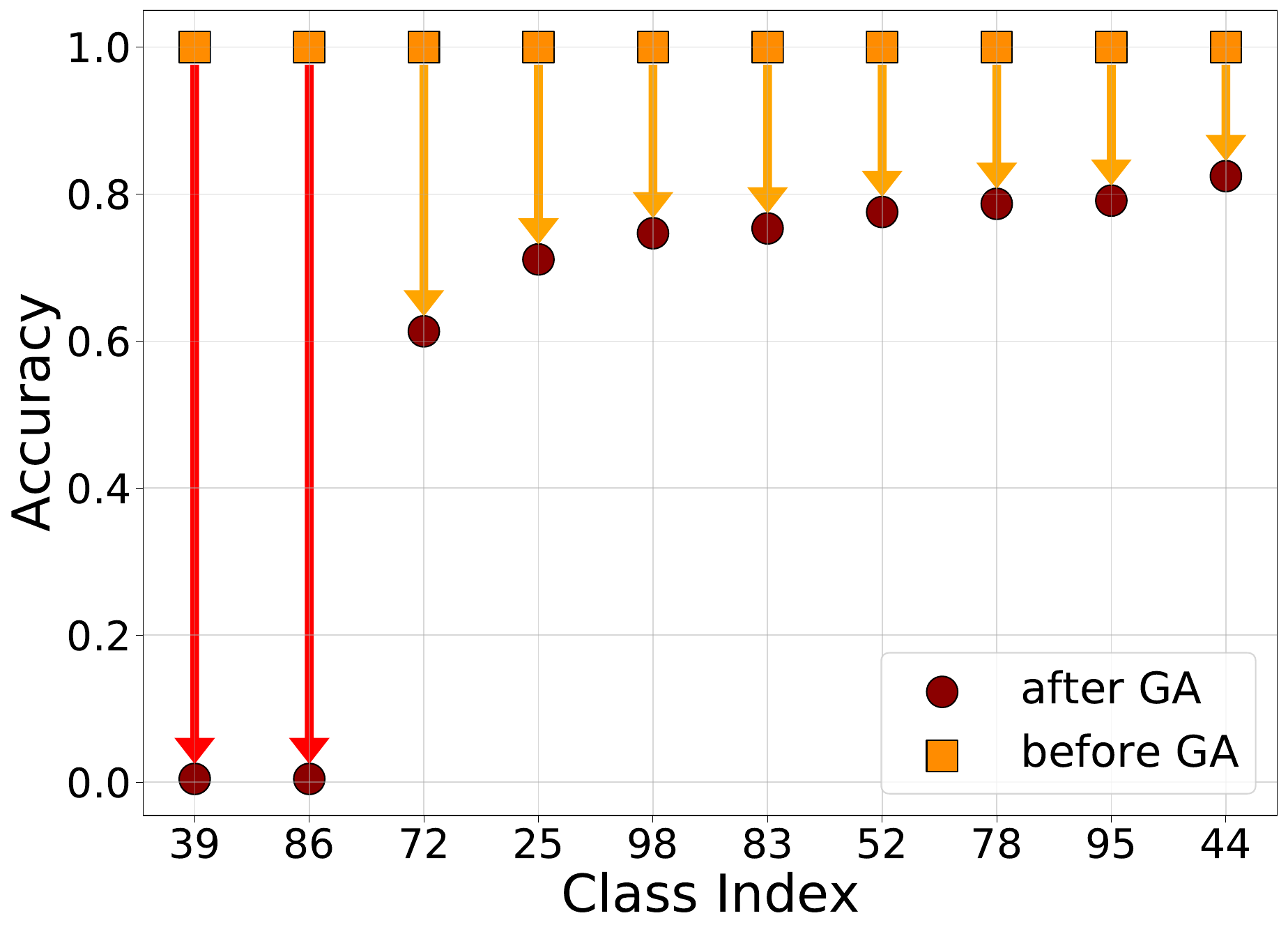}
    }
    \subfigure[household furniture]{
    \includegraphics[scale=0.1]{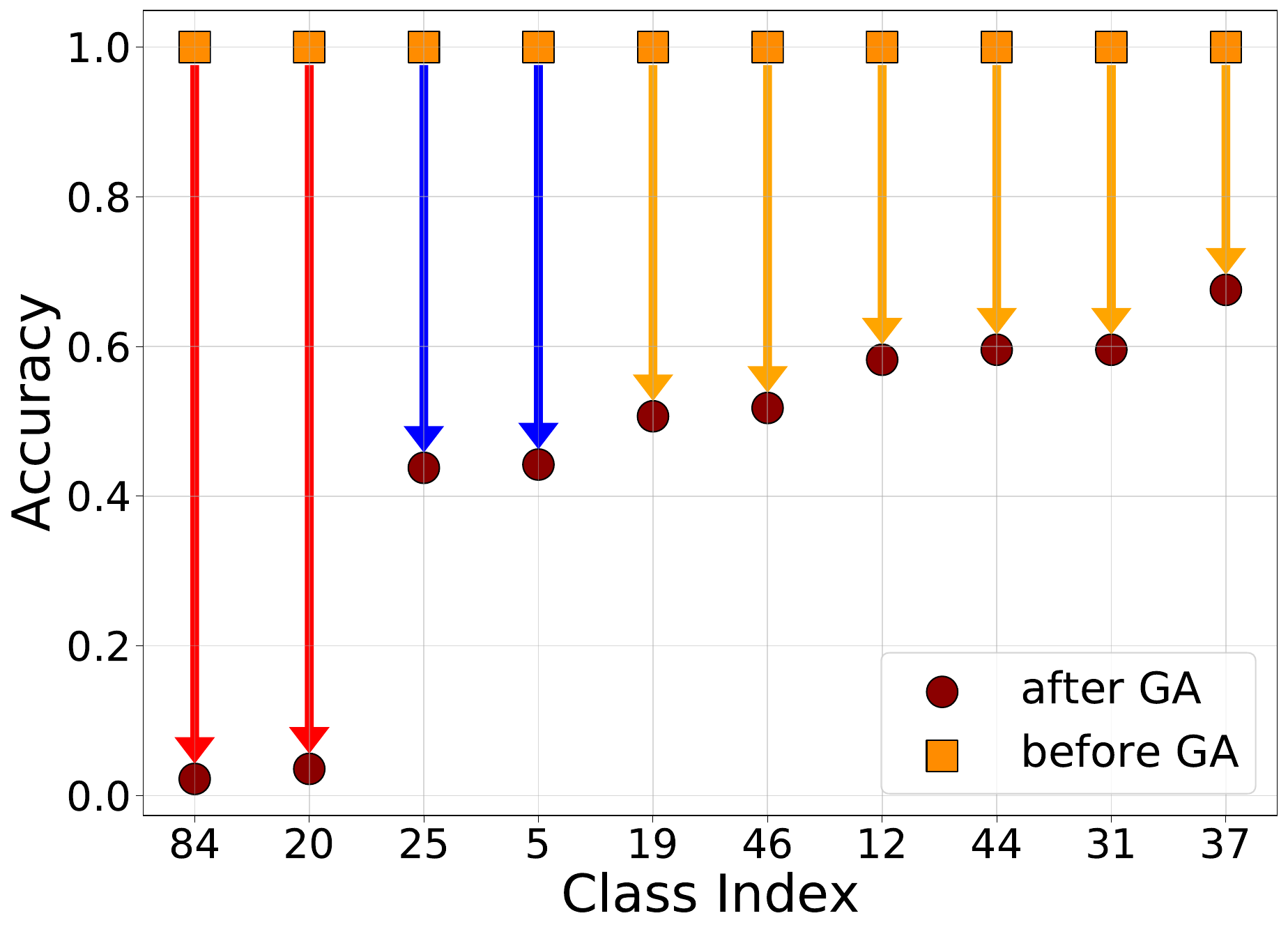}
    }
    \subfigure[insects]{
    \includegraphics[scale=0.1]{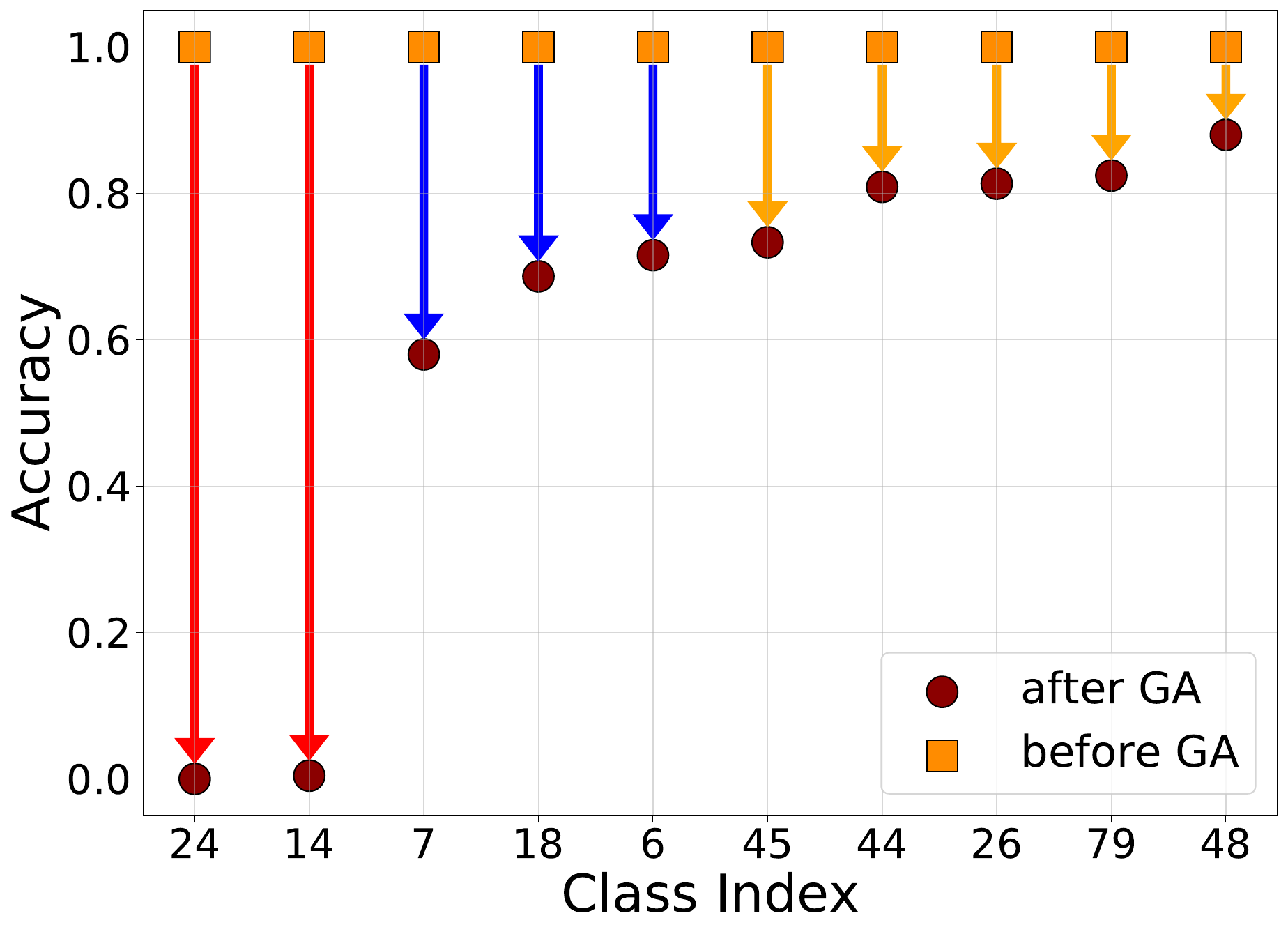}
    }\\
    \subfigure[large carnivores]{
    \includegraphics[scale=0.1]{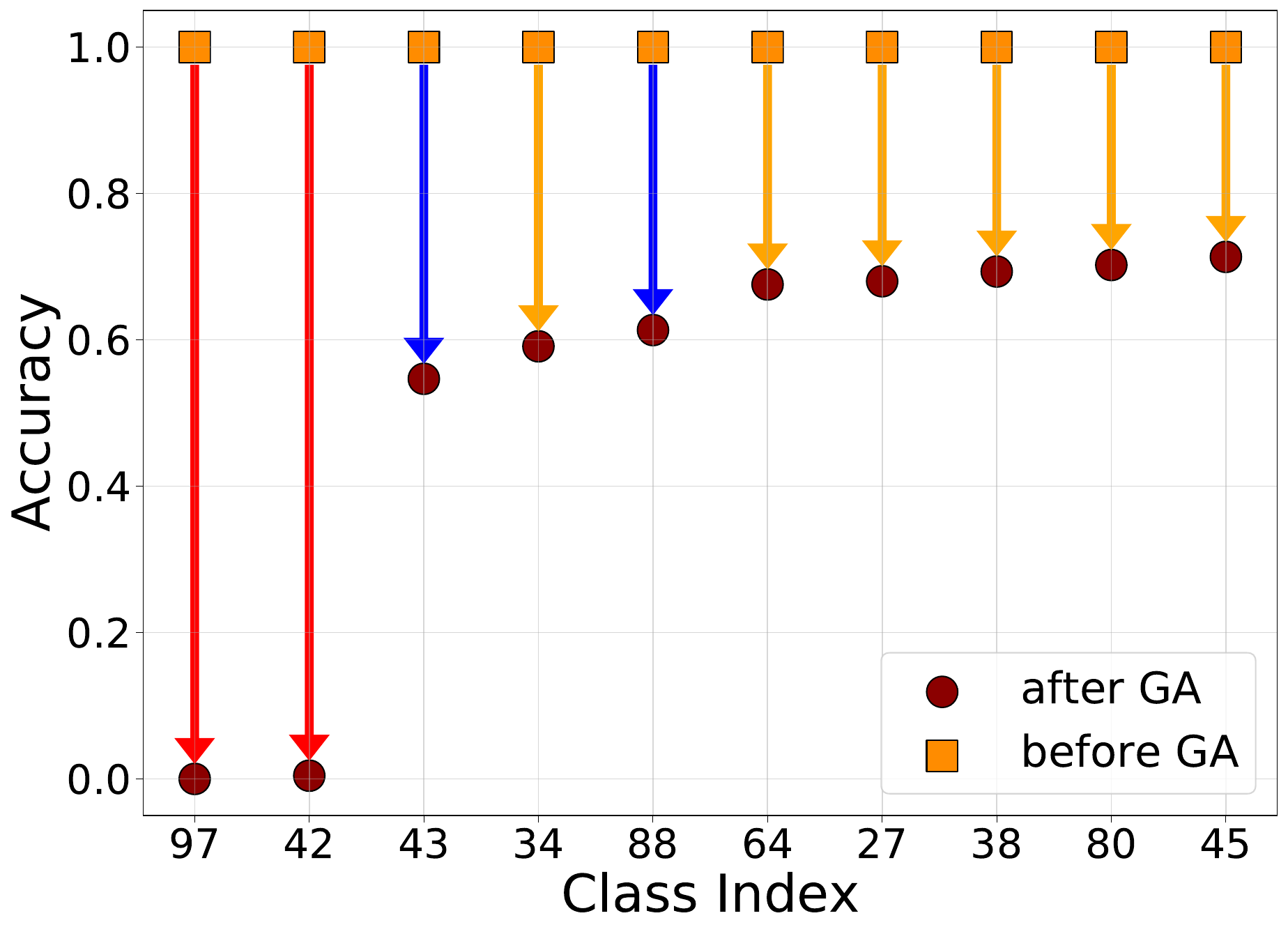}
    }
    \subfigure[large man-made outdoor things]{
    \includegraphics[scale=0.1]{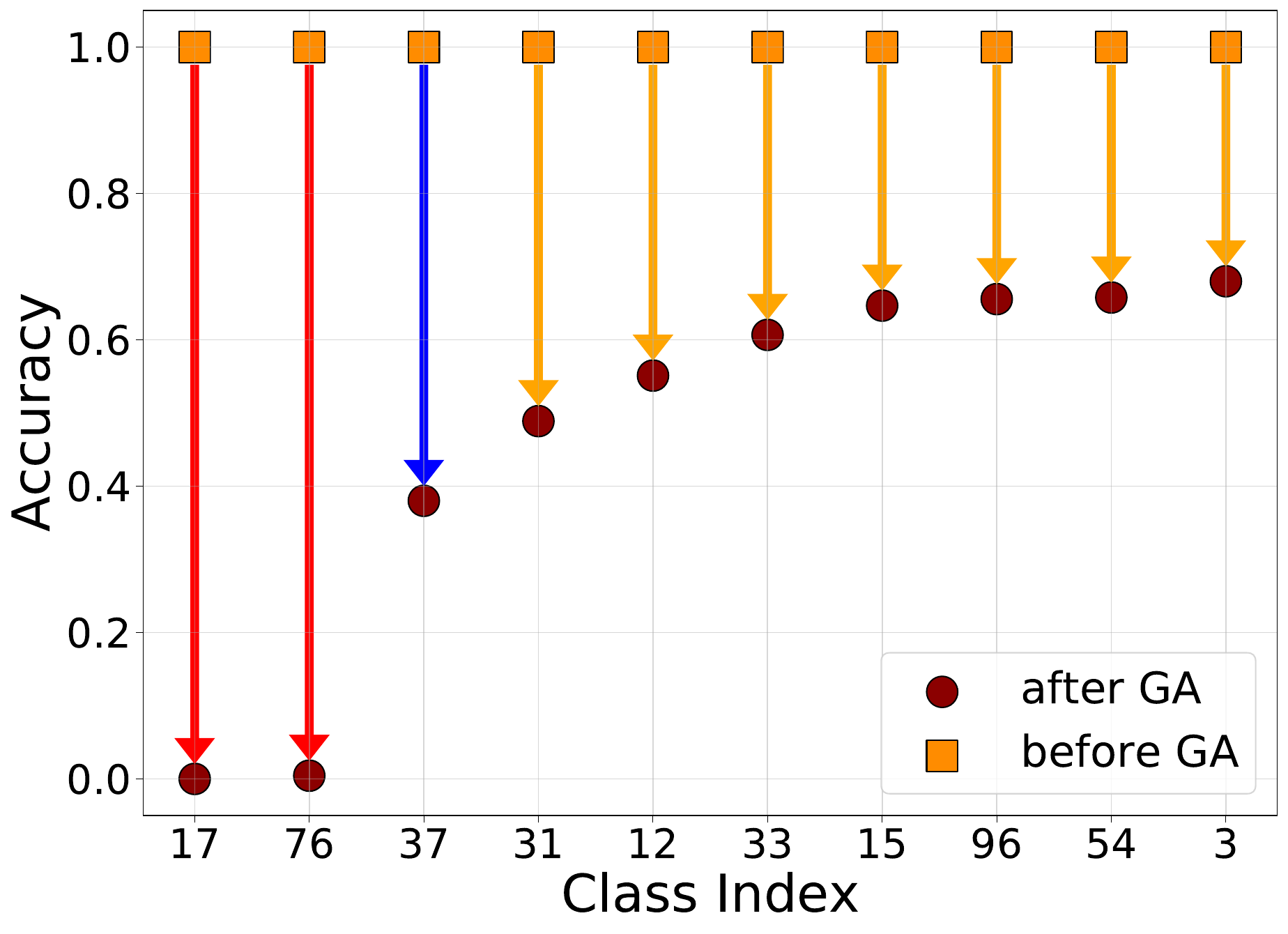}
    }
    \subfigure[large natural outdoor scenes]{
    \includegraphics[scale=0.1]{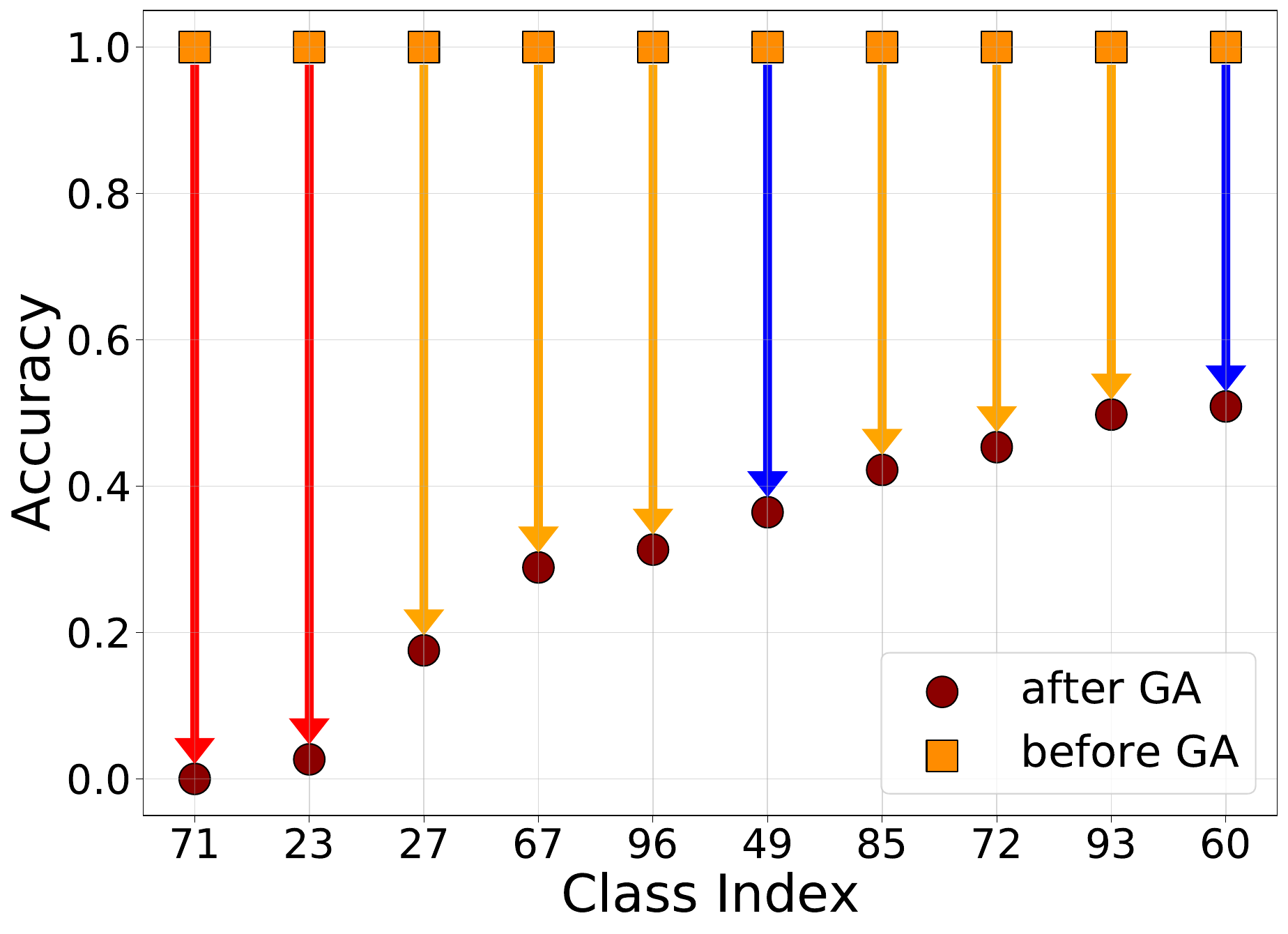}
    }
    \subfigure[large omnivores and herbivores]{
    \includegraphics[scale=0.1]{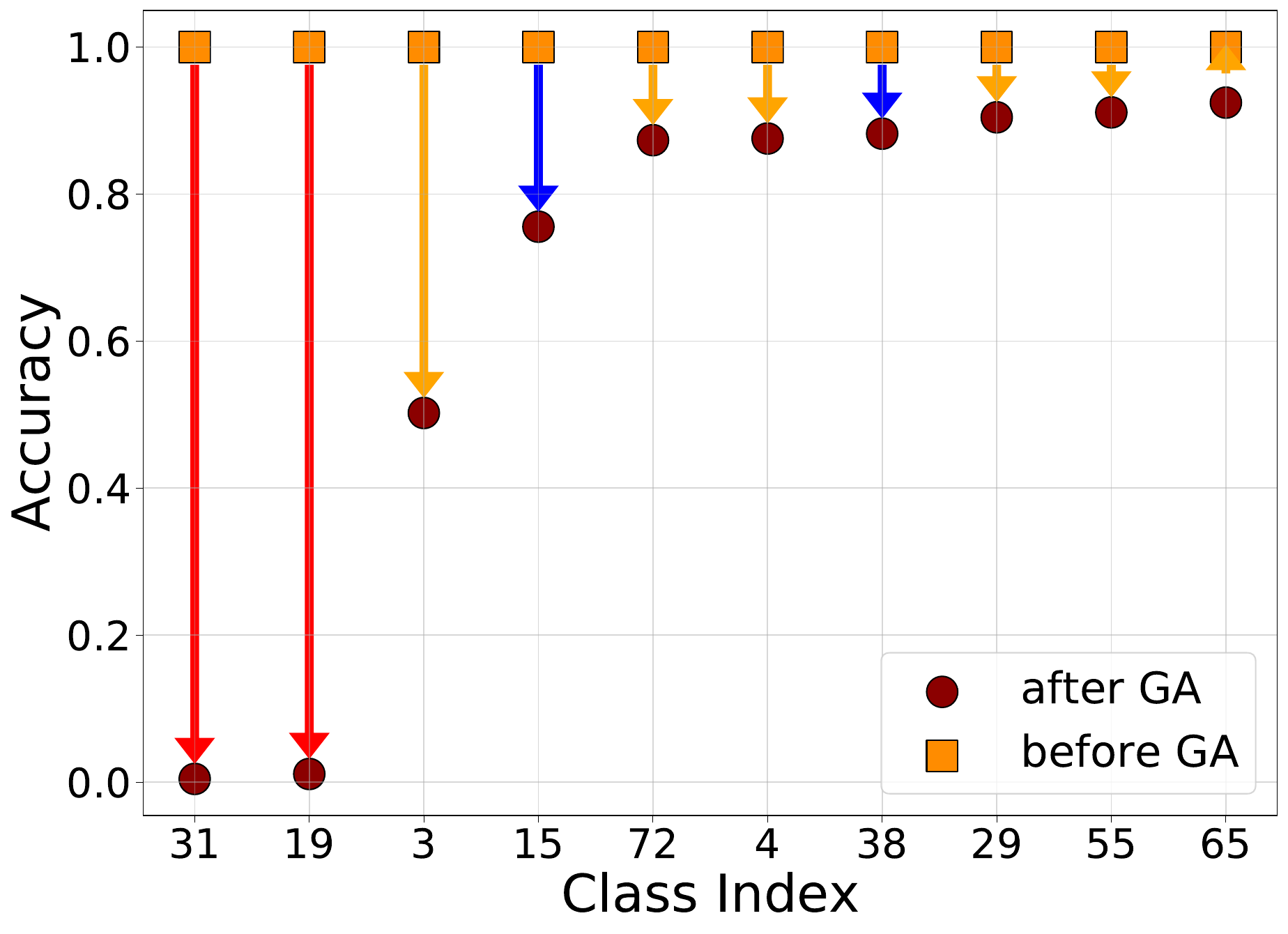}
    }\\
    \subfigure[medium-sized mammals]{
    \includegraphics[scale=0.1]{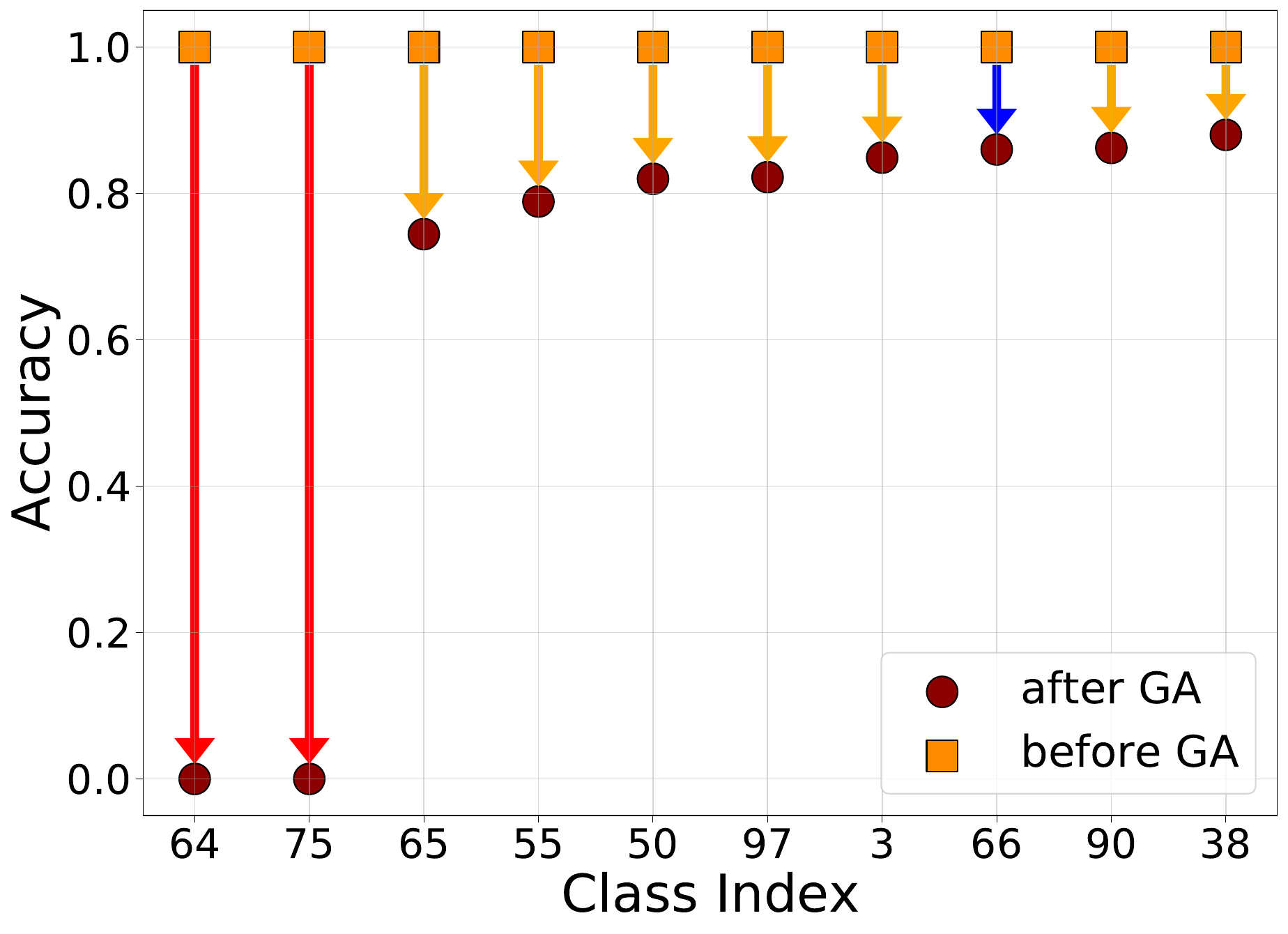}
    }
    \subfigure[non-insect invertebrates]{
    \includegraphics[scale=0.1]{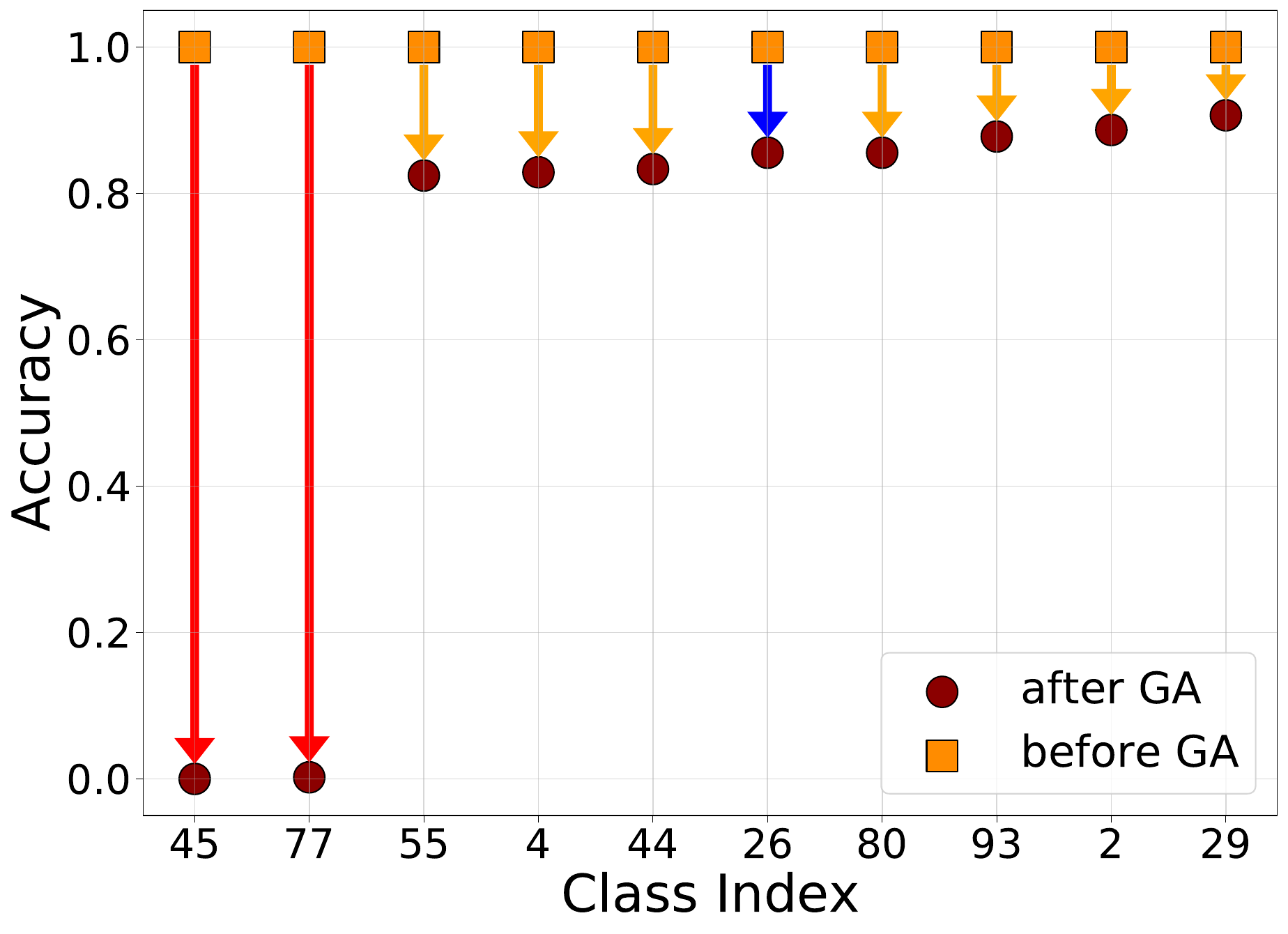}
    }
    \subfigure[people]{
    \includegraphics[scale=0.1]{fig5_4.pdf}
    }
    \subfigure[reptiles]{
    \includegraphics[scale=0.1]{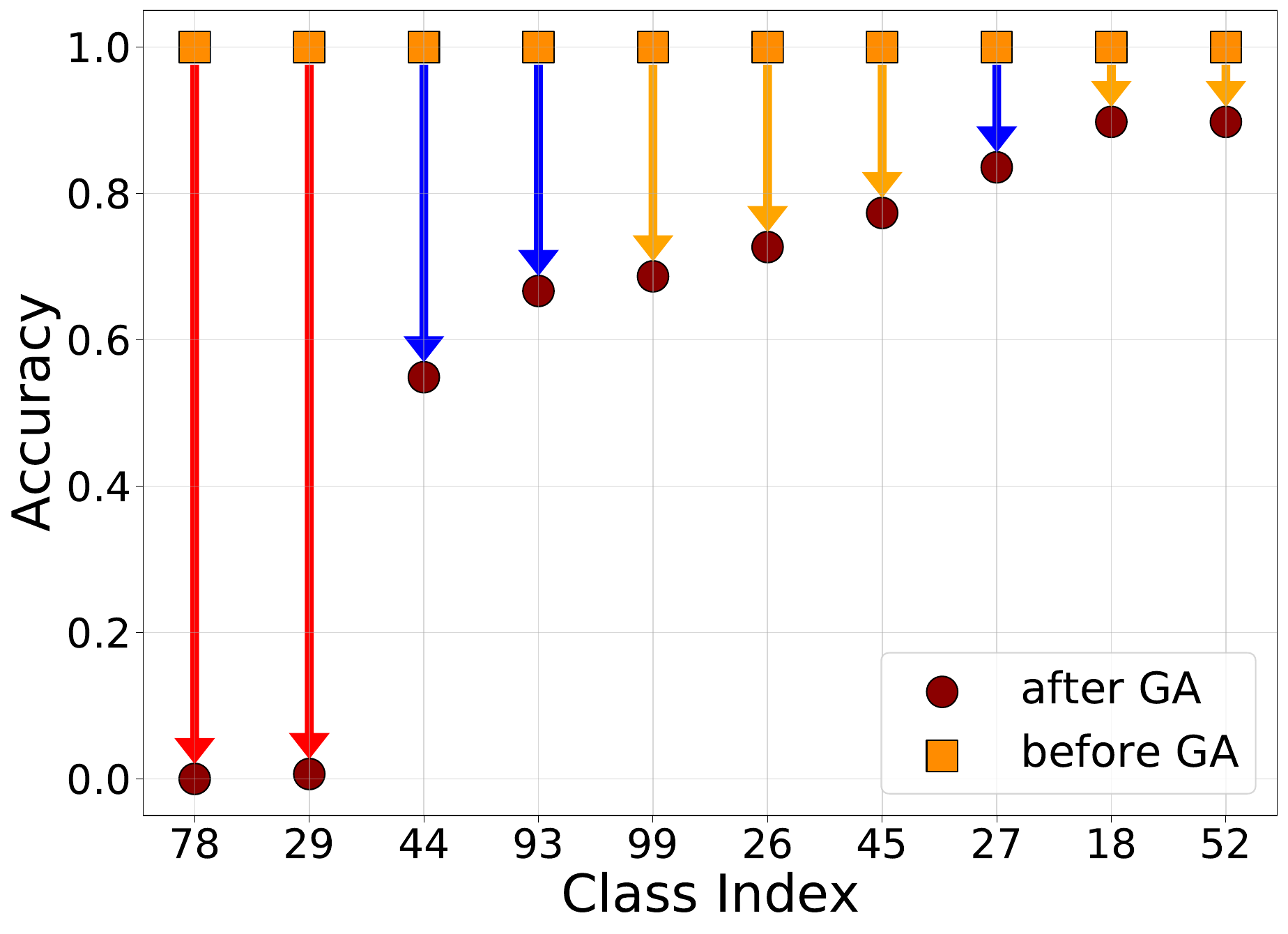}
    }\\
    \subfigure[small mammals]{
    \includegraphics[scale=0.1]{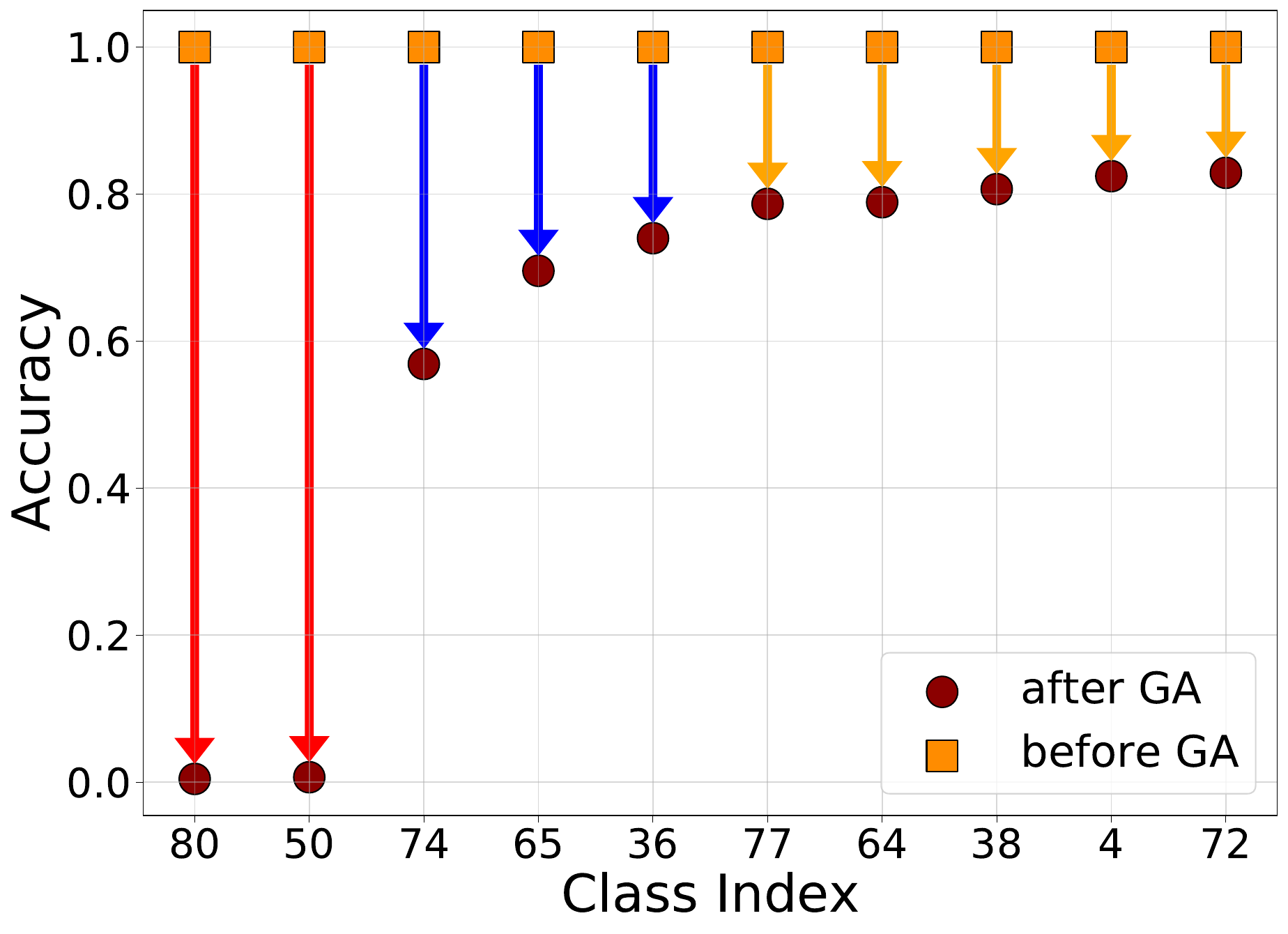}
    }
    \subfigure[trees]{
    \includegraphics[scale=0.1]{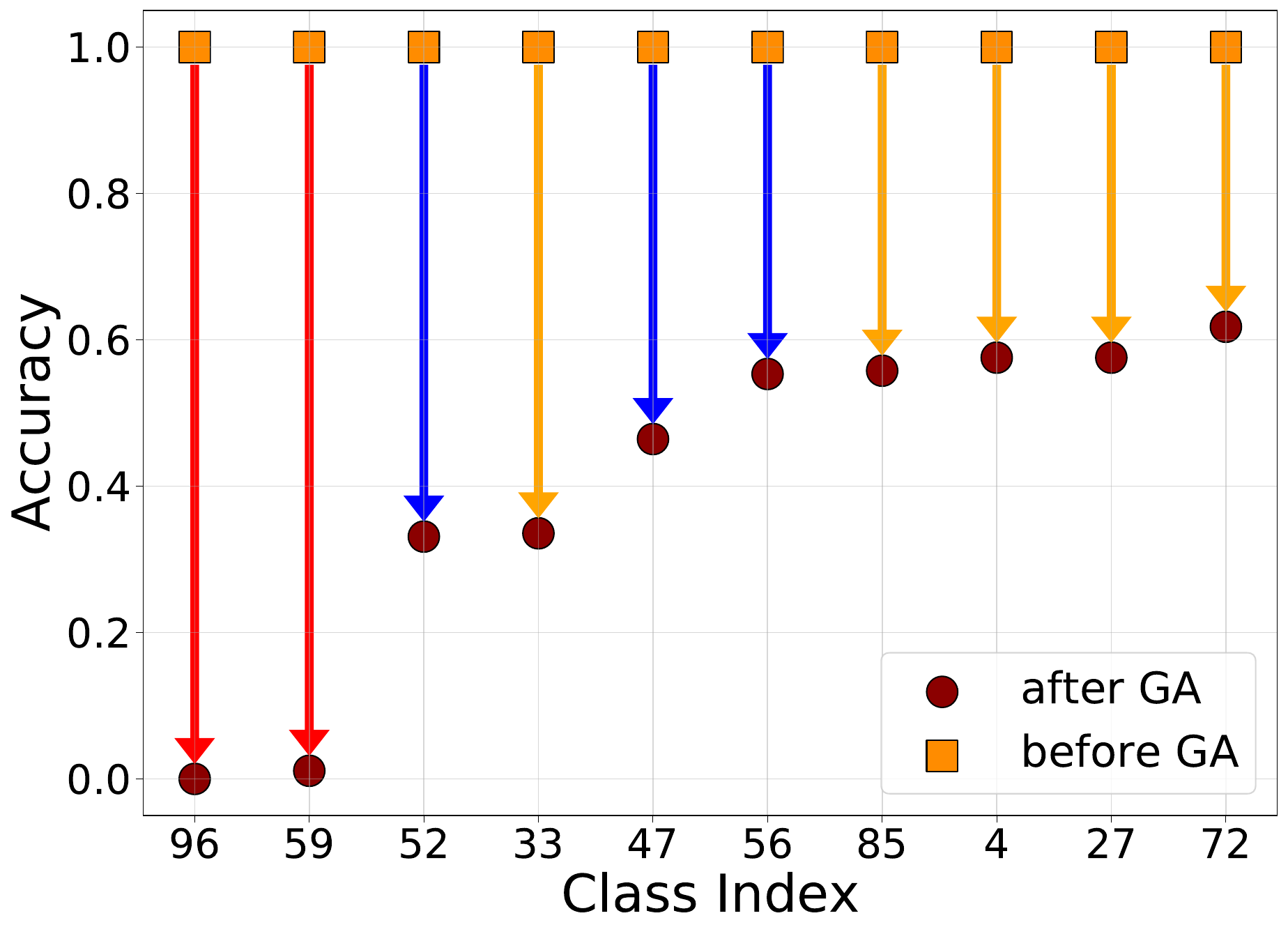}
    }
    \subfigure[vehicles 1]{
    \includegraphics[scale=0.1]{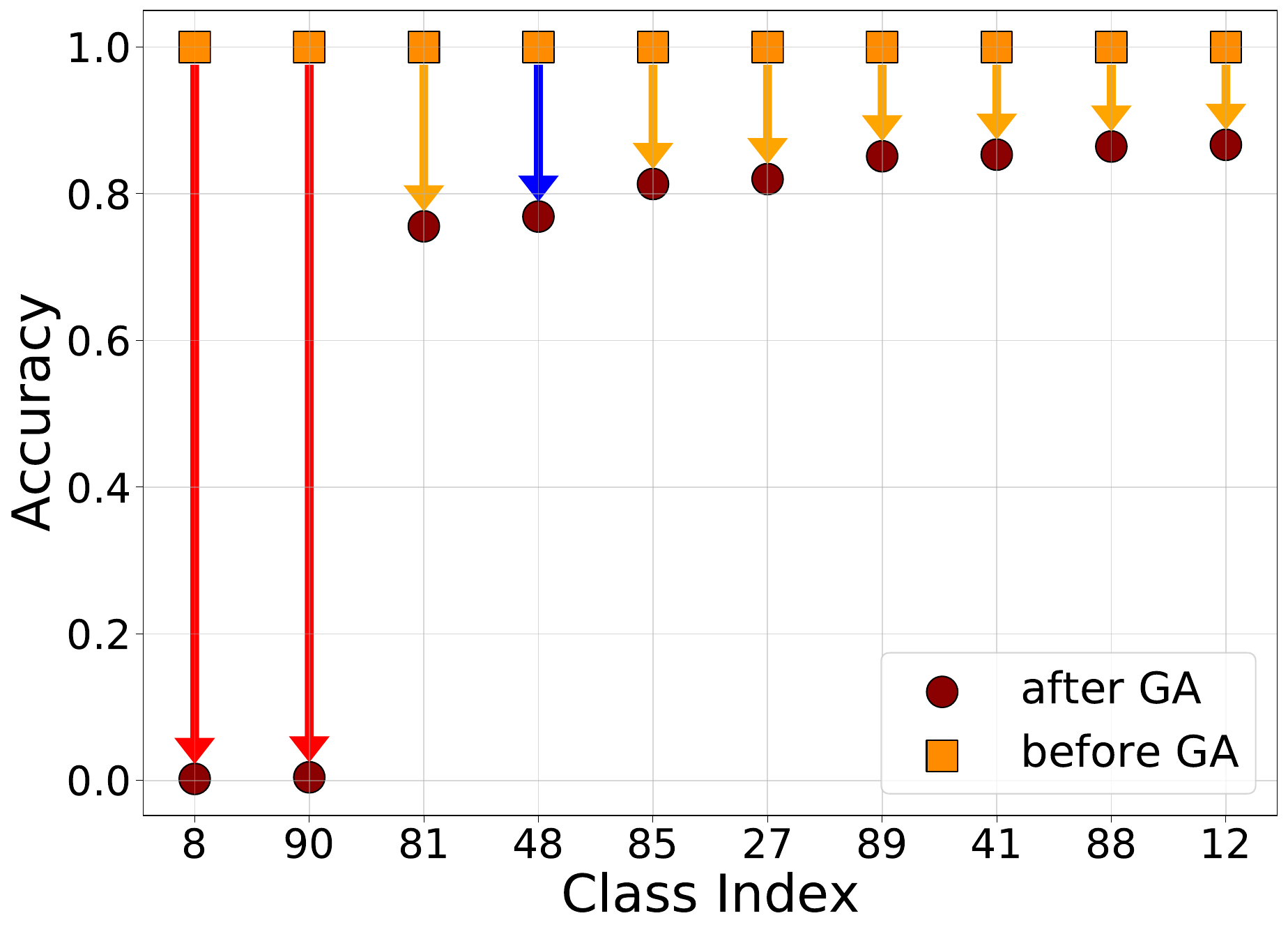}
    }
    \subfigure[vehicles 2]{
    \includegraphics[scale=0.1]{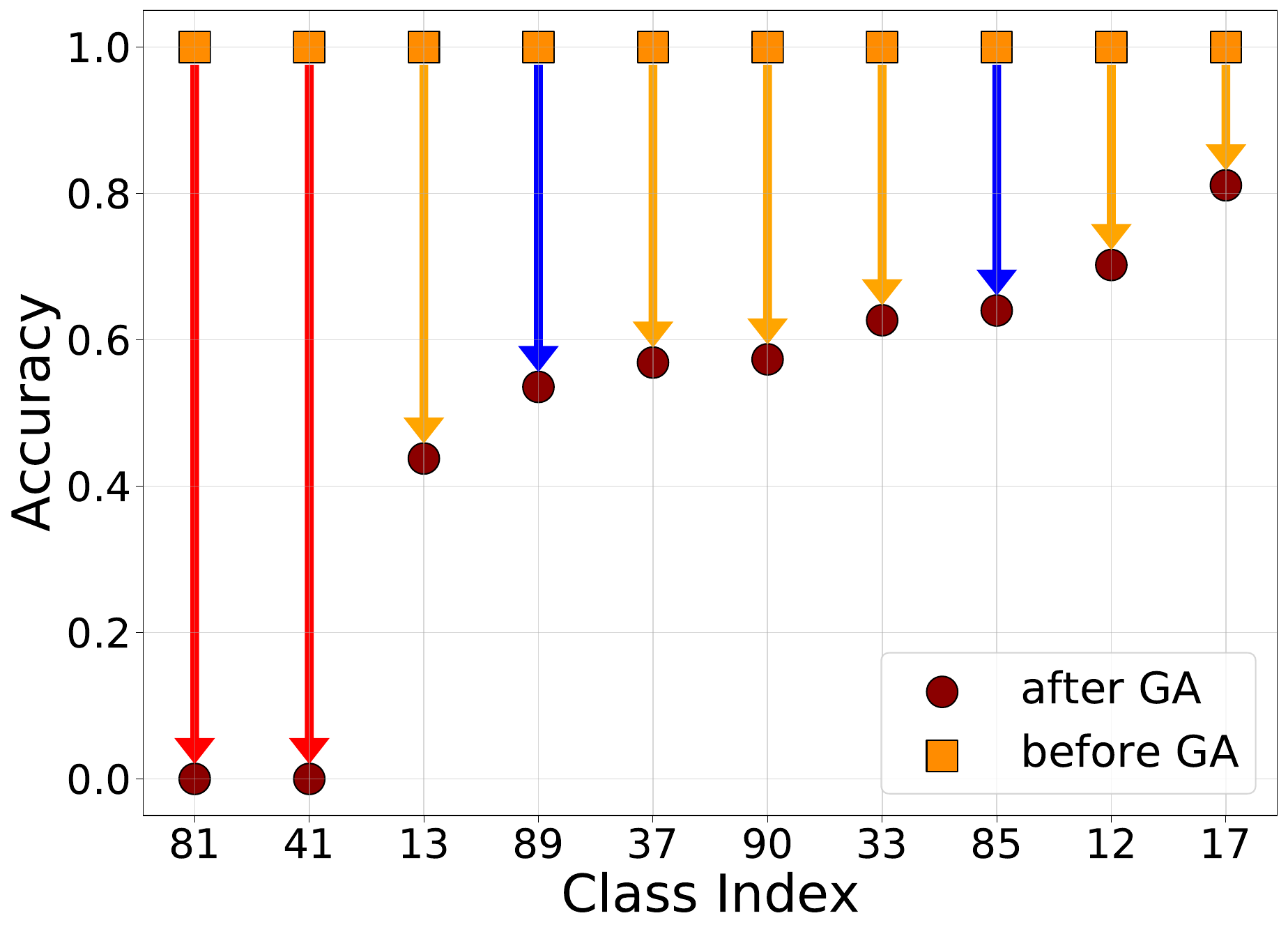}
    }
    \end{center}
    \footnotesize{Target identification results with different unlearning requests and the minimum identified forgetting data on the CIFAR-100 dataset. Note that some target concepts are not successfully identified by the identified data.}
    \caption{\textbf{Task Identification using the CIFAR-100 dataset for target mismatch forgetting.}
    }
    \vspace{-2mm}
    \label{fig:app_task_identification}
\end{figure*}

In this part, we further discuss the important factors for the achievability of the unlearning tasks. To be more specific, for the target or data mismatch forgetting, the scenario assumes that the identified forgetting data is part of the whole samples belonging to the target concept, which means there are other forgetting data included in the remaining set that need to be found. Thus, target identification is important for effective unlearning. As demonstrated in Section~\ref{sec:method_analyze}, the representation gravity can be a useful clue in forgetting dynamics to identify the other concept-aligned forgetting data. An implicit assumption is that those concept-aligned forgetting data have similar semantic features to the initially provided forgetting data, which has smaller representation distance than the retaining part of data as illustrated in Figure~\ref{fig:app_task_assumption}. Empirically, the model can have similar prediction changes on those concept-aligned forgetting data with the initial forgetting data. However, not all of the superclasses officially defined for the CIFAR-100 dataset are suitable for constructing the unlearning request, as some superclasses are not semantically separable like "aquatic mammals" and "fish". It can be found in Figure~\ref{fig:app_task_identification}, where we check the Top-10 classes with the most accuracy changes after gradient ascent for each superclass in the CIFAR-100 dataset, some concept-aligned forgetting data (class-level indicated by blue arrows) are not easily identified given the two initially provided forgetting data classes (indicated by red arrows). One interesting future problem can be how to handle the spurious correlation given the insufficient representative samples.

\begin{figure*}[t!]
    \begin{center}
    \hspace{0.04in}
    \includegraphics[scale=0.145]{fig6_tsne_cifar100_new_3_pre.jpg}
    \includegraphics[scale=0.145]{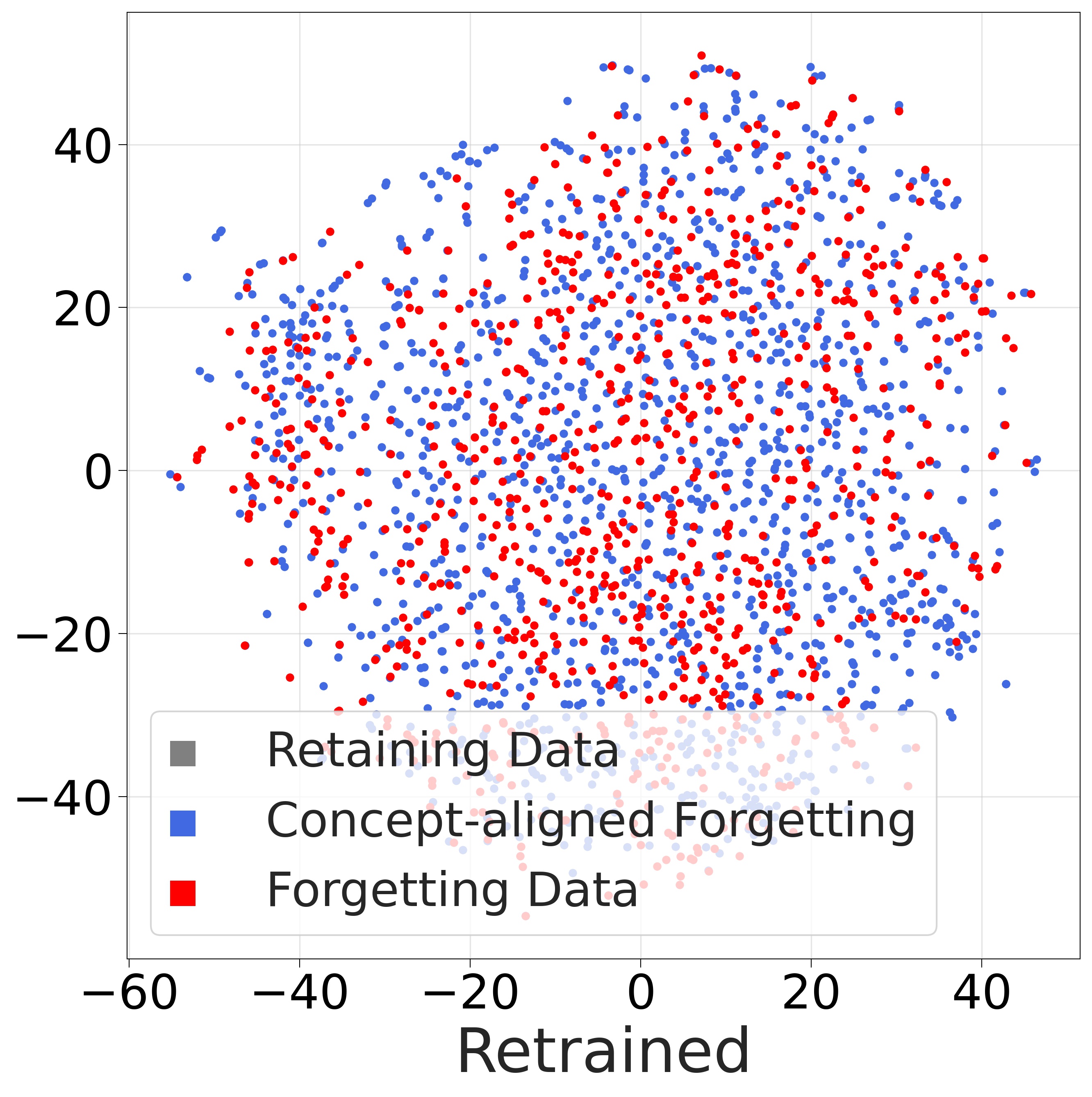}
    \hspace{0.08in}
    \includegraphics[scale=0.145]{fig6_tsne_cifar100_new_0_pre.jpg}
    \includegraphics[scale=0.145]{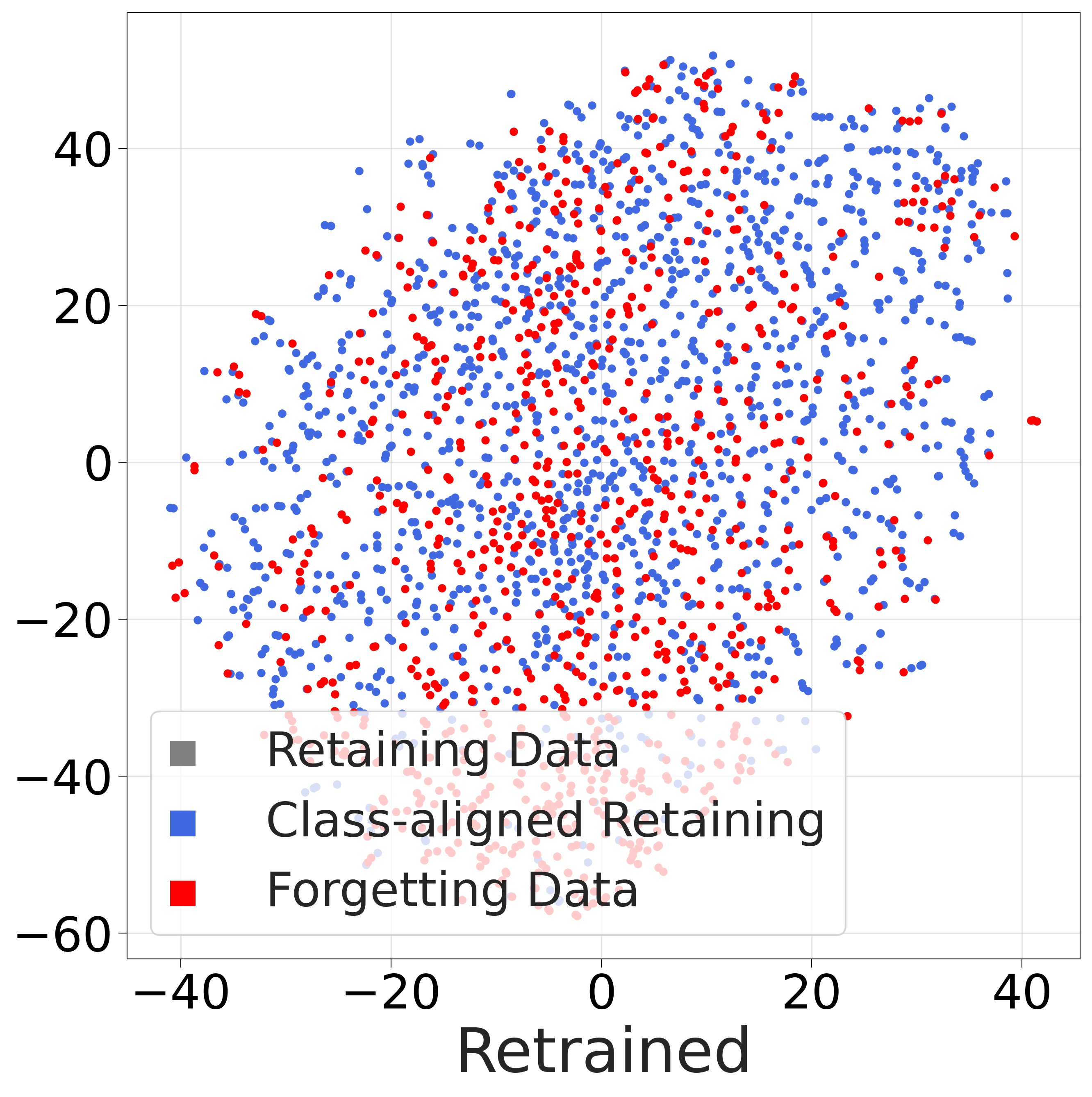}
    \\
    \hspace{0.04in}
    \includegraphics[scale=0.145]{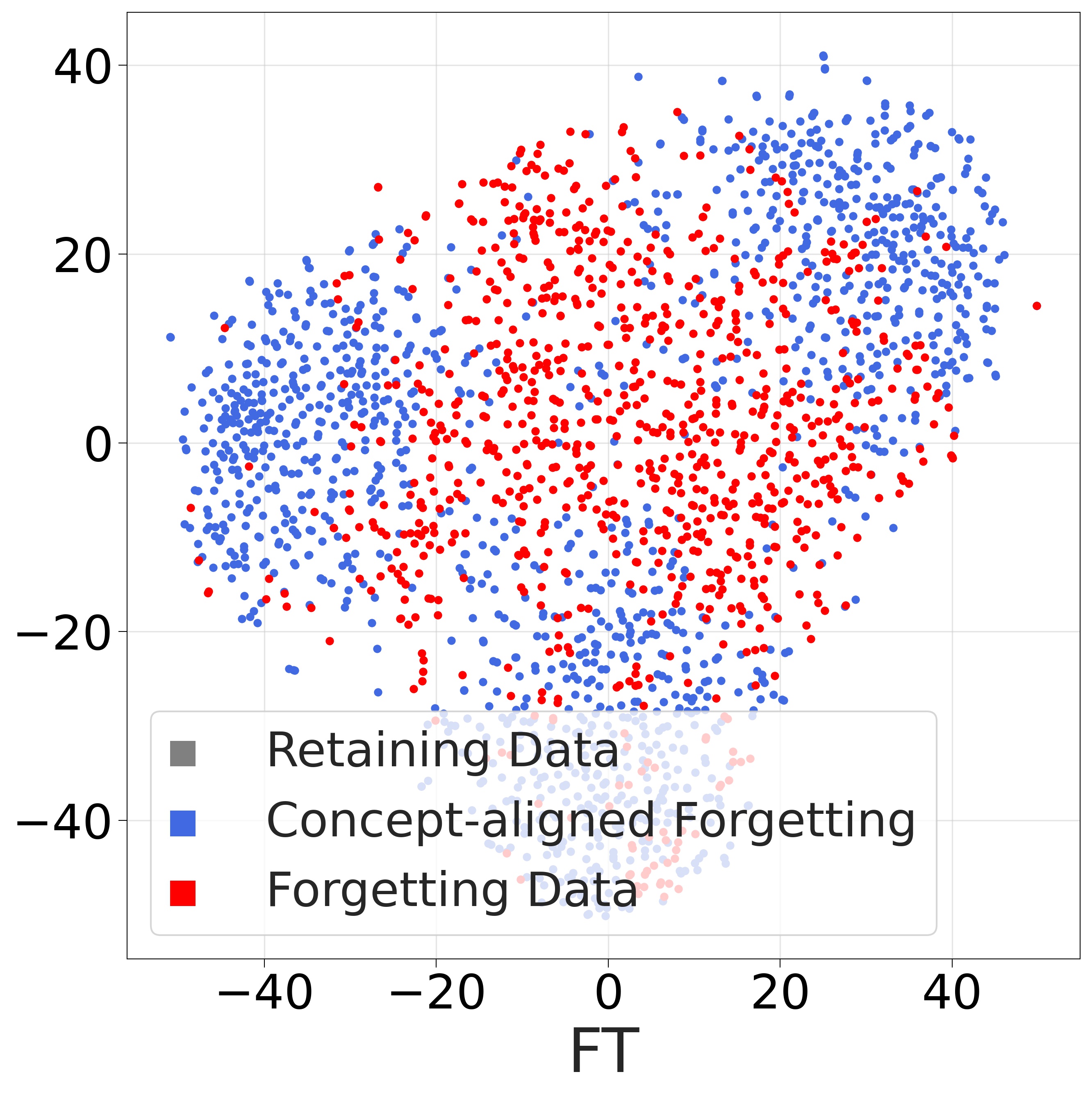}
    \includegraphics[scale=0.145]{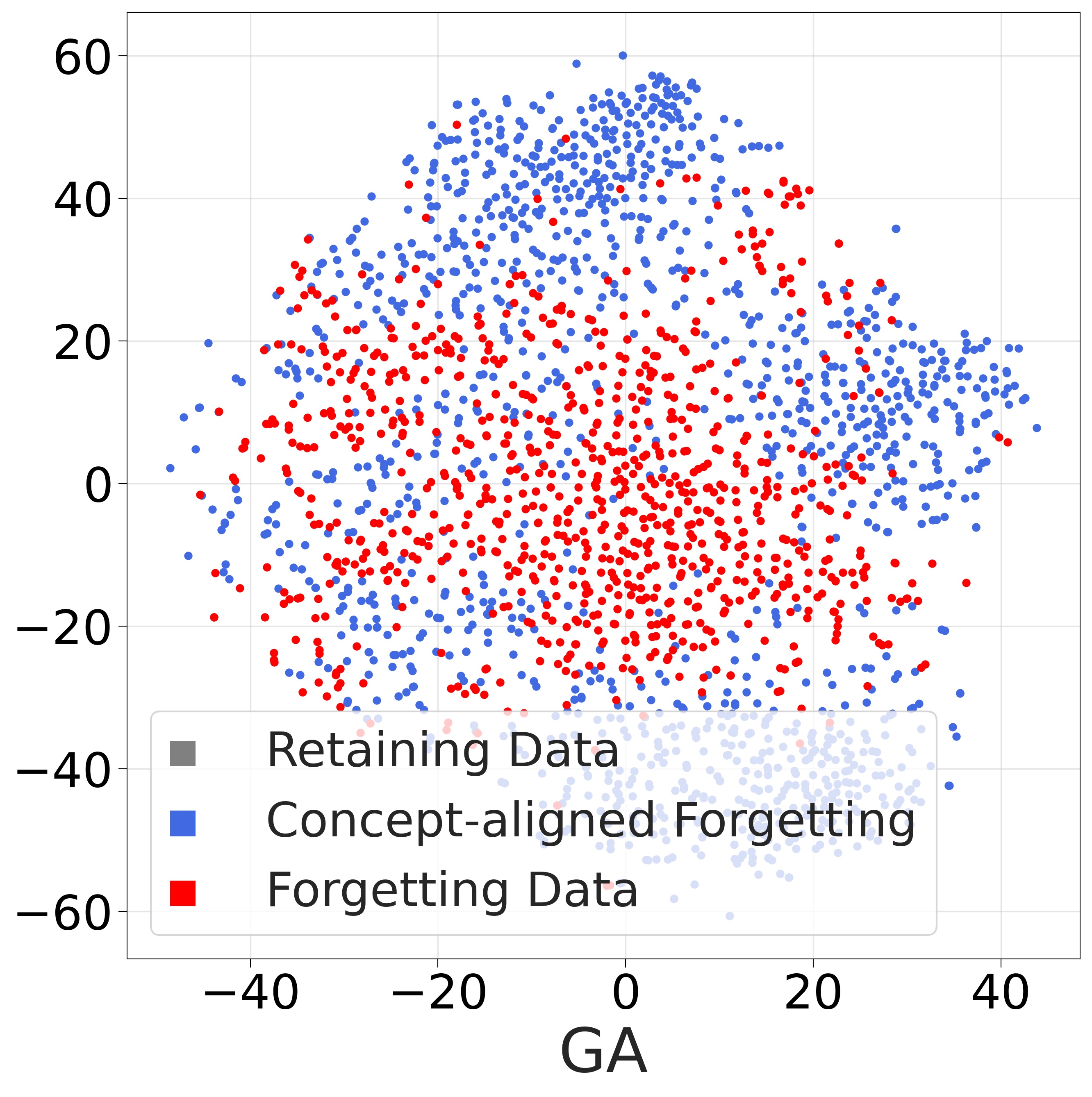}
    \hspace{0.08in}
    \includegraphics[scale=0.145]{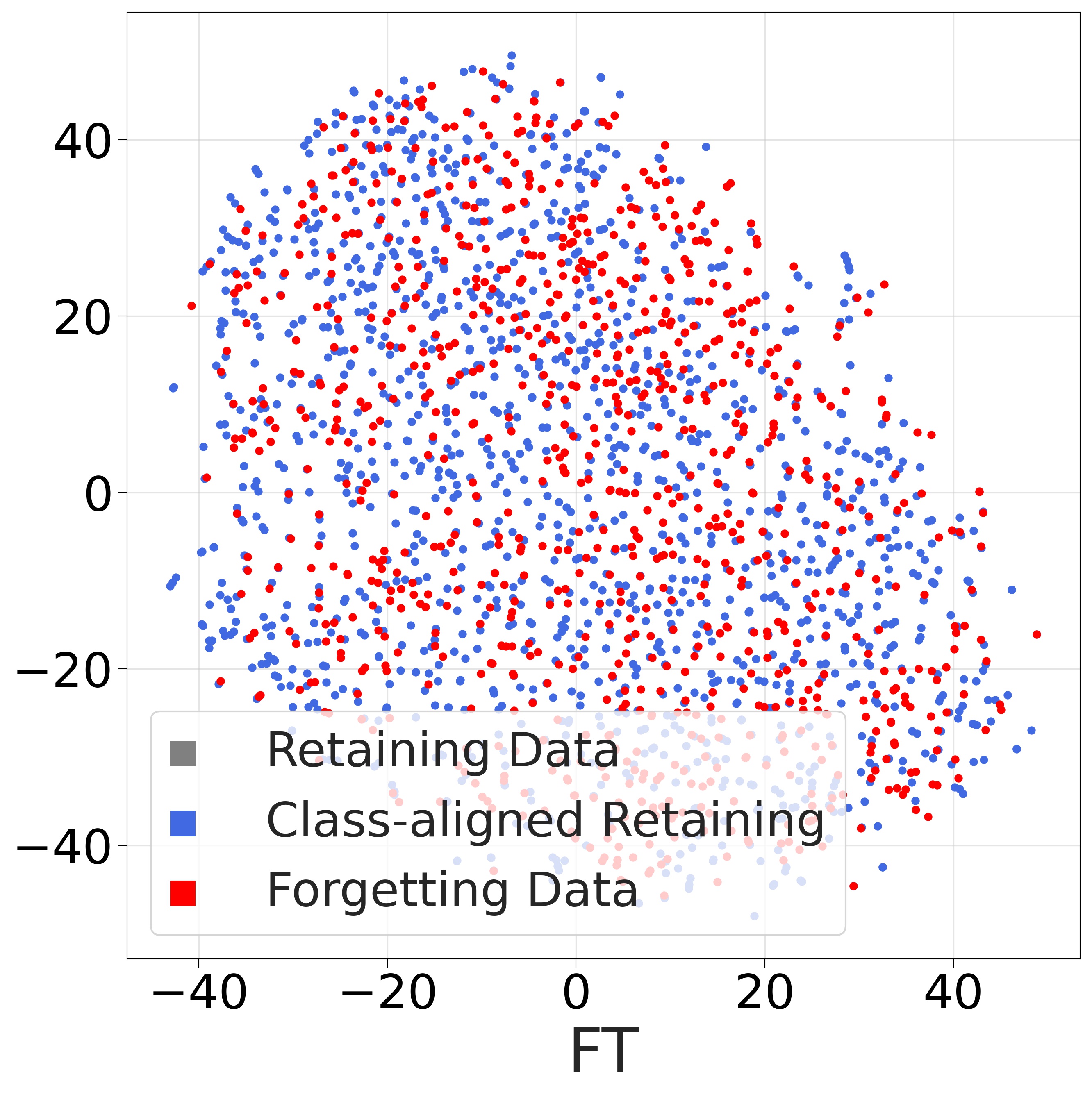}
    \includegraphics[scale=0.145]{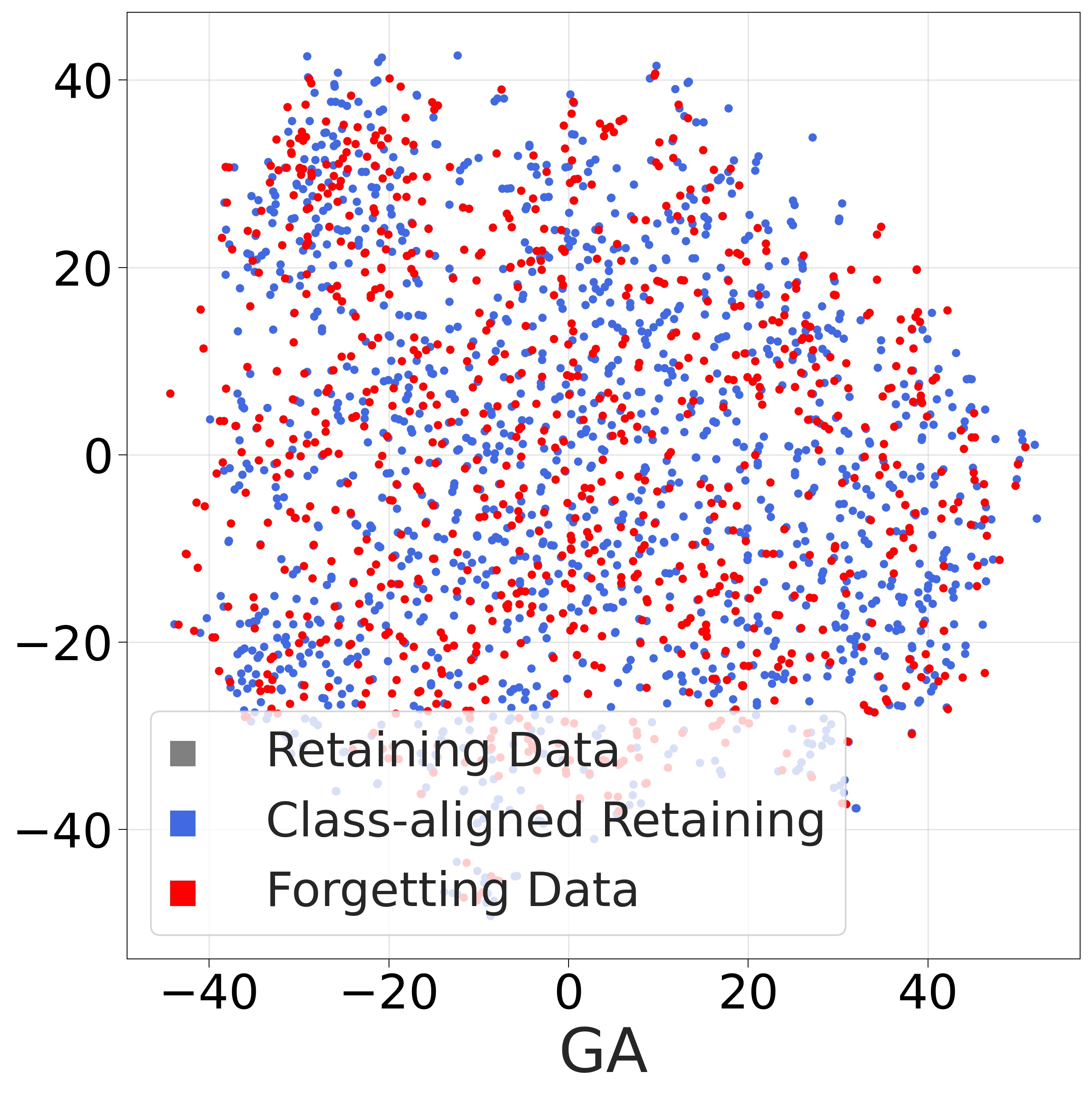}
    \\
    \end{center}
    \footnotesize{
    We present the tSNE visualization~\cite{maaten2008visualizing_tsne} of the learned features, using two representative unlearning methods, i.e., finetune (FT)~\cite{warnecke2021machine} and gradient ascent (GA)~\cite{thudi2022unrolling} with the pre-trained and retrained ones.}
    \caption{\textbf{The entangled/under-entangled feature representations visualized by tSNE.}
    }
    \label{fig:feature_app}
\end{figure*}

\subsection{Forgetting with Different Model Structures}
\label{app:exp_model}

In this part, we further check the unlearning performance of our TARF on different pre-trained model structures compared with several baselines. We choose CIFAR-100 as the pre-training classification task and conduct all matched forgetting and model mismatch forgetting. The results are summarized in Table~\ref{tab:mu_main_app_structure}. The results validate that our TARF can robustly achieve better unlearning performance across different model structures.

\begin{table*}[ht]
    \caption{Results (\%) of unlearning with different model structure. All methods are trained on the same backbone, i.e., the basis of unlearning initialization is the same (except for retraining from scratch). Values are percentages. Bold numbers are superior results. $\downarrow$ indicates smaller are better. }
    \vspace{2mm}
    \centering
    \footnotesize
    \renewcommand\arraystretch{1.0}
    \resizebox{\textwidth}{!}{
    \begin{tabular}{c|l|cccc|c|cccc|c}
        \toprule[1.5pt]
        \multirow{2}*{\textbf{CIFAR-100}} & Task & \multicolumn{5}{c|}{\textbf{All matched}} & \multicolumn{5}{c}{\textbf{Model mismatch}} \\
        \cmidrule{2-12}
        & Metric & UA & RA & TA  & MIA & Gap$\downarrow$ & UA & RA & TA  & MIA & Gap$\downarrow$ \\
        \midrule[0.6pt]
        \multirow{5}*{\shortstack{\textbf{VGG-19}}}
        & \gray Retrained & \gray0.00 & \gray97.26 & \gray73.13 & \gray100.00 & \gray-  & \gray87.44 & \gray98.22 & \gray82.12 & \gray19.89 & \gray- \\
        & FT~\cite{warnecke2021machine} & 0.00 & 90.92 & 66.86 & 100.00 & 3.15 & 95.22 & 95.17& 77.71& 7.56& 6.89 \\
        & RL~\cite{toneva2018empirical} & 0.00 & 90.29 & 66.16 & 100.00 & 3.48 & 96.22 & 95.26 & 77.71 & 98.56 & 23.71 \\
        & GA~\cite{ishida2020we} & 0.00 & 79.27 & 56.03 & 100.00 & 8.77 & 0.00 & 93.09 & 74.30 & 100.00 & 45.13 \\
        \cmidrule{2-12}
        & \textbf{TARF} (ours) &0.00 &91.96 & 67.94 & 100.00& \textbf{2.62} & 82.67 & 93.71 & 76.24 &  24.22 & \textbf{4.87} \\
        \midrule[0.6pt]
        \multirow{5}*{\shortstack{\textbf{ResNet-18}}}
        & \gray Retrained & \gray0.00 & \gray97.85 & \gray76.03 & \gray100.00 & \gray- & \gray88.22 & \gray98.58 & \gray78.50 & \gray25.78 & \gray- \\
        & FT~\cite{warnecke2021machine} & 0.66 & 96.55 & 71.97 & 100.00 & 1.51 & 98.22 & 96.79 & 80.14 & 6.78 & 8.11 \\
        & RL~\cite{toneva2018empirical} & 0.11 & 95.90 &71.57 & 100.00 & 1.63 & 94.11 & 96.70 &80.17 & 96.89 & 20.14  \\
        & GA~\cite{ishida2020we} & 1.89 & 95.26 & 69.14 & 99.89 & 2.87 & 9.33 & 95.13& 77.22 & 96.89 & 38.68  \\
        \cmidrule{2-12}
        & \textbf{TARF} (ours) & 0.00 & 96.90 & 71.51 & 100.00 & \textbf{1.37} & 86.00 & 96.54 & 74.20 & 22.78 & \textbf{2.89}  \\
        \midrule[0.6pt]
        \multirow{5}*{\shortstack{\textbf{WideResNet}}}
        & \gray Retrained & \gray0.00 & \gray97.71 & \gray76.95 & \gray100.00 & \gray-  & \gray88.11 & \gray98.37 & \gray83.61 & \gray23.56 & \gray- \\
        & FT~\cite{warnecke2021machine} & 0.67 & 96.61 & 71.29 & 100.00 & 1.86 &97.44 &95.70 &78.70 & 7.33 & 8.29 \\
        & RL~\cite{toneva2018empirical} & 0.00 & 95.86 & 71.36 & 100.00 & 1.86 &85.77 & 94.69 & 78.26 & 96.00 & 20.95 \\
        & GA~\cite{ishida2020we} & 0.44 & 91.49 & 66.29 & 100.00 & 2.26  &4.33 & 91.76 & 75.18 & 99.11 &  43.71 \\
        \cmidrule{2-12}
        & \textbf{TARF} (ours) & 0.00 & 96.51 & 71.77 & 100.00 & \textbf{1.60} & 88.00 & 95.50 & 79.06 &  22.67 & \textbf{2.11} \\
        \bottomrule[1.5pt]
    \end{tabular}
    }
    \label{tab:mu_main_app_structure}
\end{table*}

\subsection{Full Results with Different Forgetting Tasks}
\label{app:full_exp_re}

In this section, we provide the full results of Table~\ref{tab:mu_main}, which is conducted by setting different random seeds (for multiple runs) with the original trails and reported as the mean and std values for each evaluation metric. Tables~\ref{tab:mu_main_app_full_cifar10_1} to~\ref{tab:mu_main_app_full_cifar10_4} presents the performance of unlearning on CIFAR-10, and Tables~\ref{tab:mu_main_app_full_cifar100_1} to~\ref{tab:mu_main_app_full_cifar100_4} presents the performance of unlearning on CIFAR-100. The performance comparison of our TARF with other baseline across the four forgetting tasks (i.e., all matched, target mismatch, model mismatch, and data mismatch) demonstrated the general effectiveness of our algorithm framework.   

\begin{table*}[ht]
    \caption{Main Results ($\%$). Comparison with the unlearning baselines. All methods are trained on the same backbone, i.e., the basis of unlearning initialization is the same (except for retraining from scratch). Values are percentages. Bold numbers are superior results. $\downarrow$ indicates smaller are better. }
    \vspace{2mm}
    \centering
    \footnotesize
    \renewcommand\arraystretch{1.0}
    \resizebox{\textwidth}{!}{
    \begin{tabular}{c|l|cc|cc|cc|cc|cc}
        \toprule[1.5pt]
         \multirow{2}*{\textbf{CIFAR-10}} & Metric & \multicolumn{2}{c|}{\textbf{UA}} &  \multicolumn{2}{c|}{\textbf{RA}} & \multicolumn{2}{c|}{\textbf{TA}} & \multicolumn{2}{c|}{\textbf{MIA}} & \multicolumn{2}{c}{\textbf{Gap$\downarrow$}} \\
        \cmidrule{2-12}
        &  Method & mean & std & mean & std & mean & std & mean & std & mean & std  \\
        \midrule[0.6pt]
        \multirow{10}*{\shortstack{\textbf{All matched}}}
        & \gray Retrained & \gray0.00 & \gray- & \gray99.51 & \gray- & \gray94.69 & \gray- & \gray100.00 & \gray- & \gray- & \gray-  \\
        & FT~\cite{warnecke2021machine} & 4.66 & 3.59 & 98.58 & 0.04 & 92.42 &  0.06 & 100.00 &0.00 & 1.96 & 0.89\\
        & RL~\cite{toneva2018empirical} &2.23 &1.90 & 98.30 & 0.65 &91.97 &0.74 &100.00 &0.00 &1.54 &0.82  \\
        & GA~\cite{ishida2020we} & 0.34 &0.16 &95.48 &0.24 &88.52 &0.35 &99.88 &0.10 & 2.67& 0.21 \\
        & IU~\cite{izzo2021approximate} &  0.11 & 0.05 &72.50 &15.65 &68.28 & 14.10 &99.98 &0.02 &13.39 &7.41  \\
        & BS~\cite{chen2023boundary} &24.72 &0.32 &88.91 & 0.97&81.84 &0.94 &89.23 &0.56 &14.74 &0.70  \\
        & $L_{1}$-sparse~\cite{jia2023model} &0.00 &0.00 &94.18 &0.03 & 90.01& 0.24 & 100.00 &0.00 &2.50 & 0.05  \\
        & SalUn~\cite{fan2023salun} &0.48 &0.46 &88.66 &2.67 &84.48 &2.40 &100.00 &0.00 &5.39 & 1.38 \\
        & SCRUB~\cite{kurmanji2023towards} & 0.00 & 0.00& 12.92&0.01 &12.92 &0.00 &0.00 &0.00 &67.09 &0.00  \\
        \cmidrule{2-12}
        & \textbf{TARF} (ours) &0.00 &0.00 &98.22 &0.02 &92.09 & 0.14& 100.00& 0.00& \textbf{0.97} &0.03  \\
        \bottomrule[1.5pt]
    \end{tabular}
    }
    \label{tab:mu_main_app_full_cifar10_1}
\end{table*}

\begin{table*}[ht]
    \caption{Main Results ($\%$). Comparison with the unlearning baselines. All methods are trained on the same backbone, i.e., the basis of unlearning initialization is the same (except for retraining from scratch). Values are percentages. Bold numbers are superior results. $\downarrow$ indicates smaller are better. }
    \vspace{2mm}
    \centering
    \footnotesize
    \renewcommand\arraystretch{1.0}
    \resizebox{\textwidth}{!}{
    \begin{tabular}{c|l|cc|cc|cc|cc|cc}
        \toprule[1.5pt]
         \multirow{2}*{\textbf{CIFAR-10}} & Metric & \multicolumn{2}{c|}{\textbf{UA}} &  \multicolumn{2}{c|}{\textbf{RA}} & \multicolumn{2}{c|}{\textbf{TA}} & \multicolumn{2}{c|}{\textbf{MIA}} & \multicolumn{2}{c}{\textbf{Gap$\downarrow$}} \\
        \cmidrule{2-12}
        &  Method & mean & std & mean & std & mean & std & mean & std & mean & std  \\
        \midrule[0.6pt]
        \multirow{10}*{\shortstack{\textbf{Model}\\\textbf{mismatch}}}
        & \gray Retrained & \gray87.76 & \gray- & \gray99.58 & \gray- & \gray95.91 & \gray- & \gray20.57 & \gray- & \gray- & \gray-  \\
        & FT~\cite{warnecke2021machine} & 94.78 & 0.11 & 98.65& 0.12 & 93.77& 0.21 & 10.42 & 0.86 & 5.06 & 0.27 \\
        & RL~\cite{toneva2018empirical} & 48.25 & 5.43 &98.01 &0.12 & 93.03 & 0.21 &98.10 &0.64 &30.37 &1.53  \\
        & GA~\cite{ishida2020we} & 6.49 & 0.73 &86.91 & 0.08& 82.03 &0.18 &94.39 &0.59 &45.41 & 0.27 \\
        & IU~\cite{izzo2021approximate} & 15.84 & 7.86 & 85.89 & 1.45& 81.08 &1.49 &93.58 &3.71 & 43.36 &3.62  \\
        & BS~\cite{chen2023boundary} & 14.05 & 3.76&53.28 &2.51 &51.25 &1.86 & 94.90&1.06 & 59.75& 2.29 \\
        & $L_{1}$-sparse~\cite{jia2023model} & 92.25 & 0.87 & 95.01 & 0.25 & 91.67 &0.04 &17.40 &2.86 &4.14 & 1.00  \\
        & SalUn~\cite{fan2023salun} & 16.31& 7.40& 92.91& 1.05& 86.50&2.12 &99.24 &0.09 & 41.55& 2.14 \\
        & SCRUB~\cite{kurmanji2023towards} & 48.62 &0.02 & 27.86& 0.03& 28.29& 0.05& 48.64& 0.01& 51.63&0.02  \\
        \cmidrule{2-12}
        & \textbf{TARF} (ours) &89.91 &1.20 &97.73 &0.24 &92.66 & 0.17&20.36 &2.54 &\textbf{2.45} &0.46  \\
        \bottomrule[1.5pt]
    \end{tabular}
    }
    \label{tab:mu_main_app_full_cifar10_2}
\end{table*}

\begin{table*}[ht]
    \caption{Main Results ($\%$). Comparison with the unlearning baselines. All methods are trained on the same backbone, i.e., the basis of unlearning initialization is the same (except for retraining from scratch). Values are percentages. Bold numbers are superior results. $\downarrow$ indicates smaller are better. }
    \vspace{2mm}
    \centering
    \footnotesize
    \renewcommand\arraystretch{1.0}
    \resizebox{\textwidth}{!}{
    \begin{tabular}{c|l|cc|cc|cc|cc|cc}
        \toprule[1.5pt]
         \multirow{2}*{\textbf{CIFAR-10}} & Metric & \multicolumn{2}{c|}{\textbf{UA}} &  \multicolumn{2}{c|}{\textbf{RA}} & \multicolumn{2}{c|}{\textbf{TA}} & \multicolumn{2}{c|}{\textbf{MIA}} & \multicolumn{2}{c}{\textbf{Gap$\downarrow$}} \\
        \cmidrule{2-12}
        &  Method & mean & std & mean & std & mean & std & mean & std & mean & std  \\
        \midrule[0.6pt]
        \multirow{10}*{\shortstack{\textbf{Target}\\\textbf{mismatch}}}
        & \gray Retrained & \gray0.00 & \gray- & \gray99.38 & \gray- & \gray93.85 & \gray- & \gray100.00 & \gray- & \gray- & \gray-  \\
        & FT~\cite{warnecke2021machine} &52.23 &1.80 &98.43 &0.05 & 91.74 & 0.09 & 50.59 &0.15 &26.18 &0.40 \\
        & RL~\cite{toneva2018empirical} &50.63 &0.62 &98.21 &0.65 &91.51 &0.61 &53.88 &2.36 &25.06 &0.12  \\
        & GA~\cite{ishida2020we} & 41.64 &0.82 & 97.05 &0.04 & 89.68 & 0.17 &63.23 &1.10 &21.23 & 0.43  \\
        & IU~\cite{izzo2021approximate} &45.32 &0.81 &70.25 &17.82 &65.67 & 2.76 & 55.98 &2.76 & 36.66 & 9.37  \\
        & BS~\cite{chen2023boundary} & 53.78 &0.16 & 89.67&1.02 &79.34 &3.95 &66.31 &10.02 &25.36 &3.28  \\
        & $L_{1}$-sparse~\cite{jia2023model} &49.55 &0.08& 93.57 &0.05 &89.06 &0.23 & 51.33& 0.09 & 27.21 & 0.05  \\
        & SalUn~\cite{fan2023salun} &47.85 &1.22 &87.84 &3.25 &83.38 &2.94 &58.10 &2.85 &27.40 &1.10  \\
        & SCRUB~\cite{kurmanji2023towards} & 43.58 & 0.02 & 3.63 & 1.23 & 3.66& 0.78 & 56.27& 0.23& 63.62 & 0.01  \\
        \cmidrule{2-12}
        & \textbf{TARF} (ours) & 0.05  &0.02 &97.65 &0.08 &91.28 &0.47 &100.00 &0.00 & \textbf{1.09} & 0.14 \\
        \bottomrule[1.5pt]
    \end{tabular}
    }
    \label{tab:mu_main_app_full_cifar10_3}
\end{table*}

\begin{table*}[ht]
    \caption{Main Results ($\%$). Comparison with the unlearning baselines. All methods are trained on the same backbone, i.e., the basis of unlearning initialization is the same (except for retraining from scratch). Values are percentages. Bold numbers are superior results. $\downarrow$ indicates smaller are better. }
    \vspace{2mm}
    \centering
    \footnotesize
    \renewcommand\arraystretch{1.0}
    \resizebox{\textwidth}{!}{
    \begin{tabular}{c|l|cc|cc|cc|cc|cc}
        \toprule[1.5pt]
         \multirow{2}*{\textbf{CIFAR-10}} & Metric & \multicolumn{2}{c|}{\textbf{UA}} &  \multicolumn{2}{c|}{\textbf{RA}} & \multicolumn{2}{c|}{\textbf{TA}} & \multicolumn{2}{c|}{\textbf{MIA}} & \multicolumn{2}{c}{\textbf{Gap$\downarrow$}} \\
        \cmidrule{2-12}
        &  Method & mean & std & mean & std & mean & std & mean & std & mean & std  \\
        \midrule[0.6pt]
        \multirow{10}*{\shortstack{\textbf{Data}\\\textbf{mismatch}}}
        & \gray Retrained & \gray0.00 & \gray- & \gray99.53 & \gray- & \gray95.56 & \gray- & \gray100.00 & \gray- & \gray- & \gray-  \\
        & FT~\cite{warnecke2021machine} & 96.85 & 0.06 &98.62 & 0.13 & 93.47 & 0.21  & 6.93 & 0.45 & 48.23 & 0.18 \\
        & RL~\cite{toneva2018empirical} &73.62 & 2.86 &97.90 &0.22 &92.59 & 0.66 & 52.04 & 2.23 &31.55 & 1.49  \\
        & GA~\cite{ishida2020we} &9.82 & 1.13 &96.14 & 0.28 &90.46 & 0.33 & 90.46 &0.95 &6.56 & 0.67  \\
        & IU~\cite{izzo2021approximate} & 15.19 & 7.66 & 94.80 & 0.70 &89.08 & 0.46 & 92.83 & 4.26 & 8.39 & 2.69 \\
        & BS~\cite{chen2023boundary} & 16.72 &0.02 &61.01&0.21 &53.81&4.05 & 93.47& 1.24& 25.88& 1.27  \\
        & $L_{1}$-sparse~\cite{jia2023model} & 95.42 & 0.35 &  94.57 & 0.26 & 91.07 & 0.01 & 10.82 & 1.30 & 48.51 & 0.47  \\
        & SalUn~\cite{fan2023salun} &55.52 &3.76 & 92.68&1.19 & 89.25&1.22 &60.23 &3.30 &27.12 & 2.37 \\
        & SCRUB~\cite{kurmanji2023towards} & 59.46 & 0.02& 22.56&0.01 & 25.72 &0.01 & 54.08 &0.02 &63.06 &0.02  \\
        \cmidrule{2-12}
        & \textbf{TARF} (ours) &0.00 &0.00 &98.35 & 0.18&93.42 & 0.34&100.00 &0.00 &\textbf{0.83} & 0.13 \\
        \bottomrule[1.5pt]
    \end{tabular}
    }
    \label{tab:mu_main_app_full_cifar10_4}
\end{table*}

\begin{table*}[ht]
    \caption{Main Results ($\%$). Comparison with the unlearning baselines. All methods are trained on the same backbone, i.e., the basis of unlearning initialization is the same (except for retraining from scratch). Values are percentages. Bold numbers are superior results. $\downarrow$ indicates smaller are better. }
    \vspace{2mm}
    \centering
    \footnotesize
    \renewcommand\arraystretch{1.0}
    \resizebox{\textwidth}{!}{
    \begin{tabular}{c|l|cc|cc|cc|cc|cc}
        \toprule[1.5pt]
         \multirow{2}*{\textbf{CIFAR-100}} & Metric & \multicolumn{2}{c|}{\textbf{UA}} &  \multicolumn{2}{c|}{\textbf{RA}} & \multicolumn{2}{c|}{\textbf{TA}} & \multicolumn{2}{c|}{\textbf{MIA}} & \multicolumn{2}{c}{\textbf{Gap$\downarrow$}} \\
        \cmidrule{2-12}
        &  Method & mean & std & mean & std & mean & std & mean & std & mean & std  \\
        \midrule[0.6pt]
        \multirow{10}*{\shortstack{\textbf{All matched}}}
        & \gray Retrained & \gray0.00 & \gray- & \gray97.85 & \gray- & \gray76.03 & \gray- & \gray100.00 & \gray- & \gray- & \gray-  \\
        & FT~\cite{warnecke2021machine} & 0.67 & 0.01 & 96.44 & 0.12 & 72.16 &  0.19 & 100.00 &0.00 & 1.49 & 0.02\\
        & RL~\cite{toneva2018empirical} & 0.56 &0.45 & 96.00 &0.10 & 71.79 & 0.22 & 100.00 & 0.00 & 1.66 & 0.03  \\
        & GA~\cite{ishida2020we} & 1.61 &0.28 &95.00 &0.26 &68.85 &0.29 &99.89 &0.00 & 2.93& 0.07 \\
        & IU~\cite{izzo2021approximate} &  0.00 & 0.00 &39.80 &2.19 &31.09 & 1.51 &100.00 &0.00 &25.75 &0.93  \\
        & BS~\cite{chen2023boundary} &12.83 &0.05 &97.17 &0.06 &69.30 &0.64 &99.45 &0.12 &5.20&0.22  \\
        & $L_{1}$-sparse~\cite{jia2023model} &0.00 &0.00 &82.57 &0.57 & 66.35& 1.27 & 100.00 &0.00 &6.24 & 0.46  \\
        & SalUn~\cite{fan2023salun} &0.00 &0.00 &77.00 &1.66 &63.06 &0.92 & 100.00& 0.00& 8.46&0.64  \\
        & SCRUB~\cite{kurmanji2023towards} & 0.00& 0.00& 99.72&0.26 &76.69 &0.06 &100.00 &0.00 &\textbf{0.64} & 0.08 \\
        \cmidrule{2-12}
        & \textbf{TARF} (ours) &0.00 &0.00 &96.67 &0.24 &72.40 &0.14 &100.00 &0.00 & 1.21& 0.09\\
        \bottomrule[1.5pt]
    \end{tabular}
    }
    \label{tab:mu_main_app_full_cifar100_1}
\end{table*}

\begin{table*}[ht]
    \caption{Main Results ($\%$). Comparison with the unlearning baselines. All methods are trained on the same backbone, i.e., the basis of unlearning initialization is the same (except for retraining from scratch). Values are percentages. Bold numbers are superior results. $\downarrow$ indicates smaller are better. }
    \vspace{2mm}
    \centering
    \footnotesize
    \renewcommand\arraystretch{1.0}
    \resizebox{\textwidth}{!}{
    \begin{tabular}{c|l|cc|cc|cc|cc|cc}
        \toprule[1.5pt]
         \multirow{2}*{\textbf{CIFAR-100}} & Metric & \multicolumn{2}{c|}{\textbf{UA}} &  \multicolumn{2}{c|}{\textbf{RA}} & \multicolumn{2}{c|}{\textbf{TA}} & \multicolumn{2}{c|}{\textbf{MIA}} & \multicolumn{2}{c}{\textbf{Gap$\downarrow$}} \\
        \cmidrule{2-12}
        &  Method & mean & std & mean & std & mean & std & mean & std & mean & std  \\
        \midrule[0.6pt]
        \multirow{10}*{\shortstack{\textbf{Model}\\\textbf{mismatch}}}
        & \gray Retrained & \gray88.22 & \gray- & \gray98.58 & \gray- & \gray78.50 & \gray- & \gray25.78 & \gray- & \gray- & \gray-  \\
        & FT~\cite{warnecke2021machine} & 95.45 & 2.78 & 95.91& 0.89& 79.74& 0.40 & 11.56& 4.78&6.34 & 1.77 \\
        & RL~\cite{toneva2018empirical} & 87.11 & 7.00 &96.27 &0.44 &80.00 &0.17 &97.95 &1.06 &20.75 &0.61  \\
        & GA~\cite{ishida2020we} & 8.06 &1.28 &94.98 & 0.15& 77.09 & 0.13 &97.34 &0.45 &39.18 & 0.50 \\
        & IU~\cite{izzo2021approximate} & 39.95 & 5.28 & 97.22 & 0.39& 79.71 &  0.63 &83.28 &3.17 &27.08 &2.05  \\
        & BS~\cite{chen2023boundary} & 18.56& 0.56& 95.87& 0.03& 74.96& 2.68& 94.95& 0.28& 36.27& 0.87 \\
        & $L_{1}$-sparse~\cite{jia2023model} & 87.11 & 5.00 & 84.35 & 0.18 & 75.61 &0.39 &15.56 &4.45 &7.84 & 0.69  \\
        & SalUn~\cite{fan2023salun} & 74.78 &8.45 &79.98 & 1.14&71.55 &0.77 &65.61 &11.39 &19.71 & 5.44 \\
        & SCRUB~\cite{kurmanji2023towards} &0.00 &0.00 &13.64 &2.39 &12.75 &2.75 &99.45 &0.56 &78.14 &1.15  \\
        \cmidrule{2-12}
        & \textbf{TARF} (ours) &84.78 &1.90 &97.19 &0.14 &80.02 &0.15 &28.89 &2.89 &\textbf{2.37} &1.15  \\
        \bottomrule[1.5pt]
    \end{tabular}
    }
    \label{tab:mu_main_app_full_cifar100_2}
\end{table*}

\begin{table*}[ht]
    \caption{Main Results ($\%$). Comparison with the unlearning baselines. All methods are trained on the same backbone, i.e., the basis of unlearning initialization is the same (except for retraining from scratch). Values are percentages. Bold numbers are superior results. $\downarrow$ indicates smaller are better. }
    \vspace{2mm}
    \centering
    \footnotesize
    \renewcommand\arraystretch{1.0}
    \resizebox{\textwidth}{!}{
    \begin{tabular}{c|l|cc|cc|cc|cc|cc}
        \toprule[1.5pt]
         \multirow{2}*{\textbf{CIFAR-100}} & Metric & \multicolumn{2}{c|}{\textbf{UA}} &  \multicolumn{2}{c|}{\textbf{RA}} & \multicolumn{2}{c|}{\textbf{TA}} & \multicolumn{2}{c|}{\textbf{MIA}} & \multicolumn{2}{c}{\textbf{Gap$\downarrow$}} \\
        \cmidrule{2-12}
        &  Method & mean & std & mean & std & mean & std & mean & std & mean & std  \\
        \midrule[0.6pt]
        \multirow{10}*{\shortstack{\textbf{Target}\\\textbf{mismatch}}}
        & \gray Retrained & \gray0.00 & \gray- & \gray97.85 & \gray- & \gray73.72 & \gray- & \gray100.00 & \gray- & \gray- & \gray-  \\
        & FT~\cite{warnecke2021machine} &58.58 &0.40 &96.42 &0.10 & 72.31 & 0.22 & 45.94 &0.83 &28.87 &0.34  \\
        & RL~\cite{toneva2018empirical} &57.76 &1.14 &96.00 &0.10 &72.04 &0.16 &50.67 &3.69 &27.66 &1.15  \\
        & GA~\cite{ishida2020we} & 22.07 &0.69 & 96.87 &0.24 &70.52 &0.30 &90.45 &0.23 &8.95 &0.10  \\
        & IU~\cite{izzo2021approximate} &30.80 &0.18 &39.44 &2.25 &31.00 & 1.42 &63.83 &0.14 &42.03 & 0.91  \\
        & BS~\cite{chen2023boundary} &40.91 &0.47 &98.36 & 0.04&70.04  &1.38 &85.00 &0.16 & 15.03&0.18  \\
        & $L_{1}$-sparse~\cite{jia2023model} &49.87 &2.90 &82.55 &0.44 &66.35 &1.27 & 50.47&3.54 &30.52 &1.18  \\
        & SalUn~\cite{fan2023salun} &43.29 &1.60 &77.15 &1.63 & 63.30&0.93 &64.63 &1.34 &27.45 &0.10  \\
        & SCRUB~\cite{kurmanji2023towards} & 59.56 &0.09 &99.74 &0.26 &76.14 & 0.82&45.45 &0.56 &29.60 &0.02  \\
        \cmidrule{2-12}
        & \textbf{TARF} (ours) &0.29 & 0.03&97.06 &0.29 &73.27 &0.41 &100.00 &0.00 &\textbf{0.38} & 0.17 \\
        \bottomrule[1.5pt]
    \end{tabular}
    }
    \label{tab:mu_main_app_full_cifar100_3}
\end{table*}

\begin{table*}[ht]
    \caption{Main Results ($\%$). Comparison with the unlearning baselines. All methods are trained on the same backbone, i.e., the basis of unlearning initialization is the same (except for retraining from scratch). Values are percentages. Bold numbers are superior results. $\downarrow$ indicates smaller are better. }
    \vspace{2mm}
    \centering
    \footnotesize
    \renewcommand\arraystretch{1.0}
    \resizebox{\textwidth}{!}{
    \begin{tabular}{c|l|cc|cc|cc|cc|cc}
        \toprule[1.5pt]
         \multirow{2}*{\textbf{CIFAR-100}} & Metric & \multicolumn{2}{c|}{\textbf{UA}} &  \multicolumn{2}{c|}{\textbf{RA}} & \multicolumn{2}{c|}{\textbf{TA}} & \multicolumn{2}{c|}{\textbf{MIA}} & \multicolumn{2}{c}{\textbf{Gap$\downarrow$}} \\
        \cmidrule{2-12}
        &  Method & mean & std & mean & std & mean & std & mean & std & mean & std  \\
        \midrule[0.6pt]
        \multirow{10}*{\shortstack{\textbf{Data}\\\textbf{mismatch}}}
        & \gray Retrained & \gray0.00 & \gray- & \gray98.50 & \gray- & \gray80.15 & \gray- & \gray100.00 & \gray- & \gray- & \gray-  \\
        & FT~\cite{warnecke2021machine} & 90.79 & 8.18 &96.19 & 0.52 &79.80 & 0.03 &20.46 & 16.78 &43.25 & 6.10 \\
        & RL~\cite{toneva2018empirical} &93.60 & 3.82 &96.32 &0.39 &79.92 & 0.02 &65.20 &5.56 &32.73 & 2.24  \\
        & GA~\cite{ishida2020we} & 6.98 & 0.98 &97.78 & 0.14 &79.34 & 0.11 &97.53 &0.51 &2.75 &0.31  \\
        & IU~\cite{izzo2021approximate} &37.22 & 5.71&99.17 &0.21 &80.01 &1.81 &85.41 &2.41 &13.54 & 2.08 \\
        & BS~\cite{chen2023boundary} &15.71 &0.33 &98.47 &0.04 &76.02 &3.74 & 96.05&0.18 &5.86 &0.18  \\
        & $L_{1}$-sparse~\cite{jia2023model} &  89.20 & 4.67 &  85.18& 0.05&76.02 & 0.20 &12.67 & 4.36 &48.64 & 2.20  \\
        & SalUn~\cite{fan2023salun} &79.00 & 6.07& 79.92  &1.05 &71.55 &0.51 & 44.18&9.96 & 39.42& 3.62 \\
        & SCRUB~\cite{kurmanji2023towards} & 0.00& 0.00& 14.08& 2.47& 13.48& 2.48& 99.45& 0.55& 36.83& 1.37 \\
        \cmidrule{2-12}
        & \textbf{TARF} (ours) &0.00 &0.00 &95.80 &0.79 &79.55 &0.57 & 100.00&0.00 &\textbf{1.61} &0.05  \\
        \bottomrule[1.5pt]
    \end{tabular}
    }
    \label{tab:mu_main_app_full_cifar100_4}
\end{table*}

\section{Further Discussions}
\label{app:broader_limitation}

In this work, we explore the label domain mismatch in class-wise unlearning, which aims to enhance the flexibility of data regulation with the increasing concern about the trustworthiness of machine learning. Pushing forward the practical usage of machine unlearning, our research provides a broader consideration of real-world unlearning scenarios and offers significant positive social impacts. It can enhance data privacy protection by allowing individuals to effectively remove their data, ensuring some sensitive data is not used for analysis. In addition, unlearning can remove bias or discrimination by correcting flawed datasets, promoting the development of fairness or other ethical considerations. This feature also enables enterprises to adhere to data protection standards such as GDPR~\cite{rosen2011right} and CCPA~\cite{pardau2018california}, therefore promoting confidence among users.
Our newly introduced unlearning setting, which decouples the class label and the target concept, is more general and discusses the achievability of various unlearning requests, which may often be different from the taxonomy of pre-training tasks.

Although we take a step forward in more practical class-wise unlearning by considering the label domain mismatch scenarios, it is not the end of this direction and there are still many problems to be addressed.
Following the previous works~\cite{warnecke2021machine,golatkar2020eternal,jia2023model,chen2023boundary}, our work mainly focuses on the class-wise unlearning with the classification model for the exploration, future efforts can also be paid in the unlearning problem of the emerging and powerful generative models. On the technical level, although those compared unlearning methods and our framework can achieve the forgetting target, it all requires extra computational cost, and how to make it more efficient can be further studied.



\begin{thebibliography}{10}
\addtolength{\itemsep}{1.5 em}
\setlength{\itemsep}{5pt}

\bibitem{bekker2020learning}
Jessa Bekker and Jesse Davis.
\newblock Learning from positive and unlabeled data: A survey.
\newblock {\em Machine Learning}, 2020.

\bibitem{bommasani2021opportunities}
Rishi Bommasani, Drew~A Hudson, Ehsan Adeli, Russ Altman, Simran Arora, Sydney von Arx, Michael~S Bernstein, Jeannette Bohg, Antoine Bosselut, Emma Brunskill, et~al.
\newblock On the opportunities and risks of foundation models.
\newblock In {\em arXiv}, 2021.

\bibitem{bourtoule2021machine}
Lucas Bourtoule, Varun Chandrasekaran, Christopher~A Choquette-Choo, Hengrui Jia, Adelin Travers, Baiwu Zhang, David Lie, and Nicolas Papernot.
\newblock Machine unlearning.
\newblock In {\em IEEE Symposium on Security and Privacy (SP)}, 2021.

\bibitem{cao2015towards}
Yinzhi Cao and Junfeng Yang.
\newblock Towards making systems forget with machine unlearning.
\newblock In {\em IEEE symposium on security and privacy (SP)}, 2015.

\bibitem{chen2023boundary}
Min Chen, Weizhuo Gao, Gaoyang Liu, Kai Peng, and Chen Wang.
\newblock Boundary unlearning.
\newblock {\em arXiv preprint arXiv:2303.11570}, 2023.

\bibitem{chen2020self}
Xuxi Chen, Wuyang Chen, Tianlong Chen, Ye~Yuan, Chen Gong, Kewei Chen, and Zhangyang Wang.
\newblock Self-pu: Self boosted and calibrated positive-unlabeled training.
\newblock In {\em ICML}, 2020.

\bibitem{du2015modelling}
Jun Du and Zhihua Cai.
\newblock Modelling class noise with symmetric and asymmetric distributions.
\newblock In {\em AAAI}, 2015.

\bibitem{du2015convex}
Marthinus du~Plessis, Gang Niu, and Masashi Sugiyama.
\newblock Convex formulation for learning from positive and unlabeled data.
\newblock In {\em ICML}, 2015.

\bibitem{du2014analysis}
Marthinus~C du~Plessis, Gang Niu, and Masashi Sugiyama.
\newblock Analysis of learning from positive and unlabeled data.
\newblock In {\em NeurIPS}, 2014.

\bibitem{fan2024challenging}
Chongyu Fan, Jiancheng Liu, Alfred Hero, and Sijia Liu.
\newblock Challenging forgets: Unveiling the worst-case forget sets in machine unlearning.
\newblock {\em arXiv preprint arXiv:2403.07362}, 2024.

\bibitem{fan2023salun}
Chongyu Fan, Jiancheng Liu, Yihua Zhang, Dennis Wei, Eric Wong, and Sijia Liu.
\newblock Salun: Empowering machine unlearning via gradient-based weight saliency in both image classification and generation.
\newblock {\em arXiv preprint arXiv:2310.12508}, 2023.

\bibitem{french1999catastrophic}
Robert~M French.
\newblock Catastrophic forgetting in connectionist networks.
\newblock {\em Trends in cognitive sciences}, 1999.

\bibitem{gandikota2023erasing}
Rohit Gandikota, Joanna Materzynska, Jaden Fiotto-Kaufman, and David Bau.
\newblock Erasing concepts from diffusion models.
\newblock {\em arXiv preprint arXiv:2303.07345}, 2023.

\bibitem{ginart2019making}
Antonio Ginart, Melody Guan, Gregory Valiant, and James~Y Zou.
\newblock Making ai forget you: Data deletion in machine learning.
\newblock In {\em NeurIPS}, 2019.

\bibitem{golatkar2020eternal}
Aditya Golatkar, Alessandro Achille, and Stefano Soatto.
\newblock Eternal sunshine of the spotless net: Selective forgetting in deep networks.
\newblock In {\em CVPR}, 2020.

\bibitem{goodfellow2016deep}
Ian Goodfellow, Yoshua Bengio, Aaron Courville, and Yoshua Bengio.
\newblock {\em Deep learning}.
\newblock MIT Press, 2016.

\bibitem{goodfellow2014explaining}
Ian~J. Goodfellow, Jonathon Shlens, and Christian Szegedy.
\newblock Explaining and harnessing adversarial examples.
\newblock In {\em ICLR}, 2015.

\bibitem{graves2021amnesiac}
Laura Graves, Vineel Nagisetty, and Vijay Ganesh.
\newblock Amnesiac machine learning.
\newblock In {\em AAAI}, 2021.

\bibitem{guo2018curriculumnet}
Sheng Guo, Weilin Huang, Haozhi Zhang, Chenfan Zhuang, Dengke Dong, Matthew~R Scott, and Dinglong Huang.
\newblock Curriculumnet: Weakly supervised learning from large-scale web images.
\newblock In {\em ECCV}, 2018.

\bibitem{hashimoto2018fairness}
Tatsunori Hashimoto, Megha Srivastava, Hongseok Namkoong, and Percy Liang.
\newblock Fairness without demographics in repeated loss minimization.
\newblock In {\em ICML}, 2018.

\bibitem{he2016deep}
Kaiming He, Xiangyu Zhang, Shaoqing Ren, and Jian Sun.
\newblock Deep residual learning for image recognition.
\newblock In {\em CVPR}, 2016.

\bibitem{ho2020denoising}
Jonathan Ho, Ajay Jain, and Pieter Abbeel.
\newblock Denoising diffusion probabilistic models.
\newblock In {\em NeurIPS}, 2020.

\bibitem{hoofnagle2019european}
Chris~Jay Hoofnagle, Bart Van Der~Sloot, and Frederik~Zuiderveen Borgesius.
\newblock The european union general data protection regulation: what it is and what it means.
\newblock {\em Information \& Communications Technology Law}, 2019.

\bibitem{hsieh2019classification}
Yu-Guan Hsieh, Gang Niu, and Masashi Sugiyama.
\newblock Classification from positive, unlabeled and biased negative data.
\newblock In {\em ICML}, 2019.

\bibitem{ishida2020we}
Takashi Ishida, Ikko Yamane, Tomoya Sakai, Gang Niu, and Masashi Sugiyama.
\newblock Do we need zero training loss after achieving zero training error?
\newblock In {\em ICML}, 2020.

\bibitem{izzo2021approximate}
Zachary Izzo, Mary~Anne Smart, Kamalika Chaudhuri, and James Zou.
\newblock Approximate data deletion from machine learning models.
\newblock In {\em AISTATS}, 2021.

\bibitem{jia2023model}
Jinghan Jia, Jiancheng Liu, Parikshit Ram, Yuguang Yao, Gaowen Liu, Yang Liu, Pranay Sharma, and Sijia Liu.
\newblock Model sparsity can simplify machine unlearning.
\newblock In {\em NeurIPS}, 2023.

\bibitem{kirkpatrick2017overcoming}
James Kirkpatrick, Razvan Pascanu, Neil Rabinowitz, Joel Veness, Guillaume Desjardins, Andrei~A Rusu, Kieran Milan, John Quan, Tiago Ramalho, Agnieszka Grabska-Barwinska, et~al.
\newblock Overcoming catastrophic forgetting in neural networks.
\newblock {\em Proceedings of the national academy of sciences}, 2017.

\bibitem{kiryo2017positive}
Ryuichi Kiryo, Gang Niu, Marthinus~C du~Plessis, and Masashi Sugiyama.
\newblock Positive-unlabeled learning with non-negative risk estimator.
\newblock In {\em NeurIPS}, 2017.

\bibitem{koh2017understanding}
Pang~Wei Koh and Percy Liang.
\newblock Understanding black-box predictions via influence functions.
\newblock In {\em ICML}, 2017.

\bibitem{kovashka2016crowdsourcing}
Adriana Kovashka, Olga Russakovsky, Li~Fei{-}Fei, and Kristen Grauman.
\newblock Crowdsourcing in computer vision.
\newblock In {\em Found. Trends Comput. Graph. Vis.}, 2016.

\bibitem{krizhevsky2009learning_cifar10}
Alex Krizhevsky.
\newblock Learning multiple layers of features from tiny images.
\newblock In {\em arXiv}, 2009.

\bibitem{kurmanji2023towards}
Meghdad Kurmanji, Peter Triantafillou, Jamie Hayes, and Eleni Triantafillou.
\newblock Towards unbounded machine unlearning.
\newblock In {\em NeurIPS}, 2023.

\bibitem{liu2002partially}
Bing Liu, Wee~Sun Lee, Philip~S Yu, and Xiaoli Li.
\newblock Partially supervised classification of text documents.
\newblock In {\em ICML}, 2002.

\bibitem{liu2015classification}
Tongliang Liu and Dacheng Tao.
\newblock Classification with noisy labels by importance reweighting.
\newblock {\em IEEE Transactions on pattern analysis and machine intelligence}, 2015.

\bibitem{maaten2008visualizing_tsne}
Laurens van~der Maaten and Geoffrey Hinton.
\newblock Visualizing data using t-sne.
\newblock In {\em booktitle of machine learning research}, 2008.

\bibitem{madry2017towards}
Aleksander Madry, Aleksandar Makelov, Ludwig Schmidt, Dimitris Tsipras, and Adrian Vladu.
\newblock Towards deep learning models resistant to adversarial attacks.
\newblock In {\em ICLR}, 2018.

\bibitem{maini2024tofu}
Pratyush Maini, Zhili Feng, Avi Schwarzschild, Zachary~C Lipton, and J~Zico Kolter.
\newblock Tofu: A task of fictitious unlearning for llms.
\newblock {\em arXiv preprint arXiv:2401.06121}, 2024.

\bibitem{menon2015learning}
Aditya Menon, Brendan Van~Rooyen, Cheng~Soon Ong, and Bob Williamson.
\newblock Learning from corrupted binary labels via class-probability estimation.
\newblock In {\em ICML}, 2015.

\bibitem{neel2021descent}
Seth Neel, Aaron Roth, and Saeed Sharifi-Malvajerdi.
\newblock Descent-to-delete: Gradient-based methods for machine unlearning.
\newblock In {\em Algorithmic Learning Theory}, 2021.

\bibitem{pardau2018california}
Stuart~L Pardau.
\newblock The california consumer privacy act: Towards a european-style privacy regime in the united states.
\newblock {\em J. Tech. L. \& Pol'y}, 2018.

\bibitem{rosen2011right}
Jeffrey Rosen.
\newblock The right to be forgotten.
\newblock {\em Stan. L. Rev. Online}, 2011.

\bibitem{sekhari2021remember}
Ayush Sekhari, Jayadev Acharya, Gautam Kamath, and Ananda~Theertha Suresh.
\newblock Remember what you want to forget: Algorithms for machine unlearning.
\newblock In {\em NeurIPS}, 2021.

\bibitem{shah2023unlearning}
Vedant Shah, Frederik Tr{\"a}uble, Ashish Malik, Hugo Larochelle, Michael Mozer, Sanjeev Arora, Yoshua Bengio, and Anirudh Goyal.
\newblock Unlearning via sparse representations.
\newblock {\em arXiv preprint arXiv:2311.15268}, 2023.

\bibitem{shaik2023exploring}
Thanveer Shaik, Xiaohui Tao, Haoran Xie, Lin Li, Xiaofeng Zhu, and Qing Li.
\newblock Exploring the landscape of machine unlearning: A survey and taxonomy.
\newblock {\em arXiv preprint arXiv:2305.06360}, 2023.

\bibitem{simonyan2014very}
Karen Simonyan and Andrew Zisserman.
\newblock Very deep convolutional networks for large-scale image recognition.
\newblock In {\em ICLR}, 2015.

\bibitem{thudi2022unrolling}
Anvith Thudi, Gabriel Deza, Varun Chandrasekaran, and Nicolas Papernot.
\newblock Unrolling sgd: Understanding factors influencing machine unlearning.
\newblock In {\em IEEE European Symposium on Security and Privacy (EuroS\&P)}, 2022.

\bibitem{thudi2022necessity}
Anvith Thudi, Hengrui Jia, Ilia Shumailov, and Nicolas Papernot.
\newblock On the necessity of auditable algorithmic definitions for machine unlearning.
\newblock In {\em USENIX Security Symposium (USENIX Security)}, 2022.

\bibitem{toneva2018empirical}
Mariya Toneva, Alessandro Sordoni, Remi~Tachet des Combes, Adam Trischler, Yoshua Bengio, and Geoffrey~J Gordon.
\newblock An empirical study of example forgetting during deep neural network learning.
\newblock In {\em ICLR}, 2018.

\bibitem{ullah2021machine}
Enayat Ullah, Tung Mai, Anup Rao, Ryan~A Rossi, and Raman Arora.
\newblock Machine unlearning via algorithmic stability.
\newblock In {\em COLT}, 2021.

\bibitem{warnecke2021machine}
Alexander Warnecke, Lukas Pirch, Christian Wressnegger, and Konrad Rieck.
\newblock Machine unlearning of features and labels.
\newblock {\em arXiv preprint arXiv:2108.11577}, 2021.

\bibitem{xu2023machine}
Heng Xu, Tianqing Zhu, Lefeng Zhang, Wanlei Zhou, and Philip~S Yu.
\newblock Machine unlearning: A survey.
\newblock {\em ACM Computing Surveys}, 2023.

\bibitem{yao2023large}
Yuanshun Yao, Xiaojun Xu, and Yang Liu.
\newblock Large language model unlearning.
\newblock {\em arXiv preprint arXiv:2310.10683}, 2023.

\bibitem{yoon2022few}
Youngsik Yoon, Jinhwan Nam, Hyojeong Yun, Jaeho Lee, Dongwoo Kim, and Jungseul Ok.
\newblock Few-shot unlearning by model inversion.
\newblock {\em arXiv preprint arXiv:2205.15567}, 2022.

\bibitem{yu2004pebl}
Hwanjo Yu, Jiawei Han, and KC-C Chang.
\newblock Pebl: Web page classification without negative examples.
\newblock {\em IEEE Transactions on Knowledge and Data Engineering}, 2004.

\bibitem{zagoruyko2016wide}
Sergey Zagoruyko and Nikos Komodakis.
\newblock Wide residual networks.
\newblock In {\em BMVC}, 2016.

\bibitem{zhang2023forgetmenot}
Eric Zhang, Kai Wang, Xingqian Xu, Zhangyang Wang, and Humphrey Shi.
\newblock Forget-me-not: Learning to forget in text-to-image diffusion models.
\newblock {\em arXiv preprint arXiv:2211.08332}, 2023.

\bibitem{wang2024unlearning}
Qizhou Wang, Bo Han, Puning Yang, Jianing Zhu, Tongliang Liu, and Masashi Sugiyama.
\newblock Unlearning with control: assessing real-world utility for large language model unlearning.
\newblock In {\em arXiv}, 2024.

\end{thebibliography}
\end{document}